%% file: mainthesis-basic.tex
\pdfoutput=1
\documentclass[12pt]{thesis}  %12pt is larger than 11pt

\usepackage{titlesec}
   \titleformat{\chapter}
      {\normalfont\large}{Chapter \thechapter:}{1em}{}
\usepackage{tikz}
\usepackage{times}
\usepackage[USenglish]{babel}
\usepackage{subcaption}
\usepackage{graphicx}
\usepackage{cite}
\usepackage{lscape}
\usepackage{indentfirst}
\usepackage{latexsym}
\usepackage{multirow}
\usepackage{epstopdf}
\usepackage{tabls}
\usepackage{wrapfig}
\usepackage{slashbox}
\usepackage{float}
\usepackage{caption}
\usepackage{supertabular}
\usepackage[onehalfspacing]{setspace} %used for footnote spacing
\usepackage{amsmath}
\usepackage{amsfonts}
\usepackage{booktabs}
\usepackage{comment}
\usepackage{algorithm}
\usepackage{algpseudocode}
\usepackage{xr}
\usepackage{xcolor}
\usepackage{url}
\usepackage{hyperref}
\hypersetup{
    colorlinks,
    linkcolor=blue,
    citecolor=blue,
    urlcolor=blue
}

\definecolor{Red}{RGB}{255, 0, 0}
\definecolor{Blue}{RGB}{0, 0, 255}
\definecolor{DarkYellow}{RGB}{255, 213, 0}
\definecolor{DarkGreen}{RGB}{100, 213, 100}

\usepackage{pifont}% http://ctan.org/pkg/pifont
\newcommand{\cmark}{\ding{51}}%
\newcommand{\xmark}{\ding{55}}%

\newcommand{\note}[1]{{\color{black}{#1}}}
\newcommand{\mb}[1]{{\mathbf{#1}}}
\newcommand{\mc}[1]{{\mathcal{#1}}}

\usepackage{amsmath,amsfonts,amssymb,amsthm}

\newtheorem{problem}{Problem}

\newtheorem*{problem*}{Problem}

\newcommand{\DTWOCOPLAN}{\texttt{\textsc{D2CoPlan}}}
\newcommand{\EXPERT}{\texttt{\textsc{Expert}}}
\newcommand{\DG}{\texttt{\textsc{DG}}}
\newcommand{\DMP}{\texttt{\textsc{DMP}}}
\newcommand{\ORACLE}{\texttt{\textsc{Oracle}}}
\newcommand{\VS}{\textcolor{black}} % Vishnus's comments

\usepackage{algorithm}
\usepackage{algpseudocode}
\usepackage{algorithmicx}
\usepackage{mathtools}
\DeclarePairedDelimiter{\ceil}{\lceil}{\rceil}
\newcommand{\cA}{{\cal A}}
\newcommand{\cS}{{\cal S}}
\usepackage{textcomp}

\usepackage{ulem}\usetikzlibrary{shapes, arrows, automata}
\usepackage{diagbox}

\input{mySymbol.sty}
\input{pennColors.sty}
\def\Tr{\mathsf{T}}

\usepackage{soul}

\newcommand{\cleanfootnote}[1]{%
    \begingroup
    \renewcommand{\thefootnote}{}%
    \footnote{#1}%
    \addtocounter{footnote}{-1}%
    \endgroup
}

\usepackage{tcolorbox} % For creating colored boxes
\tcbuselibrary{skins, breakable} % For skinning the boxes
\newcommand{\promptbox}[2]{%
\begin{tcolorbox}[enhanced jigsaw,breakable,pad at break*=1mm, title=#1,
  colback=blue!10, fontupper={\fontfamily{lmtt}\selectfont}]
  #2
\end{tcolorbox}
}

\newcommand{\responsebox}[2]{%
\begin{tcolorbox}[enhanced jigsaw,breakable,pad at break*=1mm, title=#1,
  colback=green!10, fontupper={\fontfamily{lmtt}\selectfont}]
  #2
\end{tcolorbox}
}

\usepackage{tocloft,calc}
\renewcommand{\cftlottitlefont}{\hspace*{\fill}\large}
\renewcommand{\cftafterlottitle}{\hspace*{\fill}}
\renewcommand{\cftloftitlefont}{\hspace*{\fill}\large}
\renewcommand{\cftafterloftitle}{\hspace*{\fill}}
\renewcommand{\cftchapaftersnum}{:\ }
\renewcommand{\cftchappresnum}{\chaptername\space}
\makeatletter
\g@addto@macro\appendix{%
  \addtocontents{toc}{%
    \protect%
  }%
}
\renewcommand{\baselinestretch}{2}
\begin{document}
\pagestyle{empty}
\include{Abstract} %(must be first, required, non-numbered)
\include{Titlepage} %(must follow Abstract, required, non-numbered)
\include{Copyright} %(highly recommended, non-numbered)

%Pages from this point start at lower-case Roman number ii)
\pagestyle{plain} \pagenumbering{roman} \setcounter{page}{2}

% \phantomsection %create the correct anchor for the bookmark
% \include{Preface}  %(if present, start at lower-case Roman number ii)
%  \addcontentsline{toc}{chapter}{Preface}

% \phantomsection %create the correct anchor for the bookmark
% \addcontentsline{toc}{chapter}{Foreword}
% \include{Foreword} %(if present, lower-case Roman)

\phantomsection %create the correct anchor for the bookmark
\addcontentsline{toc}{chapter}{Dedication}
\include{Dedication} %(if present, lower-case Roman)

\phantomsection %create the correct anchor for the bookmark
\addcontentsline{toc}{chapter}{Acknowledgements}
\include{Acknowledgements} %(if present, lower-case Roman)
    \cleardoublepage

\phantomsection %create the correct anchor for the bookmark
    \addcontentsline{toc}{chapter}{Table of Contents}
    \renewcommand{\contentsname}{Table of Contents}
\renewcommand{\baselinestretch}{1}
\small\normalsize
\tableofcontents %(required, lower-case Roman)
\newpage

\phantomsection %create the correct anchor for the bookmark
\addcontentsline{toc}{chapter}{List of Tables}
    \renewcommand{\contentsname}{List of Tables}
\listoftables %(if present, lower-case Roman)
\newpage

\phantomsection %create the correct anchor for the bookmark
\addcontentsline{toc}{chapter}{List of Figures}
    \renewcommand{\contentsname}{List of Figures}
\listoffigures %(if present, lower-case Roman)
\newpage

% LIST OF ABBREVIATIONS
\phantomsection %create the correct anchor for the bookmark
\addcontentsline{toc}{chapter}{List of Abbreviations}
\include{Abbreviations-supertabular}

\newpage
\setlength{\parskip}{0em}
\renewcommand{\baselinestretch}{2}
\small\normalsize

% Assisted ObjectNav
\newcommand*{\coloredbox}[2]{%
\noindent\colorbox{#1}{%
\parbox{1.0\linewidth}{#2}}}%

%Pages from this point start at Arabic numeral 1
\setcounter{page}{1}
\pagenumbering{arabic}
\include{Chapter1}

\include{Chapter2}
\include{Chapter3}
\include{Chapter4}
\include{Chapter5}
\include{Chapter6}

\include{Chapter7}
\include{Chapter8}

\include{Chapter9}
\include{Chapter10}
\titleformat{\chapter}
{\normalfont\large}{Appendix \thechapter:}{1em}{}
\include{Appendix}

\renewcommand{\baselinestretch}{1}
\small\normalsize

\addcontentsline{toc}{chapter}{} %{Bibliography}
\bibliographystyle{unsrt}
\bibliography{bibfile}

\end{document}

%% file: Abstract.tex
%Abstract Page

\hbox{\ } \vspace{.7in}
\renewcommand{\baselinestretch}{1}
\small \normalsize

\begin{center}
\large{{ABSTRACT}}

\vspace{3em}

\end{center}
\hspace{-.15in}
\begin{tabular}{ll}
Title of Dissertation:    & {\large  ENHANCED ROBOT PLANNING AND PERCEPTION }\\
&                     {\large  THROUGH ENVIRONMENT PREDICTION} \\
% &                     {\large  THEORY AND EXPERIMENT} \\
\ \\
&                          {\large  Vishnu Dutt Sharma} \\
&                           {\large Doctor of Philosophy, 2024} \\
\ \\
Dissertation Directed by: & {\large  Professor Pratap Tokekar} \\
&               {\large  Department of Computer Science } \\
\end{tabular}

\vspace{3em}

\renewcommand{\baselinestretch}{2}
\large \normalsize

Mobile robots rely on maps to navigate through an environment. In the absence of any map, the robots must build the map online from partial observations as they move in the environment. Traditional methods build a map using only direct observations. In contrast, humans identify patterns in the observed environment and make informed guesses about what to expect ahead. Modeling these patterns explicitly is difficult due to the complexity in the environments. However, these complex models can be approximated well using learning-based methods in conjunction with large training data. By extracting patterns, robots can use not only direct observations but also predictions of what lies ahead to better navigate through an unknown environment. In this dissertation, we present several learning-based methods to equip mobile robots with prediction capabilities for efficient and safer operation.

In the first part of the dissertation, we learn to predict using geometrical and structural patterns in the environment. Partially observed maps provide invaluable cues for accurately predicting the unobserved areas. We first demonstrate the capability of general learning-based approaches to model these patterns for a variety of overhead map modalities. Then we employ task-specific learning for faster navigation in indoor environments by predicting 2D occupancy in the nearby regions. This idea is further extended to 3D point cloud representation for object reconstruction. Predicting the shape of the full object from only partial views, our approach paves the way for efficient next-best-view planning, which is a crucial requirement for energy-constrained aerial robots.  Deploying a team of robots can also accelerate mapping. Our algorithms benefit from this setup as more observation results in more accurate predictions and further improves the task efficiency in the aforementioned tasks.

In the second part of the dissertation, we learn to predict using spatiotemporal patterns in the environment. We focus on dynamic tasks such as target tracking and coverage where we seek decentralized coordination between robots. We first show how graph neural networks can be used for more scalable and faster inference while achieving comparable coverage performance as classical approaches. We find that differentiable design is instrumental here for end-to-end task-oriented learning. Building on this, we present a differentiable decision-making framework that consists of a differentiable decentralized planner and a differentiable perception module for dynamic tracking. 

In the third part of the dissertation, we show how to harness semantic patterns in the environment. Adding semantic context to the observations can help the robots decipher the relations between objects and infer what may happen next based on the activity around them. We present a pipeline using vision-language models to capture a wider scene using an overhead camera to provide assistance to humans and robots in the scene. We use this setup to implement an assistive robot to help humans with daily tasks, and then present a semantic communication-based collaborative setup of overhead-ground agents, highlighting the embodiment-specific challenges they may encounter and how they can be overcome. 

The first three parts employ learning-based methods for predicting the environment. However, if the predictions are incorrect, this could pose a risk to the robot and its surroundings. The third part of the dissertation presents risk management methods with meta-reasoning over the predictions. We study two such methods: one extracting uncertainty from the prediction model for risk-aware planning, and another using a heuristic to adaptively switch between classical and prediction-based planning, resulting in safe and efficient robot navigation. 

%% file: Titlepage.tex
%Titlepage

\thispagestyle{empty} \hbox{\ } \vspace{1.5in}
\renewcommand{\baselinestretch}{1}
\small\normalsize
\begin{center}

\large{{ENHANCED ROBOT PLANNING AND PERCEPTION\\ THROUGH ENVIRONMENT PREDICTION}}\\
\ \\
\ \\
\large{by} \\
\ \\
\large{Vishnu Dutt Sharma}%Your full name as it appears in University records.
\ \\
\ \\
\ \\
\ \\
\normalsize
Dissertation submitted to the Faculty of the Graduate School of the \\
University of Maryland, College Park in partial fulfillment \\
of the requirements for the degree of \\
Doctor of Philosophy \\
2024
\end{center}

\vspace{7.5em}

\noindent Advisory Committee: \\
\hbox{\ }\hspace{.5in}Professor Pratap Tokekar, Chair/Advisor \\
\hbox{\ }\hspace{.5in}Professor Nikhil Chopra, Dean's Representative \\
\hbox{\ }\hspace{.5in}Professor Dinesh Manocha \\
\hbox{\ }\hspace{.5in}Professor Tianyi Zhou \\
\hbox{\ }\hspace{.5in}Professor Kaiqing Zhang

%% file: Copyright.tex
%Copyright

\thispagestyle{empty}
\hbox{\ }

\vfill
\renewcommand{\baselinestretch}{1}
\small\normalsize

\vspace{.5in}

\begin{center}
\large{\copyright \hbox{ }Copyright by\\
Vishnu Dutt Sharma  %Type your name as it appears in University records
\\
2024}
\end{center}

\vfill

\newpage 

%% file: Dedication.tex
%Dedication

\renewcommand{\baselinestretch}{2}
\small\normalsize
\hbox{\ }
 
\vspace{.5in}

\begin{center}
\large{Dedication}

\null\vfill
To my brother, Ramdutt Sharma. 
\vfill\vfill

\clearpage

\newpage 

\clearpage

\null\vfill

\textit{``We are stardust brought to life, then empowered by the universe \\to figure itself out – and we have only just begun."}\\
\hfill  --- \textit{Neil deGrasse Tyson}
\vfill\vfill

\clearpage

\end{center} 

%% file: Acknowledgements.tex
%Acknowledgments

\renewcommand{\baselinestretch}{2}
\small\normalsize
\hbox{\ }
 
\vspace{.5in}

\begin{center}
\large{Acknowledgments} 
\end{center} 

\vspace{1ex}
My Ph.D. journey has been a thrilling rollercoaster, filled with exhilarating highs and challenging lows. While I officially embarked on this path five years ago, the journey truly started many years earlier, and it would not have been possible without the unwavering support of many individuals who stood by me all along. 

First and foremost I express my deepest gratitude to my advisor, Dr. Pratap Tokekar, for his tireless support and mentorship at every step of this journey. Looking back, I recognize how he encouraged me to explore my interests and gently nudged me in the right direction when necessary, helping me grow into an independent researcher. His support extended beyond academic guidance, providing me with empathy and encouragement throughout this process. As my friend and colleague Deeksha aptly put it, ``he is a great adviser and an amazing guide", and I am deeply grateful for the opportunity to have worked with him.

I am profoundly thankful to my committee members, Dr. Nikhil Chopra, Dr. Dinesh Manocha, Dr. Tianyi Zhou, and Dr. Kaiqing Zhang, for their invaluable feedbacks and insights on my dissertation. I also want to extend my gratitude to Dr. Manocha for his advice during the preliminary exam and for connecting me with the collaborators at the GAMMA lab at UMD, which led to the projects forming an essential part of this dissertation.

Heartfelt thanks to my collaborators and co-authors: Harnaik Singh Dhami, Lifeng Zhou, Anukriti Singh, Vishnu Sashank Dorbala, Qingbiao Li, Jingxi Chen, and Maymoonah Toubeh. This dissertation was made possible with their help and working with them expanded the horizons of my conceptual and practical knowledge. I am also grateful to have worked with and learned from Dr. Matthew Andrews,  Jeongran Lee, and Ilija Had\v{z}i\'{c} during my internship.

This dissertation would not have been possible without the generous financial support from several sources: the U.S National Science Foundation (grant \#1943368), the Office of Naval Research (grant \#N00014-18-1-2829), Kulkarni Foundation, Nokia Bell Labs, and Comcast Corporation. I am grateful to Ivan Penskiy and the Maryland Robotic Center for providing the necessary hardware for experiments. I also extend my thanks to IEEE RAS, the Department of Computer Science at UMD, and the  Graduate School at UMD for support in the form of travel grants.  Attending conferences with these grants allowed me to experience research on a broad scale and connect with the wider research community.

I owe my deepest gratitude to my friends who kept me going through the adversities: Akshita Jha, Siddharth Jar, Aman Gupta, Vikram Mohanty, Rajnish Aggarwal, Abhilash Sahoo, Alisha Pradhan, Biswaksen Patnaik, and Pramod Chundury. Special thanks to Akshita and Siddharth, who were always a call away, to lend an ear to all my personal and professional problems and provide kind and energizing words. I also want to thank Aman for motivating me to keep working towards starting my Ph.D. journey.

I am grateful to Dr. Pawan Goyal and Dr. Amrith Krishna, who provided me with my first opportunity to pursue academic research during my undergraduate studies. Under their guidance, I learned the fundamental skills that have carried me through my academic career. 

Many thanks to my friends and lab-mates from the RAAS Lab: Guangyao Shi, Amisha Bhaskar, Prateek Verma, Jingxi Chen, Troi Williams, Charith Reddy, Deeksha Dixit, Rui Liu, Chak Lam Shek, Zahir Mahammad, and Sachin Jadhav. I thank them for making this experience enjoyable. 

This journey would have been unimaginable without the unwavering love and support of my family. I owe them a deep gratitude for their patience, understanding, and encouragement throughout this journey.

Finally, I wish to thank the creators of Naruto. I started watching the show when I was going through a very rough patch. The character instilled in me the courage to keep going and start this journey. It transformed the anger within into acknowledgment and kindness towards myself, setting me on a path that ultimately led to this accomplishment.

%% file: Abbreviations-supertabular.tex
%List of Abbreviations

\renewcommand{\baselinestretch}{1}
\small\normalsize
\hbox{\ }

\vspace{.5in}

\begin{center}
\large{List of Abbreviations}
\end{center} 

\vspace{3pt}
\begin{supertabular}{ll}
AUV & Autonomous Underwater Vehicle \\
CNN & Convolutional Neural Network \\
DS & Dialogue Similarity \\
GA & Ground Agent \\
GC & Generative Communication \\
GNN & Graph Neural Network \\
LLM & Large Language Model \\
MAE & Masked Autoencoder \\
MLP & Multi-Layer Perceptron \\
NBV & Next Best View \\
OA & Overhead Agent \\
OSR & Oracle Success Rate \\
SCT & Success weighted by Completion Time \\ 
SLAM & Simultaneous Localization and Mapping \\
SPL & Success weighted by Path Length \\
SR & Success Rate \\
UAV & Unmanned Aerial Vehicle \\
UGV & Unmanned Ground Vehicle \\
VLM & Vision-Langugae Model \\

\end{supertabular}

%% file: Chapter1.tex
%Chapter 1

\renewcommand{\thechapter}{1}

\chapter{Introduction}
\label{chapter:intro}

\label{sec:intro:into}
Navigating through unknown environments is a fundamental capability of mobile robots and has been studied by the robotics community for a long time~\cite{kuipers2017shakey}. Onboard sensors help them perceive their surroundings and planning algorithms enable them to navigate through the environment~\cite{kuipers2017shakey}. If a robot already has a map of the environment, it can plan optimal routes between locations, even in the presence of some dynamic objects~\cite{fox_dynamic_1997}. However, the presence of a map is not always guaranteed. 
In certain scenarios, building a map from the ground up is itself the very mission assigned to a robot~\cite{yamauchi_frontier-based_1997, durrant2006simultaneous}. 

Efficient navigation requires careful selection of the next location to move to, at each time step. 
Should the robot need to traverse to a pre-defined goal, it must move to locations such that the time taken to reach the goal is minimized. If the objective is to build a map, it must move to locations that help in observing most of the environment in the fewest steps. At each step, the local observations are integrated to construct a partial global map, and a new path is planned based on the updated map. Typically, a planner produces navigation strategies either avoiding or exploring unknown areas, depending upon the task at hand and safety constraints. Unsurprisingly, the amount of information about the map acts as the bottleneck for efficient navigation. So we ask, \textit{is there a way to make the robot navigation process more efficient even with limited information}? To answer this, we look towards humans for inspiration.

 In similar circumstances where there is limited information, humans show remarkable efficiency in navigation by making guesses about the yet-unseen part of the environment. At the core of this cognitive process lies our ability to identify patterns in our surroundings. We can walk through our living room without directly looking at the floor and by just mentally visualizing the furniture in the room even if we can only see a part of it. We move through a crowd gracefully by walking faster or slower by intuitively anticipating other people's motions. 
 Locating a book, say \emph{Probabilistic Robotics}, within a library is easy once we know the positions of the volumes starting with 'O' and 'Q'. In all these instances, we rely on patterns we have identified from our prior experiences to inform our decision-making. Inspired by these feats, the central question we aim to answer in this dissertation is: \textit{can robots also reason about the regions beyond their field of view by employing similar pattern recognition capabilities to improve their navigation efficiency? }

 To answer this question, it is important to understand how humans develop these capabilities. Our brain learns these patterns through experiences and builds an internal model to facilitate their applications~\cite{shugen2002framework, konovalov2018neurocomputational, gregory1970intelligent}. The easiest solution to impart such capabilities to the robots is therefore creating such models manually for them. However, handcrafting models for these patterns prove to be a formidable task, owing to our limited knowledge of the system-environment interaction and the exact distribution of these patterns~\cite{krizhevsky2017imagenet}. In contrast, learning-based approaches, especially deep neural networks, offer a promising avenue to approximate these models using extensive training data. This dissertation introduces an array of \textit{learning-based algorithms} for this purpose, demonstrating their efficacy in enabling robots to make informed predictions about 
 unobserved environments, resulting in enhancing navigation efficiency for a variety of input views, modalities, and number of robots. Being mindful of the inherent \textit{approximations}, we concurrently develop methods to manage risks when using predictions to ensure the robots can utilize them safely. Thus, \textit{the goal of this dissertation is to equip robots
 with pattern recognition capabilities and facilitate the judicious use of predictions to enhance navigation efficiency and safety}. 

\section{Types of Patterns and Informed Decision-Making}
\label{sec:intro:patterns}
In this section, we start by introducing some patterns often harnessed by humans which prove to be of significant utility to robots for informed decision-making. Specifically, we focus on geometrical and structural, spatiotemporal, and semantic patterns, and describe their characteristics and relevance to robotic planning in the subsections below.

\subsection{Geometrical and Structural Patterns}
\label{subsec:intro:geometrical_intro}
Geometrical or structural patterns refer to recurring arrangements or configurations of shapes or structures in the environment. There is a plethora of these patterns present in our surroundings, mainly due to our affinity towards finding regularity in what we see, even in scenarios where they may not be immediately apparent. A great example of this is star constellations, where we can visualize animals and objects in a seemingly random arrangement of tiny twinkling lights.

The human brain is remarkable at recognizing and organizing visual information through the application of\textit{ Gestalt principles}~\cite{wertheimer1938laws,faulkner1998essence}. These fundamental concepts in psychology explain how we perceive complex scenes as cohesive wholes, rather than isolated elements. They not only equip us to understand our surroundings but are also expressed as regular designs of man-made objects. Consider the prevalent rectilinear design of walls, tables, and boxes, or the cylindrical shape of bottles, cups, and tumblers. Despite variations in specific details, objects often adhere to similar underlying structures. Ask a child to draw an airplane, and we would expect to see a tube with pointed ends with two triangles on the sides. Boeing and Airbus models may differ in their tail designs, but the general shape is similar. Another prevalent characteristic of man-made objects is symmetry. Together, this results in easy-to-remember and recognizable shapes all around us. 

\begin{figure}
    \centering
    \includegraphics[width=\textwidth]{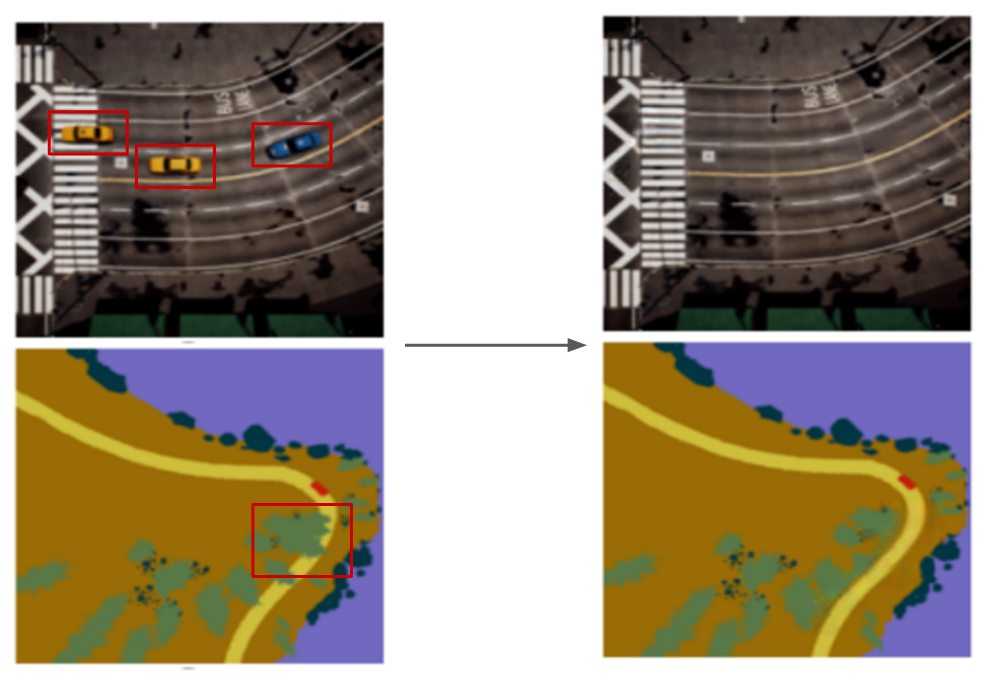}
    \caption{Motivating Application with Geometrical Structural Patterns: Limited observations and occlusion can limit the robots' planning capabilities. Geometrical and structural predictions can help them make informed decisions by predicting the map beyond direct observations. The images are from the \textit{City} and \textit{Forest} environments in AirSim.}
    \label{fig:intro:structural_motivating_eg}
\end{figure}

 \textit{We show that these principles can be exploited in robot navigation, especially when the robots work in man-made environments, by predicting the object shapes beyond the robot's field of view.} The abundance of familiar and recurring shapes presents a great opportunity for robots to infer the complete shape from partial views, making efficient navigation possible. Deploying a team of robots leads to further improvement as more information helps in more accurate predictions, making these patterns immensely useful for robot navigation. Identifying and modeling these patterns is a challenging task. Additionally, the underlying method should ensure generalizability which is crucial for the deployment of the robot to new scenarios, including the real world.

\subsection{Spatiotemporal Patterns}
\label{subsec:intro:spatiotemporal_intro}
Spatiotemporal patterns refer to the arrangement of entities in their scene, not necessarily geometrical, and the regularity in their motion. While we focus on static arrangements in geometrical and structural patterns, here we aim to couple the motion with the spatial arrangements and harness the predictable cues. For \textit{``mobile''} robots these cues can arise from the environment, as well as their own motion. 

Soccer is the most popular sport in the world. Millions of fans watch it in areas and on TV. Passing the ball, one of the most fundamental skills in soccer involves one player kicking the ball toward a teammate and the opponents try to intercept it. The motion of the can be easily predicted with Newton's laws of motion. But the players and the viewers do not have to pull their notebooks out to calculate where the ball would be after a certain time. Even those unfamiliar with Newton's laws can easily predict the trajectory of the ball as humans can anticipate the motion from their experience. Similar patterns can be found in how people move in the crowd and how cars navigate on the road and are used by us to safely and smoothly navigate through them. 

\begin{figure}[ht]
    \centering
    \includegraphics[width=0.65\textwidth]{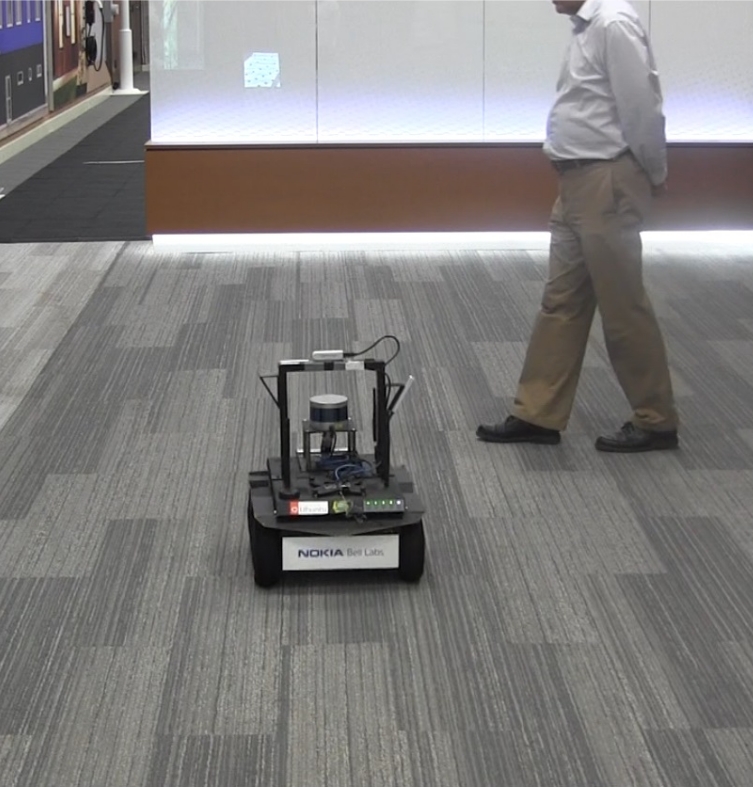}
    \caption{Motivating Application with Spatiotemporal Patterns: The motion of dynamic objects can be difficult to model. Spatiotemporal patterns can help the robots estimate the motion to avoid or track them efficiently. The image is from experiments performed at Nokia Bell Labs as an intern.}
    \label{fig:intro:spatiotemporal_motivating_eg}
\end{figure}

\textit{Robots also can learn to anticipate motion based on experience, as demonstrated in our works.} Robots can extract these patterns from moving targets, predicting their future locations for better tracking. When working with other robots, they could further utilize their spatial arrangements to coordinate with others to improve robot navigation for the whole team, especially when their communication capabilities are limited. Limited observations pose a significant challenge in these scenarios, which is further exacerbated by the need for scalable solutions which can be hard to model and learn when generating labels with optimal algorithms for large teams is impractical. 

\subsection{Semantic Patterns}
\label{subsec:intro:semantic_intro}
Semantic patterns mean a consistent arrangement or structure of words, symbols, or elements that convey meaning within a given context to express higher-level specific concepts, relationships, or information. While the first two types of patterns often come to us intuitively, semantic patterns are usually reasoned about explicitly~\cite{kahneman2011thinking}.

Human decision-making is highly dependent on semantic entities, resulting in a world full of such patterns. Looking at someone pouring coffee into the cup, one may guess they may need sugar or milk next is one such example where one can use a pattern emerging from the understanding of objects, their semantic relationship, and the activity they imply.
Language, color codings, and sounds are some examples of such patterns. 
The inherent \textit{intuitiveness} in semantic patterns paves the way to make accurate predictions about them.

\begin{figure}[ht]
    \centering
    \includegraphics[width=0.9\textwidth]{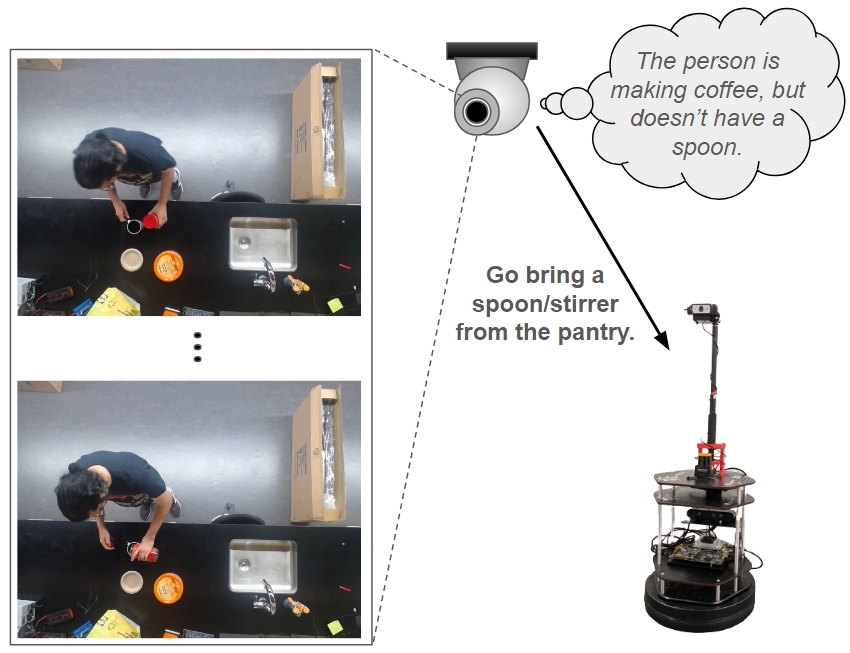}
    \caption{Motivating Application with Semantic Patterns: Lack of semantic understanding can make the robots reliant on human instructions to perform tasks for them. Semantic patterns can help the robots anticipate what a human may need, and help them proactively. The images are from our experiments at RAAS Lab, University of Maryland.}
    \label{fig:intro:semanticmotivating_eg}
\end{figure}

\textit{The ubiquity of the semantics around us means it is essential for robots to reason semantically to efficiently assist humans. It also means that similar to us, the robots can make inferences to navigate efficiently in such a world.} Using these patterns can enable robots to fully utilize the man-made signals around them and in turn effectively work alongside humans by reasoning similarly to them. Learning these patterns requires a huge amount of data and computational resources, which may not be easily available. The emergence of foundational and large language models addresses this issue and promises an easy avenue for scene understanding and human-level planning but requires bridging them together with grounding approaches to make them realizable. 

\section{Research Contributions}
This section highlights our contributions proposing methods to leverage the three types of patterns. The key proposition of these methods is to utilize learning-based approaches for modeling the patterns from data and improving planning efficiency across a variety of tasks and scenarios. Additionally, we explore methods for risk management with predictive models for the safe deployment of robots.

We start with a discussion on how geometric and structural patterns can be used to improve PointGoal navigation~\cite{anderson2018evaluation}, i.e. traversing to a pre-defined goal through an unknown environment, and active object reconstruction by making accurate predictions about unseen regions from partial observations. These works take advantage of the patterns in the static environment around them. Then we turn our attention to dynamic entities in the scene and detail our approaches to harnessing spatiotemporal patterns arising from the motion of a team of robots and the dynamic targets and their spatial arrangements for scalable coverage and tracking. Our works show how predictions are invaluable to efficient navigation for mobile robots, but erroneous predictions can be a point of concern for safety in some applications. We deal with this issue in the last subsection with our meta-reasoning approaches for the safe navigation of the robot using heuristics and uncertainty-based strategies for risk management. 

\subsection{Enhanced Perception with Structural Continuity and Closure}
\label{subsec:intro:geometrical}
This section describes our contributions to using \textbf{geometrical and structural patterns} to enhance robot perception and, as a result, planning. We start with a discussion on how geometric and structural patterns can be used to improve robot navigation for PointGoal Navigation with a 2D occupancy map, and active object reconstruction based on 3D point cloud representations, by predicting unseen regions from only partial observations. Lastly, we show that these approaches can be effectively extended to multi-agent systems by using multiple views to further improve the predictions, and in turn, the navigation efficiency.

 \begin{figure}[ht!]
    \centering
    \includegraphics[width=.80\linewidth]{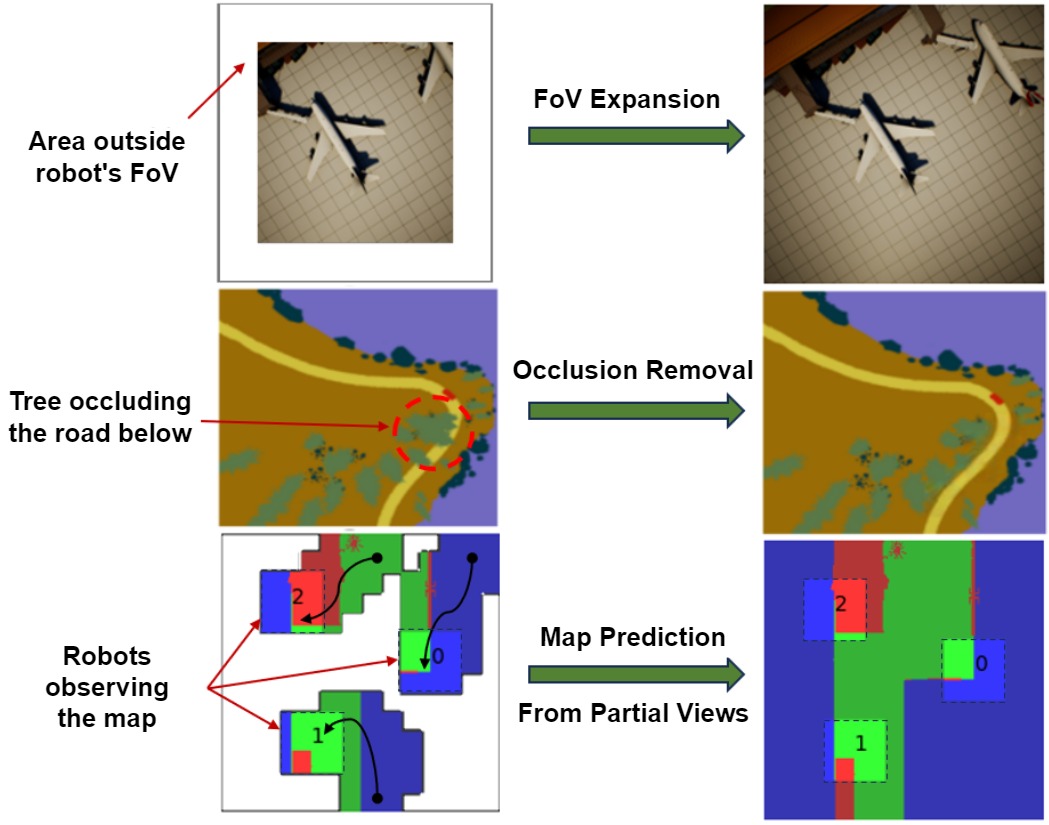}
    % \vspace{0.1pt}
    \caption{Example of some tasks that can be done with geometrical and structural patterns}
    \label{fig:ch1:fliptd_tasks}
\end{figure}

% \VS{Talk about it}
\subsubsection{PointGoal Navigation with 2D Maps}
\label{subsubsec:intro:proxmap}
\textit{PointGoal navigation} refers to the navigation task in which the robot is given a specific destination point (goal) in the environment and is required to reach it~\cite{anderson2018evaluation}. 2D overhead or bird's eye view (BEV) maps are commonly used by ground robots for this task. Typically, the robot builds the navigation maps incrementally from local observations using onboard sensors. 2D ranging LiDARs and RGB-D cameras are the most popular sensors for this task and are used to generate an occupancy map, which distinguishes the free areas from occupied or unknown areas. Sometimes, an unmanned aerial vehicle may act as a scout and observe a wider area from a height to get RGB maps, over which semantic segmentation is applied to act as an occupancy map for the navigation of the ground robot. The planner plans a path to the goal using these maps. As the robot navigates and updates the map, the path is also updated as a result. 

A conservative planner may avoid the regions of the unknown regions for safety, taking a longer time to navigate to the goal. Instead, if a robot is able to correctly predict the occupancy in the occluded regions, the robot may navigate efficiently. Recent works have shown that predicting the structural patterns in the environment through learning-based approaches can greatly enhance task efficiency~\cite{elhafsi2020map, katyal_high-speed_2021}. This is accomplished by predicting the occupancy maps in the yet unobserved regions, effectively increasing the field of view of the sensor. Figure~\ref{fig:ch1:fliptd_tasks} shows some example applications with this capability. We show that the existing foundational vision networks can accomplish this without any fine-tuning by using the concept of \textbf{continuity} learned from the computer vision datasets.

Specifically, we use Masked Autoencoders~\cite{he_masked_2022}, pre-trained on street images, for the field of view expansion on RGB images, semantic segmentation maps, and binary maps. The images and maps span both outdoor and indoor scenes and a diverse set of locations from AirSim~\cite{shah2018airsim} and AI2THOR simulator\cite{kolve2017ai2}. Two key findings stand out from our experiments in this work: (1) inferring unobserved scenes is easier for more simpler and abstract representations such as semantic segmentation and binary map as they don't require complex reasoning about textures, and (2) predictions are more accurate when made closer to the areas of direct observation and degrade as we move farther away. The limitations of such foundation models are that they are computationally heavy and may not be suitable for the structural patterns of the region with missing information. This work was \textbf{accepted at the 2024 International Conference on Robotics and Automation (ICRA 2024)}~\cite{sharma2023pre}.

 \begin{figure}[ht!]
    \centering
    \includegraphics[width=1.0\linewidth]{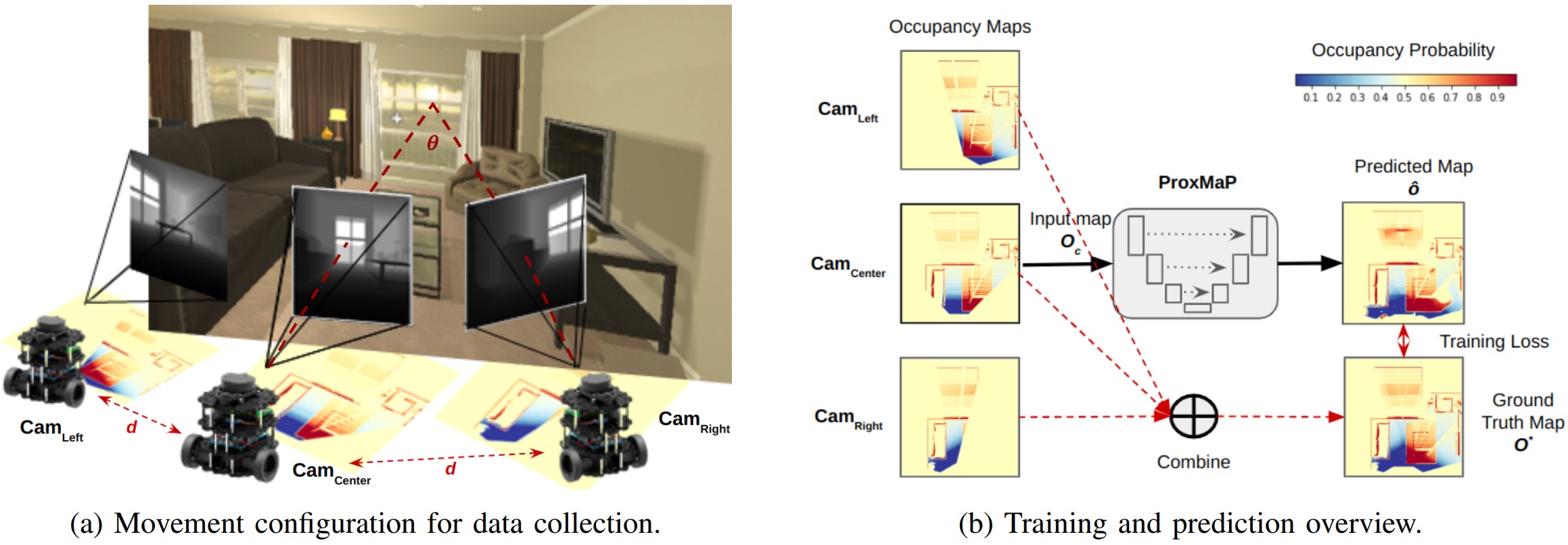}
    % \vspace{0.1pt}
    \caption{Movement configuration for data collection and training strategy for proximal occupancy map prediction (ProxMaP)}
    \label{fig:ch1:proxmap}
\end{figure}

To overcome these limitations, we look towards task-specific, convolutional neural networks (CNNs) based approaches for occupancy map prediction. Existing works using CNN for this task learn to predict occupancy in areas away from direct observations and thus may suffer from network overfitting~\cite{ramakrishnan_occupancy_2020, georgakis_uncertainty-driven_2022}. This also requires a time-consuming data collection step. To alleviate these issues, prior work has proposed a self-supervised approach with multi-camera setup~\cite{wei2021occupancy}, but this setup does not result in precision in predicted maps and is not economical. 
We use the takeaways from our previous work and focus on making predictions near the observed regions only, thus reducing overfitting and making precise predictions. This also results in a self-supervised and efficient data collection approach (as shown in Figure ~\ref{fig:ch1:proxmap}), which is also more economical than the existing self-supervised approach. We further improve navigation by adjusting the robot's speed according to the information over the path to the goal, resulting in faster navigation to the goal. This work was \textbf{accepted at the 2023 IEEE/RSJ International Conference on Intelligent Robots and Systems (IROS 2023)}~\cite{sharma2023proxmap}. 

\subsubsection{Active Object Reconstruction with 3D Point Clouds}
\label{subsubsec:intro:prednbv}
\textit{Object reconstruction} refers to the task of observing an object from multiple viewpoints and reconstructing a 3D representation of it. 3D LiDARs, which provide a sparse point cloud as output, are often used with UAVs for this task. Ideally, we can observe the object from a large constellation of viewpoints around the objects, an efficient plan requires selecting only those viewpoints which minimize the overlap with the previous observations to finish the task quickly. This is a crucial requirement when using UAVs as the limited battery capacity limits its flight time. 

\textit{Similar to the 2D situation, to incrementally build the map, the robot must choose the \textit{next-best-view} (NBV) carefully to produce an accurate reconstruction with an efficient plan}.
If the shape of the object (equivalent to the map) is already known, we can carefully select a minimal number of viewpoints resulting in a \textit{geometrical NBV approach}. But this may not always be possible and such a requirement limits the situation where the UAV can be deployed. Prior works have used learning-based approaches for this, but these networks are usually specific to an object class (e.g. houses only). This means that we must train a model for each target object. Other works have used the Gestalt principle of \textit{similarity} to bridge the gap between lack of map and geometrical NBV approach. Revisiting the example from Section~\ref{sec:intro:into}, all planes have similar shapes even though they may differ in details. Thus if we can predict the full space of the object of interest from partial views (this is known as \textit{shape completion}), we can effectively construct a geometric NBV plan.

 \begin{figure}[ht!]
    \centering
    \includegraphics[width=.90\linewidth]{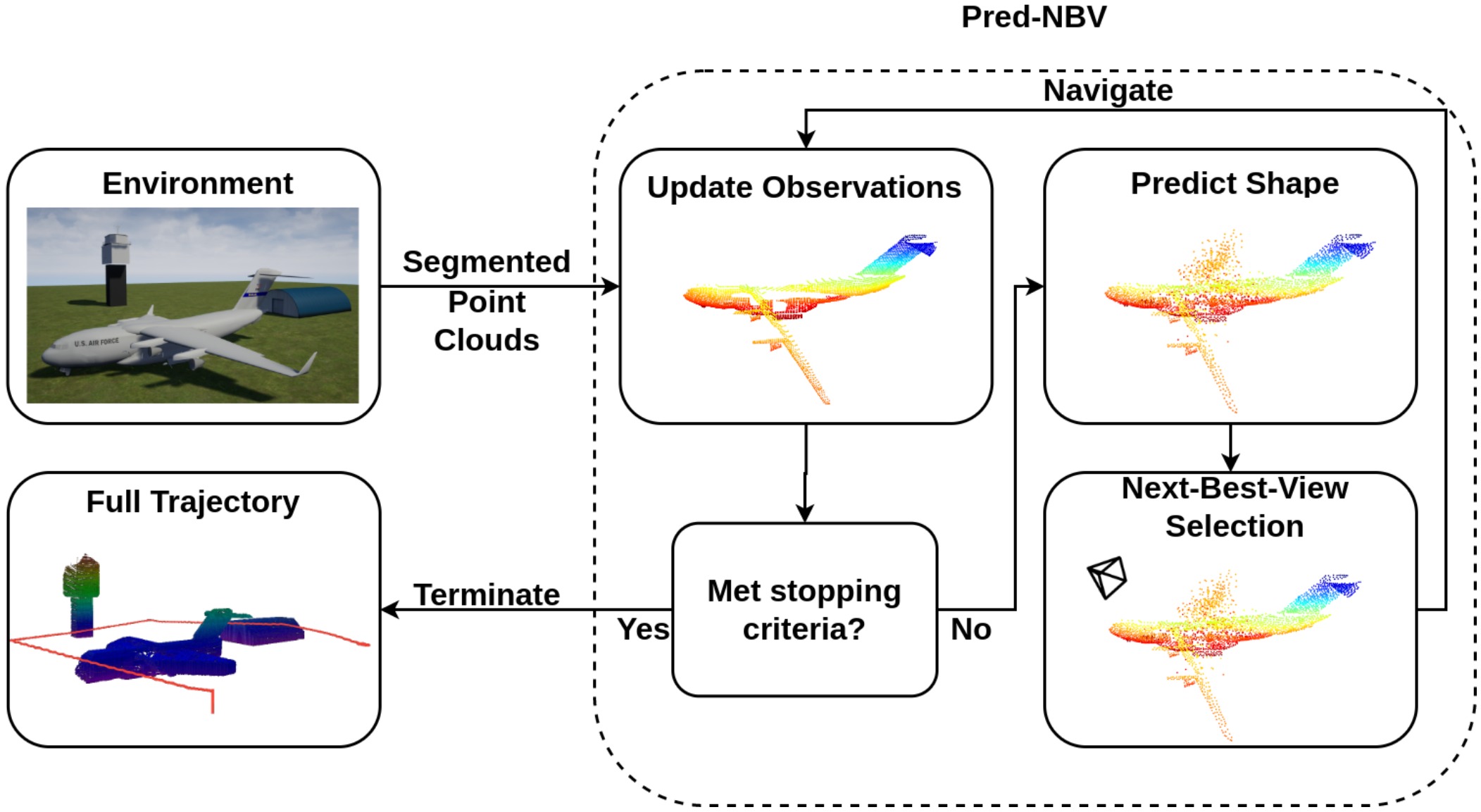}
    % \vspace{0.1pt}
    \caption{Overview of our prediction-guided next-best-view approach (Pred-NBV) for object reconstruction}
    \label{fig:ch1:prednbv}
\end{figure}

The existing works for 3D shape prediction make an implicit assumption about the partial observations and therefore cannot be used for real-world planning~\cite{yu_pointr_2021}. Also, they do not consider the control effort for next-best-view planning, which directly affects the flight time~\cite{wu20153d}. We proposed Pred-NBV~\cite{dhami2023prednbv}, a realistic object shape reconstruction method consisting of PoinTr-C, an enhanced 3D prediction model trained on the ShapeNet dataset using curriculum learning, and an information and control effort-based next-best-view method to address these issues. Figure~\ref{fig:ch1:prednbv} shows the overview of Pred-NBV. In each iteration, the robot predicts the full shape from partial observations and goes to the closest location which will result in a high information gain. After moving to the new location, the current observations are combined with the previous ones. The process repeats till we reach a termination condition. Pred-NBV achieves an improvement of 25.46\% in object coverage over the traditional methods in the AirSim simulator~\cite{shah2018airsim} and performs better shape completion than PoinTr~\cite{yu_pointr_2021}, the state-of-the-art shape completion model, even on real data obtained from a Velodyne 3D LiDAR mounted on DJI M600 Pro. This work was \textbf{accepted at the 2023 IEEE/RSJ International Conference on Intelligent Robots and Systems (IROS 2023)}~\cite{dhami2023prednbv}.

\subsubsection{Extensions to Multi-Robot Systems}
\label{subsubsec:intro:multirobot}
Previous subsections show how continuity and closure principles can improve the navigation efficiency for single robot systems by making predictions over partial observations. It is intuitive to expect further benefits when more information and context (and hence more geometrical and structural cues) are provided to the prediction model. This brings the question, \textit{what if these ideas are extended to multi-robot systems}? 

\begin{figure}[ht!]
    \centering
    \includegraphics[width=.80\linewidth]{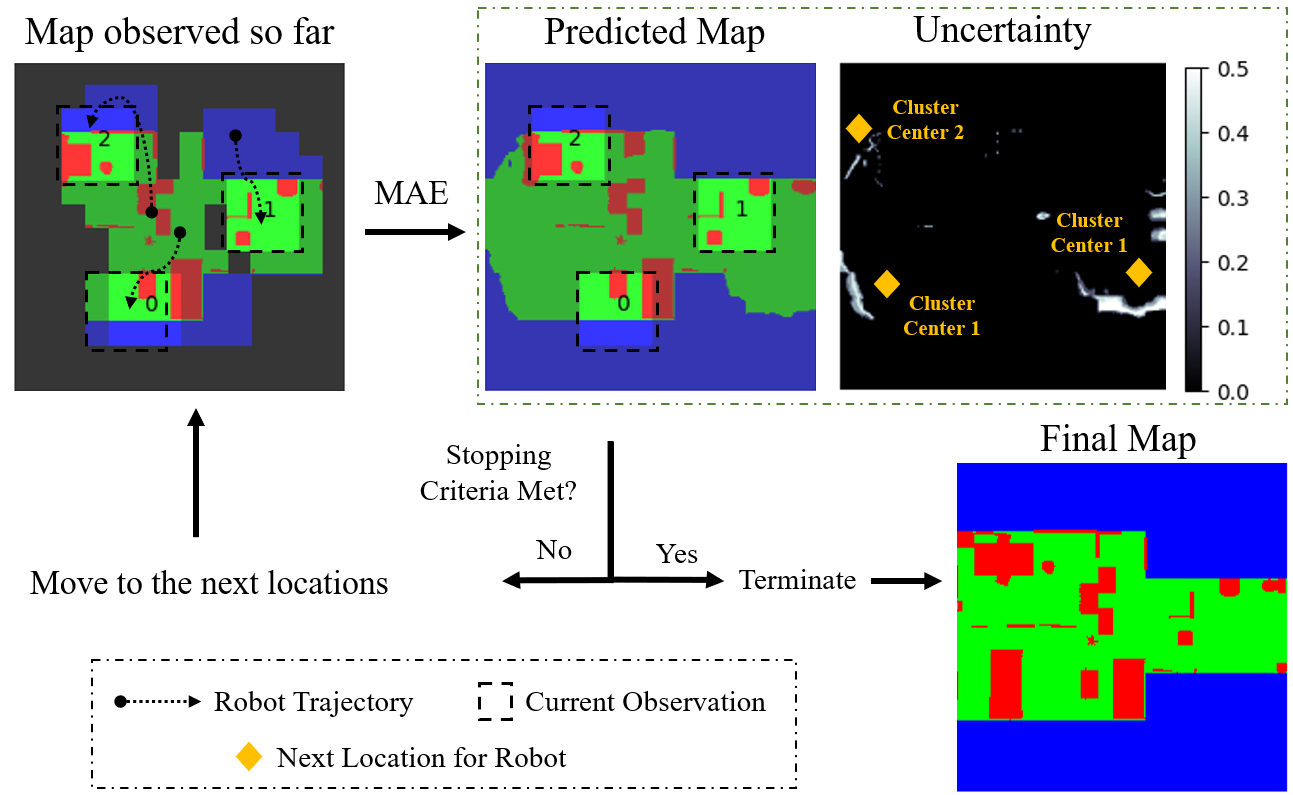}
    % \vspace{0.1pt}
    \caption{Overview of our prediction and uncertainty-driven planning approach for multi-agent coverage}
    \label{fig:ch1:fliptd_coverage}
\end{figure}

For navigation with a 2D map, we study the task of multi-agent exploration, which requires a team of robots to observe the whole map in the fewest steps. We use predictions for planning as well as mapping by inferring the unexplored regions using continuity. As observations from multiview points provide strong geometrical and structural cues, the predictions are more reliable. Whenever there is uncertainty in predictions, we navigate the robots to these regions to directly observe and reduce uncertainty. For this, we use MAE to predict the unexplored regions and extract uncertainty from them by adding visually imperceptible noise to the input. Then a centralized planner uses the K-Nearest Neighbors (KNN) algorithm to identify regions with unknowns and high uncertainty to identify regions of interest and moves the robots to them. Figure~\ref{fig:ch1:fliptd_coverage} shows an overview of this process. This method results in higher prediction accuracy in fewer steps compared to the traditional method of assigning non-overlapping regions to robots and scanning them in a sweeping fashion. These findings were \textbf{accepted at the 2024 International International Conference on Robotics and Automation (ICRA 2024)}~\cite{sharma2023pre}.

For 3D object reconstruction, we extended prior work to MAP-NBV, a prediction-guided active algorithm for 3D reconstruction with multi-agent UAV systems. We use PoinTr-C similar to Pred-NBV, but add a centralized planner to find NBV for all the UAVs together. We jointly optimize the information gain and control effort for efficient collaborative 3D reconstruction of the object. Our method achieves a 19\% improvement over the non-predictive multi-agent approach and a 17\% improvement over the prediction-based, non-cooperative multi-agent approach. This work was \textbf{accepted at the 2024 IEEE/RSJ International Conference on Intelligent Robots and Systems (IROS 2024)}~\cite{dhami_map-nbv_2023}.

\subsection{Spatiotemporal Patterns}
\label{subsec:intro:spatiotemporal}
In this section, we describe the \textbf{spatiotemporal patterns} and our contributions outlining their use for improving robot planning. Specifically, we focus on decentralized algorithms for multi-robot coverage and target tracking, putting primary emphasis on planning. The first subsection aims to harness the patterns emerging from the spatial arrangement of the robots, represented as a communication graph, and their motion through the space. In the next subsection, we additionally utilize similar patterns from moving targets.

\subsubsection{Decentralized Coordinated Coverage using Graph Neural Networks}
\label{subsubsec:intro:gnn}
\begin{figure}[ht!]
    \centering
    \includegraphics[width=.90\linewidth]{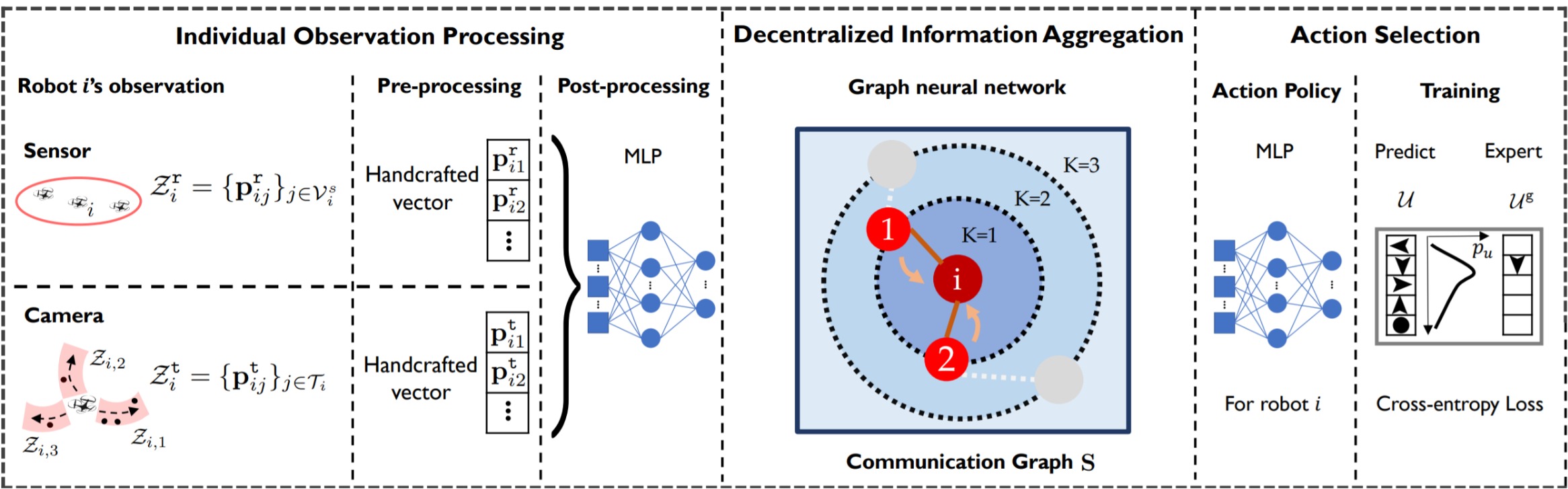}
    % \vspace{0.1pt}
    \caption{Our graph neural network (GNN) based method for decentralized multi-robot coverage}
    \label{fig:ch1:gnn}
\end{figure}

The problem of decentralized multi-robot target tracking asks for jointly selecting actions, e.g., motion primitives, for the robots to maximize the joint coverage with local communications. One major challenge for practical implementations is to make target-tracking approaches scalable for large-scale problem instances. In this work, we propose a general-purpose learning architecture toward collaborative target tracking at scale, with decentralized communications. Classical, manually designed decentralized approaches can be more scalable compared to centralized ones at the cost of reduced coverage~\cite{gharesifard2017distributed}. We investigate whether planners, designed using learning-based algorithms, can accomplish the same when provided with local observations as hand-crafted features. Particularly, our learning architecture, shown in Figure~\ref{fig:ch1:gnn}, leverages a graph neural network (GNN) to capture local interactions of the robots and learns decentralized decision-making for the robots~\cite{zhou_graph_2022}. We train the learning model by imitating an expert solution and implement the resulting model for \textit{decentralized action selection involving local observations and communications only}. 

Using GNN here based on the representation of the spatial arrangement as a graph also allows \textit{centralized training, decentralized execution}~\cite{tolstaya2020learning}, a property specific to GNNs. We demonstrate the performance of our learning-based approach in a scenario of active target tracking with large networks of robots. The simulation results show our approach nearly matches the tracking performance of the expert algorithm, and yet runs several orders faster with up to 100 robots. Moreover, it slightly outperforms a decentralized greedy algorithm but runs faster (especially with more than 20 robots). The results also exhibit our approach's generalization capability in previously unseen scenarios, e.g., larger environments and larger networks of robots. \textit{This shows that learning-based approaches can patterns for efficient planning, achieving similar coverage with lesser inference speed in comparison to the equivalent classical approaches}. This work was \textbf{accepted at the 2022 IEEE Symposium on Safety, Security, and Rescue Robotics (SSRR 2023)}~\cite{zhou_graph_2022}. 

\subsubsection{Decentralized Coverage and Tracking with Differentiable Planners}
\label{subsubsec:intro:d2coplan}
\begin{figure}[ht!]
    \centering
    \includegraphics[width=.90\linewidth]{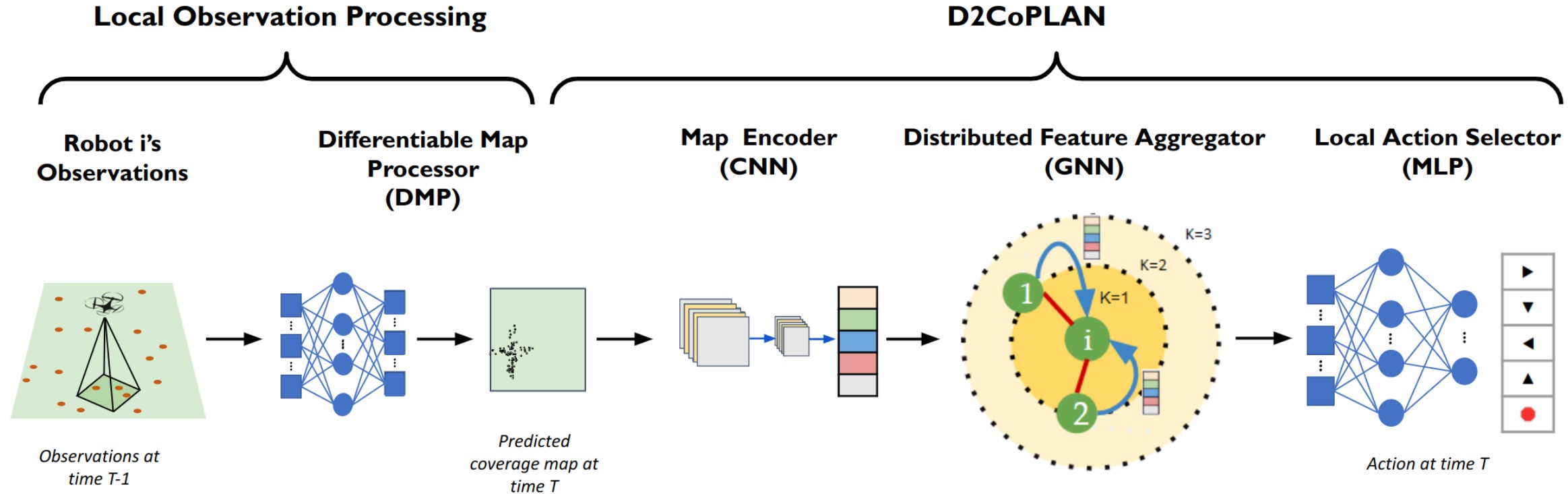}
    % \vspace{0.1pt}
    \caption{Our tracking and coverage approach using a differentiable decentralized coverage planner (\texttt{D2CoPlan})}
    \label{fig:ch1:d2coplan}
\end{figure}

 Learning-based distributed algorithms provide a scalable avenue in addition to bringing data-oriented feature generation capabilities to the table, allowing integration with other learning-based approaches. Our previous work focuses on local communication through GNN to improve planning. As we demonstrated in Section~\ref{subsec:intro:geometrical}, perception is often harder to model than planning. Thus we ask, \textit{can a learning-based planner augment the training for learning-based perception models to outperform a simple combination of the two}? Realizing this setup required an end-to-end differentiable planner that can be seamlessly combined with a perception network. To this end, we present a learning-based, differentiable distributed coverage planner (\texttt{D2CoPlan})~\cite{sharma_d2coplan_2023}, shown in Figure~\ref{fig:ch1:d2coplan}, that scales efficiently in runtime and number of agents compared to the expert algorithm and performs on par with the classical distributed algorithm. Then we combine it with a perception network to predict motion primitives for covering dynamic targets, hence solving a target tracking problem. We find that this modular combination is not only able to outperform combinations of classical and learning-based counterparts but also learns more efficiently than a single monolithic end-to-end planning network. These findings suggest that \textit{differentiable designs in perception and planning are key to the development of more powerful learning-based solutions through end-to-end, task-specific learning}. This work was published in the \textbf{proceeding of the 2023 International Conference on Robotics and Automation (ICRA 2023)}~\cite{sharma_d2coplan_2023}.

\subsection{Semantic Pattern Prediction to Assist Humans and Robots}
\label{subsec:intro:semantic}
In this section, we present our contributions utilizing \textit{semantic patterns} grounded in vision and language to assist humans and other robots. For the former, we investigate the innate world knowledge in Vision-Language Models (LLMs) and Vision-Language Models (VLMs) to anticipate what a person may need in the future and help them with the task. We specifically focus on language-based semantic patterns.

\subsubsection{Large Language Models as Anticipatory Planners for Assisting Humans}
\label{subsubsec:intro:llm}
Large Language Models (LLMs) are among the most recent significant advancements in artificial intelligence (AI). Trained using reinforcement learning from human feedback (RLHF), these models are exceptionally good at conversations with humans. Within a few months of their introduction, people have come up with a huge range of applications using LLMs as human surrogates. Language researchers are studying what LLMs learn and have revealed that they can reason about the world which carries huge implications for applications dependent on semantic patterns. 

Consider this scenario: a person wakes up in the morning and is getting ready to make coffee. They reach the kitchen counter and turn on the coffee maker. An assistive home robot observing them infers that they are making coffee but notices that there is no sugar nearby. Anticipating that they may need sugar next, the assistant sends a robot to bring sugar from the pantry and bring it to the human. This ability to help humans without them needing to ask the robot explicitly requires a world model to understand and make inferences for a wide array of human activities.

Using the capability of the LLMs to act as approximate world models, we use their capability to generate likely words to assist humans by anticipating their next actions. Here we pose the next action prediction as the next word prediction with LLMs, given the past observations about a human's activity. Due to the generative nature and lack of grounding in LLMs, they may generate many plausible actions. To ground them into real applications, we provide the context in the form of a textual description of the map. The robot action primitives are used as additional input to ensure \textit{affordance}-based actions are selected to effectively assist the humans. We build this system in and show applications in the real world for a variety of tasks in a home-like environment for image and text-based observation, VLMs, and image captioning methods.

\subsubsection{Vision Language Models as Global Context Providers for Assisting Robots}
\label{subsubsec:intro:assistive_objectnav}
Recently researchers have been exploring multi-model representations learning to combine multiple representations in the same embedding, mainly aimed at tasks to assist humans. One such joint representation that is relevant to robots is vision-language representation, which can help the agent equipped with such VLM not only comprehend the observations but also make predictions about the future state of the environment. Focusing on such \textit{semantic predictions} we propose a pipeline where an environment camera monitors the surroundings to identify the ongoing activities and directs the robot to help the human with them. Having such cameras in the environment is not unusual nowadays; cameras are often used for security purposes in industrial residential spaces. Growing interests in AI assistants have also brought devices equipped with cameras, which can provide additional guidance to the robot to help the people around. 

\begin{figure}[ht]
    \centering
    \includegraphics[width=0.9\textwidth]{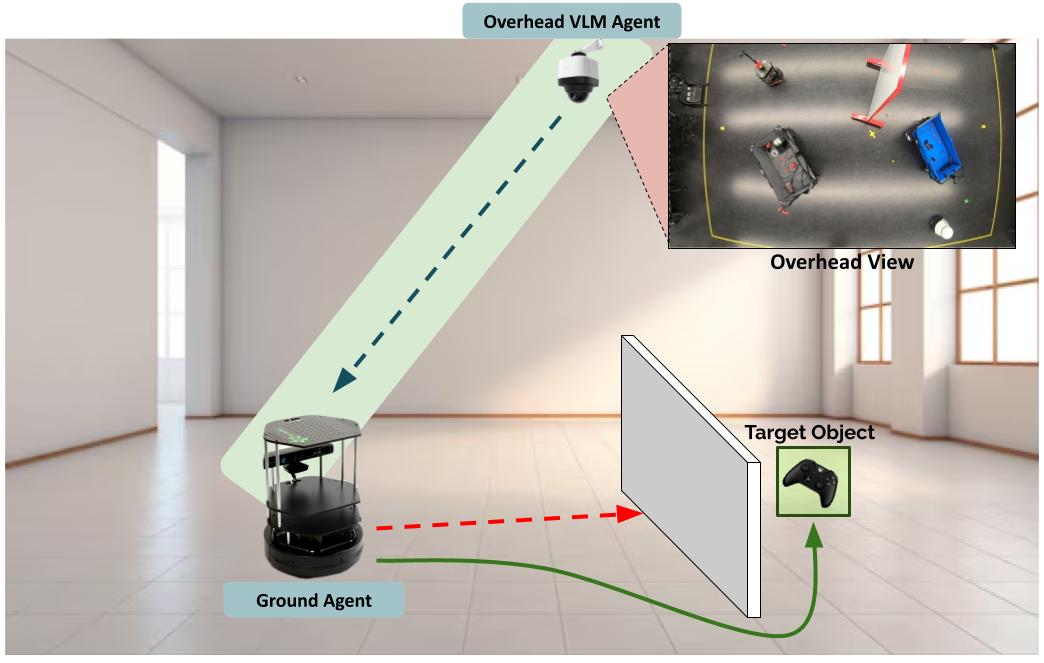}
    \caption{A VLM-based overhead agent working along with a ground robot can act as an effective assistance unit}
    \label{fig:intro:semantic_demo_overview}
\end{figure}

As depicted in Figure~\ref{fig:intro:semantic_demo_overview}, the overhead camera monitors and deciphers the activities using VLMs, acting as a \textit{overhead VLM agent}. The ground robot, or\textit{ground agent} can only see limited parts of the scene, due to limited camera field-of-view and the occlusions in the scene (such as walls). In contrast, the placement of the overhead camera allows it to observe a wider area than a ground robotic agent could. But it can not move around and help the people in the environment, which the ground agent can do. We propose a method to utilize the strengths of these agents: the overhead agent is tasked with scene understanding, activity recognition, and predicting what assistance a person in the scene may need. It can then direct the ground robot to move around and perform the appropriate actions. Since the overhead agent may not be able to see all the details from the height and walls may obstruct its view of some areas, the ground agent uses VLMs to accomplish the task. We implemented this pipeline using GPT-4 to direct a Turtlebot2 robot to help humans in a house-like environment in the real world, paving a path to assistive robots with the help of external sensors and VLM-based capabilities.

\begin{figure}[ht]
    \centering
    \includegraphics[width=0.9\textwidth]{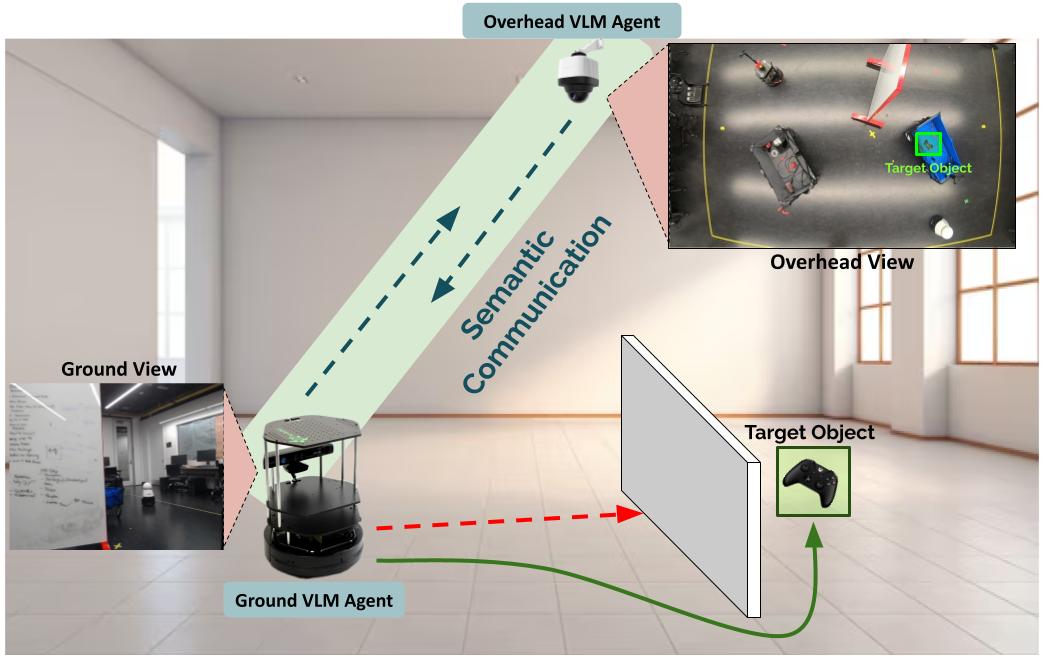}
    \caption{Two VLM-based agents, one with an overhead view and another with a ground view, can work together to find an object on a scene with generative communication}
    \label{fig:intro:semantic_gc_overview}
\end{figure}

Given the world knowledge that VLMs encompass, they are well-equipped to assist not only in the form of \textit{temporal semantic predictions}, but can also prove helpful for \textit{spatial semantic prediction} (e.g., a bowl of sugar is more likely to be in the pantry than in a bathroom). Owing to these properties, the research community has welcomed these models with open arms for many applications, including ObjectNav. ObjectNav~\cite{anderson2018evaluation} is an embodied task where a mobile robot must find an object (e.g., a fork) in an environment without a map. This task is challenging as the \textit{ground agent} has limited observation due to limited field-of-view (FoV) and obstructions, and hence its planning horizon is limited. While a VLM may provide sound reasoning about the target object and where to find it, the lack of grounding, hallucinations, and reliance on limited observations pose challenges for effective applications. To address these challenges, we propose using an environment camera, also equipped with a VLM to provide additional guidance as an \textit{overhead agent}. However, the overhead agent itself may suffer from occlusions from walls and objects or may confuse other objects with the robot, it must communicate with the GA to estimate the latter's position well and provide accurate guidance to accomplish ObjectNav.

This proposed approach, shown in Figure~\ref{fig:intro:semantic_gc_overview}, is the first instance of two agents with global and local views communicating with VLMs to the best of our knowledge. Similar prior works have focused on emergent \textbf{semantic communication} with limited vocabulary, in contrast to generative communication with unrestricted communication. To study the effect of communication, we further investigate the effect of communication length and varying degrees of communication over the task performance and show that communication indeed a crucial role in completing the task. To mitigate the adverse effects of hallucination in simple two-way communication, we propose a selective cooperation framework for this task and achieve \textit{10\%} improvement over the non-assistive, single-agent method. This work is currently \textbf{under review}.

\subsection{Meta-Reasoning to Manage Risk in Predictions}
\label{subsec:intro:metareasoning}
The accuracy of learning-based models is largely dependent on the training data. Generalizability is often a source of concern for these methods and manifests itself as a Sim2Real gap in robotic applications. An error in prediction can be dangerous to the surroundings, the humans nearby, and even to the robot itself. Trust in predictions is thus a critical issue for deploying robots in the real world. 

Existing works have explored this issue through the lens of uncertainty extraction~\cite{kendall_bayesian_2017}, interpretable designs~\cite{sharma_interpretable_2022}, and explainable methods~\cite{sakai2022explainable} among other. We contribute in this regard with implicit and explicit meta-reasoning approaches over predictions for planning, as described below.

\subsubsection{Meta-Reasoning for Risk-Aware Planning with Implicit Means}
\label{subsubsec:intro:riskaware}
Perception networks used in neural networks are generally point prediction models, i.e. for the same inputs, the network provides the same deterministic output. Bayesian neural networks, which were designed for stochastic outputs, can be computationally intensive and thus are unsuitable for deployment on resource-constrained robots. Kendall et al.~\cite{kendall_bayesian_2017} proposed Bayesian SegNet which uses Bayesian dropouts~\cite{gal2016dropout} for sementic segmentation and uncertainty extraction and show their use for detecting uncertainty in street-view images for autonomous driving. 

\begin{figure}[ht!]
    \centering
    \includegraphics[width=.90\linewidth]{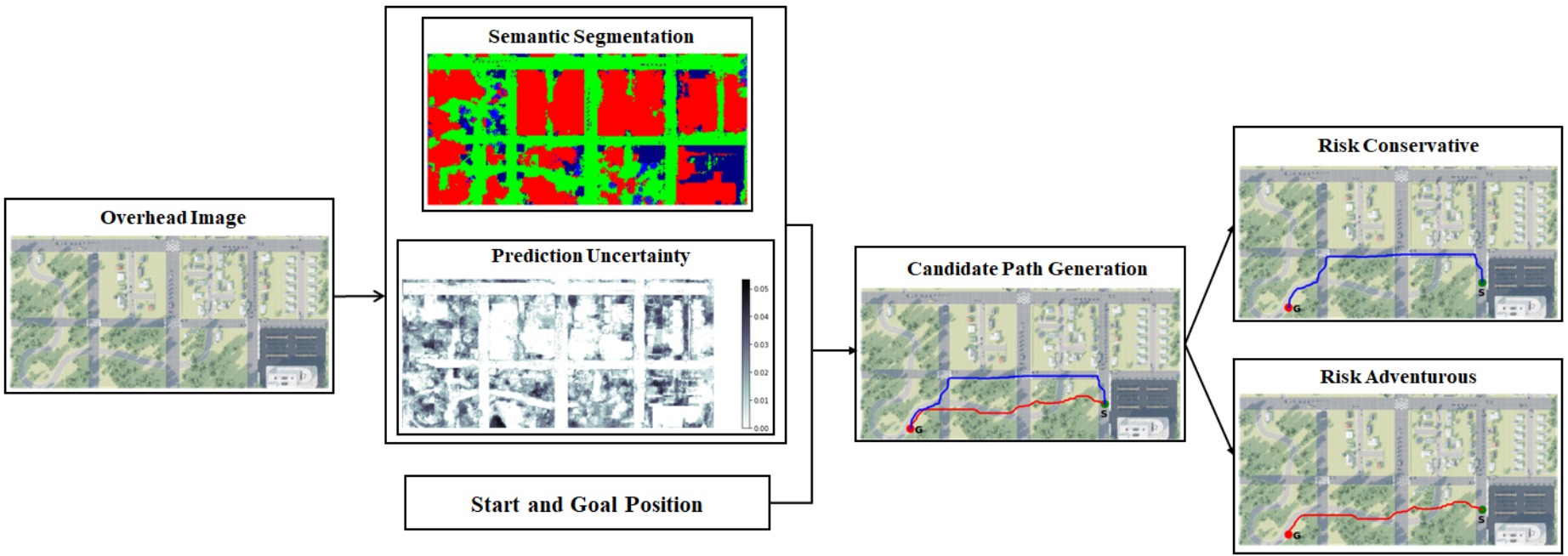}
    % \vspace{0.1pt}
    \caption{Risk-aware planning strategy using uncertainty extraction allows the user to choose between a \textcolor{blue}{conservative} and \textcolor{red}{adventurous} plans}
    \label{fig:ch1:risk_aware_planning}
\end{figure}

Our work builds on this idea and uses Bayesian SegNet on top-down images for high-level planning. In this setup, an aerial robot acts as a scout for ground robot navigation and uses Bayesian SegNet on aerial images. We train our network in the \textit{CityEnviron} environment and test it in one suburban scene. The grass patches, which are scarce in the city act as out-of-distribution entities and result in high prediction uncertainty. The semantic costmap is combined with the uncertainty map using a user-defined risk-affinity factor, which allows the user to select between risk-conservative and risk-adventurous paths. The proposed approach thus allows risk management using implicit uncertainty in the networks. This process is shown in Figure~\ref{fig:ch1:risk_aware_planning}. This work was \textbf{published at the 2020 IEEE/RSJ International Conference on Intelligent Robots and Systems (IROS 2020)}~\cite{sharma_risk-aware_2020}.

\subsubsection{Meta-Reasoning for Hybrid Planning with Explicit Means}
\label{subsubsec:intro:hybrid}
Uncertainty extraction from network prediction is a contentious issue as we do not know how to truly validate the '\textit{uncertainty}' thus obtained. This may result in a gap in trustworthiness, even if learning-based methods are more desirable and suitable for certain scenarios. The easiest solution in such scenarios can be to use the predictions only when we trust them, otherwise, switch back to traditional methods. Recent works have explored this idea, albeit with neural networks acting as the switch~\cite{dey2023learning, raj2023targeted}, which can again cause trust issues and may not be generalizable.

\begin{figure}[ht!]
  \centering
  \begin{subfigure}[b]{0.70\textwidth}
    \includegraphics[width=0.95\textwidth]{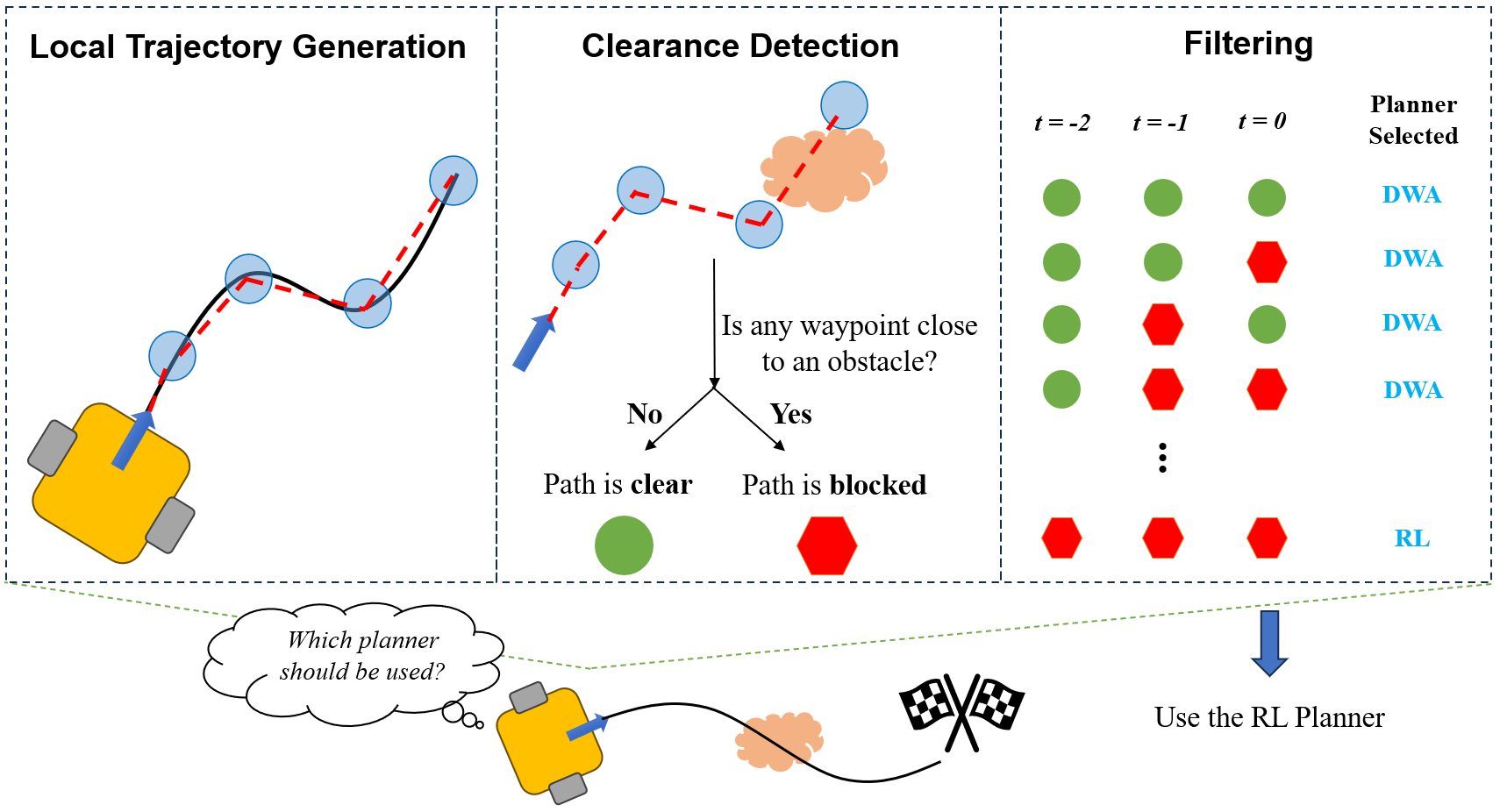}
    % \caption{Caption for Plot 1}
  \end{subfigure}
  \hfill
  \begin{subfigure}[b]{0.29\textwidth}
    \includegraphics[width=1.0\textwidth]{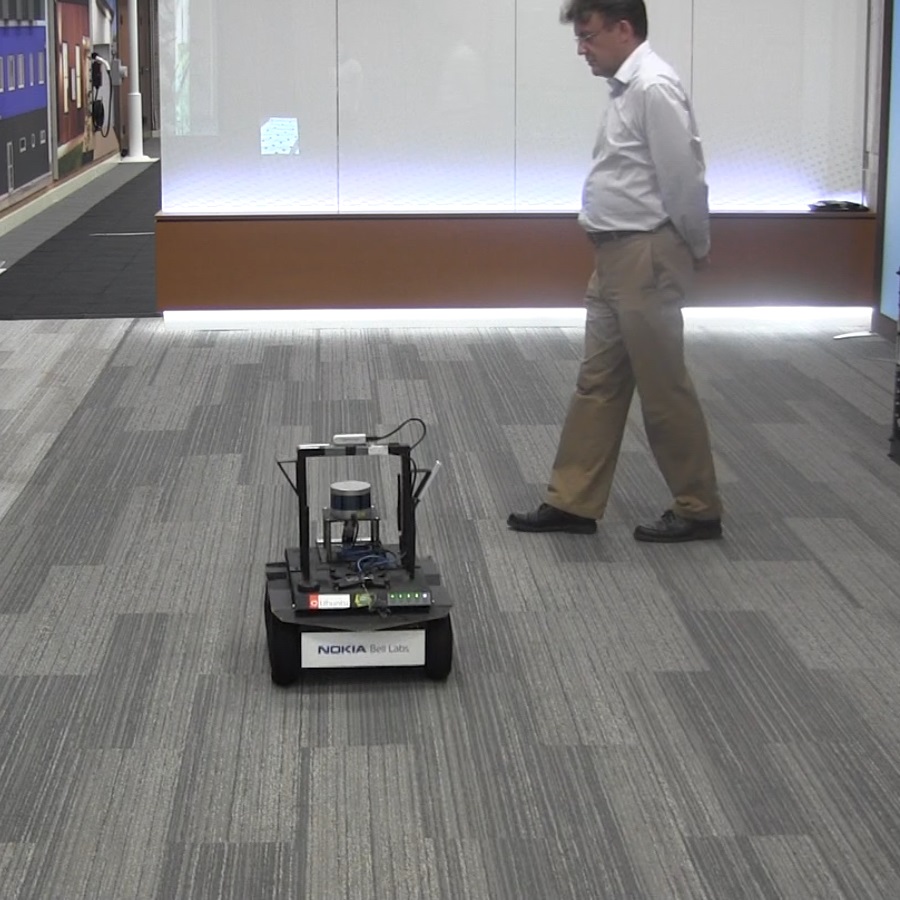}
    \vspace{12pt}
    % \caption{Caption for Plot 2}
  \end{subfigure}
  \caption{Left: Overview of the proposed hybrid local planning approach which combines the benefits of classical and AI-based local planners. Right: A real experiment scenario showing a hybrid planner in action when a human suddenly appears on the robot's path. }
  \label{fig:ch1:hybrid_planner_overview}
\end{figure}

We use a heuristic-based approach switching between a classical and a reinforcement learning (RL) based method for local planning in an indoor environment with unexpected obstacles. The classical planner, dynamic window approach (DWA)~\cite{fox_dynamic_1997} is robust and moves the robot efficiently on a smooth path. However, DWA may be slow to react to unexpected obstacles on its path and may result in a collision. SACPlanner~\cite{nakhleh_sacplanner_2023}, an RL-based local planner is more reactive in these situations but results in a jerky and inefficient motion. We thus come up with a switching approach to identify whether there is an unexpected obstacle on the path and switch between the two planners accordingly. If the path is clear, then it uses the DWA planner, resulting in fast progress towards the goal. If an obstacle is detected, the robot uses the RL-based planner to safely avoid the obstacles and switches back to the DWA if no other obstacle is on the path to the goal. This method results in faster navigation to the goal without any collision with the obstacles in various scenarios in real-world experiments. Figure~\ref{fig:ch1:hybrid_planner_overview} visualizes this process and shows an experimental situation where this framework was tested in the real world. This work is currently under revision~\cite{sharma2024hybrid}. 

\subsubsection{Organization of the Dissertation}
\label{subsubsec:intro:organization}
This dissertation is organized into 10 chapters, following this chapter.

In Chapters~\ref{chapter:proxmap}-~\ref{chapter:mapnbv}, we present methods to utilize structural and geometrical patterns for efficient navigation planning. Chapters~\ref{chapter:proxmap} and \ref{chapter:fliptd} focus on predicting unseen parts of 2D maps from partial views with task-specific training\footnote{\url{https://raaslab.org/projects/ProxMaP}} and a pre-trained masked image model\footnote{\url{https://raaslab.org/projects/MIM4Robots}}, respectively. Chapters~\ref{chapter:prednbv} and \ref{chapter:mapnbv} present methods for predicting 3D maps using partial observations and building next-base-view planning approaches with them for single-agent\footnote{\url{https://raaslab.org/projects/PredNBV}} and multi-agent systems\footnote{\url{https://raaslab.org/projects/MAPNBV}}, respectively. 

Chapter~\ref{chapter:gnn_d2coplan} concentrates on leveraging spatiotemporal patterns. First, we show how learning-based approaches can act as approximate, but scalable planners for multi-agent coverage problems\footnote{\url{https://github.com/VishnuDuttSharma/deep-multirobot-task}}. Then we present a method using a learning-based decentralized approach as a differentiable planner to efficiently train a multi-agent tracking method in an end-to-end manner, achieving better results than its counterpart composed of independently trained submodules\footnote{\url{https://raaslab.org/projects/d2coplan}}. 

In Chapter~\ref{chapter:llm_assist} we present a framework using semantic patterns with the help of VLMs and LLMs. This framework uses an overhead camera to direct or coordinate with a ground robot. We first show this framework in action with a real-world implementation to assist a human in doing everyday tasks in a house-like environment. Then we use this framework to help ground robot equipped with VLM for ObjectNav and present an analysis of the conversation between the two agents.

Chapters~\ref{chapter:risk_aware} and \ref{chapter:hybrid} present methods to manage risk in predictions. In Chapter~\ref{chapter:risk_aware} we propose an implicit measure for risk-aware path planning where the out-of-distribution data acts as the source of uncertainty. Chapter~\ref{chapter:hybrid} focuses on an explicit measure for risk management by using a heuristics-based approach to switch between classical and learning-based planners to balance navigation efficiency and the collision risk due to unexpected obstacles on the path. 

We conclude the dissertation with an overview of our research contributions and underline the future research directions. The prompts and additional results from Chapter~\ref{chapter:llm_assist} are presented in the appendix. The software and media corresponding to the work in this dissertation are available at \url{https://vishnuduttsharma.github.io/thesis/}.

%% file: Chapter2.tex
%Chapter 2

\renewcommand{\thechapter}{2}

\chapter{Structural and Geometric Pattern Prediction in 2D Occupancy Maps}
\label{chapter:proxmap}

\section{Introduction}
\cleanfootnote{Further details and results for this work are available at \url{https://raaslab.org/projects/ProxMaP}.}

To navigate in a complex environment, a robot needs to know the map of the environment. This information can either be obtained by mapping the environment beforehand, or the robot can build a map online using the onboard sensors. Occupancy maps are often used, which provide probabilistic estimates about the free (navigable) and occupied (non-navigable) areas. These estimates can be updated as the robot gains new information while navigating. 
Given an occupancy map, the robot can adjust its speed to navigate faster through high-confidence, free areas and slower through low-confidence, free areas so that it can stop before collision. The effective speed of the robot thus depends on the occupancy estimates.
Occlusions due to obstacles and limited field-of-view (FoV) of the robot lead to low-confidence occupancy estimates, which limit the navigation speed of the robot.

\begin{figure}
  \centering
  \begin{subfigure}{0.45\linewidth}
    \includegraphics[width=1.0\linewidth]{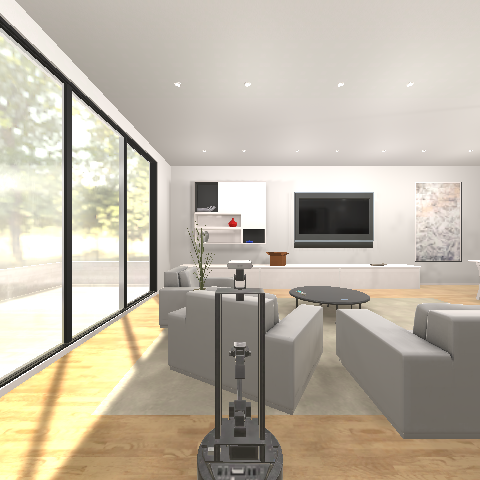}
    \caption{Third-person view of the robot in a living room}
    \label{fig:proxmap:robot_eg}
  \end{subfigure}
%   \hfill
  \begin{subfigure}{0.41\linewidth}
    \includegraphics[width=1.0\linewidth]{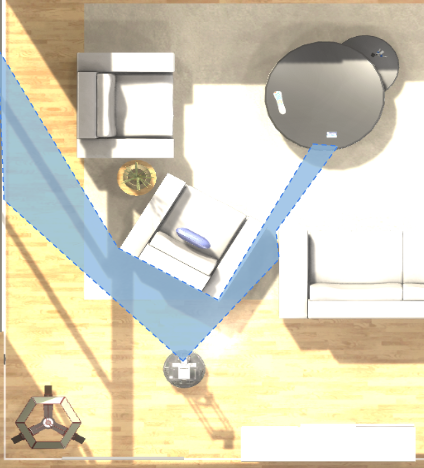}
    \caption{Top view of the robot showing visibility polygon}
    \label{fig:proxmap:camera_eg}
  \end{subfigure}
  \caption{An example situation where the robot's view is limited by the obstacles (sofa blocking the view) and the camera field of view (sofa on the right is not fully visible).}
  \label{fig:proxmap:real_bot}
\end{figure}

In this chapter, we propose training a neural network to predict occupancy in the regions that are currently occluded by obstacles, as shown in Fig.~\ref{fig:proxmap:real_bot}. Prior works learn to predict the occupancy map all around the robot i.e., simulating a $360^\circ$ FoV given the visible occupancy map within the current, limited FoV~\cite{ramakrishnan2020occupancy, georgakis2022uncertainty,katsumata2022map}.  
Since the network is trained to predict the occupancy map all around the robot, it overfits by learning the room layouts. This happens as the network must learn to predict the occupancy information about the areas such as the back of the robot, for which the robot may not have any overlapping information in its egocentric observations. This makes the prediction task difficult to learn. Furthermore, the ground truth requires mapping the whole scene beforehand, which could make sim-to-real transfer tedious. It also means that the whole environment needs to be mapped to get the ground truth data.

Our key insight is to simplify this problem by making predictions only about the proximity of the areas where the robot could move immediately. This setting has three-fold advantages: first, the prediction task is easier and relevant as the network needs to reason only about the immediately accessible regions (that are partly visible); second, the robot learns to predict obstacle shapes instead of learning room layouts, making it more generalizable; and third, ground truth is easier to obtain which can be obtained by moving the robot, making the approach self-supervised.

\begin{figure*}
  \centering
  \mbox{}\hfill 
  \begin{subfigure}{0.43\linewidth}
    \includegraphics[width=1.0\linewidth]{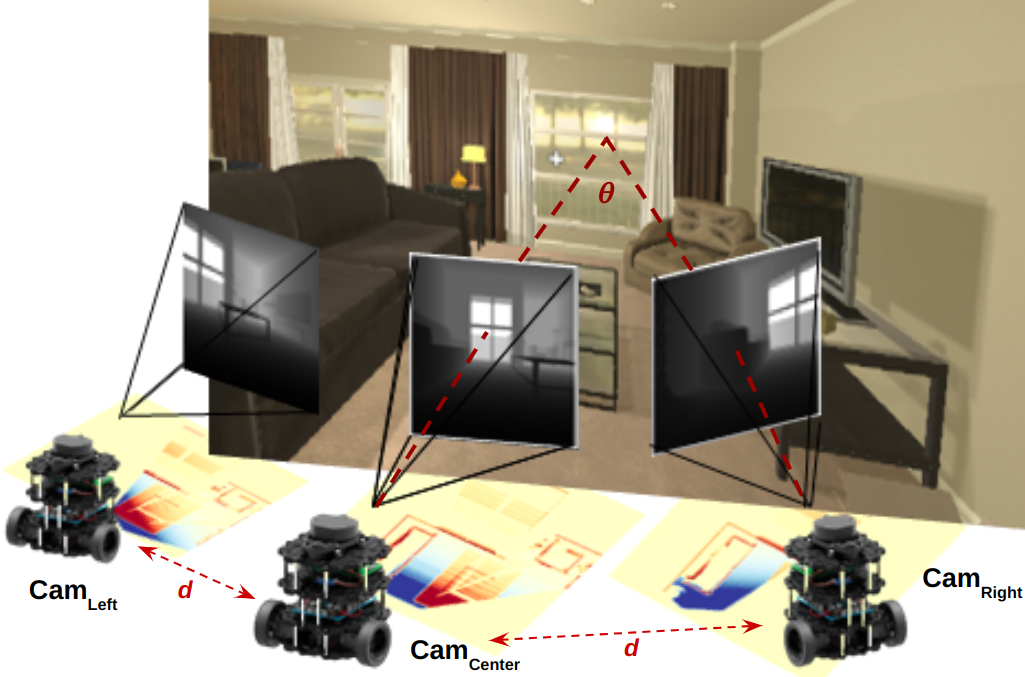}
    \caption{Movement configuration for data collection.}
    \label{fig:proxmap:over_a}
  \end{subfigure}
  \hfill
  \begin{subfigure}
  {0.45\linewidth}
    \includegraphics[width=1.0\linewidth]{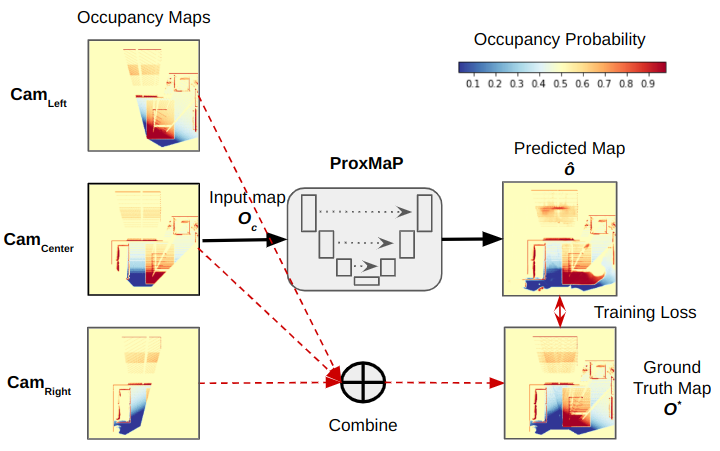}
    \caption{Training and prediction overview.}
    \label{fig:proxmap:over_b}
  \end{subfigure}
  \hfill\mbox{}
  \caption{Overview of the proposed approach. The \textcolor{red}{\textbf{training}} and \textbf{inference} flows are indicated with red and black arrows, respectively. We take the input view by moving the robot to the left and right sides (\textit{$Cam_{Left}$} and \textit{$Cam_{Right}$}), looking towards the region of interest. ProxMaP makes predictions using the \textit{$Cam_{Center}$} only, and the map obtained by combining the information from the three positions acts as the ground truth. }
  \label{fig:proxmap:overview}
\end{figure*}

Following are our main contributions in this work:
\begin{enumerate}
\item We present ProxMaP, a self-supervised proximal occupancy map prediction method for indoor navigation, trained on occupancy maps generated from AI2THOR~\cite{kolve2017ai2} simulator, and show that it makes accurate predictions and also generalizes well on HM3D dataset~\cite{ramakrishnan2021hm3d} without fine-tuning. 

\item We study the effect of training ProxMaP under various paradigms on prediction quality and navigation tasks, highlighting the role of training methods on occupancy map prediction tasks. 
We also present some qualitative results on real data showing that ProxMaP can be extended to real-world inputs.

\item We simulate the point goal navigation as a downstream task utilizing our method for occupancy map prediction and show that our method outperforms the baseline, non-predictive approach, relatively by $12.40\%$ in navigating faster, and even outperforms a robot with multiple cameras in the general setting.
\end{enumerate}

\section{Related Works}
\label{sec:proxmap:related}
Mapping the environment is a standard step for autonomous navigation. The classical methods typically treat unobserved (i.e., occluded locations) as unknown. Our focus in this work is on learning to predict the occupancy values in these occluded areas. As shown by recent works, occupancy map prediction can help the robot navigate faster~\cite{katyal2021high} and in an efficient manner~\cite{elhafsi2020map}.

Earlier works explored machine learning techniques for online occupancy map prediction~\cite{o2012gaussian,doherty2017bayesian}, but they require updating the model online with new observations. Recent works shifted to offline training using neural networks, treating map-to-map prediction as an inpainting task. Katyal et al.\cite{katyaloccupancy} compared ResNet, UNet, and GAN for 2D occupancy map inpainting with LiDAR data, finding that UNet outperforms the others. Subsequent works used UNet for occupancy map prediction with RGBD sensors, demonstrating improved robot navigation\cite{ramakrishnan2020occupancy,georgakis2022uncertainty,wei2021occupancy}. 

Offline training for these methods requires collecting ground truth data by mapping the entire training environment, which can be time-consuming and hinder real-world deployment. Moreover, these models are trained to predict occupancy for the entire surroundings of the robot, including the scene behind for which they may lack context within the current observation, which could result in the networks memorizing room layouts, affecting their generalizability. Additionally, methods relying on historical observations for predictions~\cite{rummelhard2015conditional} face data efficiency challenges during training.

As robots can actively collect data, self-supervised methods have been successful in addressing data requirements for various robotic learning tasks~\cite{stolzle2022reconstructing, hu2021safe, khurana2022differentiable, WANG2021340, Wellhausen2019where, dhami2023prednbv}. For occupancy map prediction in indoor robot navigation, Wei et al.~\cite{wei2021occupancy} proposed a self-supervised approach using two downward-looking RGBD cameras at different heights. The network predicts the combined occupancy map from the lower camera's input without manual annotation, making it data-efficient and suitable for real robots. However, it struggles to predict edge-like obstacles and requires additional data collection for fine-tuning. Moreover, tilted cameras limit the captured information ahead compared to straight, forward-looking cameras.

To this end, we propose a self-supervised method consisting of a single, forward-looking camera to maximize the information acquisition for the navigation plane, while reducing the control effort required to collect data.  Adding two cameras to the side of the robot could further reduce this effort. We design our predictor as a classification network, which can generate sharper images compared to the regression networks, as shown later in this chapter. We focus on making predictions in the proximity of the robot, reducing the likelihood of memorization and improving its generalizability by using the current view as context. 

\section{Proposed Approach: Proximal Occupancy Map Prediction (ProxMaP)}
\label{sec:proxmap:approach}
In this work, we consider a ground robot equipped with an RGBD camera in indoor environments. Two additional views are obtained by moving the robot around as shown in Fig.~\ref{fig:proxmap:over_a}. The same can also be achieved by adding extra cameras to the robot. In the following subsections, we detail the network architecture for ProxMaP, training details, and the data collection process.

\subsection{Network Architecture and Training Details}
\label{sec:proxmap:network}
We use the occupancy map generated by $Cam_{Center}$ as input and augment it using a prediction network. Our goal is to accurately predict the occupancy information about the \textit{unknown} cells in the input map. The network uses the map generated by combining information from the three robot positions as the ground truth for training and thus learns to predict occupancy in the robot's proximity. We use  UNet~\cite{ronneberger2015u} for map prediction in ProxMaP due to its ability to perform pixel-to-pixel prediction well by sharing intermediate encodings between the encoder and decoder. We use a UNet with a 5-block encoder and a 5-block decoder. For training, we convert these maps to 3-channel images representing \textit{free}, \textit{unknown}, and \textit{occupied} regions. This is done by assigning each cell to one of the 3 classes based on its probability \textit{p}: if $p \le 0.495$, the cells are treated as \textit{free}; if $p \ge 0.505$, it is treated as \textit{occupied}; and as \textit{unknown} in rest of the cases, similar to Wei et al.~\cite{wei2021occupancy}. We train the network with cross-entropy loss, a popular choice for training classification networks.

 Since previous works have used variations of UNet for occupancy map prediction training as a regression task~\cite{wei2021occupancy, stolzle2022reconstructing} and as a generative task~\cite{katyal2019uncertainty, katyal2021high}, we also train ProxMaP with these variations. We also use UNet as the building block for these approaches with $O_c$ as input and $O^*$ as the target map (Fig.~\ref{fig:proxmap:over_b}). For the regression tasks, these maps are transformed from log odds to probability maps before training. For generative tasks, we use the UNet-based pix2pix~\cite{isola2017image} network with single and three-channel input and output pairs for regression and classification, respectively.
 
 For regression, since both input and output are probability maps, we use the KL-divergence loss function for training UNet, which simplifies to binary cross-entropy (BCE) under the assumption that each occupancy map is sampled from a multivariate Bernoulli distribution parameterized by the probability of each cell. In addition, we also train a UNet with Mean Squared Error (MSE) loss for regression. 
 For training the generative models, we use $L_{1}$ loss and $L_{GAN}$ losses as suggested by Isola et al.~\cite{isola2017image}. 
 
 In the rest of the discussion, while discussing ProxMaP's variations, we will refer to the generative classification, generative regression, and discriminative regression approaches as \textit{Class-GAN}, \textit{Reg-GAN}, and \textit{Reg-UNet}, respectively.

\subsection{Data Collection}
\label{sec:proxmap:datacollection}
We use the AI2THOR~\cite{kolve2017ai2} simulator, which provides photo-realistic scenes with depth and segmentation maps. Our setup, as shown in Fig.~\ref{fig:proxmap:over_a}, includes three RGBD cameras: \textbf{$Cam_{Center}$}, positioned at the robot's height of 0.5m and location, and two additional observations from \textbf{$Cam_{Left}$} and \textbf{$Cam_{Right}$}, located at a horizontal distance of 0.3m from the original position towards left and right, respectively. Each camera is rotated by $30\deg$ to capture extra information and increase the robot's FoV. This is done to capture extra information about the scene, while also making sure that the cameras on the sides have some overlap with \textbf{$Cam_{Center}$} The rotation of the cameras virtually increases the FoV for the robot and the translation makes sure that the robot is able to learn to look around the corners rather than simply rotating at its location.

Each camera captures depth and instance segmentation images. The depth image aids in creating a 3D re-projection of the scene into point clouds, while the segmentation image identifies the ceiling (excluded from occupancy map generation) and the floor (representing the free/navigable area). The rest of the scene is considered occupied/non-navigable. The segmentation-based processing can be replaced with height-based filtering of the ceiling and floor after re-projection. All the point clouds are reprojected to a top-down view in the robot frame using appropriate rotation and translation. Maps are then limited to a $5m\times5m$ area in front of the robot and converted to $256\times256$ images to use in the network. Points belonging to obstacles increment the corresponding cell value by 1, while floor points decrement by 1. Each bin's point count is multiplied by a factor $m=0.1$ to obtain an occupancy map with log odds. To limit log-odds values, the point count is clipped to the range $[-10,10]$. The resulting map from $Cam_{Center}$, denoted $O_c$, is the network input. The ground truth map $O^*$ is constructed as a combination of the maps from the three cameras, similar to Wei et al.~\cite{wei2021occupancy}, as follows:
\begin{equation}
  O^* = \mathrm{max}\{\mathrm {abs}(O_{c}), \mathrm {abs}(O_{l}), \mathrm {abs}(O_{r})\} \cdot \mathrm {sign}(O_{c} + O_{l} + O_{r} ),
  \label{eq:gtmap}
\end{equation}
where $O_c$, $O_l$ and $O_r$ refer to the occupancy maps generated by $Cam_{Center}$, $Cam_{Left}$, and $Cam_{Right}$, respectively. These log-odds maps are converted to probability maps before being used for network training.

AI2THOR provides different types of rooms. We use living rooms only as they have a larger size and contain more obstacles compared to others. Out of the 30 such rooms, we use the first 20 for training and validation and the rest for testing. For data collection, we divide the floor into square grids of size 0.5m and rotate the cameras by $360^\circ$ in steps of $45^\circ$. Some maps do not contain much information to predict due to the robot being close to the walls. Thus, we filter out map pairs where the number of occupied cells in $O^*$ is more than $20\%$. This process provides us with $\sim$6000 map pairs for training and $\sim$2000 pairs for testing.

\section{Experiments \& Evaluation}
\label{sec:proxmap:expresults}
We report two types of results in this section. First, we present the prediction performance of the ProxMaP and its variations on our test dataset from AI2THOR. Additionally, we show prediction results on HM3D~\cite{ramakrishnan2021hm3d} to test generalizability. Then we use these networks for indoor point-goal navigation and compare them with non-predictive methods and state-of-the-art self-supervised approach~\cite{wei2021occupancy}. Finally, we present qualitative results on some real observations to highlight the potential of real-world applications. The networks were trained on a GeForce RTX 2080 GPU, with a batch size of 4 for GANs and 16 for discriminative models. Early stopping was used to avoid overfitting with the maximum number of epochs set to 300.

\subsection{Occupancy Map Prediction}
\textbf{Setup.} As our ground truth maps are generated from a limited set of observations, they may not contain the occupancy information of all the surrounding cells. Hence, we evaluate the predictions only in cells whose ground truth occupancy is known to be either occupied or free. We refer to such cells as \textit{inpainted cells}. For classification, we choose the most likely label as the output for each cell.  For regression, a cell is considered to be \textit{free} if the probability \textit{p} in this cell is lesser than $0.495$. Similarly, a cell with $p \ge 0.505$ is considered to be \textit{occupied}. The remaining cells are treated as \textit{unknown} and are not considered in the evaluations.

Prediction accuracy is a typical metric to evaluate the prediction quality. However, it may not present a clear picture of our situation due to the data imbalance caused by fewer occupied cells. Ground robots with cameras at low heights, similar to our case, are more prone to data imbalance as the robot may only observe the edges of the obstacles. Thus we also present the precision, recall, and F1 score for each class.

\begin{figure}
  \centering
\includegraphics[width=1.00\linewidth]{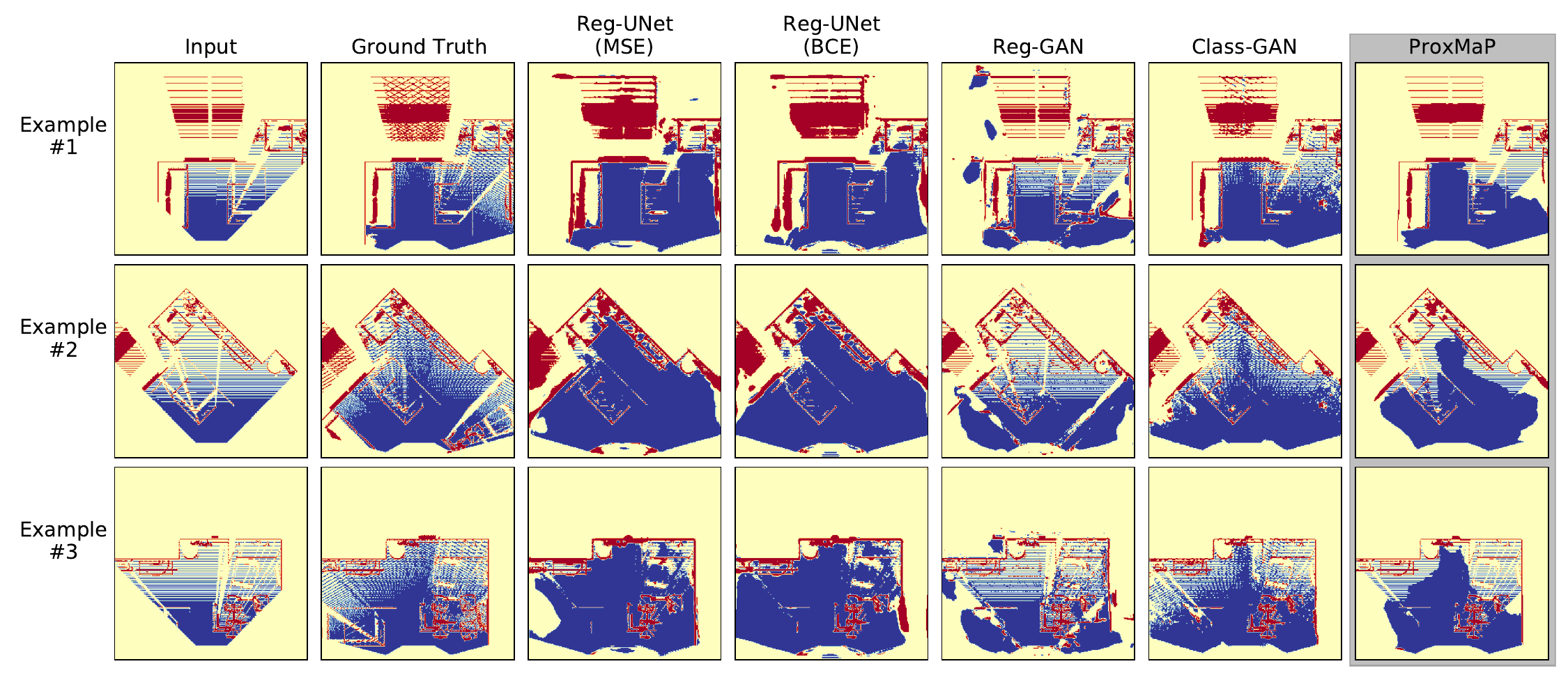}

  \caption{Results obtained by the proposed model over some examples (rows). Red, yellow, and green areas represent a \textcolor{Red}{\textbf{high}}, \textcolor{DarkYellow}{\textbf{moderate}}, and \textcolor{Blue}{\textbf{low}} chance of occupancy in an area. ProxMaP makes more accurate and precise predictions than others.}
  \label{fig:proxmap:example_plots}
\end{figure}

\textbf{Results.} Fig.~\ref{fig:proxmap:example_plots} shows the qualitative results from ProxMaP and its variants, and Table~\ref{tab:proxmap:metrics} summarizes the quantitative outcomes. The classification version of ProxMaP exhibits superior precision in predicting \textit{occupied} cells. In contrast, regression networks tend to predict surrounding areas of observed \textit{occupied} cells as \textit{occupied}, leading to higher recall but lower precision. The wider precision gap here results in a higher F1 score. The generative networks struggle due to closely mimicking patterns, including those of unknown cells. For instance, \textit{Reg-GAN} and \textit{Class-GAN} try to replicate locations of unknown cells in the ground truth, as observed in Example 3's bottom right corner. In the regression task, MSE loss outperforms BCE, which diffuses observed cells and hampers performance compared to MSE loss.

We also evaluate the generalizability of ProxMaP by testing it on similarly obtained $\sim$5000 map pairs from the HM3D dataset~\cite{ramakrishnan2021hm3d}, which contains sensor data from a realistic setting. The results are summarized in Table~\ref{tab:proxmap:metrics_habitat}. We find that discriminative regression models exhibit higher accuracy than other models. However, ProxMaP continues to lead in terms of F1-score and precision for the known classes, consistent with the results in Table~\ref{tab:proxmap:metrics}. All the metrics are lower than the results on AI2THOR due to differences in data sources, as expected.

In summary, classification networks perform better than regression networks on occupancy prediction. Generative approaches work better for classification than regression but one should be careful in using them as they tend to learn all patterns in the ground truth, including the undesirable ones. We present more qualitative results on our \href{https://raaslab.org/projects/ProxMaP/}{project webpage}.

\begin{table}\tiny
\caption{Comparison across different variations of ProxMaP over living room data from AI2THOR~\cite{kolve2017ai2} simulator. Abbreviations \textit{Reg} and \textit{Class} refer to \textit{Regression} and \textit{Classification} tasks, respectively}
\begin{center}
\resizebox{\textwidth}{!}{
\begin{tabular}{ l  c  c c c  c c c  c c c } 
% \hline
\toprule
\multirow{3}*{Method} & \multicolumn{3}{c}{F1-Score} & \multicolumn{3}{c}{Precision}  & \multicolumn{3}{c}{Recall} & \multirow{3}*{Accuracy}\\ 
\cmidrule(lr){2-4} \cmidrule(lr){5-7} \cmidrule(lr){8-10} 
{} & Free & Unknown & Occupied & Free & Unknown & Occupied & Free & Unknown & Occupied & {}\\
% \hline
\midrule
Reg-UNet (MSE) & $82.09\%$ & $90.39\%$ & $67.08\%$ & $73.91\%$ & $96.31\%$ & $60.28\%$ & $92.32\%$ & $85.15\%$ & $\textbf{75.62\%}$ & $90.94\%$ \\
Reg-UNet (BCE) & $81.37\%$ & $89.53\%$ & $65.08\%$ & $72.03\%$ & $\textbf{96.76\%}$ & $57.49\%$ & $\textbf{93.51\%}$ & $83.30\%$ & $74.99\%$ & $90.85\%$\\
Reg-GAN & $77.90\%$ & $91.14\%$ & $67.51\%$ & $81.78\%$ & $89.57\%$ & $69.19\%$ & $74.38\%$ & $92.77\%$ & $65.91\%$ & $86.92\%$ \\
Class-GAN & $82.86\%$ & $93.08\%$ & $72.22\%$ & $82.58\%$ & $92.32\%$ & $81.14\%$ & $83.14\%$ & $93.85\%$ & $65.07\%$ & $89.71\%$  \\
\textbf{ProxMaP} & $\textbf{85.43\%}$ & $\textbf{94.19\%}$ & $\textbf{76.12\%}$ & $\textbf{87.42\%}$ & $92.38\%$ & $\textbf{88.07\%}$ & $83.52\%$ & $\textbf{96.07\%}$ & $67.02\%$ & $\textbf{92.44\%}$ \\
% \hline
\bottomrule
\end{tabular}
}
\end{center}
\label{tab:proxmap:metrics}
\end{table}

\begin{table}\tiny
\caption{Generalizability of ProxMaP and variations over Habitat-Matterport3D (HM3D)~\cite{ramakrishnan2021hm3d} dataset. Abbreviations \textit{Reg} and \textit{Class} refer to \textit{Regression} and \textit{Classification} tasks, respectively}
\begin{center}
\resizebox{\textwidth}{!}{
\begin{tabular}{ l c c c c c c c c c c } 
% \hline
\toprule
\multirow{3}*{Method} & \multicolumn{3}{c}{F1-Score} & \multicolumn{3}{c}{Precision}  & \multicolumn{3}{c}{Recall} & \multirow{3}*{Accuracy}\\ 
\cmidrule(lr){2-4} \cmidrule(lr){5-7} \cmidrule(lr){8-10} 
{} & Free & Unknown & Occupied & Free & Unknown & Occupied & Free & Unknown & Occupied & {}\\
\midrule
% \midrule
Reg-UNet (MSE) & $79.99\%$ & $85.91\%$ & $56.33\%$ & $72.24\%$ & $93.52\%$ & $44.48\%$ & $89.59\%$ & $79.45\%$ & $76.79\%$ & $90.08\%$ \\
Reg-UNet (BCE) & $78.70\%$ & $84.40\%$ & $49.92\%$ & $69.70\%$ & $\textbf{94.84\%}$ & $36.96\%$ & $\textbf{90.37\%}$ & $76.02\%$ & $\textbf{76.91\%}$ & $\textbf{90.81\%}$ \\
Reg-GAN & $75.05\%$ & $88.25\%$ & $56.21\%$ & $82.38\%$ & $85.42\%$ & $51.17\%$ & $68.92\%$ & $91.28\%$ & $62.36\%$ & $83.55\%$ \\
Class-GAN & $80.27\%$ & $90.04\%$ & $70.81\%$  & $81.65\%$ & $89.04\%$ & $76.01\%$ & $78.93\%$ & $91.05\%$ & $66.28\%$ & $86.41\%$ \\
\textbf{ProxMaP} & $\textbf{81.50\%}$ & $\textbf{90.91\%}$ & $\textbf{74.11\%}$ & $\textbf{84.32\%}$ & $89.07\%$ & $\textbf{84.22\%}$ & $78.86\%$ & $\textbf{92.83\%}$ & $66.16\%$ & $87.54\%$  \\
\bottomrule
\end{tabular}
}
\end{center}
\label{tab:proxmap:metrics_habitat}
\end{table}

\subsection{Navigation Performance}
\textbf{Setup.} We use the occupancy map prediction for point-goal navigation in an unknown environment by repeated prediction and planning over the path. We also compare the discriminative and generative variation here and use only \textit{Reg-UNet(MSE)} as the discriminative regression method as it performs better than \textit{Reg-UNet(BCE)}. We compare these methods against an approach relying solely on the input map $O_c$, referred to as the \textit{baseline}, and the method proposed by Wei et al.~\cite{wei2021occupancy}, but trained with MSE loss instead of BCE as the network trained with  BCE loss (proposed by the authors) did not perform well. We obtained an F1-score of $44.01\%$ with BCE loss as compared to the F1-score of $47.61\%$ for the network trained with MSE loss. Additionally, we compare these methods against the setting where the robot is equipped with cameras on the sides, i.e., when $Cam_{Left}$ and $Cam_{Right}$ directly provide the inputs (thereby not requiring any prediction) about the scene (referred to as \textit{3-Cameras}). 

In each trial, we randomly generate a start location, a destination, and an initial yaw for the robot. At each step, the robot generates an occupancy map using the RGBD camera and maintains a global occupancy map in its memory. We utilize perceived height to filter out obstacles above the robot's height, avoiding reliance on ground truth semantic maps. We assume the robot knows its global location and yaw without error. A prediction network augments the occupancy map and the global map is updated using the same method as for ground truth generation (Eq.~\ref{eq:gtmap}). This global map is used as the cost map for path planning with Dijkstra's algorithm to find the shortest path to the destination. The robot navigates to the next prescribed waypoint, moving in 20cm steps if it is within $1^{\circ}$ of the robot's line of sight. Otherwise, the robot rotates to face the waypoint before moving. The room is discretized into a grid of square cells with sides of 0.2m. The simulation ends when the goal is within one diagonal cell at most or after the robot has moved and rotated $S_{max}$=100 times, typically sufficient to reach the goal.

For path planning, the sensed occupied and free areas are given preference over the predicted counterparts in the maps, and the costs are accordingly weighted in the cost maps. We keep the speed of the robot proportional to its confidence about the free cells on the predicted path. Thus the robot moves faster when it is confident about not colliding with an obstacle on the path. If there are possible obstacles on the path, it should keep its speed lower to be able to stop before a collision. The cost of the traversal is equal to the time taken to reach the goal.

To measure the navigation efficiency in terms of time, we use \emph{Success weighted by\\ (normalized inverse) Completion Time}~\cite{yokoyama2021success} as the metric, which is defined as \\
$SCT = \frac{1}{N} \sum_{i=1}^N S_i \frac{l_i}{max(p_i, l_i)}$,
where $N$ is the number of test episodes, $S_i$ a binary variable indicating success in the $i^{th}$ episode, $l_i$ is the traversal time on the shortest path between the source and destination, and $p_i$ is the measured traversal time by the robot in this episode. In our case, $S_i$ is 1 only when the robot is within one cell away, the number of total steps does not exceed $S_{max}$, and the simulator does not run into an error during simulation. Here, $l_i$ is found using simulation on the map generated using all the reachable positions in the environment. This map does not depend on sensing and thus is not updated during an episode. Here, $p_i$ is calculated as the time taken for traversal by the robot. A higher value of SCT is preferred, as it indicates a higher success rate and smaller difference in $p_i$ and $l_i$.

\begin{table}
\caption{SCT performance across different living rooms}
\begin{center}
\begin{tabular}{ l c} 
% \hline
\toprule
Method & SCT \\
% \hline
% \hline
\midrule
Baseline (no prediction)                 &   0.589\\
Wei et al.~\cite{wei2021occupancy} (MSE)  &   0.568\\
Reg-UNet (MSE)    &   0.629\\
Reg-GAN               &   0.592 \\
Class-GAN    &   0.632 \\
% \hline
\midrule
\textbf{ProxMaP}   &   \textbf{0.662} \\
3-Cameras (no prediction)           &   0.648\\
% \hline
\bottomrule
\end{tabular}

\end{center}
\label{tab:proxmap:nav_metrics}
\end{table}

\textbf{Results.} First, we compare the navigation efficiency by simulating navigation in different living rooms (\textit{FloorPlan221-227}). We report SCT over $N$=$102$ episodes in these rooms in Table~\ref{tab:proxmap:nav_metrics}. Most of these rooms are small and have a few situations where the robot needs to choose between multiple paths. We observe that the classification models outperform the others, reaching a maximum relative benefit of $12.39\%$ against the baseline approach. ProxMaP even outperforms the 3-cameras method by a relative margin of $2.16\%$. This happens because the prediction can also fill the gaps in the map that the 3-camera setup can not observe, resulting in a higher navigation speed over the path. The prediction method by Wei et al.~\cite{wei2021occupancy}, which relies on a higher-lower camera setup, fails to outperform the baseline in this situation. Other ProxMaP variations also outperform the baseline, highlighting the benefit of making predictions in proximity. These variations however fall short in comparison to the 3-camera setup. 

To study the navigation in a more complex situation with a higher emphasis on decision-making, we use a modified version of the living room $FloorPlan227$ 
% (Fig.~\ref{fig:proxmap:fp227_orig}) 
due to its larger size and higher number of obstacles, and therefore, more possible paths to the goals. We remove an armchair to present more path options to the robot.
We find that over $N$=$100$ episodes in this setting, ProxMaP achieves a relative improvement in SCT by $8.62\%$ over the \textit{Baseline} and $7.2\%$ improvement over Wei et al.~\cite{wei2021occupancy}. In many cases, an erroneous prediction may discourage the robot from moving on a shorter path to the goal, resulting in longer navigation time. 
A summary of these results is presented on our \href{https://raaslab.org/projects/ProxMaP}{project webpage}.
 
Overall, the classification approaches perform better in both prediction and navigation, compared to the regression counterparts. Their precise predictions and generalizability make them further suitable for occupancy map prediction.

\subsection{Predictions on Real Data}
\begin{figure}[ht!]
  \centering
\includegraphics[width=0.90\linewidth]{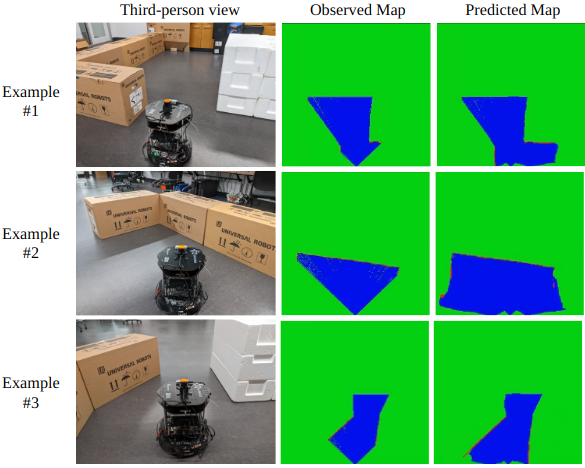}
  \caption{Prediction by ProxMaP over real-world inputs.}
  \label{fig:proxmap:real_results}
\end{figure}

 We test ProxMaP on real data using a TurtleBot2 robot 
 equipped with a Hokuyo 2D laser scanner. The FoV of the 
 scanner is limited to $90^\circ$ and its readings are used to generate the occupancy maps. We use ProxMaP over these maps to make predictions. Fig.~\ref{fig:proxmap:real_results} shows the third-person view of the robot in a maze built in our lab and the qualitative results for some interesting situations. In these maps, the \textcolor{DarkGreen}{unknown}, \textcolor{blue}{free}, and \textcolor{red}{occupied} regions are shown in green, blue, and red, respectively. Similar to our training conditions, the scanner at a low height can only see the edges of the obstacles. In these situations, we show that even when a part of the obstacle is visible, ProxMaP can estimate its shape very well, while also predicting the nearby free regions. We perform these predictions offline due to the prediction latency but will explore the real-time inference tools in future work. 

\section{Conclusion}
\label{sec:proxmap:conclusion}

We presented ProxMaP, a self-supervised method for predicting occupancy maps in the robot's proximity to aid navigation in the presence of obstacles. Our approach demonstrates higher prediction precision, generalizability to realistic and real inputs, and improved robot navigation time by adjusting robot speed based on observed and predicted information. Furthermore, ProxMaP shows the potential to outperform equivalent multi-camera setups. We also explored different variations of ProxMaP, finding that classification-based approaches yield superior predictions, both qualitatively and quantitatively, compared to regression-based methods, and also enable the robot to navigate faster by inferring about nearby regions occluded by obstacles.

The proposed method can be further extended to study the effect of different placements of the additional camera on robot navigation. The method may also benefit by leveraging other sensor and input modalities for prediction and planning. Our preliminary studies on including semantic information show promising results and will be explored in future work. 

%% file: Chapter3.tex
%Chapter 3

\renewcommand{\thechapter}{3}

\chapter{Structural and Geometric Pattern Prediction in 2D Images and Maps}
\label{chapter:fliptd}

\section{Introduction}
\cleanfootnote{Further details and results for this work are available at \url{https://raaslab.org/projects/MIM4Robots}.}
As discussed in the previous chapter, the traditional approaches for mobile robot navigation rely on the direction observations and the map built so far from them to complete the task at hand, such as PointGoal navigation, ObjectGoal navigation, and exploration~\cite{anderson2018evaluation}. 2D top-down maps or Bird's Eye Views (BEVs) are commonly used for the navigation and exploration of mobile robots through unknown areas. In the case of ground robots, occupancy maps are one such widely used representation, but depending upon the tasks and the availability of the sensors, map representations based on RGB images and semantic segmentation may also be used Aerial robots also use top-down representations for navigation and exploration or to help others when working in tandem with other aerial or ground robots~\cite{albani2017field, sharma2023d2coplan, hood2017bird, sharma2020risk}.  

Recent works across the wider robotics community have started exploring learning-based approaches to augment the robot's onboard data about the environment, e.g., occupancy map, point cloud, etc., to accomplish tasks~\cite{katyal2018occupancy, katyal2019uncertainty, saroya_online_2020, dhami2023prednbv, sharma2023proxmap}. These methods learn the patterns in representations and can predict the yet-unobserved regions based on partially observed environments. The prediction can then be used to make informed decisions for safer and more efficient motion planning.

\begin{figure}[h]
  \centering
  \includegraphics[width=0.99\linewidth]{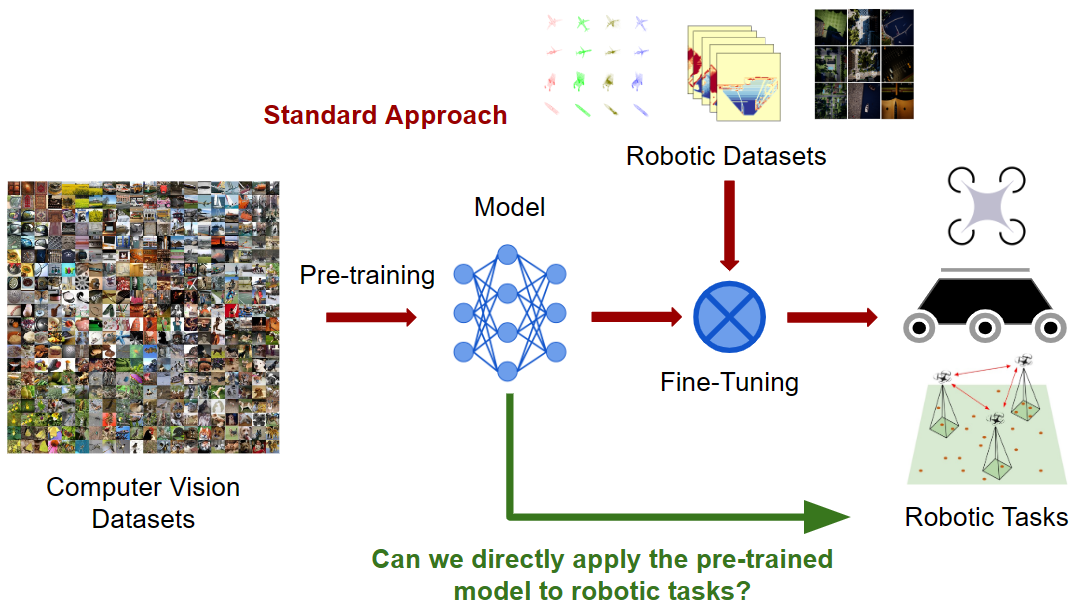}

  \caption{Traditional approach of leveraging models trained on huge computer vision datasets can be applied to robotic tasks reliant on top-down images, albeit with some task-specific fine-tuning. We show that this is not necessary and some models, such as MAE~\cite{he2022masked} can be applied directly to these robotics tasks.}
  \label{fig:fliptd:overview}
\end{figure}

Learning-based methods, simpler to ProxMaP presented in Chapter~\ref{chapter:proxmap}, require extensive datasets, which are challenging to get in robotics applications compared to computer vision. Simulators can generate virtual data, but face a sim2real gap. Many computer vision methods trained on large datasets may not directly apply to robot applications due to distributional differences in image representation; computer vision datasets are mainly comprised of first-person views, captured from a height often taller than the camera mounted on ground robots. Fine-tuning often seems to be the solution but requires similarity between pre-training and fine-tuning tasks, which can be challenging for robot navigation representations such as top-down images, semantic maps, and occupancy maps.

The recent emergence of self-supervised foundational models, which are trained on huge datasets, aims to achieve generalizability by leveraging a diverse distribution of datasets. This approach is premised on the belief that it should prompt the model to reason about fundamental concepts such as shapes and textures. However, the datasets used may not necessarily include the same distribution of images that we expect to observe during robot navigation.

This raises the question: \emph{Can we apply pre-trained computer vision models directly on robotics tasks such as navigation and exploration without fine-tuning?} Surprisingly, the answer is \ul{yes}. We substantiate this assertion with a masked image model that learns to reconstruct an image using representation learning. Specifically, Masked Autoencoder (MAE)~\cite{he2022masked}, which randomly patches the image to learn local correlation and reconstruct the masked parts. We show how, despite being trained on first-person view ~\cite{deng2009imagenet} images, it can make reasonable predictions about the unseen areas in \textit{top-down} RGB, semantic, and occupancy maps, which improves 2D planning for efficient robot navigation. We find that there is no need to fine-tune MAE on specific tasks for improvement, making it further appealing for robotics applications that may not have adequate training data.

Specifically, we make the following contributions in this chapter:
\begin{itemize}
    \item We study MAE as an expainting network for top-down images across RGB, semantic maps, and binary maps, and present quantitative and qualitative results across various degrees of increasing field-of-view for indoor and outdoor images.
    \item We present a novel uncertainty-driven exploration method for 2D semantic map reconstruction using MAE and compare it to non-predictive approaches to highlight the benefits of structural pattern prediction.
    \item We show that MAE can be effectively applied for a case study of single robot navigation aided by occupancy prediction, resulting in more efficient operation compared to a standard, non-predictive baseline method.
\end{itemize}

\begin{figure}[h]
  \centering
  \includegraphics[width=0.99\linewidth]{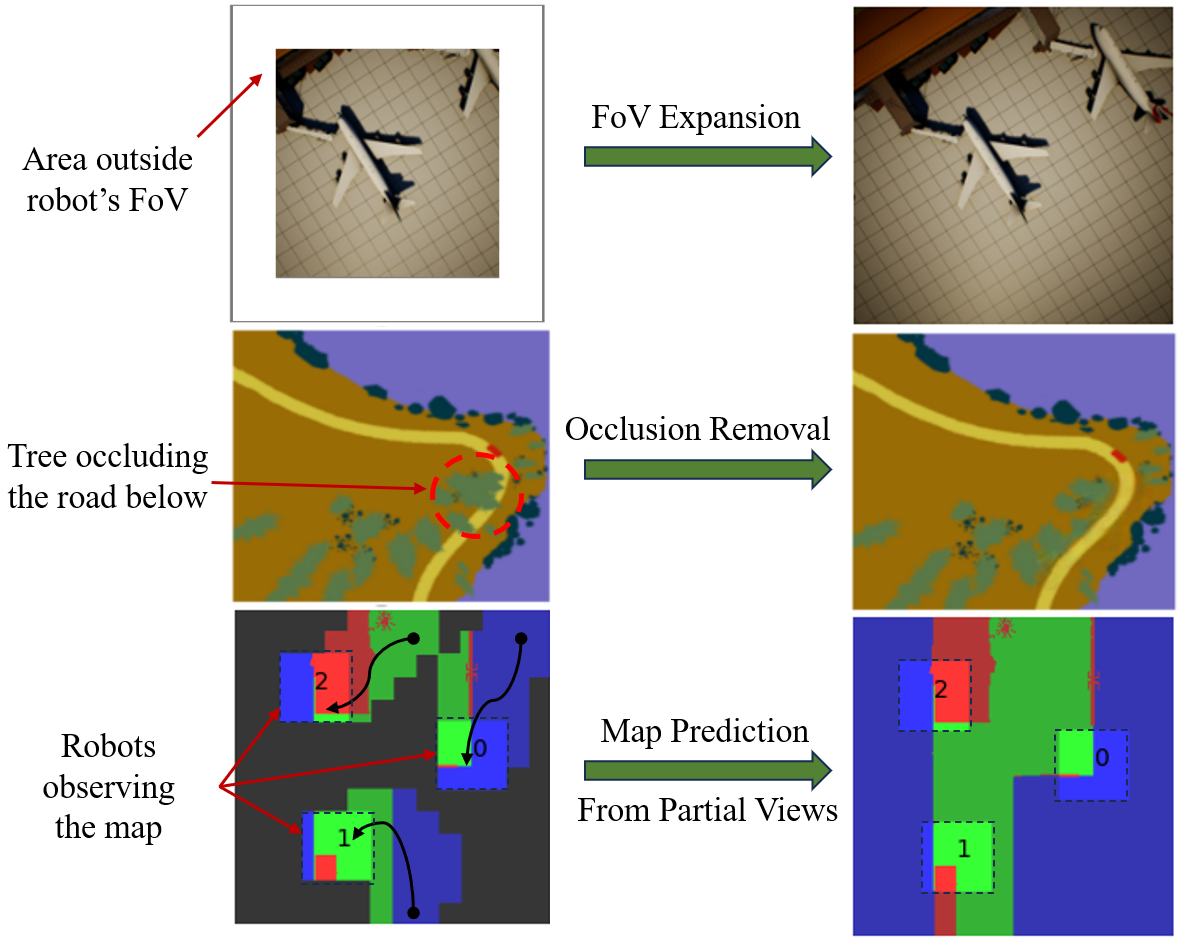}

  \caption{Example of robotics tasks solved with the help of Masked Autoencoder.}
  \label{fig:fliptd:task_examples}
\end{figure}

Our work highlights how foundational self-supervised learning algorithms like masked image model (MAE) can be used for robot tasks by choosing appropriate modalities without any fine-tuning, and paves the way for further improvement to the existing capabilities by task-specific tuning of these models. Coupled with its applicability to a variety of robotics applications, as shown in Fig.~\ref{fig:fliptd:task_examples}, MAE could potentially be the free-lunch all-around solution for 2D map-based navigation.

\section{Related Works}
\subsection{Mapping for Robot Navigation}
Top-down images and map representations are vital for robot navigation and exploration. Navigating through an unknown map by Simultaneous Mapping and Localization (SLAM), which utilizes the robot's past observations, has been a cornerstone of robotics for robotics. A top-down semantic map is another representation of interest for robotic applications. These maps are useful for semantic goal navigation~\cite{georgakis2021learning, georgakis2022cross}. Top-down images are also beneficial for aerial robot tasks such as surveying and scouting~\cite{del2021unmanned, mohamed2020unmanned}. The maps obtained by the aerial robots can be used to help the ground robots navigate. Semantic maps are obtained from such images to identify navigable and non-navigable areas for the ground robot.

Recent works in this domain have sought to improve task efficiency by \textit{predicting} the unobserved regions of the map to plan ahead. 2D Occupancy map, a top-down representation, has been the focus of many of these works, showing improvement in navigation, exploration distance, and time ~\cite{katyal2018occupancy, ramakrishnan2020occupancy, wei2021occupancy, sharma2022occupancy}. Katyal et al.~\cite {9561034} show these benefits for high-speed navigation, highlighting the importance of predictions. While the predictions are limited to the perception module, it can also enhance planning by extracting uncertainty from the predictions~\cite{katyal2019uncertainty, georgakis2022uncertainty}. The idea of uncertainty extraction also proves helpful in heterogeneous robot teams for risk-aware planning~\cite{sharma2020risk}. The key challenge with all these systems is that they need to be trained on the appropriate modalities, for which sufficient data may not be available, leading us to ask if there exist pre-trained models that can be used in these applications without much training effort, or better, without any fine-tuning at all?

\subsection{Self-supervised masked encoding}

In recent times, various approaches like BEiT~\cite{bao2021beit}, iBOT~\cite{zhou2021ibot}, and ADIOS~\cite{shi2022adversarial} have drawn inspiration from masked language models. These methods have demonstrated remarkable competitiveness in the realm of self-supervised learning (SSL). All three techniques leverage vision transformers and propose strategies to "inpaint" images that have been partially obscured by random masks in various ways. The idea of map prediction is similar to this, and existing works for robotic applications rely on generative models~\cite{koh2022simple, ren2022look, rombach2021geometry}, which require training or fine-tuning networks on simulation data to get accurate results.

Masked Autoencoder (MAE) uses Vision Transformer (ViT) encoder~\cite{dosovitskiy2020image} and is trained to use only the visible patches of an image to predict the missing patches, similar to the training strategy of BERT~\cite{devlin-etal-2019-bert}. MAE uses linear projections and position encodings for feature representation and is trained with mean squared error (MSE) between the reconstructed and original images in the pixel space, but only for masked patches. While MAE is also trained on RGB images only from the ImageNet-1K dataset~\cite{deng2009imagenet}, the underlying ViT architecture allows it to reason about other modalities as shown by MultiMAE~\cite{bachmann2022multimae}. Therefore, we use MAE for our study and show its effectiveness for prediction and inpainting across various modalities in top-down images useful for robotic tasks, without any finetuning.

 \begin{figure}[h]
  \centering
  \includegraphics[width=0.65\linewidth]{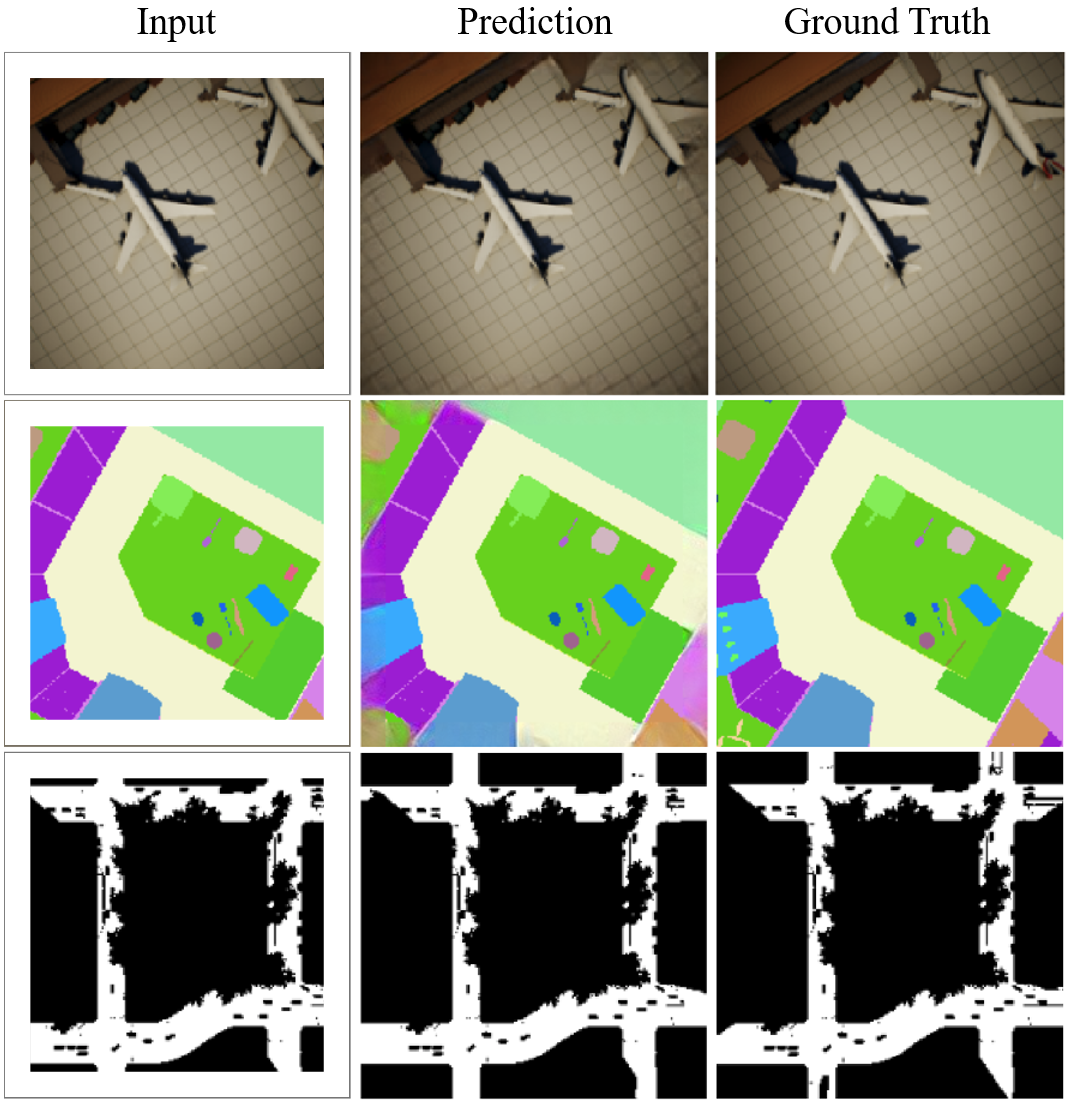}

  \caption[Examples showing that the masked autoencoder can be used to expand the effective FoV in top-down RGB, semantic, and binary images without fine-tuning.]{Masked Autoencoder can be used to expand the effective FoV in top-down RGB, semantic, and binary images without fine-tuning.}
  \label{fig:fliptd:fov_all_modal}
\end{figure}

\begin{figure}[h]
  \centering
  \includegraphics[width=0.65\linewidth]{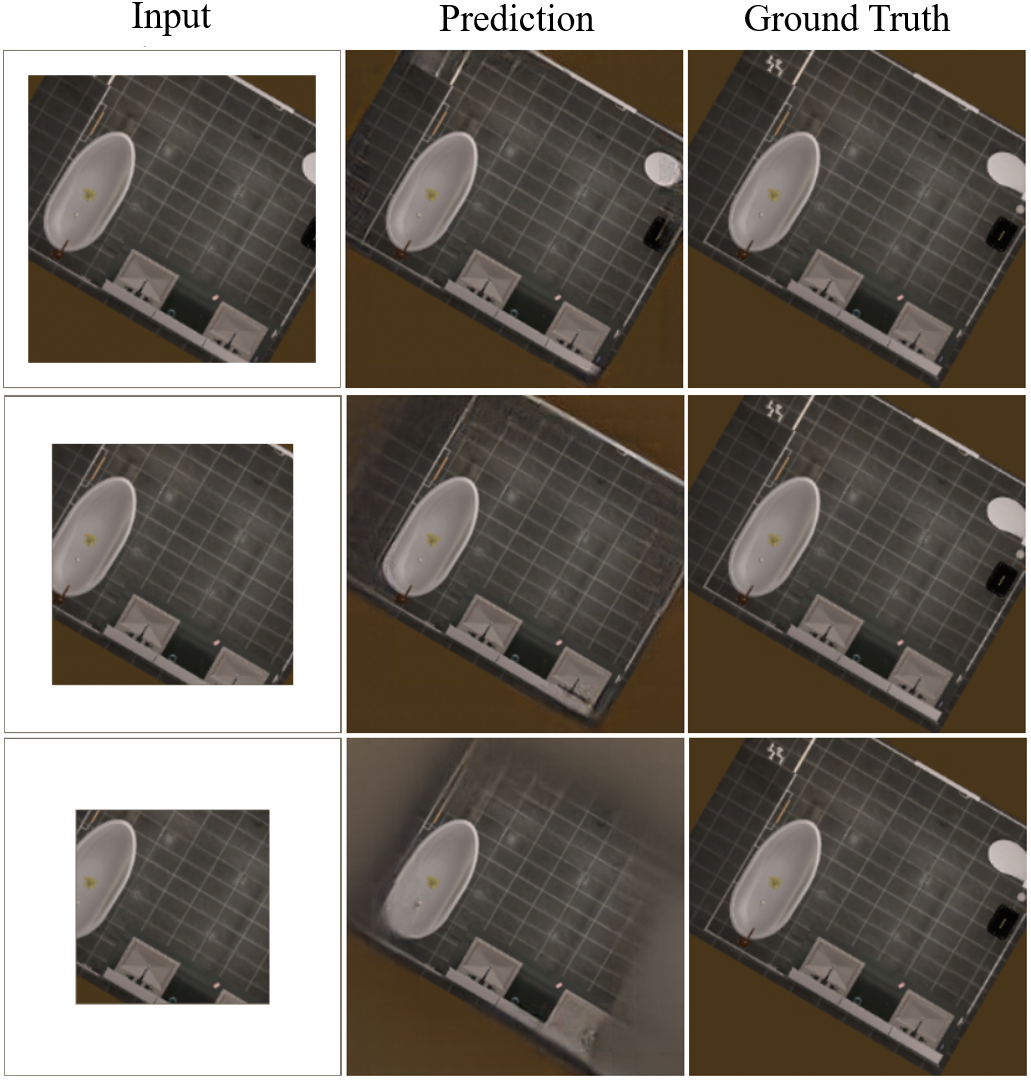}

  \caption{Results of expanding FoV for indoor images in three masking scenarios. The corner of the bathtub and room is accurately predicted based on the symmetry of the lines.  }
  \label{fig:fliptd:fov_indoors}
\end{figure}

\section{Proposed Approach: MAE as Zero-Shot Predictor}
\label{sec:fliptd:method}
We aim to determine whether the pre-trained masked autoencoder can effectively predict unobserved regions on 2D maps, represented as top-down RGB images, semantic maps, or binary maps. We focus on three tasks relevant to robot navigation and exploration, with detailed descriptions provided in the following subsections.

\subsection{FoV Expansion and Navigation}
\label{subsec:fliptd:method_fov_expansion}
Katyal et al.~\cite{katyal2018occupancy} conducted a comprehensive study on various convolutional networks to augment the effective Field of View (FoV) of the robot for predicting unexplored occupancy maps in the robot's surroundings. In their subsequent work, they demonstrated that the prediction of future occupancy maps can improve high-speed navigation ~\cite{katyal2019uncertainty}. This research employed U-Net~\cite{ronneberger2015u} as an image-to-image translation network for occupancy map prediction, finding the basis of subsequent research to further enhance robot navigation and exploration~\cite{ramakrishnan2020occupancy, wei2021occupancy, georgakis2022uncertainty, sharma2023proxmap}.

In this study, we primarily investigate the FoV expansion task, as shown in Fig.~\ref{fig:fliptd:fov_all_modal} and~\ref{fig:fliptd:fov_indoors}. Instead of employing raw occupancy maps, we opt for semantic segmentation maps and binary maps, modalities that are eventually used by conventional robotic planners. Additionally, we study RGB images, a modality consistent with the one used for MAE training and relevant to aerial mapping and surveying applications. This allows us to examine (a) whether MAE can work well on a different camera view, and (b) how other modalities, i.e., semantic and binary maps, perform during inference when compared with the one used for training MAE. Furthermore, we extend the original study by evaluating MAE performance in both indoor and outdoor environments. The inputs to MAE are provided as 3-channel images, with labels replaced by corresponding colors in semantic and binary maps. Subsequently, the colors in the output images are reconverted to labels by substituting them with the label associated with the closest color in the input images. These labels are then utilized as the assigned classes for evaluation.

\subsection{Multi-Agent Uncertainty Guided Exploration}
\label{subsec:fliptd:method_multi_agent_exploration}
Uncertainty-guided navigation and exploration, as proposed in previous works~\cite{katyal2019uncertainty, sharma2020risk, georgakis2022uncertainty} aims to enhance active robot exploration by combining the uncertainty-driven exploration technique with image inpainting networks. The eventual goal is to efficiently map the whole environment. These tasks, however, limit themselves to single-agent applications. We propose a novel approach along these lines for a multi-agent setup with pre-trained MAE, without any architectural changes such as dropout injection~\cite{gal2015dropout}.

Our approach draws on concepts from bootstrapping~\cite{efron1992bootstrap} and adversarial attacks on neural networks~\cite{szegedy2014intriguing}. By injecting minimal random noise into the input image, we obtain predictions on perturbed inputs from MAE. Despite resulting in imperceptible visual changes. We repeat this procedure multiple times to get $n$ predictions on such \textit{bootstrapped} inputs from MAE and find variance across each pixel, summed over the channels, as the uncertainty in prediction. Conceptually, pixels with high variance indicate regions where MAE lacks strong structural cues from visible input, necessitating direct observations from the robot. 

\begin{figure}[ht]
  \centering
  \includegraphics[width=0.65\linewidth]{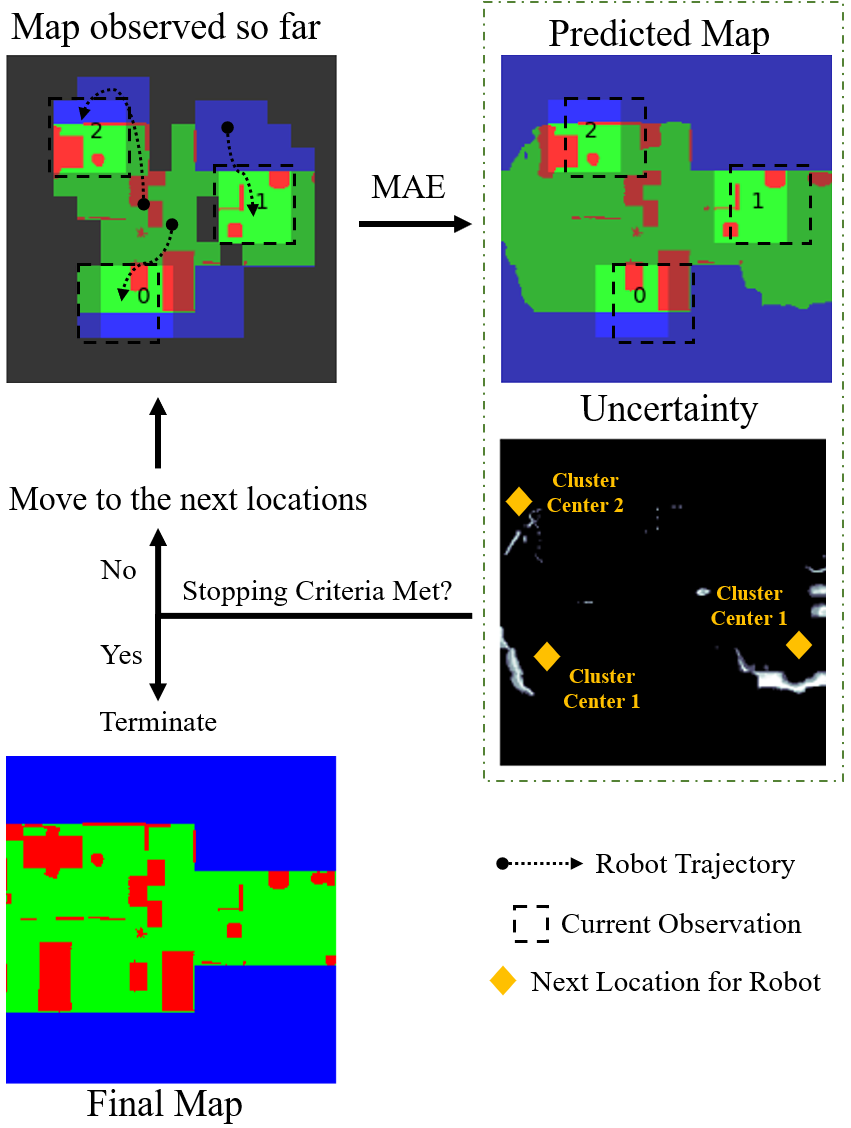}
  \caption{An overview of the MAE-based multi-agent exploration pipeline.}
  \label{fig:fliptd:multiagent_explore_overview}
\end{figure}

We put this premise to the test by looking at the prediction accuracy at each step of the exploration. To execute exploration, unexplored locations and those with high uncertainty are subsequently grouped together to identify distinct regions for potential exploration. The robots are assigned to this cluster based on their proximity to the cluster center while ensuring that no two robots are assigned to the same region. We stop the exploration when the cluster centers stabilize. Fig.~\ref{fig:fliptd:multiagent_explore_overview} shows an overview of this process. This study addresses two critical questions: (a) how to extract uncertainty from MAE, a point-prediction network, and (b) can predictions be leveraged to fill gaps in unexplored maps when resource constraints, such as battery limitations for aerial robots, hinder complete coverage?

We compared the following algorithms for this task: 
\begin{itemize}
    \item  \textbf{Boustrophedon Cell Decomposition Algorithm (Lawnmower)}: Proposed by Choset et al.~\cite{choset1998coverage}, this algorithm divides the regions into $n$ contiguous scanlines of similar size, each assigned to one robot. Each robot scans the designated area for coverage. For this method, we position the robots at the start of the respective scanlines to streamline the process.

    \item \textbf{KMeans Clustering (KMeans-U)}: Here, we apply the KMeans algorithm (with $n$ centers) to the Cartesian coordinates of the unexplored grid cells to identify the centers of the unexplored regions. The robots are then assigned to these regions based on proximity to the cluster and move towards them. At each step, the robots observe the region below and include it in the known map. Then we repeat the clustering process to find centers for the remaining unexplored areas.

    \item \textbf{KMeans Clustering followed by Reconnaissance (KMeans-R)}: Employing KMeans directly may lead to unexplored regions at the center of the map when the cluster centers stabilize (which is a stopping criterion). To address this, we introduce an additional step of relocating all the robots to the center of the map after stabilization. This results in enhanced coverage at the expense of time.

    \item \textbf{KMeans Clustering on Unknown and Uncertain Regions (KMeans-U$^2$)}: In this method, we extract uncertainty from MAE and use the locations with non-zero variance, along with those yet unexplored, for clustering. The procedure for assigning clusters to robots follows a similar approach as in the earlier methods.
\end{itemize}

Here we aim to assess the prediction capabilities of MAE and thus make predictions on the map explored so far at each step. We compare the distributions of coverage to reach $95\%$ prediction accuracy to find which algorithm is more efficient in predicting the unexplored map. 

%%%%%%%%%%%%%%%%%%%%%%%%%%%%%%%%%%%%%%%%%%%%%%%%%

\subsection{Navigation with Prediction}
\label{subsec:fliptd:method_navigation}
For autonomous navigation, it is crucial to know the map of the environment. The classical methods treat the unexplored area as unknown and build a costmap on the basis of only the current observation. 
The robot can traverse to the edge of the frontier before needing another observation to plan the path ahead. Effectively, the sensor range of the robot defines the maximum distance it can traverse at once.
 
Previous works have shown that predicting future occupancy can result in faster navigation~\cite{katyal2021high} and smoother control~\cite{elhafsi2020map}. These works train neural networks to make these predictions, using synthetically generated data and real-world data obtained by running a robot around. While the former may run into Sim2Real issues, the data collection with the latter is an arduous process. We test if pre-trained MAE could instead be used for prediction while side-stepping the data issues.

For this, we use the predictions of MAE with a standard path planning algorithm on multiple 
indoor floor plans. The robot starts from its initial position where the rest of the area in the map is hidden; using MAE we reconstruct the unseen map and update the path at every next step. The predicted unseen map acts as an estimate of the occupancy ahead, which helps to reconstruct an informed costmap for navigation.
This helps the robot cover a greater distance at once by moving to the frontier of every step. 
 
\begin{figure}[h]
  \centering
  \includegraphics[width=0.60\linewidth]{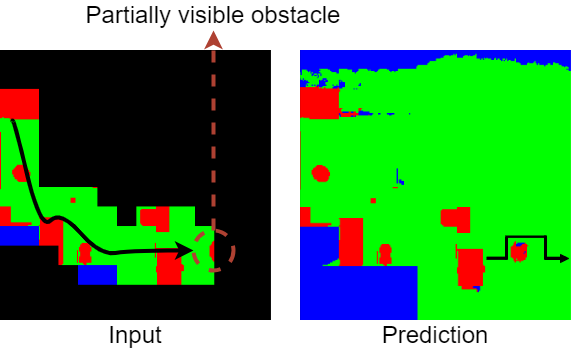}

  \caption{Left: The area robot has explored till now. Right: Prediction of obstacle (red) shape aiding robot path planning.}

  \label{fig:fliptd:fig_nav}
\end{figure}

Figure~\ref{fig:fliptd:fig_nav} shows one such example where the robot is able to plan a shorter path ahead since the predictions help to know the shape of the obstacle before the robot actually explores that area. 

We compare our prediction-based approach with a non-predictive approach and calculate the number of steps (and observations) needed to reach a pre-defined goal.
%%%%%%%%%%%%%%%%%%%%%%%%%%%%%%%%%%%%%%%%%%%%%%%%%
%%%%%%%%%%%%%%%%%%%%%%%%%%%%%%%%%%%%%%%%%%%%%%%%%%

\section{Experiments and Evaluation}
In this section, we describe the experimental setup and our findings for each task defined in Section~\ref{sec:fliptd:method}. Throughout our experiments, we utilize the MAE based on ViT-Large trained on ImageNet-1K dataset~\cite{deng2009imagenet}. 

\subsection{FoV Expansion}
\label{subsec:fliptd:eva_fov_expansion}

To study FoV expansion, we mask the periphery of the given image by different amounts. MAE uses patches of size $16\times16$ pixels, and masking a patch requires all the pixels in the patch to be masked. We mask the images with 1-3 patches on each side, resulting in an expansion of \textbf{1.17x}, \textbf{1.4x}, and \textbf{1.75x} to the robot's perceptual range, i.e., the number of pixels in a direction if the robot is at the center. 
As the data used by Katyal et al.~\cite{katyal2018occupancy} is not publicly available, we perform an evaluation on the dataset collected from two photorealistic simulated environments, consisting of diverse indoor and outdoor scenes. 

\textbf{Indoor Data:} For indoor environment, we use AI2-THOR~\cite{kolve2017ai2} which has 120 indoor scenes such as kitchens, living rooms, bathrooms, etc. We collect 1444 RGB and segmentation images with a top-down camera of a field of view of 80 degrees and rotated at intervals of 30 degrees (e.g., 30, 60, 90, etc.). 

\textbf{Outdoor Data:} For outdoor images were taken from AirSim VALID dataset ~\cite{9197186} which consist of scenes from cities, suburbs, and mountains among others, captured at different altitudes from an aerial robot. We sample 1000 images from this dataset for this study. For these environments, we also evaluate MAE on binary images, consisting of navigable and non-navigable regions, as a stand-in for occupancy maps.

We evaluate the RGB predictions for the FoV increase on the following metrics typically used to quantify visual similarity: (1) Frechet Inception Distance (FID), (2) Structural Similarity Index Measure (SSIM), (3) Peak Signal-to-Noise Ratio (PSNR), and (4) Mean Squared Error (MSE). For the semantic and binary images, we use mean Intersection-over-Union (mIoU) as the key metric but also provide the results for some of the aforementioned metrics since we use MAE to predict visually similar images for these modalities.

\textbf{Results:} Table~\ref{tab:fliptd:rgb} summarizes the results for RGB images for both types of environments. We find that increasing the FoV results in worse results than expected since MAE, an inpainting network can not reliably predict the outside areas without much context. 1.75x expansion is the extreme case where the predictions get blurry. Figure~\ref{fig:fliptd:fov_all_modal} and Figure~\ref{fig:fliptd:fov_indoors} show some examples in RGB outdoor and indoor scenes respectively and highlight this effect. 

Table~\ref{tab:fliptd:semantic} and Table~\ref{tab:fliptd:binary} summarize results for semantic and binary maps. The mIoU is very high for 1.17x expansion and goes down with increasing FoV. The effect is worse indoors as it contains many more classes (270) compared to outdoors (30) and thus may not reliably perform color-to-label matching. Also, small objects are within the scene and on the periphery, and MAE can not expand them without seeing some part of them. Note that the mIoU here is not weighted by the labels' population size. Predictions on binary maps are relatively more robust since the size of objects in each class and the difference in color mapping are larger than the semantic maps. These results present an encouraging picture for a network that was not trained on such images. We note that Katyal et al.~\cite{katyal2018occupancy} report a maximum SSIM of 0.523, 0.534, and 0.504 on real-world data for similar expansion factors. MAE results in better SSIM on both semantic segmentation and binary maps in comparison. Katyal et al.~\cite{katyal2018occupancy} report higher numbers, 0.899, 0.0818, and 0.760, on synthetic data which is similar to the distribution used for training their network. We find MAE on semantic segmentation maps still achieves higher SSIM. However, MAE with binary maps do not achieve similar performance, but still produce good results despite being trained on a different modality and camera view.

\begin{table}[h!]
  \centering
    \caption{Results for increasing the FoV in RGB images}
    \begin{tabular}{lc|cccc}
        \toprule
         \textbf{Setup} & \textbf{Expansion} & \textbf{FID $\downarrow$} & \textbf{SSIM $\uparrow$} & \textbf{PSNR $\uparrow$} & \textbf{MSE $\downarrow$}  \\ 
        \midrule
        \multirow{2}*{Indoor} & 1.17x & \textbf{17.83} & \textbf{0.94} & \textbf{27.76} & \textbf{13.76} \\
        {} & 1.40x & 41.79 & 0.86 & 22.23 & 32.42 \\
        % \addlinespace[0.1cm]
        {} & 1.75x & 76.59 & 0.78 & 19.18 & 52.98 \\
        \midrule
        \multirow{2}*{Outdoor} &  1.17x & \textbf{53.66} & \textbf{0.84} & \textbf{26.38} & \textbf{33.59} \\
        {} &  1.40x & 77.91 & 0.69 & 22.79 & 49.91 \\
        % \addlinespace[0.1cm]
        {} &  1.75x & 116.09 & 0.55 & 19.98 & 67.80 \\
        \bottomrule
    \end{tabular}
  \label{tab:fliptd:rgb}
\end{table}

\begin{table}[h]
  \centering
    \caption{Results for increasing the FoV in Semantic segmentation images}
    \begin{tabular}{lc|ccccc}
        \toprule
         \textbf{Setup} & \textbf{Expansion} & \textbf{mIoU $\uparrow$} & \textbf{FID $\downarrow$} & \textbf{SSIM $\uparrow$} & \textbf{PSNR $\uparrow$}\\ 
        \midrule
        \multirow{2}*{Indoor} &  1.17x & \textbf{0.86} & \textbf{43.48} & \textbf{0.94} & \textbf{23.06}\\
        {} & 1.40x & 0.55 & 75.42 & 0.84 & 17.33  \\
        % \addlinespace[0.1cm]
        {} & 1.75x & 0.34 & 110.01 & 0.78 & 14.90 \\
        \midrule
        \multirow{2}*{Outdoor}  & 1.17x & \textbf{0.90} & \textbf{42.63} & \textbf{0.94} & \textbf{25.96} \\
        {} & 1.40x & 0.73 & 73.03 & 0.86 & 21.39  \\
        % \addlinespace[0.1cm]
        {} & 1.75x &  0.57 & 118.56 & 0.79 & 18.80 \\
        \bottomrule
    \end{tabular}
  \label{tab:fliptd:semantic}
\end{table}

\begin{table}[h]
  \centering
  \caption{Results for increasing the FoV in  Binary images from Outdoor environment}
    \begin{tabular}{c|ccccc}
        \toprule
         \textbf{Expansion} & \textbf{mIoU $\uparrow$} & \textbf{FID $\downarrow$} & \textbf{SSIM $\uparrow$} & \textbf{PSNR $\uparrow$}\\ 
        \midrule
        1.17x & \textbf{0.90} &\textbf{ 51.87} & \textbf{0.95} & \textbf{30.36} \\
        1.40x & 0.78 & 88.44 & 0.76 & 22.05 \\
        % \addlinespace[0.1cm]
        1.75x & 0.64 & 120.94 & 0.56 & 17.81 \\
        \bottomrule
    \end{tabular}
  \label{tab:fliptd:binary}
\end{table}

\begin{figure}[h]
  \centering
  \includegraphics[width=0.65\linewidth]{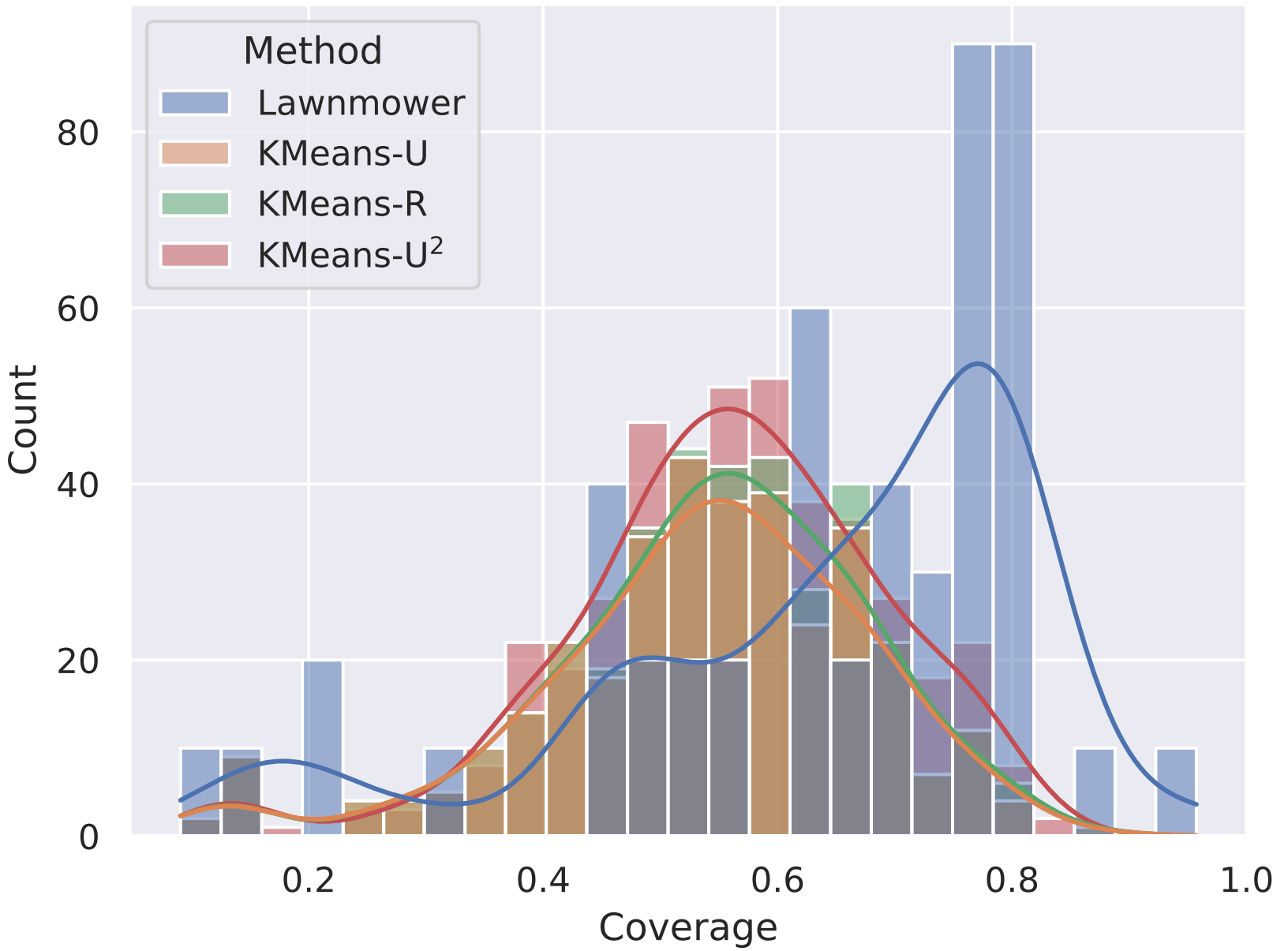}
  \caption{Comparison between the multi-agent exploration algorithms to reach at least $95\%$ accuracy in prediction of the unexplored map.}
  \label{fig:fliptd:multiagent_explore_histplot}
\end{figure}

\subsection{Multi-Agent Uncertainty Guided Exploration}
\label{subsec:fliptd:eval_multi_agent_exploration}
For this task, we use 3-channel semantic map representations, consisting of free, occupied, and out-of-boundary regions, using color-to-label matching on the MAE prediction for labeling, as described in Section~\ref{subsec:fliptd:method_fov_expansion}. In our experiments, we utilize 50 living room scenes from the ProcTHOR~\cite{deitke2022️} framework. We convert their ground truth semantic segmentation maps for the 3-class labeling. These labeled images are transformed into 3-channel RGB images, with free, occupied, and out-of-boundary regions represented by green, red, and blue colors, respectively.

We use $n=3$ aerial robots and conduct 10 experiments in each room, resulting in 500 runs total for each algorithm. We select the initial positions of the robots randomly. We assume that the area of the room to be explored is known beforehand and that the robots fly at a height taller than the obstacles and do not collide with each other. Each map is represented as an image with dimensions $224 \times 224$ pixels, and each robot can observe an area of size $48\times48$ pixels around it. We treat this as a centralized task, and the observations from all the robots are combined for decision-making.

\textbf{Results:} To compare the methods, we look at the distribution of coverage to reach at least 95\% accuracy in predicting the whole map given the partial observation. We visualize our findings in Fig.~\ref{fig:fliptd:multiagent_explore_histplot}. As shown, most runs with Lawnmower need to cover around 75-85\% of the area. This happens due to the naive movement pattern of the robots with Lawnmower and thus the robots do not benefit from the inpainting capability of MAE. All KMeans algorithms, on the other hand, are able to take advantage of it and therefore most runs with them need only 50-60\% coverage to reach the same accuracy. KMeans-U$^2$ is especially denser here as it guides the robots to areas with uncertainty, reducing the chances of incorrect predictions. We note that some heavy-tailed behavior is observed in these plots as some rooms are very simple, and a few predictions may be enough to make good predictions in them. Additionally, spawning robots at the start of the scanlines with Lawnmower places them far apart initially, an advantage other algorithms do not enjoy. This results in Lawnmower sometimes getting better accuracy with less coverage in a simple environment.

These findings highlight an intriguing observation about regions with regularly shaped objects: most shapes can be reasonably inferred by looking only at a part of them. As a result, areas with such objects may not be as beneficial for exploration after partial observation, as the large unexplored regions are. KMeans-U$^2$ performs better as it prioritizes exploring unexplored regions only when it can make a confident (low variance) estimate about objects based on the partial view. The effectiveness of this approach hinges on a prediction model's accuracy in making precise predictions, a task which our experiments have shown MAE excels at. 

\subsection{Navigation with prediction}
\label{subsec:fliptd:eva_navigation}
For this application, we use a setup similar to the previous task. Specifically, we represent the occupancy map as a 3-channel semantic map and use color-to-label matching on the MAE predictions. We select 5 large living room scenes from ProcTHOR~\cite{deitke2022️} and choose 20 start-goal pairs on them, located far apart. The map size, robot's field of view, and prediction input are similar to Section~\ref{subsec:fliptd:eval_multi_agent_exploration}. The unseen area is predicted by MAE. Using the predicted segmented map, we generate a costmap and use A$^*$ path planning algorithm for navigation. This process is repeated till the robot reaches the goal.

\textbf{Results:} 
Across twenty generated paths, our method takes on an average $10.5$ steps with a standard deviation of $2.9$ steps, whereas the traditional method takes $21.6$ steps with a standard deviation of $7.3$ steps. It is worth noting that with predictions, the larger frontier helps the robot estimate the shape of an obstacle beforehand, based on partial views, which leads to a reduction of $48\%$ in the total number resulting in efficient navigation.

\section{Conclusion}
In this work we show how MAE, a self-supervised network, pre-trained on first-person-view images can be applied to prediction-augmented robotic tasks reliant on top-down maps, without any fine-tuning. Our experiments show its applicability across various robotic tasks, involving different types of input modalities. A key takeaway from our work is that such models are capable of reasoning about regular geometric shapes and directly benefit robots in an environment filled with such patterns. We hope our analysis paves the way for further studies and the development of applications based on such powerful models. 

Our work focuses on the efficacy of a pre-trained model and is especially suitable for applications suffering from a lack of training data. We expect improvement in results with task-specific fine-tuning in future works. A drawback of MAE is that it requires the mask to be composed of square patches. While this can be attained with some innovative engineering when the required masks are irregular, other models supporting unrestricted masking might be more suited for this job. Whether they are able to retain the benefits of MAE or not will be explored in our future work.

%% file: Chapter4.tex
%Chapter 4

\renewcommand{\thechapter}{4}

\chapter{Structural and Geometrical Pattern Prediction in 3D Point Clouds}
\label{chapter:prednbv}

\section{Introduction}
\cleanfootnote{The work presented here is a result of equal contribution from Vishnu Dutt Sharma and Harnaik Singh Dhami. Further details and results for this work are available at \url{http://raaslab.org/projects/PredNBV/}.}
As shown in Chapters~\ref{chapter:proxmap} and~\ref{chapter:fliptd}, predicting the 2D maps from partial observations can help in navigating efficiently through the scene or to reconstruct it. Compared to them, 3D map representation, especially those for objects, presents new challenges in the form of increased structural and geometrical complexities.  The goal of this chapter is to improve the efficiency of mapping and reconstructing an object of interest in 3D with a mobile robot. This is a long-studied and fundamental problem in the field of robotics~\cite{bajcsy2018revisiting}. In particular, the commonly used approach is Next-Best-View (NBV) planning. In NBV planning, the robot seeks to find the best location to go to next and obtain sensory information that will aid in reconstructing the object of interest. A number of approaches for NBV planning have been proposed over the years~\cite{delmerico2018comparison}. In this chapter, we show how to leverage the recent improvements in perception due to deep learning to improve the efficiency of 3D object reconstruction with NBV planning. In particular, we present a 3D shape prediction technique that can predict a full 3D model based on the partial views of the object seen so far by the robot to find the NBV. Notably, our method works ``in the wild'' by eschewing some common assumptions made in 3D shape prediction, namely, assuming that the partial views are still centered at the full object center.

There are several applications where robots are being used for visual data collection. Some examples include inspection for visual defect identification of civil infrastructure such as bridges~\cite{shanthakumar2018view,dhamiGATSBI}, ship hulls~\cite{kim_2009_IROS} and aeroplanes~\cite{ropek_2021}, digital mapping for real estate~\cite{46965,ramakrishnan2021hm3d}, and precision agriculture~\cite{dhami2020crop}. The key reasons why robots are used in such applications are that they can reach regions that are not easily accessible to humans, and we can precisely control where the images are taken from. However, existing practices for the most part require humans to specify a nominal trajectory for the robots that will visually cover the object of interest. Our goal in this work is to automate this process.  

\begin{figure}[ht!]
    \centering
    \includegraphics[width=\linewidth]{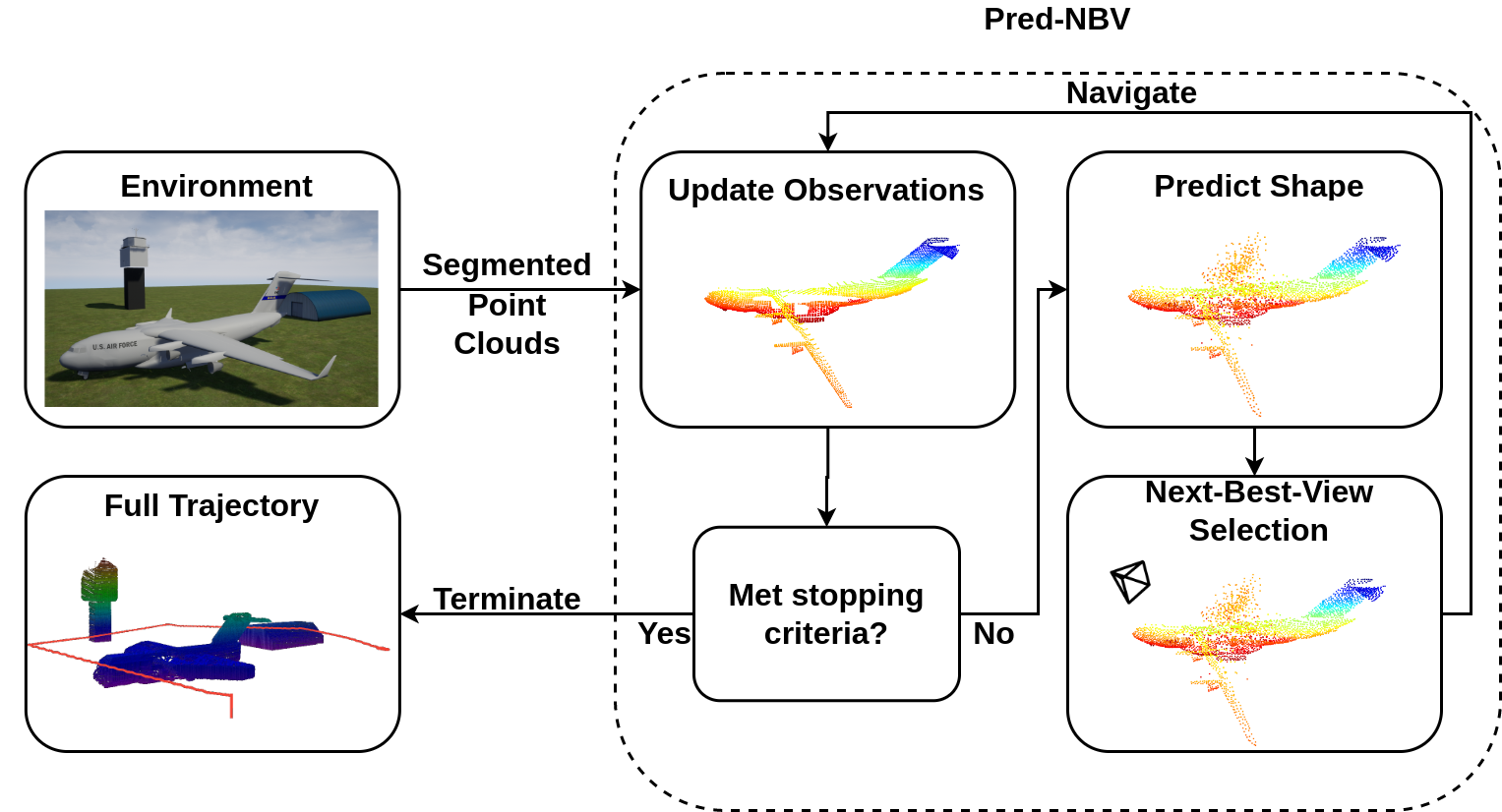}
    \caption[Overview of Prediction-based Next Best View (PredNBV).]{Overview of the proposed approach.}
    \label{fig:prednbv:overview}
\end{figure}

The NBV planning method is the commonly used approach to autonomously decide where to obtain the next measurement. NBV planning typically uses geometric cues such as symmetry~\cite{debevec1996modeling} or prior information~\cite{breyer2022closed} for deciding the next best location. In this work, we do not rely on such assumptions but instead leverage the predictive power of deep neural networks for 3D shape reconstruction.

Recent works have explored predictions as a way of improving these systems by anticipating the unknowns with prediction and guiding robots' motion accordingly. This approach has been studied for robot navigation, exploration, and manipulation~\cite{ramakrishnan2020occupancy,georgakis2022uncertainty,Wei2021,yan2019data} with the help of neural network-based methods that learn from datasets. While 2D map presentation works have shown the benefits of robotic tasks in simulation as well as the real world, similar methods for 3D prediction have been limited to simplistic simulations. The latter approaches generally rely on synthetic datasets due to the lack of realistic counterparts for learning.

Strong reliance on data results in the neural network learning implicit biases. Some models can make predictions only for specific objects~\cite{3drecgan}. Others require implicit knowledge of the object's center, despite being partially visible~\cite{yu2021pointr}. These situations are invalid in real-world, mapless scenarios and may result in inaccurate shape estimation. Many shape prediction works assume the effortless motion of the robot~\cite{wu20153d}, whereas an optimal path for a robot should include the effort required to reach a position and the potential information gain due to time and power constraints. Monolithic neural networks replacing both the perception and planning components present an alternative, but they tend to be specific to the datasets used for training and may require extensive finetuning for real-world deployment. Lack of transparency in such networks poses another challenge that could be critical for the safe operation of a robot when working alongside humans.

To make 3D predictive planning more realistic, efficient, and safe, we propose a method consisting of a 3D point cloud completion model, that relaxes the assumption about implicit knowledge of the object's center using a curriculum learning framework~\cite{bengio2009curriculum}, and an NBV framework, that maximizes the information gain from image rendering and minimizes the distance traveled by the robot. Furthermore, our approach is modular making it interpretable and easy to upgrade.

We make the following contributions to this work:
\begin{itemize}
    \item We use curriculum learning to build an improved 3D point cloud completion model, which does not require the partial point cloud to be centered at the full point cloud's center and is more robust to perturbations than earlier models. We show that this model, termed \textit{PoinTr-C}, outperforms the base model, PoinTr~\cite{yu2021pointr}, by at least $\textbf{23.06}\%$ and show qualitative comparison on ShapeNet~\cite{shapenet2015} dataset, and real point cloud obtained with a Velodyne 3D LiDAR mounted on DJI M600 Pro.
    \item We propose a next-best-view planning approach that performs object reconstruction without any prior information about the geometry, using predictions to optimize information gain and control effort over a range of objects in a model-agnostic fashion.
    \item We show that our method covers on average \textbf{25.46\%} more points on all models evaluated for object reconstruction in AirSim~\cite{airsim2017fsr} simulations compared to the non-predictive baseline approach, \textit{Basic-Next-Best-View}~\cite{aleotti2014global} and performs even better for complex structures such as airplanes.
\end{itemize}

We share the qualitative results, project code, and visualization from our method on our project website\footnote{Project webpage: \url{http://raaslab.org/projects/PredNBV/}}.

\section{Related Work}\label{sec:prednbv:rel_work}
Active reconstruction in an unknown environment can be accomplished through next-best-view (NBV) planning, which has been studied by the robotics community for a long time~\cite{scott2003view}. In this approach, the robot builds a partial model of the environment based on observations and then moves to a new location to maximize the cumulative information gained. The NBV approaches can be broadly classified into information-theoretic and geometric methods. The former builds a probabilistic occupancy map from the observations and uses the information-theoretic measure~\cite{delmerico2018comparison} to select the NBV. The latter assumes the partial information to be exact and determines the NBV based on geometric measures~\cite{tarabanis1995survey}.

The existing works on NBV with robots focus heavily on information-theoretic approaches for exploration in 2D and 3D environments~\cite{kuipers1991robot,vasquez2014volumetric}. Subsequent development for NBV with frontier and tree-based approaches was also designed for exploration by moving the robot towards unknown regions~\cite{yamauchi1997frontier, gonzalez2002navigation, adler2014autonomous, bircher2018receding}. Prior works on NBV for object reconstruction also rely heavily on information-theoretic approaches to reduce uncertainty in pre-defined closed spaces~\cite{morooka1998next, vasquez2009view}. Geometric approaches require knowing the model of the object in some form and thus have not been explored to a similar extent.
Such existing works try to infer the object geometry from a database or as an unknown closed shape~\cite{banta2000next, kriegel2013combining} and thus may be limited in application.

In recent years, prediction-based approaches have emerged as another solution. One body of these approaches works in conjunction with other exploration techniques to improve exploration efficiency by learning to predict structures in the environment from a partial observation. This is accomplished by learning the common structures in the environment (buildings and furniture, for example) from large datasets. This approach has recently gained traction and has been shown to work well for mobile robot navigation with 2D occupancy map representations~\cite{ramakrishnan2020occupancy, Wei2021, sharma2023proxmap}, exploration~\cite{georgakis2022uncertainty}, high-speed maneuvers~\cite{Katyal2021}, and elevation mapping~\cite{yang2022real}.

Similar works on 3D representations have focused mainly on prediction modules. Works along this line have proposed generating 3D models from novel views using single RGB image input~\cite{Hani2020}, depth images~\cite{Yang_2019}, normalized digital surface models (nDSM)~\cite{Alidoost2019}, point clouds~\cite{yu2021pointr, xie2020grnet, yuan2018pcn}, etc. The focus of these works is solely on inferring shapes based on huge datasets of 3D point clouds~\cite{shapenet2015}. They do not discuss the downstream task of planning. A key gap in these works is that they assume a canonical representation of the object, such as the center of the whole object, to be provided either explicitly or implicitly. Relaxing this assumption does not work well in the real world where the center of the object may not be estimated accurately, discouraging the adoption of 3D prediction models for prediction-driven planning.

Another school of work using 3D predictions combines the perception and planning modules as a neural network. These works, aimed at predicting the NBV to guide the robot from partial observations, were developed for simple objects~\cite{POP2022160}, 3D house models~\cite{Peralta2020}, and a variety of objects~\cite{zeng2020pc} ranging from remotes to rockets. Peralta et al.~\cite{Peralta2020} propose a reinforcement-learning framework, which can be difficult to implement due to sampling complexity issues. The supervised-learning approach proposed by Zeng et al.~\cite{zeng2020pc} predicts the NBV using a partial point cloud, but the candidate locations must lie on a sphere around the object, restricting the robot's planning space. Moreover, monolithic neural networks suffer from a lack of transparency and real-world deployment may require extensive fine-tuning of the hyperparameters. Prediction-based modular approaches solve these problems as the intermediate outputs are available for interpretation and the prediction model can be plugged in with the preferred planning method for a real environment.

A significant contribution of our work is to relax the implicit assumption used in many works that the center and the canonical orientation of the object under consideration are known beforehand, even if the 3D shape completion framework uses partial information as the input. A realistic inspection system may not know this information and thus the existing works may not be practically deployable.

\section{Problem Formulation}\label{sec:prednbv:prob_form}

We are given a robot with a 3D sensor onboard that explores a closed object with volume $\mathcal{V} \in \mathbb{R}^3$. The set of points on the surface of the object is denoted by $\mathcal{S} \in \mathbb{R}^3$. The robot can move in free space around the object and observe its surface. The surface of the object $s_i \subset \mathcal{S}$ perceived by the 3D sensor from the pose $\phi_i \subset \Phi$ is represented as a voxel-filtered point cloud. We define the relationship between the set of points observed from a view-point $\phi_i$ with a function $f$, i.e., $s_i = f(\phi_i)$. The robot can traverse  a trajectory $\xi$ that consists of view-points $\{\phi_1, \phi_2, \ldots, \phi_m \}$. The surface observed over a trajectory is the union of surface points observed from the consisting viewpoints, i.e. $s_{\xi} = \bigcup_{\phi \in \xi} f(\phi)$. The distance traversed by the robot between two view-points $\phi_i$ and $\phi_j$ is denoted by $d(\phi_i, \phi_j)$.

Our objective is to find a trajectory $\xi_i$ from the set of all possible trajectories $\Xi$, such that it observes the whole voxel-filtered surface of the object while minimizing the distance traversed.  
\begin{align}
    \xi^* = \argmin_{\xi \in \Xi} \sum_{i=1}^{| {\xi}| - 1} d(\phi_i, \phi_{i+1}),~ 
    \textit{such that} \bigcup_{\phi_i \in \xi} f(\phi_i) = \mathcal{S}.
\end{align}

In unseen environments, $\mathcal{S}$ is not known apriori, hence the optimal trajectory can not be determined. We assume that the robot starts with a view of the object. If not, we can always first explore the environment until the object of interest becomes visible. 

\section{Proposed Approach: Prediction-based Next-Best-View (Pred-NBV)}\label{sec:prednbv:approach}
We propose \textit{Pred-NBV}, a prediction-guided NBV method for 3D object reconstruction highlighted in Fig.~\ref{fig:prednbv:overview}. Our method consists of two key modules: (1) \textit{PoinTr-C}, a robust 3D prediction model that completes the point cloud using only partial observations, and (2) an NBV framework that uses prediction-based information gain to reduce the control effort for active object reconstruction. We provide the details in the following subsection.

\subsection{\textit{PoinTr-C}: 3D Shape Completion Network}
Given the current set of observations $v_o \in \mathcal{V}$, we predict the complete volume using a learning-based predictor $g$, i.e., $\hat{\mathcal{V}} = g(v_o)$.

To obtain $\hat{\mathcal{V}}$, we use PoinTr~\cite{yu2021pointr}, a transformer-based architecture that uses 3D point clouds as the input and output. 

PoinTr uses multiple types of machine learning methods to perform shape completion. It first identifies the geometric relationship in low resolution between points in the cloud by clustering. Then it generates features around the cluster centers, which are then fed to a transformer~\cite{vaswani2017attention} to capture the long-range relationships and predict the centers for the missing point cloud. Finally, a coarse-to-fine transformation over the predicted centers using a neural network outputs the missing point cloud. 

This model was trained on the ShapeNet~\cite{shapenet2015} dataset and outperforms the previous methods on a range of objects. However, PoinTr was trained with implicit knowledge of the center of the object. Moving the partially observed point cloud to its center results in incorrect prediction from PoinTr.

To improve predictions, we fine-tune PoinTr using a curriculum framework, which dictates training the network over easy to hard tasks by increasing the difficulty in steps during learning~\cite{bengio2009curriculum}. Specifically, we fine-tune PoinTr over increasing perturbations in rotation and translation to the canonical representation of the object to relax the assumption about implicit knowledge of the object's center. We use successive rotation-translation pairs of $(25^\circ, 0.0)$, $(25^\circ, 0.1)$, $(45^\circ, 0.1)$, $(45^\circ, 0.25)$, $(45^\circ, 0.5)$, $(90^\circ, 0.5)$, $(180^\circ, 0.5)$, and $(360^\circ, 0.5)$  for curriculum training.
We assume that the object point cloud is segmented well, which can be achieved using distance-based filters or segmentation networks.

\subsection{Next-Best View Planner}
Given the predicted point cloud $\mathcal{\hat{V}}$ for the robot after traversing the trajectory $\xi_{t}$, we generate a set of candidate poses $\mathcal{C} = \{ \phi_1, \phi_2, ..., \phi_m \}$ around the object observed so far. Given $v_o$, the observations so far, we define the objective to select the shortest path that results in observing at least $\tau \%$ of the maximum possible information gain over all the candidate poses. Considering $\mathcal{\hat{V}}$ as an exact model, we use a geometric measure to quantify the information gained from the candidate poses. Specifically, we define a projection function $I(\xi)$, over the trajectory $\xi$, which first identifies the predicted points distinct from the observed point cloud over the trajectory, then apply a hidden point removal operator on them~\cite{katz2007direct}, without reconstructing a surface or estimating a normal, and lastly, find the number of points that will be observed if we render an image on the robot's camera. Thus, we find the NBV from the candidate set $\mathcal{C}$ as follows:
\begin{align*}
    \phi_{t+1} = \argmin_{\phi \in \mathcal{C}} d(\phi, \phi_{t}),
    ~\textit{such that}~ \frac{I(\xi_t \cup \phi)}{\max_{\phi \in \mathcal{C}} I(\xi_t \cup \phi)} \geq \tau.
\end{align*}

We find the $d(\phi_i, \phi_j)$, using RRT-Connect~\cite{kuffner2000rrt}, which incrementally builds two rapidly-exploring random trees rooted at $\phi_i$ and $\phi_j$ through the observed space to provide a safe trajectory. After selecting the NBV, the robot follows this trajectory to reach the prescribed view-point. We repeat the prediction and planning process until the ratio of observations in the previous step and the current step is $0.95$ or higher.

To generate the candidate set $\mathcal{C}$, we first find the distance $d_{max}$ of the point farthest from the center of the predicted point cloud $\mathcal{\hat{V}}$ and z-range. Then, we generate candidate poses on three concentric circles: one centered at $\mathcal{\hat{V}}$ with radius $1.5 \times d_{max}$ at steps of $30^\circ$, and one $0.25 \times \text{z-range}$ above and below with radius $1.2 \times d_{max}$ at steps of $30^\circ$. We use $\tau = 0.95$ for all our experiments.

\section{Experiment and Evaluation}
\label{sec:prednbv:eval}
In this section, we evaluate the \textit{Pred-NBV} pipeline. We start with a qualitative example followed by a comparison of the individual modules against respective baseline methods. The results show that \textit{Pred-NBV} is able to outperform the baselines significantly using large-scale models from the Shapenet~\cite{shapenet2015} dataset and with real-world 3D LiDAR data.

\subsection{Qualitative Example}
\begin{figure}[ht!]
    \centering
    \begin{subfigure}[b]{.7\columnwidth}%
        \includegraphics[width = \textwidth]{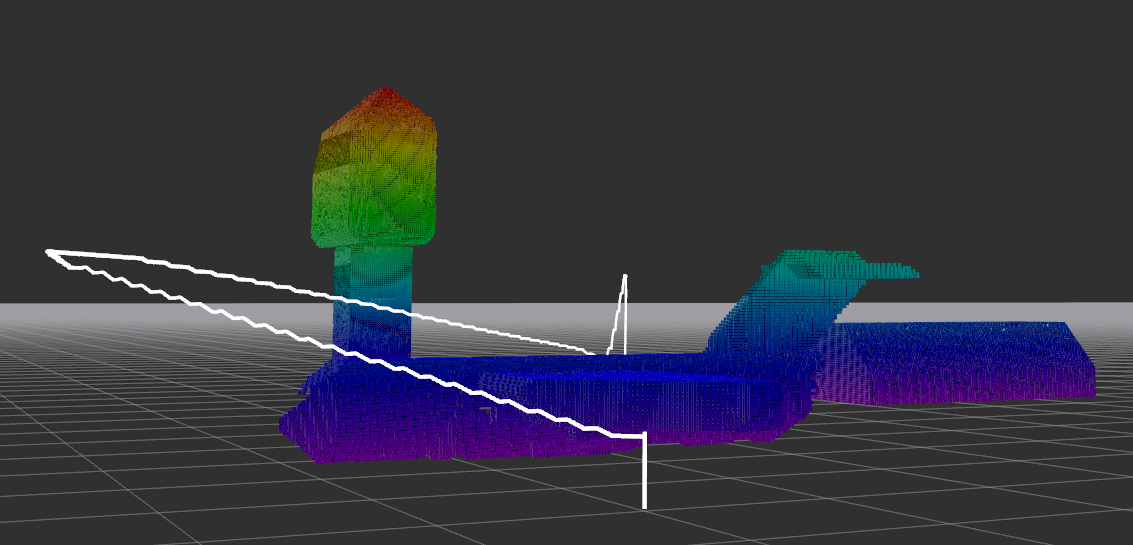}%
    \end{subfigure}%
    \hfill%
    \begin{subfigure}[b]{.5\columnwidth}%
        \includegraphics[width = \textwidth]{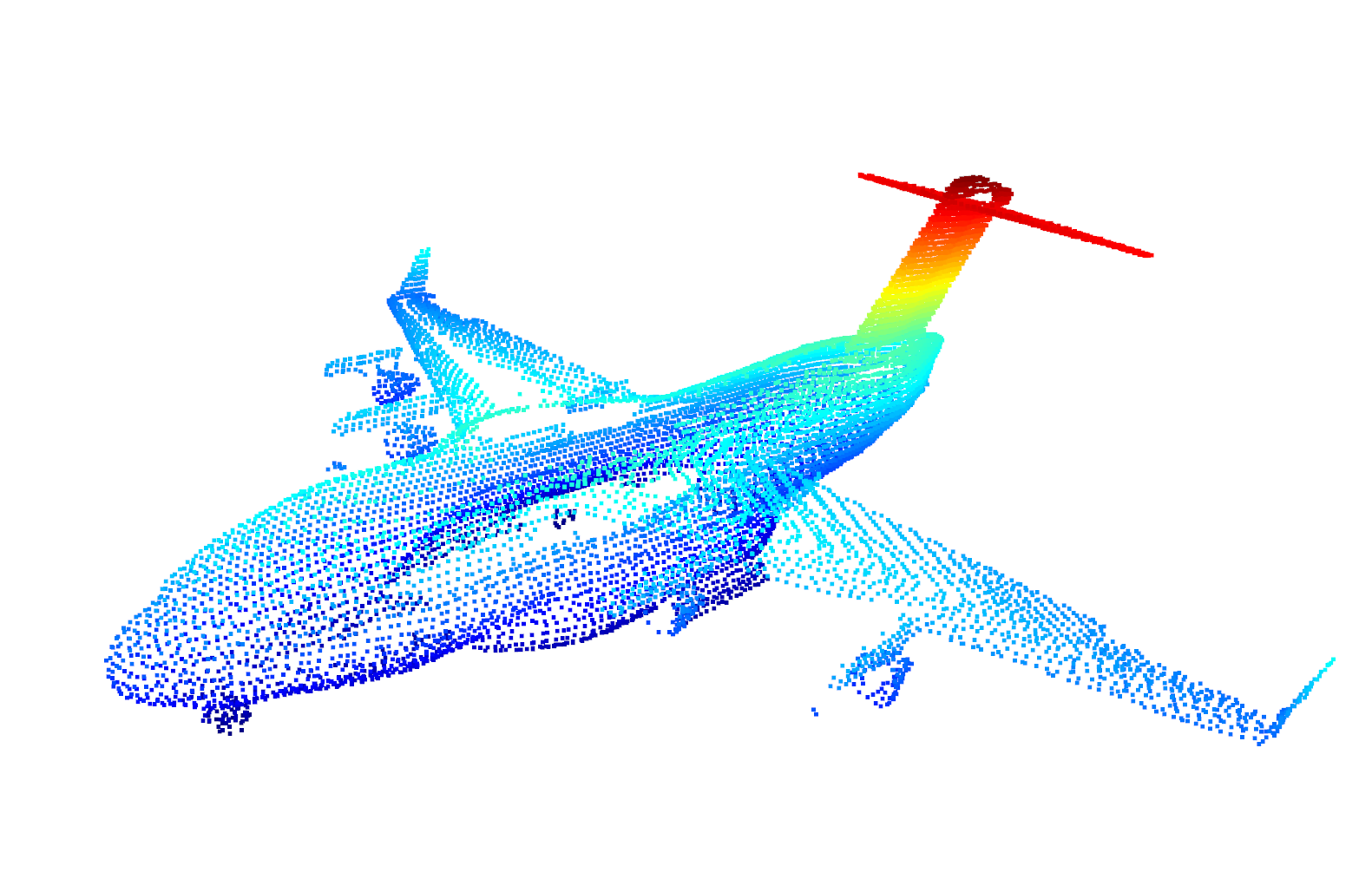}%
    \end{subfigure}%
    \caption{Flight path and total observations of C17 Airplane after running our NBV planner in AirSim simulation.}
    \label{fig:prednbv:observations_airsim}
\end{figure}

 Fig.~\ref{fig:prednbv:observations_airsim} show the reconstructed point cloud of a C17 airplane and the path followed by a UAV in AirSim~\cite{airsim2017fsr}. We create candidate poses on three concentric rings at different heights around the center of the partially observed point cloud. The candidate poses change as more of the object is visible. 

 As shown in Fig.~\ref{fig:prednbv:plane_res}, \textit{Pred-NBV} observes more points than the NBV planner without prediction in the same time budget. 

\subsection{3D Shape Prediction}
\subsubsection{Setup}

We train \textit{PoinTr-C} on a 32-core, 2.10Ghz Xeon Silver-4208 CPU and Nvidia GeForce RTX 2080Ti GPU with 11GB of memory. The network is fine-tuned over the ShapeNet\cite{shapenet2015} dataset, trained with perturbation as described in Section~\ref{sec:prednbv:approach}. Similar to PoinTr~\cite{yu2021pointr}, we use Chamfer distance (CD) and Earth Mover's Distance (EMD), permutation-invariant metrics suggested by Fan et al.~\cite{fan2017point}, as the loss function for training \textit{PoinTr-C}. For evaluation we use two versions of Chamfer distance: CD-$l_1$ and CD-$l_2$, which use L1 and L2-norm, respectively, to calculate the distance between two sets of points, and F-score which quantifies the percentage of points reconstructed correctly.

\subsubsection{Results}
Table~\ref{tab:prednbv:base_vs_best_mean} summarizes our findings regarding the effect of perturbations. \textit{PoinTr-C} outperforms the baseline in both scenarios. It only falters in CD-$l_2$ in the ideal condition, i.e., no augmentation. Furthermore, \textit{PoinTr-C} doesn't undergo large changes in the presence of augmentations, making it more robust than PoinTr. The relative improvement for \textit{PoinTr-C} is at least $23.05\%$ (F-Score).

\begin{table}[h]
    \centering
    \caption{Comparison between the baseline model (PoinTr) and \textit{PoinTr-C} over test data with and without perturbation. Arrows show if a higher ($\uparrow$) or a lower ($\downarrow$) value is better.}
    \begin{tabular}{llrrr}
        \toprule
        Perturbation  & Approach  &  F-Score $\uparrow$   &  CD-$l_1$ $\downarrow$  & CD-$l_2$ $\downarrow$\\
        \midrule
        \multirow{2}*{\xmark} & PoinTr\cite{yu2021pointr} & 0.497 & 11.621 & \textbf{0.577}\\
        {} & \textit{PoinTr-C} & \textbf{0.550} & \textbf{10.024} & 0.651\\
        \midrule
        \multirow{2}*{\cmark} & PoinTr\cite{yu2021pointr} & 0.436 & 16.464 & 1.717\\
        {} & \textit{PoinTr-C} & \textbf{0.550} & \textbf{10.236} & \textbf{0.717}\\
        \bottomrule
        \end{tabular}
        \label{tab:prednbv:base_vs_best_mean}
\end{table}

\begin{figure}
    \centering
    \begin{subfigure}[b]{0.25\columnwidth}%
        \includegraphics[width=\linewidth]{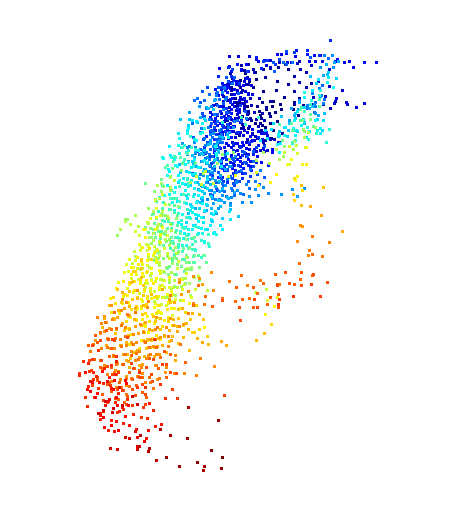}
        \subcaption{Input}
    \end{subfigure}%
    \hfill%
    \begin{subfigure}[b]{0.25\columnwidth}%
        \includegraphics[width=\linewidth]{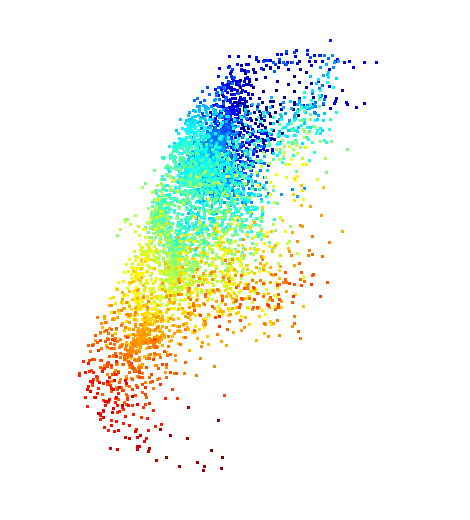}
        \subcaption{PoinTr~\cite{yu2021pointr}}
    \end{subfigure}%
    \hfill%
    \begin{subfigure}[b]{0.25\columnwidth}%
        \includegraphics[width=\linewidth]{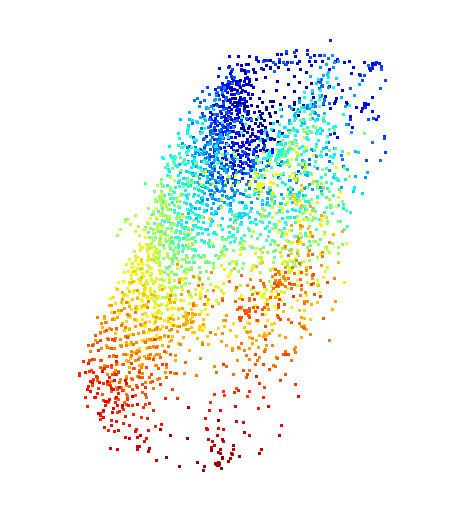}
        \subcaption{PoinTr-C}
    \end{subfigure}%
    \caption{Results over the real-world point cloud of a car obtained using LiDAR (Interactive figure available on \href{http://raaslab.org/projects/PredNBV/}{our webpage}).}
    \label{fig:prednbv:qualitative_real_colored}
\end{figure}

We provide a qualitative comparison of the predictions from the two models for various objects under perturbations on our \href{http://raaslab.org/projects/prednbv/}{project webpage}. Fig.~\ref{fig:prednbv:qualitative_real_colored} shows the results for a real point cloud of a car (visualized from the camera above the left headlight) obtained with a Velodyne LiDAR sensor. The results show the predictions from PoinTr are scattered around the center of the visible point cloud, whereas \textit{PoinTr-C} makes more realistic predictions. Our webpage provides interactive visualizations of these points 
clouds for further inspection.

\subsection{Next-Best-View Planning}
\subsubsection{Setup}

We use Robot Operating System (ROS) Melodic and AirSim~\cite{airsim2017fsr} on Ubuntu 18.04 to carry out the simulations. We equipped the virtual UAV with a RGBD camera. AirSim's built-in image segmentation is used to segment out the target object from the rest of the environment. We created a ROS package to publish a segmented depth image containing pixels only belonging to the bridge based on the RGB camera segmentation. This segmented depth image was then converted to a point cloud. We use the MoveIt~\cite{coleman2014reducing} software package based on the work done by Köse~\cite{tahsinko86:online} to implement the RRT connect algorithm. MoveIt uses RRT connect and the environmental 3D occupancy grid to find collision-free paths for point-to-point navigation.

\subsubsection{Qualitative Example}

We evaluate \textit{Pred-NBV} on 20 objects from 5 ShapeNet classes:  airplane, rocket, tower, train, and watercraft. We selected these classes as they represent larger shapes suitable for inspection. Fig.~\ref{fig:prednbv:overview} shows the path followed by the UAV using \textit{Pred-NBV} for the C-17 airplane simulation. There are non-target obstacles in the environment, such as a hangar and air traffic control tower. \textit{Pred-NBV} finds a collision-free path that selects viewpoints targeting the maximum coverage of the airplane.

\subsubsection{Comparison with Baseline}\label{sec:prednbv:sim:baseline}
We compare the performance of \textit{Pred-NBV} with a baseline NBV method~\cite{aleotti2014global}. The baseline selects poses based on frontiers in the observed space using occupancy grids. We modified the baseline to improve it for our application and make it comparable to \textit{Pred-NBV}. The modifications include using our segmentation for the occupancy grid so that frontiers are weighted toward the target object. We also set the orientation of the selected poses towards the center of the target object similar to how \textit{Pred-NBV} works. The algorithms had the same stopping criteria as \textit{Pred-NBV}.

\begin{figure}
    \centering
    \begin{subfigure}[b]{\columnwidth}%
        \hspace{0.8cm}
        \includegraphics[width = .75\textwidth]{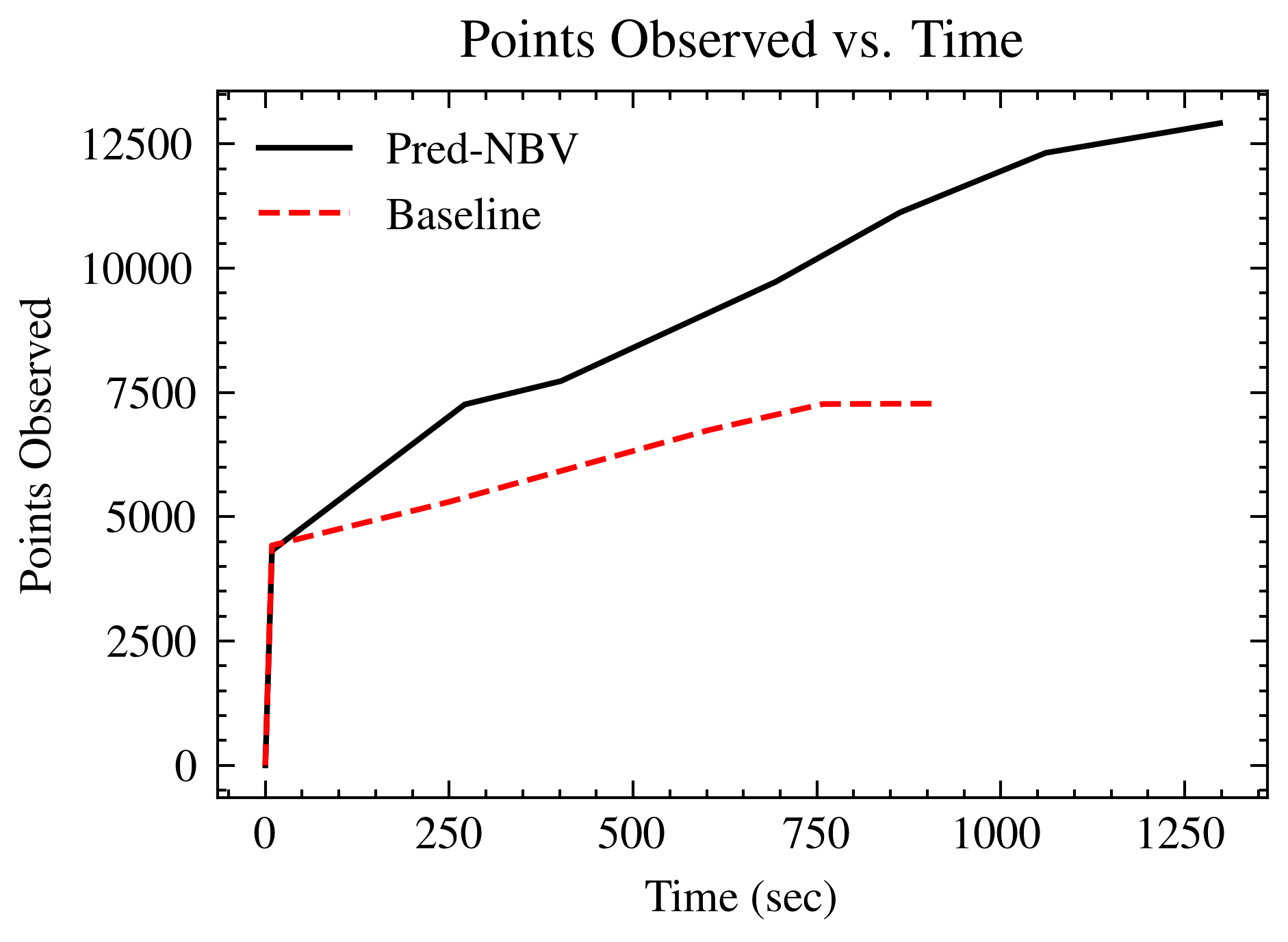}%
    \end{subfigure}%
    \hfill%
    \begin{subfigure}[b]{\columnwidth}%
        \hspace{1.0cm}
        \includegraphics[width = .725\textwidth]{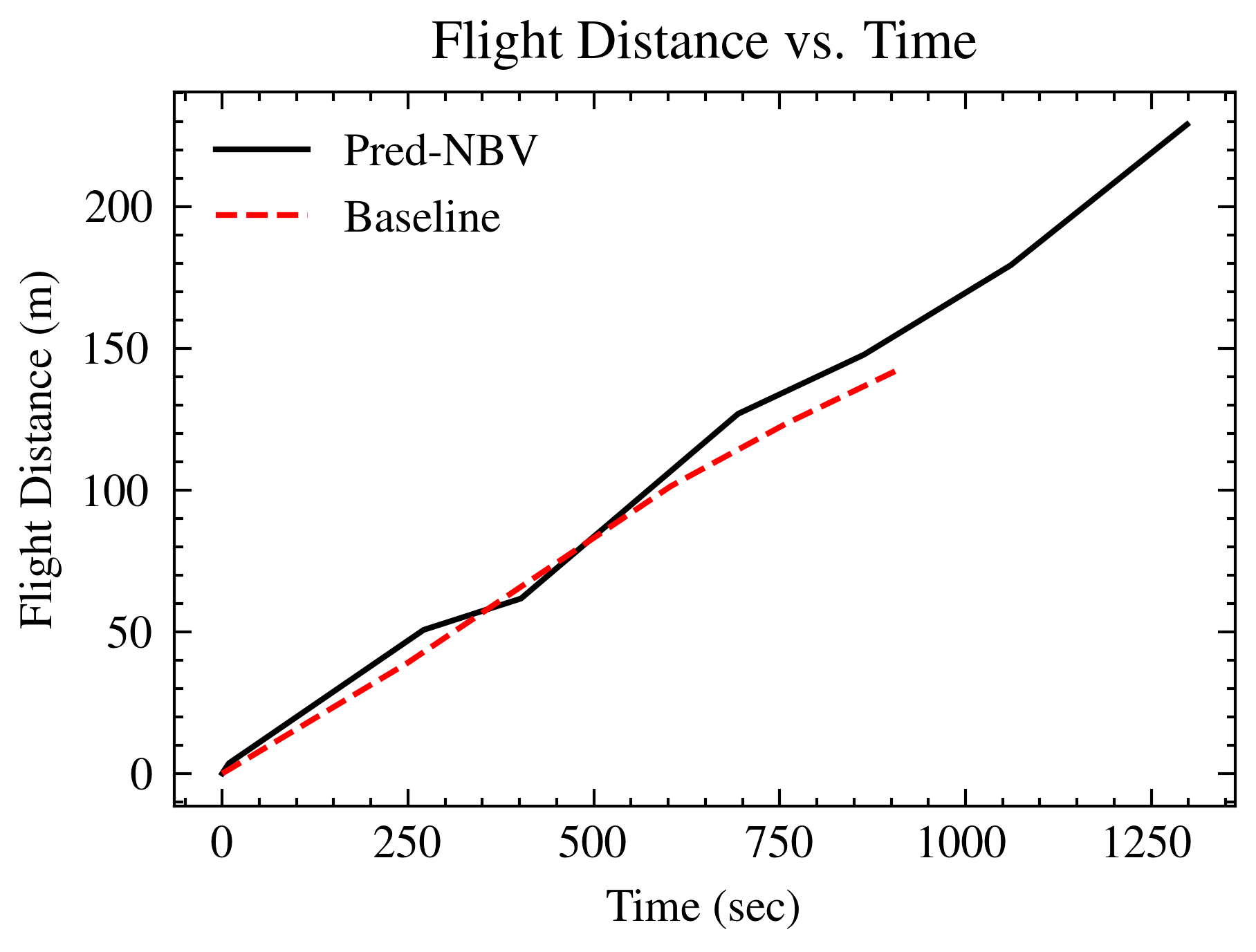}%
    \end{subfigure}%
    \caption{Comparison between \textcolor{black}{\textbf{Pred-NBV}} and the \textcolor{red}{\textbf{baseline NBV algorithm}}~\cite{aleotti2014global} for a C-17 airplane.}
    \label{fig:prednbv:plane_res}
\end{figure}

We see in Table~\ref{tab:prednbv:airsim_results} that our method observes on average 25.46\% more points than the baseline for object reconstruction across multiple models from various classes. In Fig.~\ref{fig:prednbv:plane_res}, we show that \textit{Pred-NBV} observes more points per step than the baseline while not flying further per each step. 

\begin{table}[ht!]
    \centering
    \caption{Points observed by \textit{Pred-NBV} and the baseline NBV method~\cite{aleotti2014global} for all models in AirSim.}
    \begin{tabular}{llrrr}
    
        \toprule
        \multirow{2}{*}{ Class} & \multirow{2}{*}{Model} & Points Seen & Points Seen & \multirow{2}{*}{Improvement} \\
        & & \textit{Pred-NBV} & Baseline & \\
        \midrule
        \multirow{5}{*}{Airplane}  
            & 747 & \textbf{11922} & 9657 & 20.99\% \\
            & A340 & \textbf{8603} & 5238 &  48.62\% \\
            & C-17 & \textbf{12916} & 7277 & 55.85\% \\
            & C-130 & \textbf{9900} & 7929 & 22.11\% \\
            & Fokker 100 & \textbf{10192} & 9100 & 11.32\%\\
        \midrule
        \multirow{5}{*}{Rocket} 
            & Atlas & \textbf{1822} & 1722 & 5.64\% \\
            & Maverick & \textbf{2873} & 2643 & 8.34\% \\
            & Saturn V & \textbf{1111} & 807 & 31.70\% \\
            & Sparrow & \textbf{1785} & 1639 & 8.53\% \\
            & V2 & \textbf{1264} & 1086 & 15.15\% \\
        \midrule
        \multirow{5}{*}{Tower} 
            & Big Ben & \textbf{4119} & 3340 & 20.89\% \\
            & Church & \textbf{2965} & 2588 & 13.58\% \\
            & Clock & \textbf{2660} & 1947 & 30.95\% \\
            & Pylon & \textbf{3181} & 2479 & 24.80\% \\
            & Silo & \textbf{5674} & 3459 & 48.51\% \\
        \midrule
        \multirow{2}{*}{Train}
            & Diesel & \textbf{3421} & 3161 & 7.90\% \\
            & Mountain & \textbf{4545} & 4222 & 7.37\% \\
        \midrule
        \multirow{3}{*}{Watercraft} 
            & Cruise & \textbf{4733} & 3522 & 29.34\% \\
            & Patrol & \textbf{3957} & 2306 & 52.72\% \\
            & Yacht & \textbf{9499} & 6016 & 44.90\% \\
        \bottomrule
    \end{tabular}
    \label{tab:prednbv:airsim_results}
\end{table}

\section{Conclusion}\label{sec:prednbv:con}
We propose a realistic and efficient planning approach for robotic inspection using learning-based predictions. Our approach fills the gap between the existing works and the realistic setting by proposing a curriculum-learning-based point cloud prediction model, and a distance and information gain aware inspection planner for efficient operation. Our analysis shows that our approach is able to outperform the baseline approach in observing the object surface by 25.46\%. Furthermore, we show that our predictive model is able to provide satisfactory results for real-world point cloud data. We believe the modular design paves the path to further improvement by enhancement of the constituents. 

In this work, we use noise-free observations but show that \textit{Pred-NBV} has the potential to work well on real, noisy inputs with pre-processing. In future work, we will explore making the prediction network robust to noisy inputs and with implicit filtering capabilities. We used a geometric measure for NBV in this work and will extend it to information-theoretic measures using an ensemble of predictions and uncertainty extraction techniques in future work. 

%% file: Chapter5.tex
%Chapter 5

\renewcommand{\thechapter}{5}

\chapter{Structural and Geometric Pattern Prediction with Planning for Multi-Robot Systems}
\label{chapter:mapnbv}

\section{Introduction}
\cleanfootnote{The work presented here is a result of equal contribution from Vishnu Dutt Sharma and Harnaik Singh Dhami. Further details and results for this work are available at \url{http://raaslab.org/projects/MAPNBV/}.}

\begin{figure}[ht!]
\centering
\begin{subfigure}{.47\columnwidth}
    \includegraphics[width=1.0\linewidth]{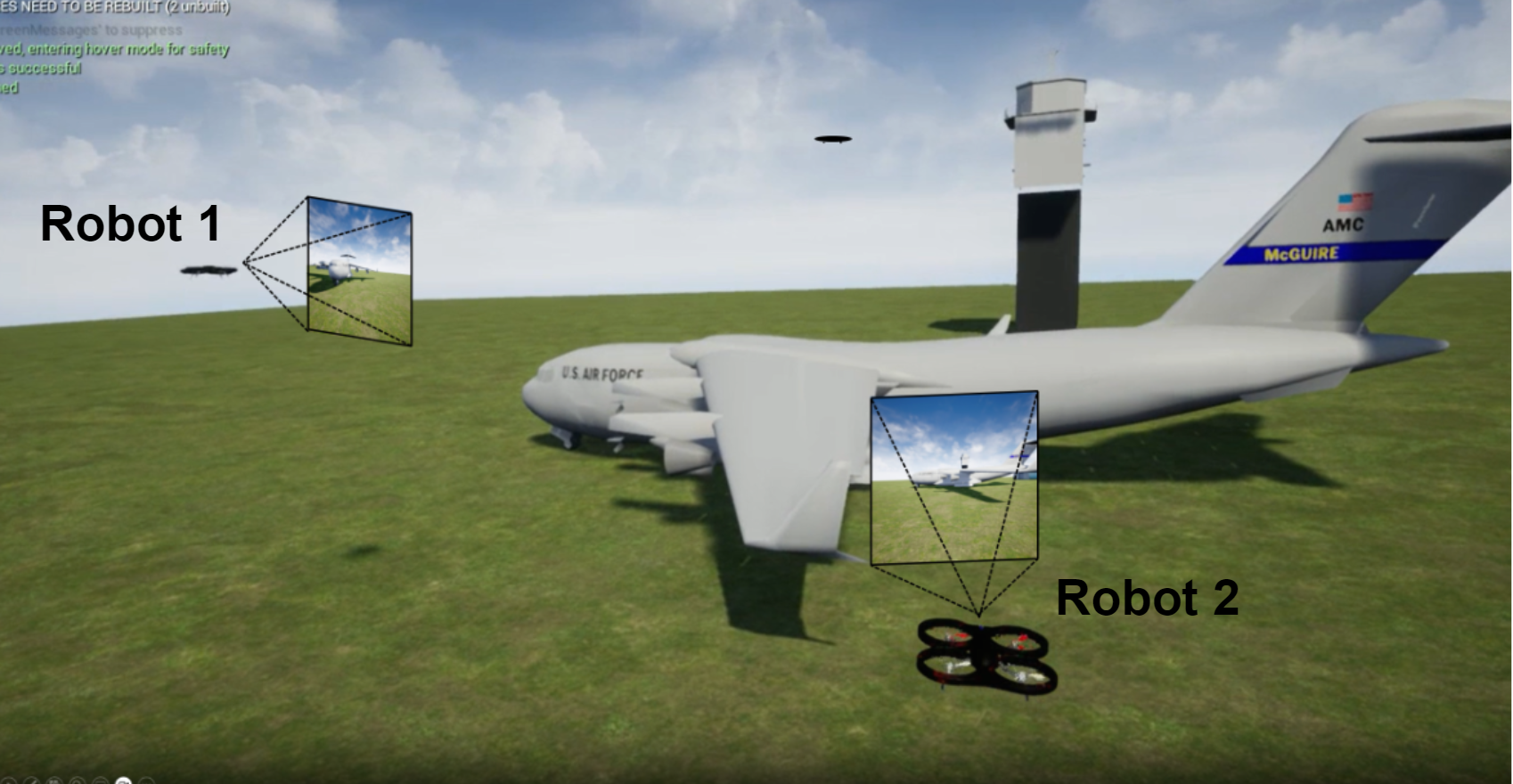}
        \caption{C-17 and the robots in AirSim simulation}
        \label{fig:mapnbv:airsimScreengrab}
\end{subfigure}%
\hfill
\begin{subfigure}{.47\columnwidth}
        \includegraphics[width=0.7\linewidth]{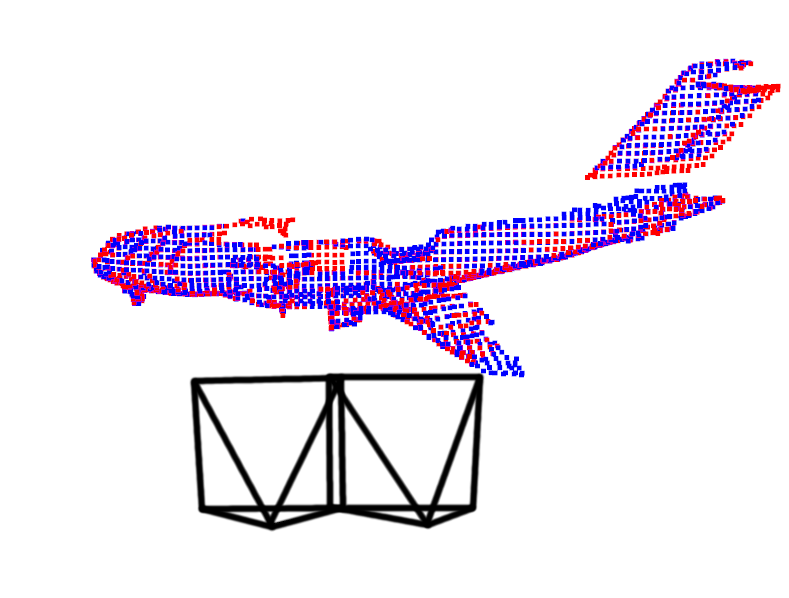}
        \caption{Initial observations by \textcolor{red}{robot 1} and \textcolor{blue}{robot 2}}
        \label{fig:mapnbv:c17Observed}
\end{subfigure}
\medskip
\begin{subfigure}{.47\columnwidth}
        \includegraphics[width=1.0\linewidth]{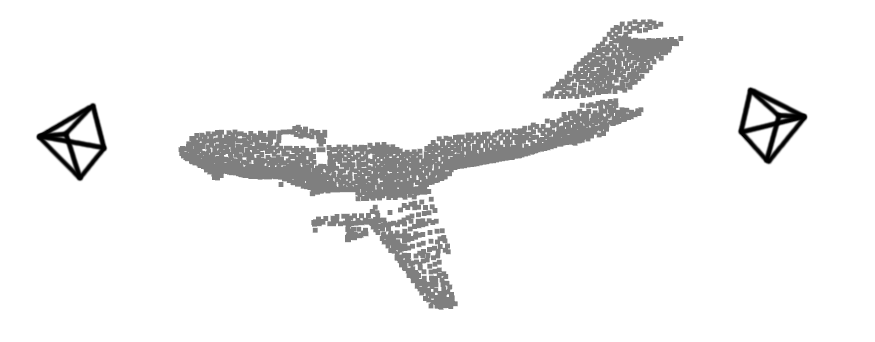}
        \caption{Poses selected by frontiers baseline based on \textcolor{gray}{observations}}
        \label{fig:mapnbv:c17BaselineNBV}
\end{subfigure}%
\hfill
\begin{subfigure}{.47\columnwidth}
        \includegraphics[width=0.97\linewidth]{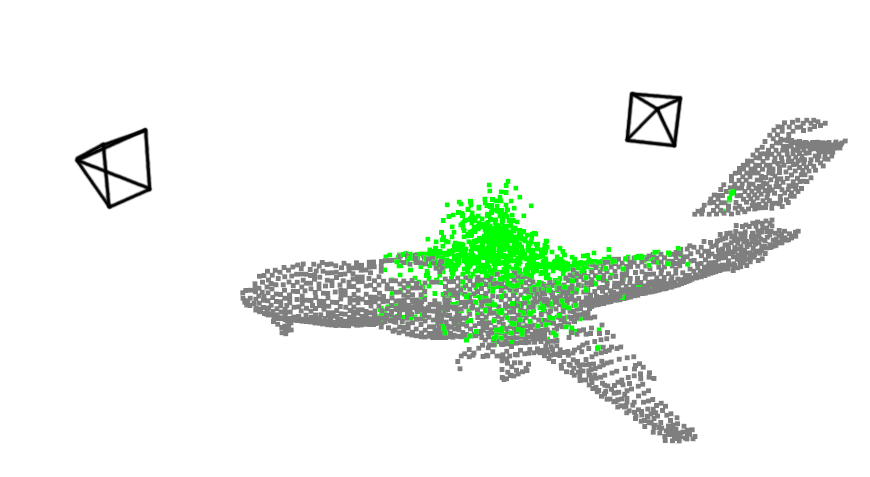}
        \caption{Poses selected by \textit{MAP-NBV} based on \textcolor{gray}{observations}}
        \label{fig:mapnbv:c17MAPNBV_poses}
\end{subfigure}

\medskip
\begin{subfigure}{.47\columnwidth}
        \includegraphics[width=1.0\linewidth]{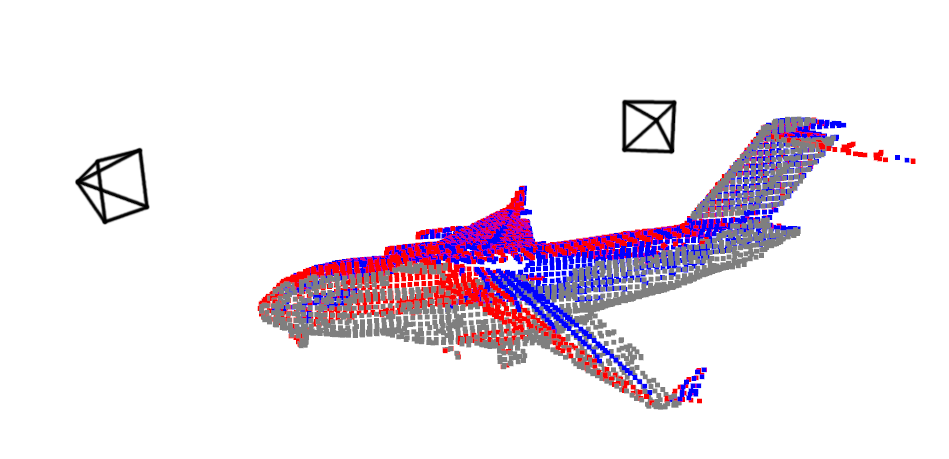}
        \caption{Observations after the first \textit{MAP-NBV} iteration}
        \label{fig:mapnbv:c17BaselineNBV_obs}
\end{subfigure}%
\hfill
\begin{subfigure}{.47\columnwidth}
        \includegraphics[width=0.97\linewidth]{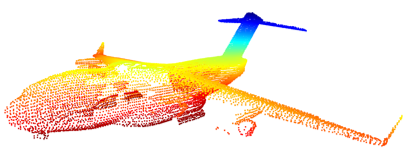}
        \caption{Full point cloud observed by \textit{MAP-NBV} after termination}
        \label{fig:mapnbv:c17MAPNBV}
\end{subfigure}

\caption{MAP-NBV uses \textcolor{green}{predictions} to select better NBVs for a team of robots compared to the non-predictive baseline approach.}
\label{fig:mapnbv:c17predicted}
\label{fig:mapnbv:pointsObserved}
\end{figure}

As shown in the previous chapter, using 3D shape prediction based Prediction Pred-NBV~\cite{dhami2023prednbv}, estimating the unseen parts of the objects with point cloud completion networks can improve NBV planning, and hence the reconstruction efficiency, for a single UAV. These findings lead us to ask: \textbf{Can prediction improve the efficiency of multi-agent object reconstruction, given that multiple robots can themselves provide good coverage of the object?}

Employing a team of UAVs for this problem is an intuitive solution as multiple UAVs can simultaneously cover multiple viewpoints. NBV planning can be used to move the UAVs, gather new observations, and fill the gaps in the object representation. An efficient reconstruction, however, requires not only a correct estimation of the missing information but also coordination among the UAVs to minimize redundancies in the observations. A naive extension of methods designed for a single agent may result in significant overlaps in the observations by the team, necessitating coordination among all the robots. Prior works have shown that for target coverage problems (which object reconstruction is) \textit{explicit coordination} plays an important role in developing an efficient solution~\cite{corah2019communication}. However, this prior work only focused on scenarios where the coordination used past observation and not predictions. This begs the question: \textbf{How does coordination, in perception and planning, affect multi-agent object reconstruction when each robot has access to predictions?}

To answer these questions, we make the following contributions in this work:
\begin{enumerate}
    \item We propose a decentralized, multi-agent, prediction-based NBV planning approach, named\textit{ MAP-NBV}, for active 3D reconstruction of various objects with a novel objective combining visual information gain and control effort.\looseness=-1
    
    \textit{MAP-NBV} uses partial point clouds and predicts what the rest of the point cloud would be (Figure~\ref{fig:mapnbv:c17predicted}) and exploits the submodular nature of the objective to coordinate in a decentralized fashion.
    \item We show that predictions effectively improve the performance by \textbf{19.41\%} over non-predictive baselines that use frontier-based NBV planning~\cite{aleotti2014global} in AirSim~\cite{shah2018airsim} simulations.
    \item We also show that \textit{MAP-NBV} results in at least \textbf{17.12\%} better reconstruction than non-cooperative prediction-guided method with experiments using the ShapeNet~\cite{shapenet2015} dataset and performs comparable to a centralized approach. 
    
\end{enumerate}
We share the qualitative results and release the project code from our method on our project website.\footnote{\url{http://raaslab.org/projects/MAPNBV/}}

\begin{figure*}[ht!]{}
    \centering
    \includegraphics[width = 1.0\columnwidth]{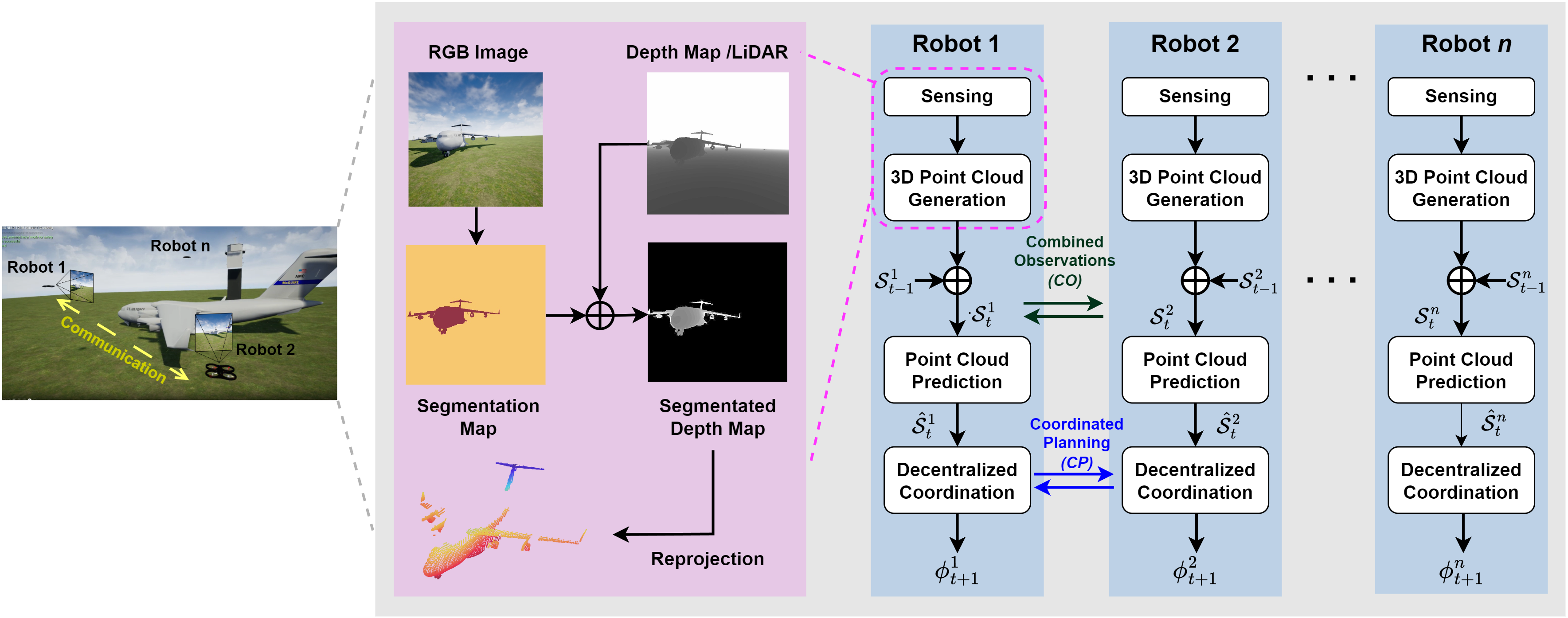}
    \caption{Algorithm Overview for MAP-NBV: Each robot runs the same algorithm including perception, prediction, and planning steps. The robots that communicate with each other can share observations and coordinate planning, whereas robots in isolation (e.g., Robot n) perform individual greedy planning.}
    \label{fig:mapnbv:algoOverview}
\end{figure*}

\section{Related Work}\label{sec:mapnbv:rel_work}

The use of robots for data acquisition purposes is an extensively studied topic for various domains. Their usage range from infrastructure inspection~\cite{ozaslaninspection} and environment monitoring~\cite{dunbabin2012environmental,sung2019competitive} for real-world application to the real-world digitization for research datasets and simulations~\cite{46965,ramakrishnan2021hm3d,ammirato2017dataset}. When the environment is unknown, active methods such as NBV~\cite{connolly1985determination} are used to construct an object model on the fly by capturing additional observations. A majority of the works on NBV planning use information-theoretic measures~\cite{delmerico2018comparison} for selection to account for uncertainty in observations~\cite{delmerico2018comparison,kuipers1991robot,vasquez2014volumetric}. The widely used frontier and tree-based exploration approaches also utilize uncertainty about the environment for guiding the robot motion~\cite{yamauchi1997frontier, gonzalez2002navigation, adler2014autonomous, bircher2018receding}. Some works devise geometric methods that make inferences about the exact shape of the object of interest and try to align the observations with the inferred model~\cite{tarabanis1995survey, banta2000next, kriegel2013combining}. Prediction-based NBV approaches have emerged as another alternative in recent years, where a neural network takes the robot and/or the environment state as the input and NBV pose or velocity as the output~\cite{johns2016pairwise, mendoza2020supervised, zeng2020pc, dhami2023prednbv}.

Directly extending single-robot NBV approaches to multi-robot systems may result in sub-optimal performance due to significant overlap in observations. This issue led to the development of exploration algorithms specifically for multi-robot systems~\cite{burgard2005coordinated, amanatiadis2013multi, hardouin2020next} with information-theoretic measures for determining NBV. 

Some recent works on multi-robot systems have explored the use of predictions for improvement in task efficiency. Almadhoun et al.~\cite{ almadhoun2021multi} designed a hybrid planner that switches between a classical NBV approach and a learning-based predictor for NBV selection but uses the partial model obtained by robot observations only.
Wu et al.~\cite{wu2019plant} use a point cloud prediction model for plants to use the predicted point cloud as an oracle leading to better results than the traditional approaches. This method uses entropy-based information gain measures for NBV and is designed for plant phenotyping with robotic arms. These methods do not consider the control effort required which is important for UAVs with energy constraints when deployed for observing large objects such as airplanes and ships. Also, these works employ information theoretic NBV approaches. We aim to explore a prediction-based approach for geometric NBV selection.

In this work, we used point cloud predictions similar to Pred-NBV~\cite{dhami2023prednbv} and built a decentralized multi-robot NBV planner. The prediction on the point cloud makes the pipeline modular and interpretable, allowing for improvements by enhancing individual modules. We select NBV based on information gain, as well as control effort, making our approach more grounded in the real world.

\section{Problem Formulation}\label{sec:mapnbv:prob_form}
We are given a team of \textit{n} robots, each equipped with an RGB-D camera or LiDAR sensor. 
This team is tasked with navigating around a closed object of volume $\mathcal{V} \in \mathbb{R}^3$ and observes its surface as a set of 3D points, $\mathcal{S}$.
At any given time $t$, the set of surface points $S^i_t$ observed by the robot $r_i$ from the view-point $\phi^i_t \in \Phi$ at time $t$ is represented as a voxel-filtered point cloud and the relationship between them is defined as $S^i_t = f(r_i, \phi^i_t)$. Each robot $r_i$ follows a trajectory $\xi_{i}$, which consists of a sequence of viewpoints aimed at maximizing the coverage of $\mathcal{S}$ while minimizing redundancy among observed points. 

The distance traveled by a robot between two poses $\phi_i$ and $\phi_j$ is represented by $d(\phi_i, \phi_j)$. The point cloud observed by the team of robots is the union of the surface points observed by the individual robots over their respective trajectories, i.e., $S_{\Bar{\xi}} =  \bigcup_{i=1}^n \bigcup_{\phi \in \xi_{i}} f(r_i, \phi)$ and $\Bar{\xi}$ represents the set of trajectories for each robot, i.e., $\Bar{\xi} = \{\xi_{1}, \xi_{2},..., \xi_{n}\}$.

The objective is to find a set of feasible trajectories $\Bar{\xi}^* = \{ \xi_{1}^*, \xi_{2}^*, ..., \xi_{n}^*\}$, such that the team observes the whole voxel-filtered surface $\mathcal{S}$, while also minimizing the total distance traveled by the robots on their respective trajectories.
\begin{align}
    \Bar{\xi}^* = \argmin_{\Bar{\xi}} \sum_{i=1}^n \sum_{j=1}^{| {\xi_{i}}-1|} d(\phi_j^i, \phi_{j+1}^i)\\ 
    \textit{such that}~~ \bigcup_{i=1}^n \bigcup_{\phi \in \xi_{i}} f(r_i, \phi) = \mathcal{S} 
\end{align}

Given a finite set of trajectories, if the object model, $\mathcal{S}$, is known, we can find the optimal set of trajectories through an exhaustive search. As the object model is not known apriori in an unknown environment, the optimal solution can not be found beforehand. Thus, each robot needs to determine the NBV based on the partial observations of the team to reconstruct the object's surface. Here we assume that each robot can observe the object at the start of the mission, which can be accomplished by moving the robots till they see the object. While a centralized server can help find an optimal assignment solution, a limited communication range can make a centralized solution infeasible. Thus, we define this problem as a decentralized one; each robot solves this objective but the communicating robots can collaborate and coordinate with their neighbors.

\section{Proposed Approach: Multi-Agent Prediction-based Next-Best-View (MAP-NBV)}\label{sec:mapnbv:approach}
In this section, we present \textit{Multi-Agent Pred-NBV (MAP-NBV)}, a prediction-guided NBV approach for a team of robots. Figure~\ref{fig:mapnbv:algoOverview} shows the overview of our process, which consists of two parts: (1) \textit{3D Model Prediction}, where we combine the observations from the neighboring robots to build a partial model of the object and use PoinTr-C~\cite{dhami2023prednbv}, a 3D point cloud completion network, to predict the full shape of the objects, 
and (2) \textit{Decentralized Coordination} which combines the observations from the communicating robots and solves an NBV objective to maximize information gain and minimize the control effort with sequential greedy assignment. Robots that do not communicate with anyone effectively run an individual greedy algorithm. This approach is detailed in Algorithm~\ref{algo:mapnbv}. Apart from being feasible and scalable, this approach also reduces the computation complexity resulting in fast runtime.

\begin{algorithm}
\caption{MAP-NBV Algorithm (for robot $r_i$)}
\begin{algorithmic}[1]
    \State \textbf{Inputs:} Initial positions $\phi^i_0$; Stopping threshold $\tau$; Information gain threshold $\lambda$
    \State \textbf{Output:} 3D Point Cloud $\mathcal{S}$
    \State \textbf{Initialization}: $novelty$ $\gets 0$; $t \gets 0$; $S^i_{-1} \gets \emptyset$, $\xi_i \gets \{\phi^i_0\}$
    \While{$novelty \le \tau$} \label{algo:line:stopping_condition}
        \State $\mathcal{S}^i_t \gets \bigcup_{k \in Neighbors} f(r_i, \phi^i_t)$ $\bigcup \mathcal{S}^i_{t-1}$  \label{algo:line:combined_observations}
        \State $\hat{\mathcal{S}}^i_t \gets$ getPredictionFromPoinTr-C($\mathcal{S}^i_t$) \label{algo:line:pc_completion}
        \State $\mathcal{C} \gets$ generateCandidatePoses($\hat{\mathcal{S}}^i_t, \mathcal{S}^i_t$) \label{algo:line:cand_pose_generation}

        \State $\mathcal{I}_{seen} \gets \bigcup_{k \in Neighbors; k < i} I (\xi_k)$ \label{algo:line:infogather}

        \State $\hat{\mathcal{I}} \gets \{ \mathcal{I}(\xi_i \cup \phi) - \mathcal{I}_{seen}; \forall  \phi \in \mathcal{C} \}$  \label{algo:line:potential_I_calculation}

        \State $\phi^i_{t+1} \gets \argmin_{\phi \in \mathcal{C}} d(\phi, \phi^i_{t+1})$, s.t. $\frac{\hat{\mathcal{I}}(\phi)}{\max \hat{\mathcal{I}}} \ge \lambda$ \label{algo:line:objective}
        
        \State $\xi_i \gets \xi_i \cup \phi^i_{t+1}$ \label{algo:line:traj_update}
        
        \State Broadcast($\xi_i$) \label{algo:line:broadcast}
        
        \State $novelty \gets \frac{| \mathcal{S}^i_t |}{| \mathcal{S}^i_{t-1}|}$ \label{algo:line:novelty_update}
        \State $t \gets t+1$ \label{algo:line:time_update}
    \EndWhile
    \State $\mathcal{S} \gets \bigcup_{i=1}^n \mathcal{S}^i_{t-1}$  \label{algo:line:final_pcd}
\end{algorithmic}
\label{algo:mapnbv}
\end{algorithm}

\subsection{3D Model Prediction (Line~\ref{algo:line:combined_observations}-\ref{algo:line:pc_completion})} 
To start, we use the RGB images to segment out the object and the depth sensors to generate the point cloud from the current observations for each robot, giving us segmented point clouds. This allows the algorithm to focus on only the target infrastructure as opposed to also including other obstacles. For the robots that can communicate with each other (e.g., Robot 1 and Robot 2 in Fig~\ref{fig:mapnbv:algoOverview}), each segmented point cloud per robot is transformed into a global reference frame and concatenated together into a single point cloud (Line~\ref{algo:line:combined_observations}). This point cloud represents the entire communication subgraph's observations of the target object at the current timestamp. The point cloud concatenation can be replaced with a registration algorithm~\cite{huang2021comprehensive}, but we use concatenation due to its ease of use. Lastly, this current timestamp's point cloud is then concatenated with previous observations to get an up-to-date observation point cloud.

In order to get an approximation $\mathcal{\hat{S}}$ of the full model $\mathcal{S}$, we use PoinTr-C~\cite{dhami2023prednbv}. The combined observed point cloud of the object at time $t$, $\mathcal{S}_t$ goes as input to PoinTr-C and it predicts the full object point cloud $\mathcal{\hat{S}}_t$ (Line~\ref{algo:line:pc_completion}). PoinTr-C requires isolating the object point clouds from the scene. This can be realized with the help of distance-based filters and state-of-the-art segmentation networks\cite{kirillov2023segment} without any fine-tuning. An example of an input point cloud and a predicted point cloud, both from individual observations and combined observations, is shown in Figure~\ref{fig:mapnbv:c17predicted}.

\subsection{Decentralized Coordination (Line~\ref{algo:line:cand_pose_generation}-\ref{algo:line:broadcast})}
\textbf{Next Best View Planning}. We use the predicted point cloud $\mathcal{S}_t$ as an approximation of the ground truth point cloud for NBV planning. For this, we first generate a set of candidate poses around the partially observed object (Line~\ref{algo:line:cand_pose_generation}). From these, we select a set of poses as NBVs for each robot, based on information gain and control effort. For a set of $k$ candidate viewpoints, we define the information gain, $\hat{\mathcal{I}}$, as the expected number of new, unique points the robots will observe after moving to these viewpoints. The control effort is defined as the total distance traversed by the robots to reach the viewpoints.

The number of new points varies with each iteration as robots move to new locations, observing more of the object's surface. While PoinTr-C predicts the point cloud for the whole object, the robots can observe only points on the surface close to it. Hence, before counting the number of new points, we apply hidden point removal~\cite{katz2007direct} to the predicted point cloud. We represent this relationship between the number of points observed and the trajectories traversed till time for a robot $r_i$ as $I(\xi_i)$, where $\xi = \{\phi^i_0, \phi^i_t, ..., \phi^i_t\}$ represents the trajectory for the robot $r_i$ till time $t$ consisting of the set of viewpoints the robot has traversed. To balance the information gain and control effort, we use a hyperparameter $\lambda$. First, we find the estimated information gain for each candidate pose $\hat{\mathcal{I}}$ (Line~\ref{algo:line:potential_I_calculation}) by treating the $\hat{\mathcal{S}}$ as the actual full object model. Each robot selects the candidate pose closest to the current pose where it achieves at least $\lambda \%$ of the maximum possible information gain over all the candidate poses (Line~\ref{algo:line:objective}).

To find the control effort, we use RRT-Connect~\cite{kuffner2000rrt} to find the path between the robot $r_i$'s current location $\phi^i_t$ to each candidate pose $\phi$ and use its length as $d(\phi, \phi^i_t)$. 
The candidate poses are generated similarly to Pred-NBV~\cite{dhami2023prednbv}, i.e., on circles at different heights around the center of the predicted object point cloud (Line~\ref{algo:line:cand_pose_generation}).
One circle is at the same height as the predicted object center with radius $1.5 \times d_{max}$, where $d_{max}$ is the maximum distance of a point from the center of the predicted point cloud. The other two circles are located above and below this circle $0.25 \times \text{z-range}$ away, with a radius of $1.2 \times d_{max}$. The viewpoints are located at steps of $30^\circ$ on each circle.

\begin{figure}[ht!]
    \centering
    \includegraphics[width=0.90\linewidth]{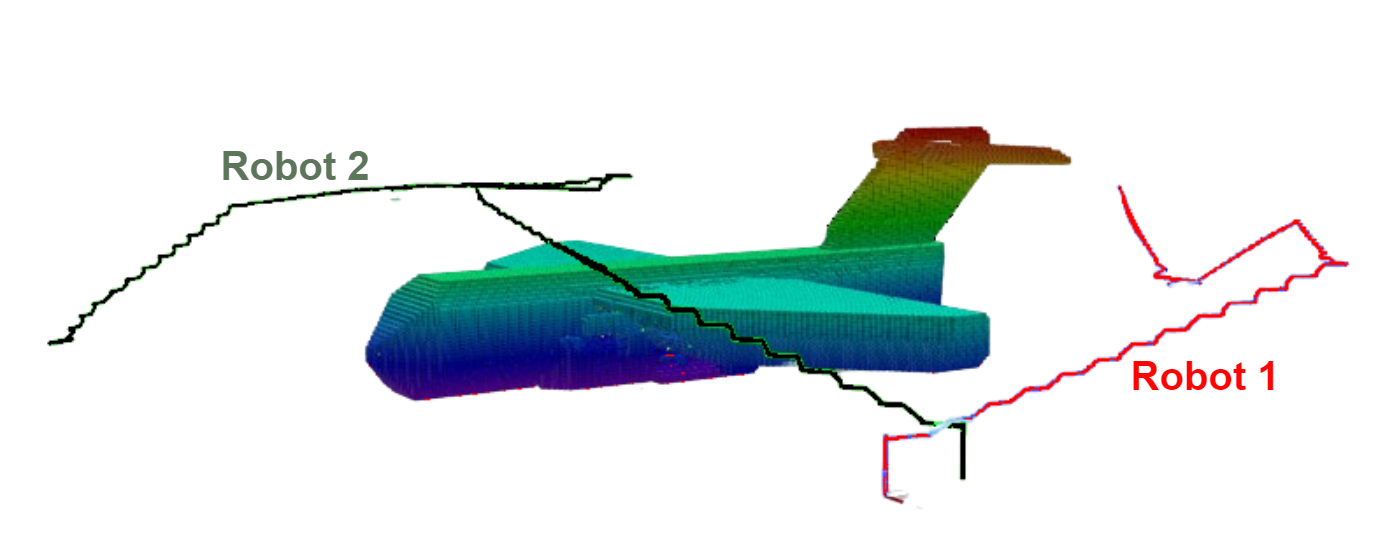}
    \caption[Flight paths of the two robots during C-17 simulation for MAP-NBV.]{Flight paths of the two robots during C-17 simulation.}
    \label{fig:mapnbv:flightPath}
\end{figure}

\textbf{Sequential Greedy Assignment}. To effectively and efficiently use multiple robots, we coordinate among them for NBV assignment. We devise a decentralized coordination strategy to accommodate the dynamic communication graph effectively. We also leverage the submodular nature of the problem and use a sequential greedy assignment which results in a near-optimal solution~\cite{calinescu2011maximizing}. With each communications subgraph, the robots select their NBVs in the order determined by their IDs (lower to higher). Each robot first considers the trajectories of its neighbors with lower IDs to find the information gain attained by their movement $\mathcal{I}_{seen}$ (Line~\ref{algo:line:infogather}). For the robot with the lowest ID, $\mathcal{I}_{seen}$ is empty. Then we greedily choose the candidate position that would result in the information gain above a threshold $\lambda$ while minimizing the distance traveled. This candidate's position for the next iteration 
$\phi_{t+1}^i$ is the NBV for $r_i$. We add this pose to the robot's trajectory (Line~\ref{algo:line:traj_update}) and broadcast this information with the neighbors (Line~\ref{algo:line:broadcast}) to minimize overlaps.

In our experiments, we consider multi-hop communication, thus each robot can coordinate with every robot on its communication subgraph. Some robots may not be within any other robot's communication range (e.g., Robot \textit{n} in Fig~\ref{fig:mapnbv:algoOverview}). For such robots, this strategy effectively turns into a greedy prediction-guided assignment similar to Pred-NBV~\cite{dhami2023prednbv}. At each iteration, we calculate $novelty$, i.e., the ratio of the number of points observed over subsequent iterations (Line~\ref{algo:line:novelty_update}). The robot stops if $novelty$ is above a predefined threshold $\tau$ (we set $\tau = 0.95$ in our experiments).
In the end, all the robots assemble at the same location and combine their observations as the object model $\mathcal{S}$ (Line~\ref{algo:line:final_pcd}).

\section{Experiments and Evaluation}\label{sec:mapnbv:eval}
We design experiments to answer the two key research questions: (1) can point cloud prediction improve multi-agent object reconstruction? and (2) how does coordination between agents affect the reconstruction? To answer the first question, we compare \textit{MAP-NBV} with a non-predictive frontier-based baseline. For the second question, we compare \textit{MAP-NBV} with a centralized and a non-coordinated variation of \textit{MAP-NBV}. We first describe the experiment setups used to answer these questions and then discuss the results obtained.

\subsection{Setup}
\label{subsec:mapnbv:setup}
\textbf{AirSim}: We use AirSim~\cite{airsim2017fsr} simulator as it allows us to load desired object models and get photo-realistic inputs while also supporting multiple robots. We use Robot Operating System (ROS) Melodic to run the simulations on Ubuntu 18.04. We spawn multiple UAVs close to each other looking towards the object. We equip each UAV with a depth camera and an RGB camera. Each UAV publishes a segmented image using AirSim's built-in segmentation. This segmented image is used along with a depth map to remove the background and isolate the depth map for the object. We then convert these segmented depth images into 3D point clouds.  For collision-free point-to-point planning, we use the MoveIt~\cite{coleman2014reducing} package implementing the work done by Köse~\cite{tahsinko86:online}.

\begin{figure}[htp]
% First column
\begin{minipage}[b]{0.43\linewidth}
    \centering
    % Airplane
    \begin{subfigure}[b]{\linewidth}
        \centering
        \includegraphics[width=\textwidth]{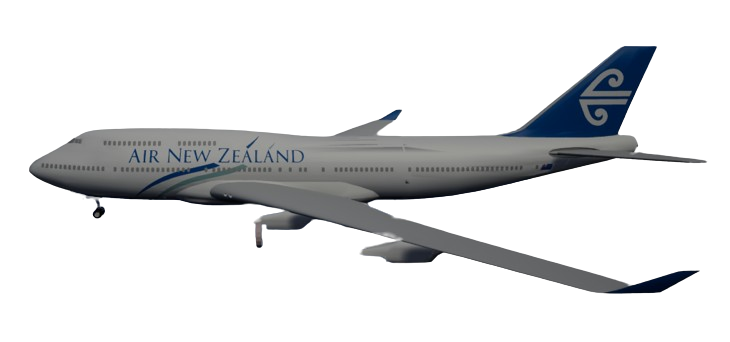}
        \caption{Airplane}
        \label{fig:mapnbv:airplane}
    \end{subfigure}

    % Train
    \begin{subfigure}[b]{\linewidth}
        \centering
        \includegraphics[width=\linewidth]{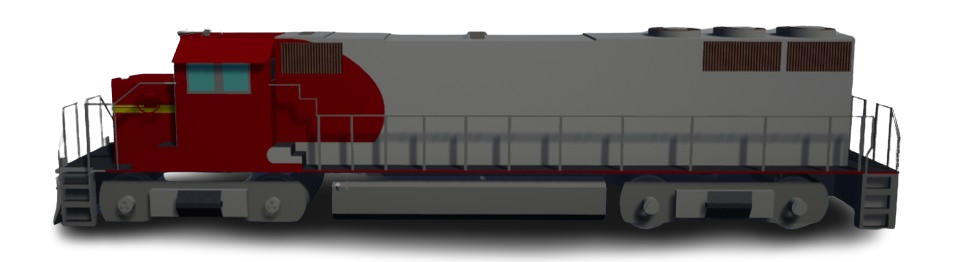}
        \caption{Train}
        \label{fig:mapnbv:train}
    \end{subfigure}

    % Boat
    \begin{subfigure}[b]{\linewidth}
        \centering
        \includegraphics[width=\linewidth]{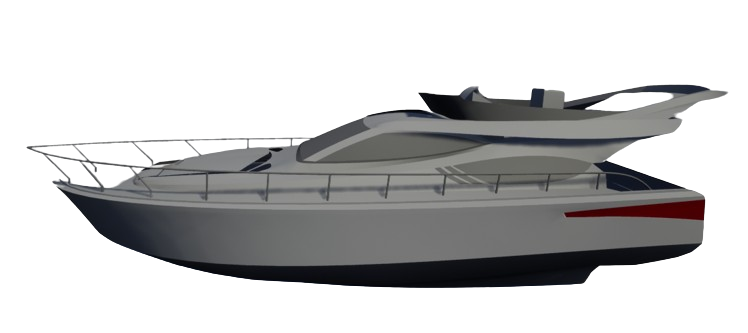}
        \caption{Boat}
        \label{fig:mapnbv:boat}
    \end{subfigure}
\end{minipage}
% \hfill
% Second column
\begin{minipage}[b]{0.56\linewidth}
    \centering
    % Rocket
    \begin{subfigure}[b]{0.41\linewidth}
        \centering
        \includegraphics[width=0.80\textwidth]{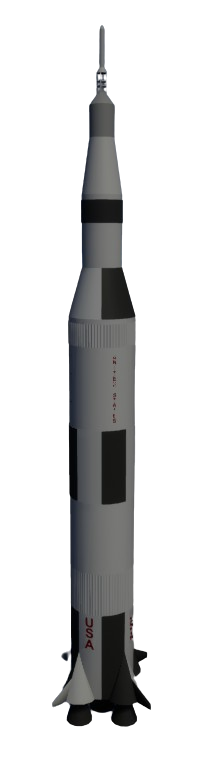}
        \caption{Rocket}
        \label{fig:mapnbv:rocket}
    \end{subfigure}
% \end{minipage}    
% \begin{minipage}[b]{0.29\textwidth}    
    % Tower
    \begin{subfigure}[b]{0.55\linewidth}
        \centering
        \includegraphics[width=0.90\textwidth]{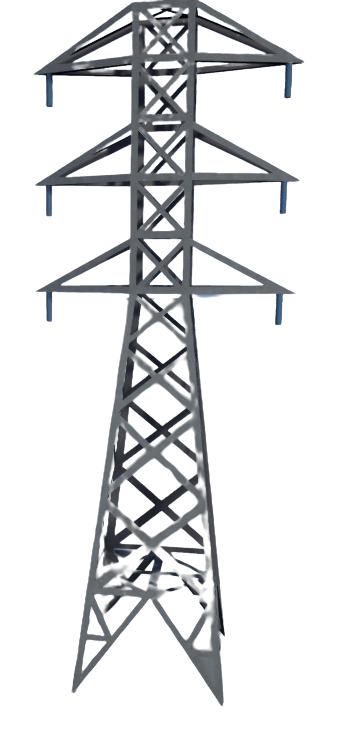}
        \caption{Tower}
        \label{fig:mapnbv:tower}
    \end{subfigure}
\end{minipage}

\caption[Examples of the 5 simulation model classes used for the multi-agent object reconstruction task using MAP-NBV.]{Examples of the 5 simulation model classes.}
\label{fig:mapnbv:simModels}
\end{figure}

\textbf{ShapeNet}: The AirSim setup helps us evaluate the coverage in a realistic setting, but it lacks a ground truth point cloud to compare the reconstruction quality. Hence, we run simulations with the ShapeNet~\cite{shapenet2015} dataset to obtain a ground truth and compare the reconstruction quality across different methods. We simulate the UAVs as free-floating cameras which can teleport to any location. These cameras do not have a roll and pitch and can only change their yaw. We use the Open3D~\cite{Zhou2018} library to load an object mesh (normalized) and select a random starting position around it. The UAVs are placed close to each other at this location. We render the depth image from the object mesh and reproject it to the 3D point cloud. Unlike the previous setup, we do not need segmentation as there is no background in this setup. 
To evaluate the reconstruction quality, we convert the mesh to a point cloud by uniformly sampling points on it and use it as the ground truth to quantify the reconstruction quality with Directional Chamfer Distance with $\ell_2$-norm (CD-$\ell_2$) (from ground truth point cloud to the observed point cloud). We use Euclidean distance as the distance metric $d(.,.)$ in this setup.

\subsection{Qualitative Example}
We evaluate \textit{MAP-NBV} on the same 20 objects that were used in Pred-NBV to allow a direct comparison. 
The 20 objects consist of five different ShapeNet classes:  airplane, rocket, tower, train, and watercraft. Examples of each class are shown in Figure~\ref{fig:mapnbv:simModels}. These classes represent diverse shapes and infrastructures that are regularly inspected. Figure~\ref{fig:mapnbv:flightPath} shows the path followed by two UAVs as given by \textit{MAP-NBV} in the C-17 airplane simulation. This environment includes other obstacles that are not of interest but still need to be accounted for in collision-free path planning. \textit{MAP-NBV} finds a collision-free path for both UAVs while targeting the maximum coverage of the C-17 airplane.

\subsection{Can Point Cloud Prediction Improve Reconstruction?}\label{sec:mapnbv:sim:MAbaseline}
We compared the performance of \textit{MAP-NBV} with a modified baseline NBV method~\cite{aleotti2014global} designed for multi-agent use on 20 objects, as listed in Table~\ref{tab:mapnbv:airsim_multiagent_results}. The baseline method employs frontiers to select the next-best views. Frontiers are points located at the edge of the observed space near unknown areas. We utilized the same modifications described in Pred-NBV~\cite{dhami2023prednbv}. Specifically, we used our segmented point cloud to choose frontiers near the target object. To ensure that the UAVs always face the target object, the orientation of all poses selected by the baseline aligns with the center of the observed target object point clouds.

We further adapted this baseline method to function in a multi-agent setting. The pose for the first UAV is selected in the exact same manner as in the single-agent baseline. For each subsequent UAV, the remaining best pose is chosen, as long as it does not fall within a certain distance threshold compared to the previously selected poses in the current iteration of the algorithm.

Both \textit{MAP-NBV} and the baseline algorithm employ the same stopping criteria. The algorithm terminates if the total points observed in the previous step exceed 95\% of the total points observed in the current step. We run both the algorithms in the AirSim setup. Our evaluation, presented in Table~\ref{tab:mapnbv:airsim_multiagent_results}, demonstrates that \textit{MAP-NBV} observes, on average, 19.41\% more points than the multi-agent baseline for object reconstruction across all 20 objects from the five different model classes. In our simulations, we utilized 2 UAVs for both algorithms.

Furthermore, the \textit{MAP-NBV} algorithm can be readily extended to accommodate more than just 2 robots. By incorporating additional UAVs, the algorithm can effectively leverage the collaborative efforts of a larger multi-agent system to improve object reconstruction performance and exploration efficiency. However, in our current evaluation, we utilized 2 UAVs for both algorithms due to limited computational resources. The simulations were computationally intensive, and our computer experienced significant slowdowns with just 2 robots in the simulation. Despite this limitation, the promising results obtained with 2 UAVs suggest that scaling up the algorithm to include more robots has the potential to yield even more significant performance improvements.

\begin{figure}[ht!]
    \centering
    \includegraphics[width=0.65\linewidth]{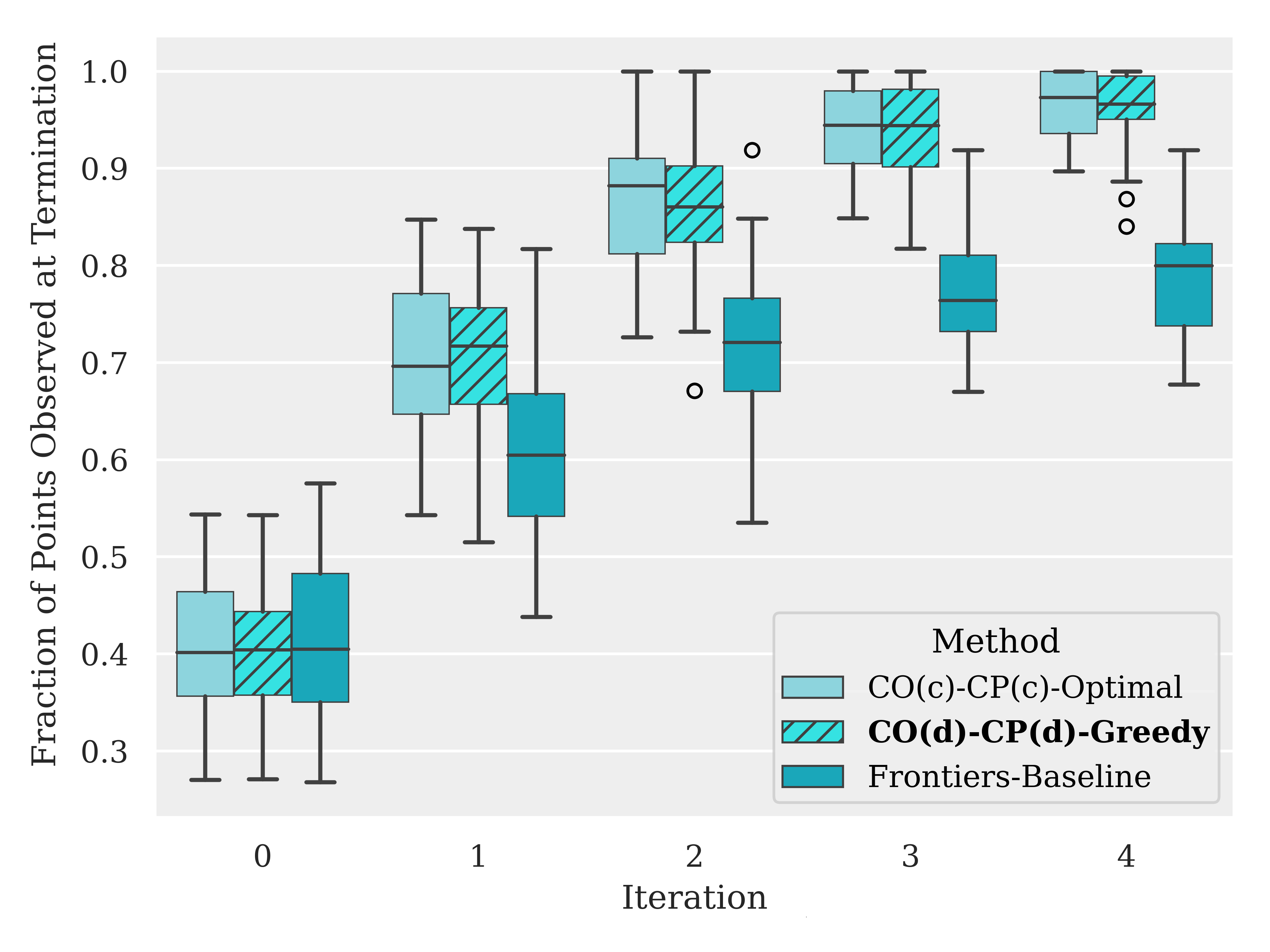}
    \caption{MAP-NBV (\texttt{CO(d)-CP(d)-Greedy}) performs comparably to the optimal solution (\texttt{CO(c)-CP(c)-Optimal}; Section~\ref{sec:mapnbv:COCPdesc}),  and much better than the frontiers-baseline in AirSim experiments.\looseness=-5 }
    \label{fig:mapnbv:plane_res}
\end{figure}

Additionally, Figure~\ref{fig:mapnbv:plane_res} illustrates that \textbf{\textit{MAP-NBV} observes more points per step than the multi-agent baseline}.

\begin{table}[ht!]
    \centering

    \caption{\textit{MAP-NBV} results in a better coverage compared to the multi-agent baseline NBV method~\cite{aleotti2014global} for all models in AirSim upon algorithm termination.}
    
    \begin{tabular}{llrrr}%{p{0.9cm}p{1.3cm}rrr}
    \toprule
    \multirow{2}{*}{Class} & \multirow{2}{*}{Model} & \multicolumn{2}{c}{Points Seen} & \multirow{2}{*}{Improvement} \\
    \cmidrule(lr){3-4}
    & & \textit{MAP-NBV} & MA Baseline & \\
        \midrule
        \multirow{5}{*}{Airplane}  
            & 747 & \textbf{16140} & 13305 & 19.26\% \\
            & A340 & \textbf{10210} & 8156 &  22.37\% \\
            & C-17 & \textbf{13278} & 10150 & 26.70\% \\
            & C-130 & \textbf{6573} & 5961 & 9.77\% \\
            & Fokker 100 & \textbf{14986} & 13158& 12.99\%\\
        \midrule
        \multirow{5}{*}{Rocket} 
            & Atlas & \textbf{2085} & 1747 & 17.64\% \\
            & Maverick & \textbf{3625} & 2693 & 29.50\% \\
            & Saturn V & \textbf{1041} & 877 & 17.10\% \\
            & Sparrow & \textbf{1893} & 1664 & 12.88\% \\
            & V2 & \textbf{1255} & 919 & 30.91\% \\
        \midrule
        \multirow{5}{*}{Tower} 
            & Big Ben & \textbf{4294} & 3493 & 20.57\% \\
            & Church & \textbf{7884} & 6890 & 13.46\% \\
            & Clock & \textbf{3163} & 2382 & 28.17\% \\
            & Pylon & \textbf{2986} & 2870 & 3.96\% \\
            & Silo & \textbf{5810} & 4296 & 29.96\% \\
        \midrule
        \multirow{2}{*}{Train}
            & Diesel & \textbf{4013} & 3233 & 21.53\% \\
            & Mountain & \textbf{5067} & 4215 & 18.36\% \\
        \midrule
        \multirow{3}{*}{Watercraft} 
            & Cruise & \textbf{5021} & 3685 & 30.69\% \\
            & Patrol & \textbf{4078} & 3683 & 10.18\% \\
            & Yacht & \textbf{11678} & 10341 & 12.14\% \\
        \bottomrule
    \end{tabular}
    
    \label{tab:mapnbv:airsim_multiagent_results}
\end{table}

\subsection{How does coordination affect reconstruction?}\label{subsec:mapnbv:ablation}
For studying the effect of coordination we randomly select 25 objects, 5 from each of the ShapeNet classes mentioned in Section~\ref{sec:mapnbv:sim:MAbaseline} for the ShapeNet setup (Section~\ref{subsec:mapnbv:setup}).

\label{sec:mapnbv:COCPdesc}
In a multi-agent setting, the communicating robots can combine their observation (\textbf{CO}) or they may choose to rely on their observations (\textbf{IO}) for point cloud predictions. After the prediction, they may choose to collaborate for planning (\textbf{CP}) or make individual decisions without relying on others (\textbf{IP}). To highlight the effect of these coordination strategies on reconstruction quality, for a team of $n$ robots and $k$ candidate viewpoints, we compare MAP-NBV with two contrasting coordination approaches: 
\begin{itemize}
    \item \textbf{\texttt{CO(c)-CP(c)-Optimal}}: This approach relies on combined observation and coordinated planning in a centralized manner. Here, a central server aggregates the observations, performs prediction, and finds the NBVs for each robot. For selecting NBVs, the algorithm evaluates all robot-candidate pose assignments and chooses the optimal setting, i.e.,  which results in maximum joint information gain. Among all possible permutations of robots that result in the optimal setting, we select the one that minimizes the maximum displacement for any robot. This process has a runtime complexity of $\mathcal{O}(k^{n})$.
    \item \textbf{\texttt{IO-IP}}: In this approach, each robot operates individually and does not share observations or coordinate with others regardless of the communication range. This is effectively a naive extension of Pred-NBV to a multi-agent setup. The runtime complexity here is $\mathcal{O}({k})$.
\end{itemize}

Following the notations above, \textbf{\uline{MAP-NBV is \texttt{CO(d)-CP(d)-Greedy}}}, i.e., combined observations (decentralized) and coordinated planning (decentralized) with greedy assignment. \textit{Decentralized} indicates that the observations and planning are shared among only those robots that form a communication subgraph. \textit{MAP-NBV} thus has a runtime complexity of $\mathcal{O}(n \cdot k)$ in the worst case only, i.e., when all the robots are connected. We study these algorithms for teams of 2, 4 and 6 robots. 

\begin{figure}[ht!]
    \centering
    \begin{subfigure}{\columnwidth}%
        \centering
        \includegraphics[width = 0.5\textwidth]{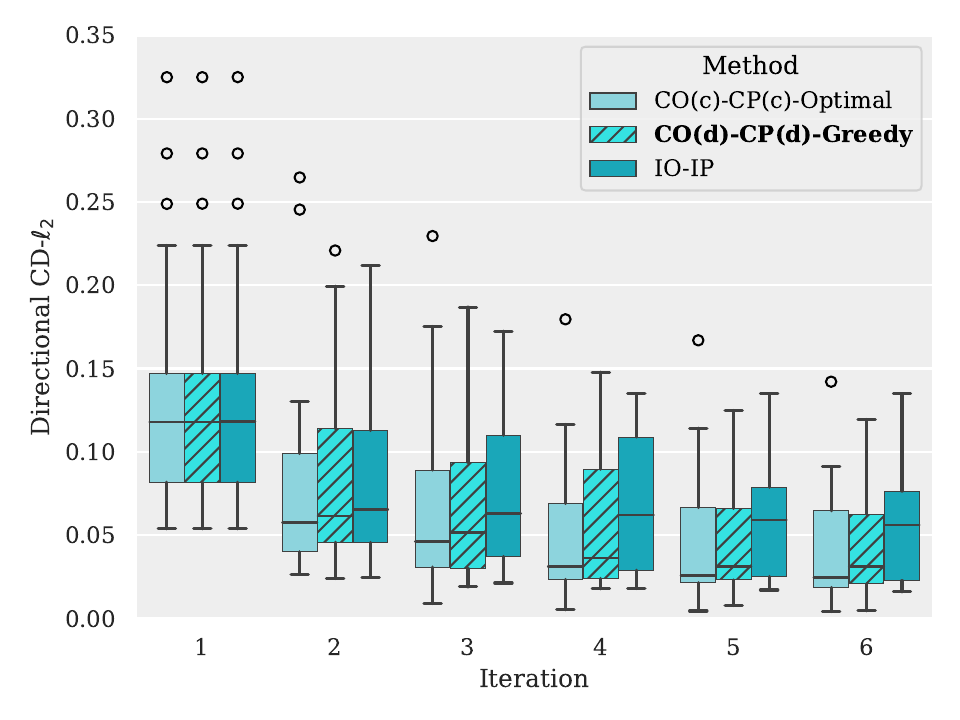}%
        \subcaption{2 Robots}
    \end{subfigure}%
    \hfill%
    \begin{subfigure}{\columnwidth}%
        \centering
        \includegraphics[width = 0.5\textwidth]{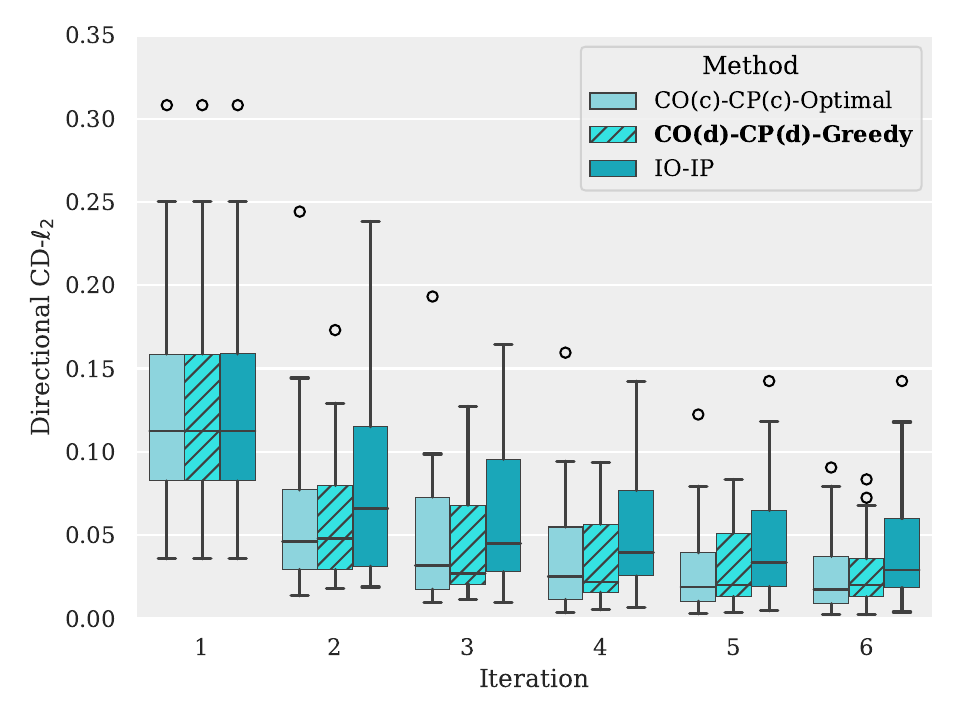}%
        \subcaption{4 Robots}
        
    \end{subfigure}%
    \hfill%
    \begin{subfigure}{\columnwidth}%
        \centering
        \includegraphics[width = 0.5\textwidth]{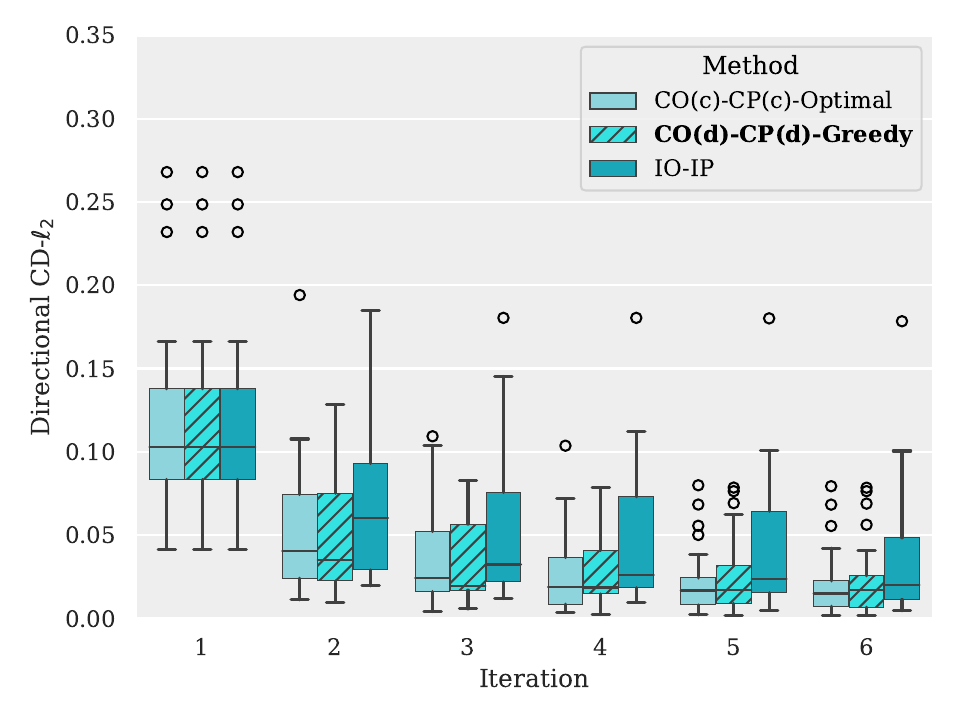}%
        \subcaption{6 Robots}
    \end{subfigure}%
    \caption{Directional CD-$\ell_2$ for teams of 2, 4, and 6 robots on ShapeNet models~\cite{shapenet2015} with different coordination strategies.}
    \label{fig:mapnbv:ablation}
\end{figure}

At each iteration, we combine the observations from all robots and compute Direction CD-$\ell_2$ over it. This thus represents the reconstruction quality if the mission terminated at that iteration. We found that 6 iterations were enough for all robots to reach the stopping criteria in each setting. Our findings are shown in Figure~\ref{fig:mapnbv:ablation}.

We observed that coordination and information sharing play a crucial role in improving the reconstruction quality. \texttt{CO(c)-CP(c)-Optimal} exhibits the best reconstruction over time, as expected, owing to shared observations which lead to better estimation of the partial point cloud, and coordinated planning, which minimizes overlap in information gain. In early iterations, the limited observations may lead to an imprecise prediction, which may result in inefficient NBV assignments. Still, over time the observation coverage increases leading to better predictions and performance over the other algorithms. We observed similar trends for the AirSim setup as well as shown in Figure~\ref{fig:mapnbv:plane_res}.

However, \texttt{CO(c)-CP(c)-Optimal} requires large computation time and does not scale well in comparison with \textit{MAP-NBV} \texttt{(CO(d)-CP(d)-Greedy)}. On a Ubuntu 20.04 system with 32-core, 2.10Ghz Xeon Silver-4208 CPU and Nvidia GeForce RTX 2080Ti GPU, \textit{MAP-NBV} was 2x faster for 4 robots and 10x faster for 6 robots compared to \texttt{CO(c)-CP(c)-Optimal}. Even with faster execution, \textit{MAP-NBV} performs comparably to \texttt{CO(c)-CP(c)-Optimal}, making it more attractive than the former. Additionally, the decentralized nature makes \textit{MAP-NBV} a more feasible algorithm than \texttt{CO(c)-CP(c)-Optimal} which requires centralization. \texttt{IO-IP}, where the robots do not share observations or coordinate plans, exhibits the worst improvement over time, highlighting that \textbf{the coordination plays a crucial role in multi-agent object reconstruction}. In fact, \textit{MAP-NBV} achieves a relative improvement of \textbf{17-22\%} in directional CD-$\ell_2$ over \texttt{IO-IP} after termination.

Interestingly, we observe that as the number of robots increases, the gap between these algorithms decreases. This is expected as more robots lead to more coverage at any time. Since the robots start the mission at close distances in our experiments, we found that \texttt{IO-IP} still exhibits worse improvement as the robots suffer significant overlaps in observations. Thus coordination still provides benefits and faster runtime of \textit{MAP-NBV} makes it a more suitable choice than the centralized alternative, even if centralization is feasible. 
We also performed experiments with a centralized, greedy selection variant of \textit{MAP-NBV} and found it performs only marginally better than \textit{MAP-NBV}'s performance. We share these findings on our \href{https://raaslab.org/projects/MAPNBV/}{project webpage} due to lack of space.

\subsection{Qualitative Real-World Experiment}
\begin{figure*}[ht!]
    \centering
    \begin{subfigure}[c]{.24\textwidth}%
        \includegraphics[width = \textwidth]{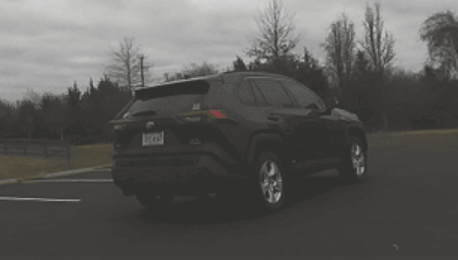}%
        \caption{}
    \end{subfigure}%
    \hfill
    \begin{subfigure}[c]{.24\textwidth}%
        \includegraphics[width = \textwidth]{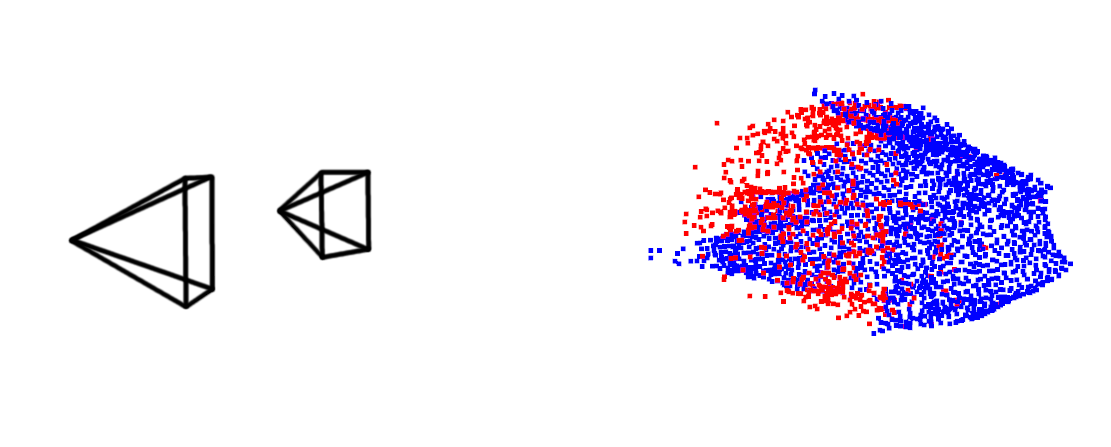}%
        \caption{}
    \end{subfigure}%
    \hfill
    \begin{subfigure}[c]{.24\textwidth}%
        \includegraphics[width = \textwidth]{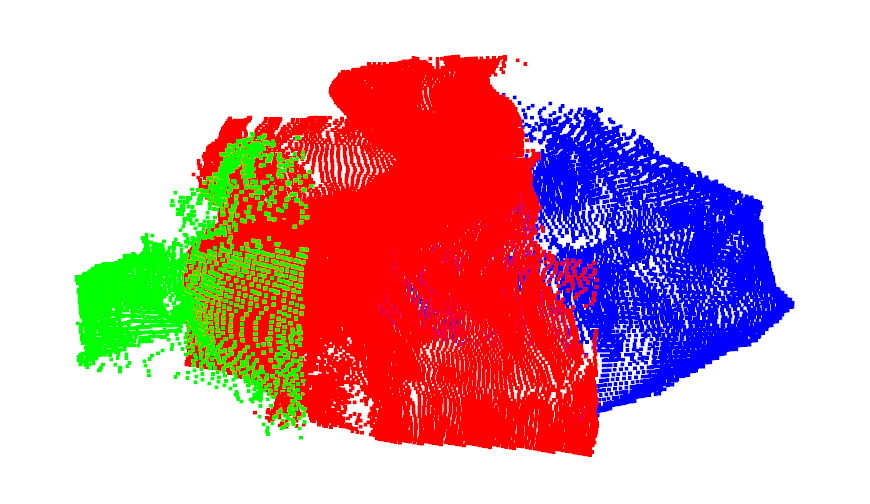}%
        \caption{}
    \end{subfigure}%
    \hfill
    \begin{subfigure}[c]{.24\textwidth}%
        \includegraphics[width = \textwidth]{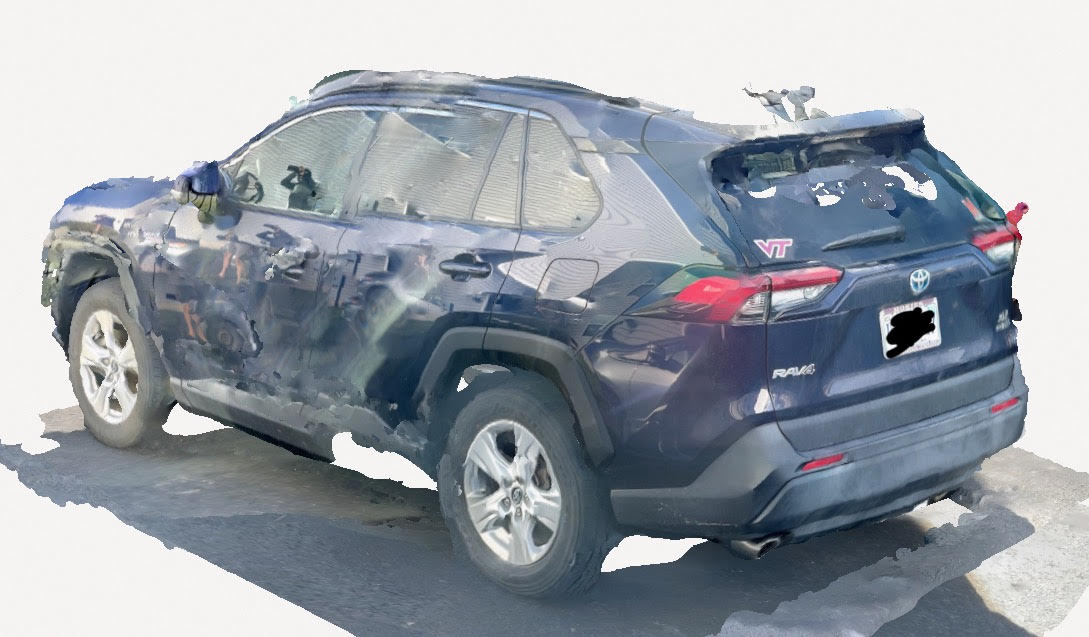}%
        \caption{}
    \end{subfigure}
    \caption{Real-World MAP-NBV experiment. (a) RGB Image. (b) \textcolor{blue}{Observations}, \textcolor{red}{Predictions}, and MAP-NBV poses. (c) \textcolor{blue}{Initial}, \textcolor{red}{Drone 1}, and \textcolor{green}{Drone 2} points after MAP-NBV iteration. (d) Reconstruction.}
    \label{fig:mapnbv:zed_experiment}
\end{figure*}

We also conducted a real-world experiment to evaluate the feasibility of implementing MAP-NBV on hardware. A single iteration of the MAP-NBV pipeline for 2 robots was run using a ZED camera. Depth images and point clouds were captured from the ZED of a Toyota RAV4 car. The car was extracted from the point clouds and used as input for the MAP-NBV pipeline. Then, depth images and point clouds were captured from the poses MAP-NBV outputted. Due to the lower quality of the ZED camera, an iPhone 12 Pro Max with built-in LiDAR was used to capture data to create a high-resolution reconstruction shown in Figure~\ref{fig:mapnbv:zed_experiment}d.

\section{Conclusion}\label{sec:mapnbv:con}
We present a multi-agent, decentralized, prediction-guided NBV planning approach for active 3D reconstruction. This method can be helpful in a variety of applications including civil infrastructure inspection. We show that our method can faithfully reconstruct the object point clouds efficiently compared to non-predictive multi-agent methods and other prediction-guided approaches. Our NBV planning objective considers both information gain and control effort, making it more suitable for real-world deployment given the flight time limit imposed on UAVs by their battery capacity. 

We are currently working on a bandwidth-aware extension of \textit{MAP-NBV} and the preliminary studies show encouraging results (shared on our \href{https://raaslab.org/projects/MAPNBV/}{webpage}). In this work, we focus solely on geometric measures for information gain. Many existing works on NBV have developed sophisticated information theoretic measures. We will explore combining both types of measures in our future work.

%% file: Chapter6.tex
%Chapter 6

\renewcommand{\thechapter}{6}

\chapter{Spatiotemporal Pattern Prediction for Multi-Robot Coordination and Tracking}
\label{chapter:gnn_d2coplan}
In Chapters \ref{chapter:proxmap}, \ref{chapter:fliptd}, \ref{chapter:prednbv} and \ref{chapter:mapnbv}, we show structural and geometric patterns can be harnessed to improve the robot's knowledge about the map and its planning capabilities, even with few direct observations. The methods presented in these chapters focused on static objects. When the objects under consideration start moving, the robot needs to estimate \textit{when} and \textit{where} the object in motion will be in the future. This problem can be further exacerbated when multiple robots need to \textit{track} multiple targets as they must also coordinate with each other for effective planning. In this section, we show how we can leverage \textit{spatiotemporal patterns}
arising from the movement patterns of the targets and the robots' spatial arrangement can improve the tracking of multiple targets with a multi-robot team. We first show how to build a learning-based decentralized planner for coverage planning when the target's locations are known to achieve a scalable and performant planning algorithm. Then we adapt this architecture to make it differentiable and show how it can help in learning a better multi-target tracking method compared to its `plug-and-play' counterparts. 

\section{Learning Decentralized Coordination with Graph Neural Networks}

\subsection{Introduction} \label{subsec:gnn:intro}
\note{Multi-robot target tracking finds a wealth of applications in robotics. Typical examples include monitoring~\cite{tokekar2013tracking},  patrolling~\cite{zengin2007real},  surveillance~\cite{grocholsky2006cooperative}, and search and rescue~\cite{kumar2017opportunities}. Such applications ask for teams of robots that act as mobile sensors to jointly plan their actions to optimize tracking objectives (e.g., the number of tracked targets or the uncertainty reduction in the targets' positions). In this subsection, we focus on tracking objective functions that are submodular, i.e., the functions that have the diminishing returns property. } Examples of such functions include information-theoretic metrics such as entropy and mutual information~\cite{krause2008near} and geometric metrics such as the visibility region~\cite{ding2017multi}. The problems of maximizing submodular functions are generally NP-hard. The most well-known approach for tackling these problems is the greedy algorithm that runs in polynomial time and yields a constant factor approximation guarantee~\cite{nemhauser1978analysis,fisher1978analysis}. 

The greedy algorithm cannot be directly implemented in scenarios where the robots can only communicate locally due to a limited communication range. To address the issue of local communication, some decentralized versions of the greedy algorithm were designed, where only neighboring information is utilized to choose actions for the robots for optimizing submodular objectives~\cite{atanasov2015decentralized,williams2017decentralized,gharesifard2018distributed,grimsman2018impact}. For example, building on the local greedy algorithm~\cite[subsection~4]{fisher1978analysis}, Atanasov et al. designed a decentralized greedy algorithm that achieves 1/2 approximation bound for multi-sensor target tracking~\cite{atanasov2015decentralized}. Specifically, the algorithm greedily selects an action for each robot in sequential order, given all the actions selected so far.  
However, with limited communication, the robots may not have access to all the previously selected actions. To this end, a few decentralized submodular maximization algorithms were devised to execute a sequential greedy algorithm over directed acyclic graphs that may not be connected~\cite{gharesifard2018distributed,grimsman2018impact}. 
Other decentralized greedy approaches include the ones that utilize a consensus-based mechanism
to bring robots to an agreement by communicating local greedy selections with neighbors over multiple hops~\cite{williams2017decentralized,qu2019distributed}. However, these algorithms may take a considerable amount of time to reach a consensus.

\note{In this subsection, we aim to explore learning-based methods to learn policies by imitating expert algorithms~\cite{choudhury2017adaptive} (e.g., the greedy algorithm~\cite{fisher1978analysis}) for multi-robot target tracking.} Particularly, we choose the graph neural network (GNN) as the learning paradigm given its nice properties of decentralized communication architecture that captures the neighboring interactions and the transferability that allows for the generalization to previously unseen scenarios~\cite{ruiz2021graph,Gama19-Architectures}. Also, GNN has recently shown success in various multi-robot applications such as formation control~\cite{Tolstaya19-Flocking, khan2019graph}, path finding~\cite{li2019graph}, and task assignment~\cite{wang2020learning}. Specifically, 
Tolstaya et al. implemented GNN to learn a decentralized flocking controller for a swarm of mobile robots by imitating a centralized flocking controller with global information. Similarly, Li et al. applied GNN to find collision-free paths for large networks of robots from start positions to goal positions in obstacle-rich environments~\cite{li2019graph,li2020message}. This method exhibits a near-expert performance and is well-generalized to larger teams of robots. 
Their results demonstrated that, by mimicking a centralized expert solution, their decentralized path planner exhibits a near-expert performance, utilizing local observations and neighboring communication only,  which also can be well generalized to larger teams of robots.
The GNN-based approach was also investigated to learn solutions for combinatorial optimization in a multi-robot task scheduling scenario~\cite{wang2020learning}.

\note{\noindent \textbf{Contributions.} To this end, we design a GNN-based learning framework that enables robots to communicate and share information with neighbors and selects actions for the robots to optimize target tracking performance. We train such a learning network to perform as close as possible to the greedy algorithm by imitating the behavior of the greedy algorithm. Different from classical algorithms (e.g., greedy algorithms), GNN can be seamlessly integrated with other networks such as convolutional neural networks (CNN) or multilayer perceptron (MLP) to process richer data representations like images or sensor measurements. Classical algorithms may not be able to either handle such data modalities or handle them efficiently. For example, (decentralized) greedy algorithms cannot directly use the raw image data or sensor measurements for decision-making. Therefore, we devise a GNN-based learning framework that takes raw sensor measurements as inputs and leverages GNN for scalable and fast feature sharing to generate robots' actions.

Specifically, we make the following contributions:
\begin{itemize}
    \item We formulate the problem of applying GNN for multi-robot target tracking with local communications (Problem~\ref{prob:learning}); 
    \item We design a GNN-based learning framework that processes robots' local observations and aggregates neighboring information to select actions for the robots (i.e., the solution to Problem~\ref{prob:learning}).
    \item We demonstrate the performance of the GNN-based learning framework such as near-expert behavior, transferability, and fast running time in the scenario of active target tracking with large networks of robots. Specifically, our method covers around 89\% of targets compared to the expert algorithm and runs several orders faster. 
\end{itemize} }

%%%%%%%%%%%%%%%%%%%%%%%%%%%%%%%%%%%%%%%%%%%%%%%%%%%%%%%%%%%%%%%%%%%%%%%%%%%%%%
\subsection{Problem Formulation} \label{subsec:gnn:problem}
We present the problem of \note{\textit{decentralized action selection for multi-robot target tracking} (see Fig.~\ref{fig:gnn:uav_tracking})}. Particularly, at each time step, the problem asks for selecting an action for each robot to optimize a \note{target tracking objective} using local information only. Specifically, the robots' actions are their candidate motion primitives, the team objective can be the number of targets covered, and for each robot, the local information is the set of targets covered by it and its neighbors. The goal of this work is to design GNN to learn such decentralized planning for the robots by \textit{imitating} an expert algorithm. We start with introducing the framework of \note{decentralized target tracking} and GNN, and then formally define the problem. 

\subsubsection{Decentralized Target Tracking} \label{subsubsec:gnn:framework}

\begin{figure}
\centering
\includegraphics[width=0.7\columnwidth]{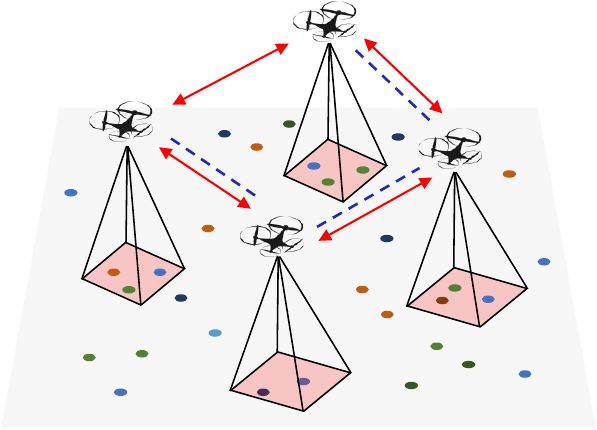}
\caption[An overview of the multi-robot target tracking setup: a team of aerial robots, mounted with down-facing cameras, aims at covering multiple targets (depicted as colorful dots) on the ground. The red arrow lines and the blue dotted lines show inter-robot observations and communications. The red squares represent the fields of view of the robots' cameras.]{Multi-robot target tracking: a team of aerial robots, mounted with down-facing cameras, aims at covering multiple targets (depicted as colorful dots) on the ground. The red arrow lines and the blue dotted lines show inter-robot observations and communications. The red squares represent the fields of view of the robots' cameras.}
\label{fig:gnn:uav_tracking}
\end{figure}

\paragraph{Robots} We consider a team of $N$ robots, denoted by $\mathcal{V}=\{1,2,\cdots, N\}$. At a given time step, the relative position between any two robots $i$ and $j$ in the environment is denoted by $\mb{p}_{ij}^\texttt{r}, i,j\in \mc{V}$. The global positions of the robots are not required.

 \begin{figure}[t]
\centering{
{\includegraphics[width=0.7\columnwidth,trim= 0cm .0cm 0 0cm,clip]{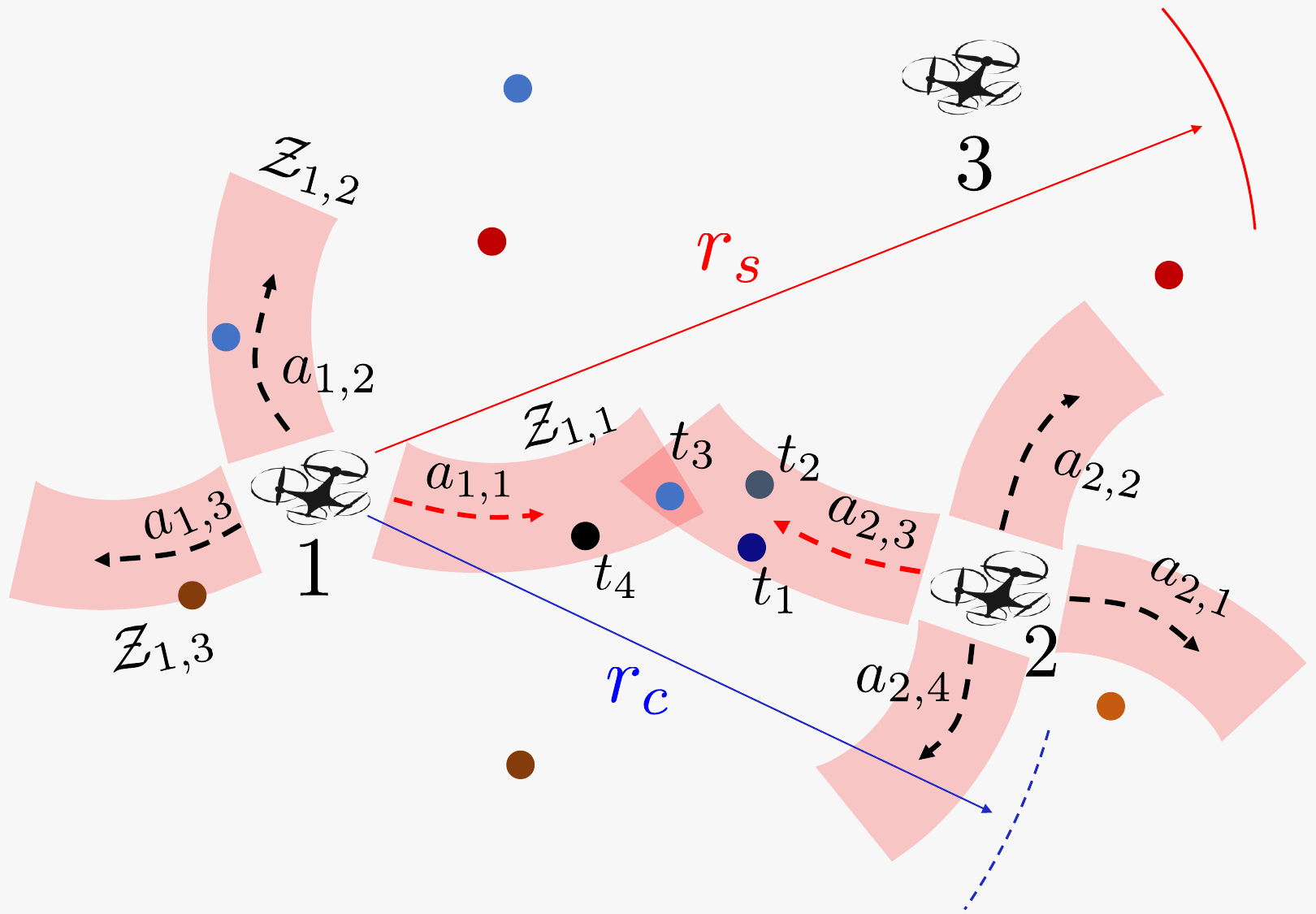}}
\caption[Overview of the decentralized target tracking problem. At a given time step, each robot observes the robots within its sensing range and chooses an action from a set of motion primitives to cover some targets using its camera. Each robot communicates with those robots within its communication range.]{Robots' observations, motion primitives, and communications: at a given time step, each robot observes the robots within its sensing range $r_s$; e.g., robot 1 observes robot 2 and robot 3.  Each robot has a set of motion primitives (depicted as dotted arrow curves), from which it can choose one to cover some targets using its camera. For example, robot 1 has 3 motion primitives, $\{a_{1,1}, a_{1,2}, a_{1,3}\}$, and it follows  $a_{1,1}$ to cover 2 targets, $\{t_3,t_4\}$. Robot 2 has 4 motion primitives, $\{a_{2,1}, a_{2,2}, a_{2,3}, a_{2,4}\}$, and it chooses $a_{2,3}$ to cover 3 targets, $\{t_1, t_2, t_3\}$. However, in combination, these two motion primitives jointly cover 4 targets, $\{t_1, t_2, t_3, t_4\}$. In addition, each robot communicates with those robots within its communication range $r_c$; e.g., robot 1 communicates with robot 2. \label{fig:gnn:uav_trajectory}}
}
\end{figure}

\paragraph{Action set} We denote a set of \textit{available actions} for each robot $i$ as $\mc{A}_i, i\in\mc{V}$. At a given time step, the robot can select at most $1$ action from its available action set; e.g., in a motion planning scenario, $\mc{A}_i$ is robot $i$'s motion primitives, and the robot can select only $1$ motion primitive to execute at each time step. For example, in Fig.~\ref{fig:gnn:uav_trajectory}, there are 2 robots, where robot 1's action set is $\mc{A}_1 = \{a_{1,1}, a_{1,2}, a_{1,3}\}$ (and robot 1 selects action $a_{1,1}$ to execute), and robot 2's action set is $\mc{A}_2 = \{a_{2,1}, a_{2,2}, a_{2,3}, a_{2,4}\}$ (and robot 2 chooses $a_{2,3}$ to execute). Denote the joint action set of all robots as $\mc{A} \triangleq \bigcup_{i\in \mathcal{V}} \mc{A}_i$. Also, denote a valid selection of actions for all robots as $\mc{U}\subseteq \mc{A}$. For example, we have $\mc{U}=\{a_{1,1}, a_{2,3}\}$ for the two robots in Fig.~\ref{fig:gnn:uav_trajectory}.

\paragraph{Observation} We consider each robot $i$ to be equipped with a sensor (e.g., a LiDAR sensor) to measure the positions relative to the robots within its sensing range (see Fig.~\ref{fig:gnn:uav_trajectory}). Without loss of generality, we assume all robots' sensors have the same \textit{sensing range} $r_s$. For each robot $i$, we denote the set of robots within its sensing range as $\mc{V}_i^s$. Then the sensor observation of each robot $i$, i.e., the relative positions between robot $i$ and $\mc{V}_i^s$, can be represented by $\mc{Z}_i^\texttt{r} = \{\mb{p}_{ij}^\texttt{r}\}_{j\in\mc{V}_i^s}$. 

In addition, each robot is mounted with a camera that perceives a part of the environment within its field of view (see Fig.~\ref{fig:gnn:uav_tracking}). Using the camera, the robot can observe some objects and measure their relative positions in the environment, once it selects an action to execute. For example, in Fig.~\ref{fig:gnn:uav_trajectory}, when robot $1$ selects motion primitive $a_{1,1}$, it can sweep and cover a set of targets $\{t_3,t_4\}$, and the corresponding observation $\mc{Z}_{1,1}$ is the relative positions to targets $\{t_3,t_4\}$. 
Notably, each action of the robot corresponds to an observation. Thus, given a time step, we denote the (\textit{possible}) camera observation\footnote{We call it as the possible observation, since the robot can select at most $1$ action to execute at a time step.} of each robot $i$ by $\mc{Z}_i^{\texttt{t}}$, which is the collection of the observations by the robot's available actions $\mc{A}_i$. For example, in Fig.~\ref{fig:gnn:uav_trajectory}, the camera observation of robot $1$  is $\mc{Z}_1^{\texttt{t}}=\{\mc{Z}_{1,1}, \mc{Z}_{1,2}, \mc{Z}_{1,3}\}$. Particularly, we denote those objects that can be covered by robot $i$ as $\mc{T}_i$ and the corresponding relative positions as $\{\mb{p}_{ij}^\texttt{t}\}_{j\in\mc{T}_i}$. Then the robot $i$'s camera observation can be represented by $\mc{Z}_i^\texttt{t} = \{\mb{p}_{ij}^\texttt{t}\}_{j\in\mc{T}_i}$.
Finally, we define the observation of each robot $i$ by $\mc{Z}_i$, which contains the observations of the sensor and camera on it, i.e.,  $\mc{Z}_i = \{\mc{Z}_i^\texttt{r}, \mc{Z}_i^{\texttt{t}}\}$.

\paragraph{Communication} Each robot $i\in\mc{V}$ communicates only with those robots within a prescribed \textit{communication range}. Without loss of generality, we consider all robots have the same \textit{communication range} $r_c$ (see Fig.~\ref{fig:gnn:uav_trajectory}). That way, we introduce an (undirected) \textit{communication graph} at a given time step as $\mc{G} = (\mc{V}, \mc{E}, \mc{W})$ with nodes the robots $\mathcal{V}$, edges $\mathcal{E} \subseteq \mc{V} \times \mc{V}$ the communication links, and weights of the edges $\mc{W}: \mc{E} \to \mathbb{R}$ denoting the strength of communications. The graph $\mc{G}$ is distance-based and $(i,j) \in \mc{E}$ if and only if $\|\mb{p}_{ij}^\texttt{r}\|_2\leq r_{c}$. We denote the  1-hop neighbors of robot $i$ by $\mathcal{N}_i$, which are the robots within the range $r_c$. We denote the adjacency matrix of graph $\mc{G}$ by $\mb{S} \in \mathbb{R}^{N \times N}$ with $[\mb{S}]_{ij} =s_{ij} = 1$ if $(i,j) \in \mc{E}$ and $0$ otherwise. Notably, the connectivity of graph $\mc{G}$ is \textit{not} required.

\paragraph{Objective function} \note{We consider a target tracking objective function $f: 2^{\mc{A}} \to \mathbb{R}$ to be monotone non-decreasing and submodular in the robots' actions $\mc{U}$.  For example, $f$ can be the number of targets covered~\cite{tokekar2014multi}. }
As shown in Fig.~\ref{fig:gnn:uav_trajectory}, the number of targets covered by the selected actions (motion primitives), $\mc{U}=\{a_{1,1}, a_{2,3}\}$, is $f(\mc{U}) = 4$.

\subsubsection{Graph Neural Networks} \label{subsubsec:gnn:gnn} 
\paragraph{Graph Shift Operation} We consider each robot $i, i\in \mc{V}$ has a feature vector $\mb{x}_i \in \mathbb{R}^F$, indicating the processed information of robot $i$. By collecting the feature vectors $\mb{x}_i$ from all robots, we have the feature matrix for the robot team $\mc{V}$ as: 
\begin{equation} \label{eqn:featureMatrix}
    \mb{X} 
    = \begin{bmatrix}
        \mb{x}_1^{\Tr} \\
        \vdots \\
        \mb{x}_N^{\Tr}
      \end{bmatrix} = [\mb{x}^1, \cdots, \mb{x}^F] \in \mathbb{R}^{N\times F},
\end{equation}
where $\mb{x}^f \in  \mathbb{R}^N, f \in [1, \cdots, F]$ is the collection of the feature $f$ across all robots $\mc{V}$; i.e., $\mb{x}^f = [\mb{x}_1^f, \cdots, \mb{x}_N^f]^{\Tr}$ with $\mb{x}_i^f$ denoting the feature $f$ of robot $i, i\in\mc{V}$. We conduct \textit{graph shift operation} for each robot $i$ by a linear combination of its neighboring features, i.e., $\sum_{j\in \mc{N}_i} \mb{x}_j$. Hence, for all robots $\mc{V}$ with graph $\mc{G}$, the feature matrix $\mb{X}$ after the shift operation becomes $\mb{S} \mb{X}$ with:   
\begin{equation} \label{eqn:graphShift}
    [\mb{S} \mb{X}]_{if} 
        = \sum_{j = 1}^{N} [\mb{S}]_{ij} [\mb{X}]_j^f
        = \sum_{j \in \mc{N}_{i}}
           s_{ij} \mb{x}_j^f, 
\end{equation}
Here, the adjacency matrix $\mb{S}$ is called the \emph{Graph Shift Operator} (GSO)~\cite{Gama19-Architectures}. 

\paragraph{Graph convolution} With the shift operation, we define the \textit{graph convolution} by a linear combination of the \textit{shifted features} on graph $\mc{G}$ via $K$-hop communication exchanges \cite{Gama19-Architectures,li2019graph}: 
\begin{equation} \label{eqn:graphConvolution}
    \mc{H}(\mb{X}; \mb{S}) = \sum_{k=0}^{K} \mb{S}^{k} \mb{X} \mb{H}_{k},
\end{equation}
where $\mb{H}_{k} \in \mathbb{R}^{F \times G}$ represents the coefficients combining $F$ features of the robots in the shifted feature matrix $\mb{S}^{k} \mb{X}$, with $F$ and $G$ denoting the input and output dimensions of the graph convolution. Note that, $\mb{S}^{k} \mb{X} = \mb{S}(\mb{S}^{k-1} \mb{X}) $ is computed by means of $k$ communication exchanges with $1$-hop neighbors. 

\paragraph{Graph neural network} Applying a point-wise non-linearity $\sigma: \mathbb{R} \to \mathbb{R}$ as the activation function to the graph convolution (eq.~\eqref{eqn:graphConvolution}), we define \textit{graph perception} as: 
\begin{equation} \label{eqn:graphPerception}
    \mc{H}(\mb{X}; \mb{S}) = \sigma(\sum_{k=0}^{K} \mb{S}^{k} \mb{X} \mb{H}_{k}).
\end{equation}

Then, we define a GNN module by cascading $L$ layers of graph perceptions (eq.~\eqref{eqn:graphPerception}):
\begin{equation} \label{eqn:convGNN}
    \mb{X}^{\ell} = \sigma \big[ \mc{H}^{\ell}(\mb{X}^{\ell-1};\mb{S}) \big] \quad \text{for} \quad \ell = 1,\cdots,L,
\end{equation}
where the output feature of the previous layer $\ell-1$, $\mb{X}^{\ell-1} \in \reals^{N \times F^{\ell-1}}$, is taken as input to the current layer $\ell$ to generate the output feature of layer $l$, $\mb{X}^{\ell}$. Recall that the input to the first layer is $\mb{X}^{0} = \mb{X}$ (eq.~\eqref{eqn:featureMatrix}). 
The output feature of the last layer $\mb{X}^{L} \in \mathbb{R}^{N \times G}$, obtained via $K$-hop communications and multi-layer perceptions, will be used to predict action set $\mc{U}$ for all robots to the following problem.

Notably, GNN can represent rich classes of mappings (functions) from the input feature to the output feature. Since the graph convolution is a linear operator (eq.~\eqref{eqn:graphConvolution}), the function's characteristics mainly depend on the property of the activation function $\sigma$ (in eq.~\eqref{eqn:convGNN}). For example, if the activation function is concave (or submodular), the function represented by GNN is also concave (or submodular).

\subsubsection{Problem Definition}~\label{subsubsec:gnn:prob_definition}
\begin{problem*}[\note{Decentralized action selection for multi-robot target tracking}]\label{pro:dis_resi_sub}
At each time step, the robots $\mc{V}$, by exchanging information with neighbors only over the communication graph $\mc{G}$, select an action to each robot $i \in \mathcal{V}$ to maximize \note{a submodular target tracking objective function} $f$:
\begin{align} \label{eq:dis_resi_eq}
\begin{split}
& \max_{\mc{U}\subseteq \mc{A}}\;\; f(\mc{U})\\
& \emph{\text{s.t. }}\; |\mc{U}\cap \mc{A}_i|= 1, ~\text{for all } i\in \mathcal{V}.\\
\end{split}
\end{align}
The constraint follows a partition matroid constraint to ensure that each robot selects $1$ action per time step (e.g., $1$ motion primitive among a set of motion primitives). 
\end{problem*}

Eq.~\eqref{eq:dis_resi_eq} can be interpreted as a submodular maximization problem with a partition matroid constraint and decentralized communication. This problem is generally NP-hard even if we assume the centralized communication~\cite{fisher1978analysis}. That is because finding the \textit{optimal} action set requires to exhaustive search and evaluate the quality of \textit{all possible} valid action sets $\mc{U} \subseteq \mc{A}$. Clearly, this exhaustive search method has combinatorial complexity and quickly becomes intractable as either the number of robots or the number of available actions of the robots increase. The most well-known approach for tackling this type of problem (with centralized communication) is the (centralized) greedy algorithm~\cite{fisher1978analysis}. The advantage of the centralized greedy algorithm is two-fold: (i), it is efficient as it runs in polynomial time (with $O(|\mc{A}|^2)$ complexity); (ii), it achieves at least $1/2$--approximation of the optimal. In addition, since the $1/2$--approximation bound is computed based on the worst-case performance, the centralized greedy algorithm can typically perform much better (on average) in practice; e.g., it performs close to the exhaustive search (see Figure~\ref{fig:gnn:all_scale}-c in subsection~\ref{subsubsec:gnn:evalau}). Since the centralized greedy algorithm is much cheaper than the exhaustive search and performs comparatively, we use the centralized greedy algorithm as our \textit{expert} algorithm to generate ground-truth training data.  

Particularly, we aim to train a learning network to perform as well as the centralized greedy algorithm for solving Eq.~\eqref{eq:dis_resi_eq}, while involving the communications among neighboring robots only. For such a learning network, GNN can be a great fit given its decentralized communication protocol. Hence, the goal is to utilize GNN to learn an action set $\mc{U}$ to Eq.~\eqref{eq:dis_resi_eq} by imitating the action set selected by the centralized greedy algorithm. More formally, we define the problem of this work as follows. 

\begin{figure*}
\centering
\includegraphics[width=1.0\textwidth]{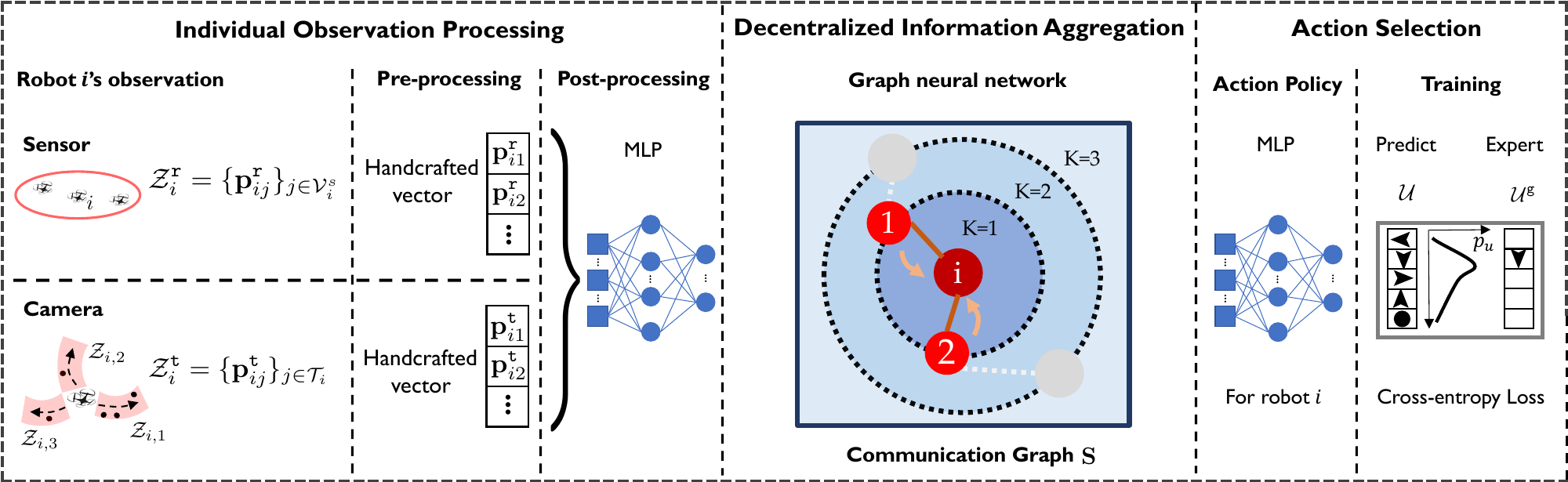}
\caption[Overview of our learning-based planner for decentralized target tracking. It consists of three main modules: (i), the Individual Observation Processing module processes the local observations and generates a feature vector for each robot; (ii), the Decentralized Information Aggregation module utilizes the GNN to aggregate and fuse the features from $K$-hop neighbors for each robot; (iii), the Decentralized Action Selection module selects an action for each robot by imitating an expert algorithm.]{Our decentralized learning architecture includes three main modules: (i), the Individual Observation Processing module processes the local observations and generates a feature vector for each robot; (ii), the Decentralized Information Aggregation module utilizes a GNN to aggregate and fuse the features from $K$-hop neighbors for each robot; (iii), the Decentralized Action Selection module selects an action for each robot by imitating an expert algorithm.     
\label{fig:gnn:learn_arhitect}}
\end{figure*}

\begin{problem}
Design a GNN-based learning framework to learn a mapping $\mc{M}$ from the robots' observations $\{\mc{Z}_i\}_{i\in\mc{V}}$ and the communication graph $\mc{G}$ to the robots' action set $\mc{U}$, i.e., $\mc{U} = \mc{M}(\{\mc{Z}_i\}_{i\in\mc{V}}, \mc{G})$, such that $\mc{U}$ is as close as possible to action set selected by the centralized greedy algorithm, denoted by $\mc{U}^\texttt{g}$.  
\label{prob:learning}
\end{problem}

We describe in detail our learning architecture for solving Problem~\ref{prob:learning} in the next subsubsection.  

%%%%%%%%%%%%%%%%%%%%%%%%%%%%%%%%%%%%%%%%%%%%%%%%%%%%%%%%%%%%%%%%%%%%%%%%%%%%%%

\subsection{Proposed Approach: GNN-based Decentralized Coverage Planner} \label{subsec:gnn:archit}
We design a learning architecture that consists of three main components---the Individual Observation Processing (Subsection~\ref{subsubsec:gnn:info_process}), the Decentralized Information Aggregation (Subsection~\ref{subsubsec:gnn:info_aggre}), and the Decentralized Action Selection module (Subsection~\ref{subsubsec:gnn:action_select}).
Next, we describe in detail these key components, illustrated in Fig.~\ref{fig:gnn:learn_arhitect}, as follows.

\subsubsection{Individual Observation Processing} \label{subsubsec:gnn:info_process}
Recall that each robot $i$'s observation $\mc{Z}_i$ includes its sensor
observation $\mc{Z}_i^\texttt{r}$ and its camera observation $\mc{Z}_i^\texttt{t}$ (see Subsection~\ref{subsubsec:gnn:framework}). We process the observation $\mc{Z}_i$ to generate a feature vector $\mb{x}_i$ for each robot $i$ in two steps. 
\begin{itemize}
    \item Step 1 is a pre-processing step. Since the sensor
    observation $\mc{Z}_i^\texttt{r}$ stores the relative robot positions $\{\mb{p}_{ij}^\texttt{r}\}_{j\in\mc{V}_i^s}$, we reshape (handcraft) it as a vector $\mb{x}_{i,1}^{-}: = [\mb{p}_{i1}^{\Tr}, \cdots, \mb{p}_{i|\mc{V}_i^s|}^{\Tr}]^{\Tr}$. Similarly, for the camera observation $\mc{Z}_i^\texttt{t}$ that contains the relative positions to the objects covered, $\{\mb{p}_{ij}^\texttt{t}\}_{j\in\mc{T}_i}$, we reshape it as vector $\mb{x}_{i,2}^{-}: = [\mb{p}_{i1}^{\Tr}, \cdots, \mb{p}_{i|\mc{T}_i|}^{\Tr}]^{\Tr}$.   
    Finally, by concatenating $\mb{x}_{i,1}^{-}$ and $\mb{x}_{i,2}^{-}$, we generate a pre-processed feature vector $\mb{x}_{i}^{-}$ for robot $i$, i.e., $\mb{x}_{i}^{-} = [(\mb{x}_{i,1}^{-})^{\Tr}, (\mb{x}_{i,2}^{-})^{\Tr}]^{\Tr}$. \footnote{We assign the same dimension to feature vector $\mb{x}_{i}^{-}$ for all the robots. In particular, for each robot $i$, we use relative positions of $10$ nearest robots in $\{\mb{p}_{ij}^\texttt{r}\}_{j\in\mc{V}_i^s}$ and relative positions of $20$ nearest targets in $\{\mb{p}_{ij}^\texttt{t}\}_{j\in\mc{T}_i}$. If robot $i$ has less than 10 nearby robots measured or covers less than 20 targets, we simply add some dummy values (e.g., -1) in the feature vector to maintain the same dimension.}
    \item Step 2 is a post-processing step where the pre-processed feature vector $\mb{x}_{i}^{-}$ is fed into a multi-layer perceptron (MLP) module to generate the robot's feature vector $\mb{x}_i$, i.e., $\mb{x}_i= \text{MLP}(\mb{x}_{i}^{-})$.    
\end{itemize}
The feature vectors of the robots are then exchanged and fused through neighboring communications (Subsection~\ref{subsubsec:gnn:info_aggre}).  

\subsubsection{Decentralized Information Aggregation} \label{subsubsec:gnn:info_aggre}
Each robot $i$ communicates its feature (or processed information) with its neighbors $\mc{N}_i$ over multiple communication hops. As shown in Subsubsection~\ref{subsubsec:gnn:gnn}, for each robot $i$, we use GNN to aggregate and fuse the feature vectors through $K$-hop communication exchanges among neighbors (eq.~\eqref{eqn:graphConvolution}). Thus, the output of the GNN (i.e., $\mb{X}^{L}$ in eq.~\eqref{eqn:convGNN}) is a hyper-representation of the fused information of the robots and their $K$-hop neighbors. The output is then taken as input to the action selection module, described in Subsection~\ref{subsubsec:gnn:action_select}, to generate an action for each robot. Notably, since only neighboring information is exchanged and fused, GNN renders a decentralized decision-making architecture.     

\subsubsection{Decentralized Action Selection} \label{subsubsec:gnn:action_select}
We aim at selecting an action set $\mc{U}$ for the robots $\mc{V}$ ($1$ action per robot) to maximize the \note{team tracking performance} $f(\mc{U})$. To this end, we use an MLP for each robot $i$ to train an action selection module. More specifically, each robot applies an MLP that takes the aggregated features as input and selects an action for the robot as output. We consider all robots to carry the same MLP, resembling a weight-sharing scheme.  
The actions of the robots are selected based on a supervised learning approach, as in Subsection~\ref{subsubsec:gnn:imitation_learn}.

\subsubsection{Supervised Learning} \label{subsubsec:gnn:imitation_learn}
We train our learning architecture by a supervised learning approach, i.e., to mimic an expert algorithm (the centralized greedy algorithm). Specifically, during the training stage, we have access to the action set $\mc{U}^\texttt{g}$ selected by the centralized greedy algorithm, the corresponding observations on the robots  $\{\mc{Z}_i\}_{i\in\mc{V}}$, and the corresponding communication graph $\mc{G}$. Thus, the training set $\mc{D}$ can be constructed as a collection of these data, i.e., $\mc{D}:= \{(\{\mc{Z}_i\}_{i\in\mc{V}}, \mc{G}, \mc{U}^{\texttt{g}})\}$. Over the training set $\mc{D}$, we train the mapping $\mc{M}$ (defined in Problem~\ref{prob:learning}) so that a cross-entropy loss $\mc{L}(\cdot, \cdot)$, representing the difference between the output action set $\mc{U}$ and the greedy action set $\mc{U}^{\texttt{g}}$ is minimized. That is, 
\begin{equation}
    \min_{\text{PM}, ~\{\mb{H}_{\ell, k}\}, ~\textbf{MLP}}~ \sum_{(\{\mc{Z}_i\}_{i\in\mc{V}}, ~\mc{G}, ~\mc{U}^{\texttt{g}}) \in \mc{D}} \mc{L}(\mc{M}(\{\mc{Z}_i\}_{i\in\mc{V}}, \mc{G}), \mc{U}^{\texttt{g}}),
    \label{eqn:cross_entropy}
\end{equation}
where we optimize over the learnable parameters of the processing method, named PM (e.g., CNN, MLP) to process the robots' observations, the set of learnable parameter matrices\\ $\{\mb{H}_{\ell, k}\}_{l\in \{1,\cdots,L\}, ~k\in\{0,\cdots,K\}}$ in GNN to aggregate and fuse the neighboring information over multiple communication hops, and the learnable parameters of MLP to select actions for the robots. 

Notably, by the decentralized learning architecture where the parametrization is operated locally on each robot, the number of learnable parameters is independent of the number of the robots $N$. This decentralized parametrization offers a perfect complement to supervised learning which requires the availability of expert solutions that can be costly in large-scale settings. For example, even though the centralized greedy algorithm is much more efficient than the exhaustive search, it still takes considerable time to generate a solution when the number of robots or their actions is large, given its running time grows quadratically in the number of robots' actions (i.e., with $O(|\mc{A}|^2)$ complexity). However, leveraging the decentralized parametrization, we only need to train over the small-scale cases and generalize the trained models such as the distributed computation (i.e., PM and MLP) and the neighboring information exchange (i.e., GNN) to the larger-scale settings, as it will be demonstrated in Subsection~\ref{subsubsec:gnn:evalau}. In other words, once trained, the learned models can be implemented in other cases, including those with different communication graphs and varying numbers of robots. 

%%%%%%%%%%%%%%%%%%%%%%%%%%%%%%%%%%%%%%%%%%%%%%%%%%%%%%%%%%%%%%%%%%%%%%%%%%%%%%

\subsection{Experiments and Evaluation}
~\label{subsec:gnn:simulation}
We present the evaluations of our method in scenarios of \textit{active target tracking with large networks of robots}. In particular, we compare our method with other baseline algorithms in terms of both running time and tracking quality.
In these comparisons, we test the trained learning models in the cases with various team sizes. 
The code of implementation is available online.\footnote{\url{https://github.com/VishnuDuttSharma/deep-multirobot-task}} The experiments are conducted using a 32-core, 2.10Ghz Xeon Silver-4208 CPU and an Nvidia GeForce RTX 2080Ti GPU with 156GB and 11GB of memory, respectively. 

Next, we first describe the active target tracking scenario, the specifications of the learning architecture, and the compared baseline algorithms. Then we present the evaluations.

\subsubsection{Multi-Robot Active Target Tracking}
We consider $N$ aerial robots that are tasked to track $M$ \note{mobile targets on the ground}. Each robot uses its sensor to obtain the relative positions to those robots that are within its sensing range $r_s$. 
The sensing range $r_s$ is set to be $r_s = 20$ units. 
   
The camera on each robot $i$ has a square field of view $d_o \times d_o$, and each robot $i$ has $5$ available motion primitives, $\mc{A}_i = \{\texttt{forward, backward,}$ $\texttt{left, right, idle}\}$\footnote{\note{The designed learning architecture is generic and can be extended to handle other action spaces (e.g., robots have different action sets as shown in Figure~\ref{fig:gnn:uav_trajectory}) or robot
trajectories as long as the centralized expert algorithm it learns from operates with the corresponding action spaces or robot
trajectories.}}.  Once the robot selects a motion primitive from $\mc{A}_i\setminus\texttt{idle}$, it flies a distance $d_m$ along that motion primitive. If the robot selects the \texttt{idle} motion primitive, it stays still (i.e., $d_m = 0$). Hence, each motion primitive corresponds to a rectangular tracking region with length $d_m+d_o$ and width $d_o$. The tracking width (or the side of the field of view) is set to be $d_o = 6$ units. The flying length $d_m$ is set to be $d_m = 20$ units for all robots selecting the non-\texttt{idle} motion primitive.

We set the robot's communication range as $r_c = 10$ units and the communication hop as $K=1$ (i.e., each robot $i$ communicates only with its $1$-hop neighbors $\mc{N}_i$). \note{We let robots fly at different heights so that collisions do not occur during their movement.} 
The objective function is considered to be the number of targets covered, given all robots selecting motion primitives.  

\subsubsection{Supervised Learning Specification}
We apply the centralized greedy algorithm~\cite{fisher1978analysis} as the expert algorithm to generate a ground-truth data set. In each problem instance, the size of the environment, a square, depends on the number of robots, i.e., its side length is specified as $100 \times \sqrt{N/20}$. We randomly generate the positions of the robots and targets in the environment and utilize the centralized greedy algorithm to select an action set for the robots. Notably, each instance includes the robots' observations $\{\mc{Z}_i\}_{i\in\mc{V}}$, the communication graph $\mc{G}$ (represented by its adjacency matrix $\mb{S}$), and the greedy (or ground-truth) action set $\mc{U}^\texttt{g}$. The ground-truth data set comprises $120,000$ instances for varying numbers of robots and the corresponding environments. In particular, we scale the size of the environment proportionally to the number of robots but keep the target density the same. Here, the target density is captured by the percentage of the cells in the grid occupied by the targets. We set the target density as 2.5\%  in all cases. 

The robot's observations include the relative positions of the 10 nearest robots within its sensing range and the 20 nearest targets that can be covered. \note{We experimentally found these hyperparameters such as the target density and the numbers of nearest robots and targets are suitable across different environmental scales.} The data is randomly shuffled at training time and divided into a training set (60\%), a validation set (20\%), and a testing set (20\%). 

Our learning architecture consists of a 3-layer MLP with 32, 16, and 8 hidden layers as the Individual Observation Processing module, a 2-layer GNN with 32 and 128 hidden layers as the Decentralized Information Aggregation module, and a single-layer MLP as the Decentralized Action Selection module. \note{For each robot, the network outputs a probability of selecting each action in the action set. All the robots have identical action sets}. This learning network is implemented in PyTorch v1.6.0 and accelerated with CUDA v10.1 APIs. We use a learning rate scheduler with cosine annealing to decay the learning rate from $5\times10^{-3}$ to $10^{-6}$ over 1500 epochs with batch size 64. This architecture and training parameters are selected from multiple parameter search experiments.

\subsubsection{Compared Algorithms}
We compare our method, named \texttt{GNN} with three other algorithms. The algorithms differ in how they select the robots' motion primitives. The first algorithm is the centralized greedy algorithm (the expert algorithm), named \texttt{Centrl-gre}. The subsubsecond algorithm is an optimal algorithm, named \texttt{Opt} which attains the optimal solution for eq.~\eqref{eq:dis_resi_eq} by exhaustive search. Particularly, for $N$ robots, each with $5$ motion primitives, \texttt{Opt} needs to evaluate $5^N$ possible cases to find the optimal solution.  Evidently, \texttt{Opt} is viable only for small-scale cases. Hence, \texttt{Opt} is used for comparison only when the number of robots is small (e.g., $N\leq10$). The third algorithm is a random algorithm, named \texttt{Rand}, which randomly (uniformly) selects one motion primitive for each robot.
The fourth algorithm is a decentralized greedy algorithm, named \texttt{Decent-gre}, which applies the standard greedy algorithm~\cite{nemhauser1978analysis} to select an action with the maximal marginal gain for each robot among the robot and its \note{1-hop} neighbors\footnote{\note{Notably, \texttt{Decent-gre} is different from the distributed (sequentially) greedy algorithms in~\cite{gharesifard2018distributed,grimsman2018impact} where each robot selects an action based on the actions of its previous neighbors (i.e., its neighbors that have already selected actions). This is to ensure \texttt{Decent-gre} and \texttt{GNN} are compared with the same communication setting.}}.
With the same settings, these four algorithms are compared in terms of the \note{team's tracking quality, i.e., the number of targets covered by all robots}, and the running time, across $1000$  trials. \note{Notably, the running time for the \texttt{Decent-gre} is calculated as the maximum time taken by any robot. For GNN, it is the inference time on GPU.}

\begin{figure*}[t]
\centering
\begin{subfigure}[c]{.48\textwidth}
    \includegraphics[width=1.00\columnwidth]{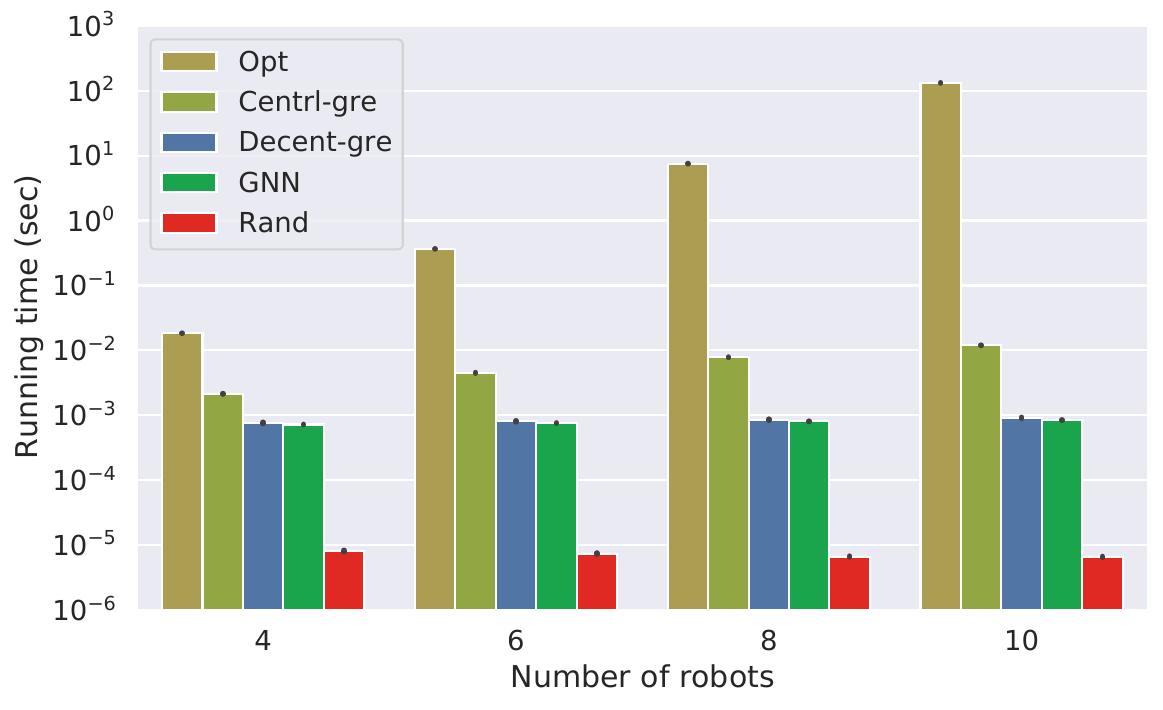}
    \caption{Small-scale comparison}
\end{subfigure}
\begin{subfigure}[c]{.48\textwidth}
    \includegraphics[width=1.00\columnwidth]{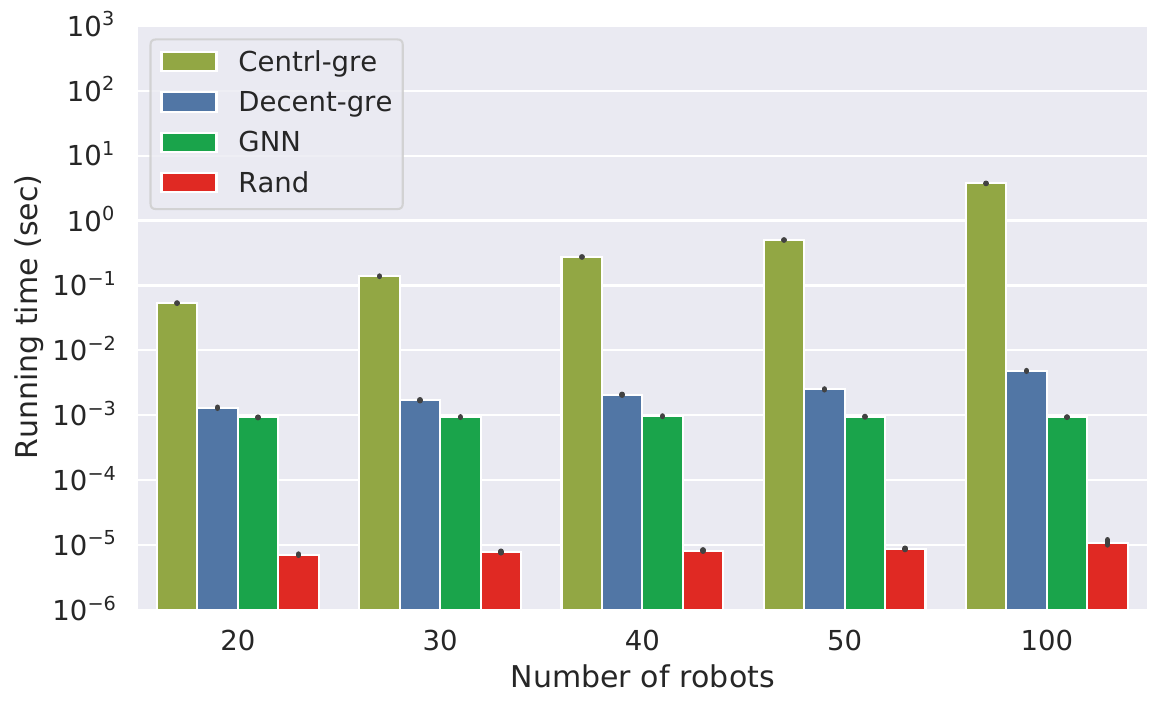}
    \caption{Large-scale comparison}
\end{subfigure}

\begin{subfigure}[c]{.48\textwidth}
    \includegraphics[width=1.00\columnwidth]{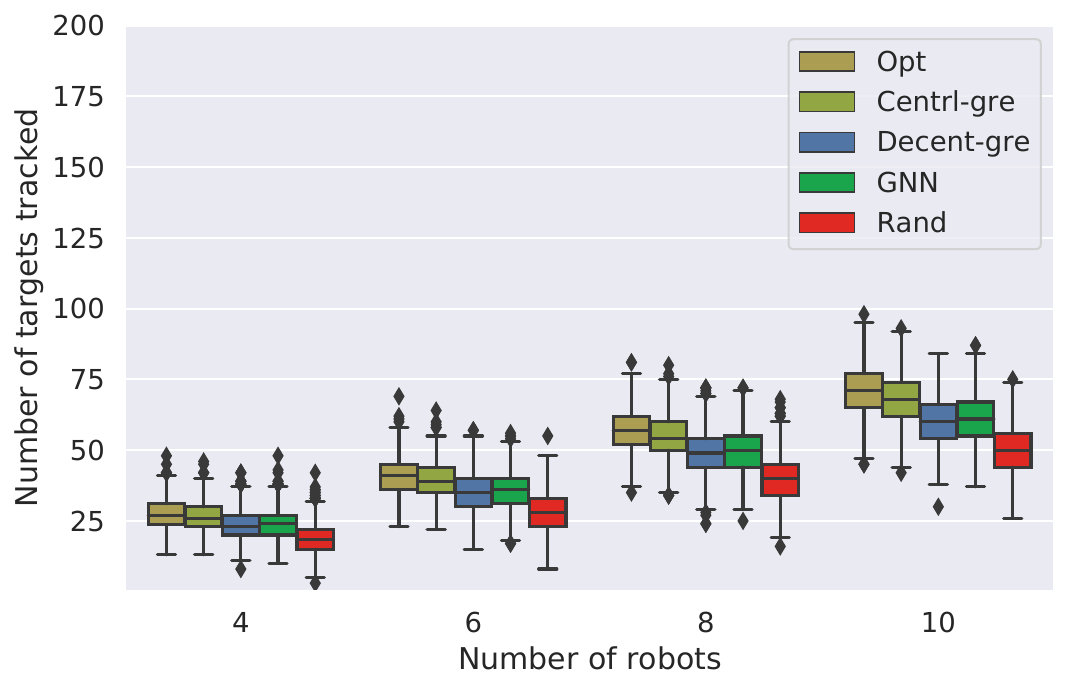}
    \caption{Small-scale comparison}
\end{subfigure}
\begin{subfigure}[c]{.48\textwidth}
    \includegraphics[width=1.00\columnwidth]{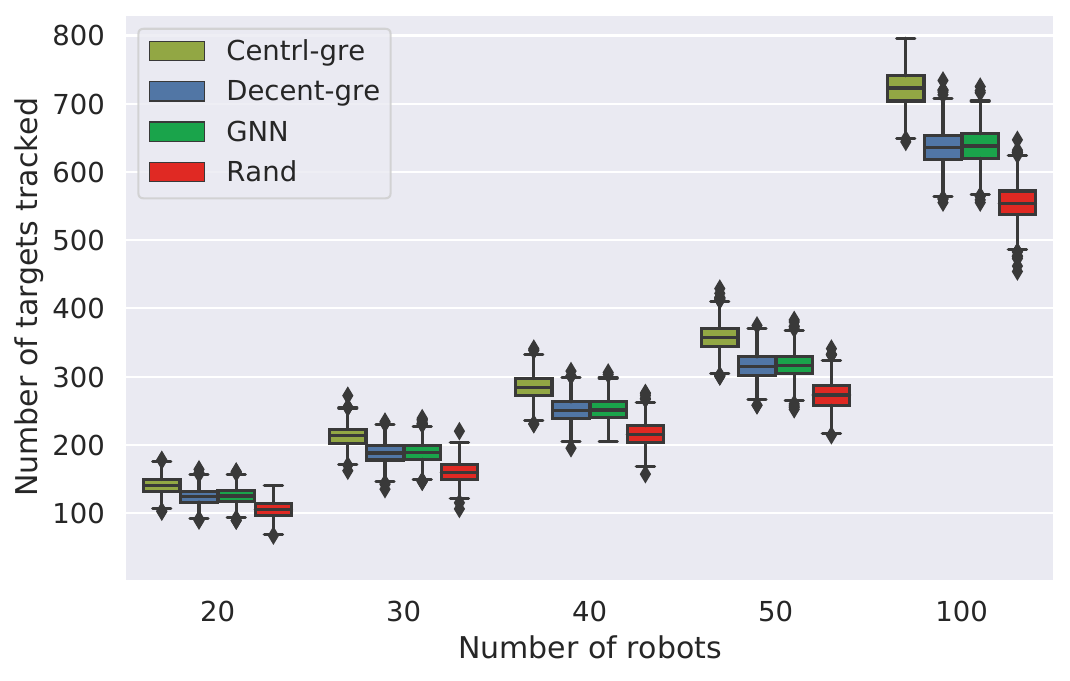}
    \caption{Large-scale comparison}
\end{subfigure}
\caption{Comparison of \texttt{Opt}, \texttt{Centrl-gre}, \texttt{Decent-gre}, \texttt{GNN}, and \texttt{Rand} in terms of running time (plotted in $\log$ scale) and the number of targets covered. (a) \& (d) are for small-scale comparison averaged across 1000 Monte Carlo trials. (b) \& (e) are for large-scale comparison averaged across 1000 Monte Carlo trials. }
\label{fig:gnn:all_scale}
\end{figure*}

\subsubsection{Evaluations} \label{subsubsec:gnn:evalau}
\paragraph{Small-scale comparison} We first compare \texttt{GNN} with \texttt{Opt}, \texttt{Centrl-gre}, \texttt{Decent-gre} and \texttt{Rand} in small-scale cases, as shown in Figures~\ref{fig:gnn:all_scale}-(a) \& (c). Since \texttt{Opt} is only feasible for small-scale scenarios, we set the number of robots $N=4, 6, 8, 10$. Particularly, we train \texttt{GNN} with the number of robots $N=20$ and test its performance over $N=4, 6, 8, 10$.

We observe \texttt{GNN} has a superior running time as shown in Figure~\ref{fig:gnn:all_scale}-(a): it runs considerably faster than both \texttt{Centrl-gre} and \texttt{Opt}: around 0.5 order faster than the former and 1.5 orders faster than the latter with $4$ robots. \note{It also runs slightly faster than \texttt{Decent-gre}}\footnote{\note{The averaged running times of \texttt{Opt}, \texttt{Centrl-gre}, \texttt{Decent-gre}, and \texttt{GNN} with 4 robots are 18ms, 2.14ms, 0.77ms, and 0.72ms, respectively.}}. This superiority becomes more significant as the number of robots increases. In addition, \texttt{GNN} has an average running time of less than 1 ms, regardless of the number of robots, which is due to its decentralized decision-making protocol. Despite the faster running time, \texttt{GNN} retains a tracking performance \note{close to \texttt{Centrl-gre} and \texttt{Opt}, slightly better than \texttt{Decent-gre}, and better than \texttt{Rand}: it covers in average more than 89\% of the number of targets covered by \texttt{Centrl-gre} and more than 85\% of that by \texttt{Opt}, and covers more targets than both \texttt{Decent-gre} and \texttt{Rand},} as shown in Figure~\ref{fig:gnn:all_scale}-(c)\footnote{\note{In the small-scale experiment, the averaged number of the targets covered for \texttt{Opt}, \texttt{Centrl-gre}, \texttt{Decent-gre}, \texttt{GNN}, and \texttt{Rand} are 49.12, 47.21, 42, 42.67, and 34.23, respectively.}}.

Figures~\ref{fig:gnn:all_scale}-(a) \& (c) also demonstrate the generalization capability of \texttt{GNN} in smaller-scale scenarios: even though it is trained with $20$ robots, it maintains both fast running time and the tracking performance close to  \texttt{Centrl-gre} and \texttt{Opt} with smaller number of robots, e.g., $N=4, 6, 8, 10$. Another interesting observation is that \texttt{Centrl-gre} covers a similar number of targets as \texttt{Opt}, and yet runs several orders faster when $N\geq 8$. This demonstrates the rationality of choosing \texttt{Centrl-gre} as the expert algorithm in this target tracking scenario.

\paragraph{Large-scale comparison} We compare \texttt{GNN} with \texttt{Centrl-gre}, \texttt{Decent-gre}, 
and \texttt{Rand} in the large-scale scenarios where the number of robots is set as $N=20, 30, 40, 50, 100$. \texttt{Opt} is not included in this comparison due to its long evaluation time. For example, it takes almost two days to evaluate $1000$ instances, each with $5^{10}$ possible cases, for 10 robots. 
\texttt{GNN} is trained with the number of robots $N = 20$ and is tested over $N=20, 30, 40, 50, 100$. The results are reported in Figures~\ref{fig:gnn:all_scale}-(b) \& (d). Similarly, we observe \texttt{GNN} runs around 1.5 to 2.5 orders faster than \texttt{Centrl-gre} with the running time less than 1 ms for all $N=20, 30, 40, 50, 100$ (Figures~\ref{fig:gnn:all_scale}-(b)). 
It also runs \note{faster than \texttt{Decent-gre}, which runs slower as the number of robots increases. 
Although \texttt{GNN} runs faster, it achieves a tracking performance close to \texttt{Centrl-gre}, slightly better than \texttt{Decent-gre}, and better than \texttt{Rand} (Figure~\ref{fig:gnn:all_scale}-(d)).} Additionally, these results verify \texttt{GNN}'s generalization capability in larger-scale scenarios: it is trained with $20$ robots, and yet, can be well generalized to a larger number of robots, e.g., $N=30, 40, 50, 100$.

\begin{table}
\centering
{\renewcommand{\arraystretch}{1.5}
\begin{tabular}{|l|*{4}{c|}}\hline
\backslashbox{Train}{Test}
&\makebox[2em]{20}&\makebox[2em]{30}&\makebox[2em]{40} &\makebox[2em]{50}\\\hline
\makebox[6em]{20} &89.34\% &88.98\% &88.56\% &88.61\%\\\hline
\makebox[6em]{30} &89.45\% &88.93\% &88.70\% &88.66\%\\\hline
\makebox[6em]{40} &89.33\% &88.78\% &88.54\% &88.62\%\\\hline
\makebox[6em]{50} &89.38\% &88.87\% &88.57\% &88.72\%\\\hline
\end{tabular}}
\caption{Percentage of the number of targets covered (the average across 1000 trials) by \texttt{GNN}trained and tested with varying numbers of robots.}
\label{tab:gnn:train_test}
\end{table}

\paragraph{Generalization evaluation} We further verify \texttt{GNN}'s generalization capability by evaluating the tracking quality of \texttt{GNN} trained and tested with varying numbers of robots. Specifically, we train \texttt{GNN} with $N = 20,30,40, 50$ robots and test it on $N = 20, 30, 40, 50$ robots. The evaluation results are reported in Table~\ref{tab:gnn:train_test} where the tracking quality is captured by the percentage of the number of targets covered with respect to the number of the targets covered by \texttt{Centrl-gre}.  
We observe that \texttt{GNN} trained and tested with the \textit{same} and \textit{different} number of robots cover a similar percentage of the targets (around 89\%), which further demonstrates the generalization capability of \texttt{GNN}. 

To summarize, in the evaluations above, \texttt{GNN} provides a significant computational speed-up, and, yet, still attains a target tracking quality that nearly matches that of \texttt{Centrl-gre} and \texttt{Opt}. \note{Moreover, \texttt{GNN} achieves a slightly better tracking quality than \texttt{Decent-gre} but runs faster (especially with more than $20$ robots) and thus scales better. In addition, \texttt{GNN} has a better tracking quality than \texttt{Rand}.} Further, \texttt{GNN} exhibits the capability of being able to well generalize to previously unseen scenarios. Particularly, it can be trained in a smaller-scale environment, which typically has a cheaper computational overhead. Then the trained policies can be applied to larger-scale environments.

\subsection{Conclusion} ~\label{subsec:gnn:conclusion}
We worked towards choosing actions for the robots with local communications to maximize a team's tracking quality. 
We devised a supervised learning approach that selects actions for the robots by imitating an expert solution. 
Particularly, we designed a GNN-based learning network that maps the robots' individual observations and inter-robot communications to the robots' actions. We demonstrated the near-expert performance, generalization capability, and fast running time of the proposed approach.

This work opens up a number of future research avenues. 
\note{An ongoing work is to extend the designed GNN-based learning architecture in a distributed fashion and analyze the number of messages shared and communication costs. In addition, we will incorporate the attention mechanism~\cite{vaswani2017attention} that 
enables robots to learn when to communicate~\cite{liu2020when2com}, who to communicate with~\cite{liu2020who2com}, and what to communicate~\cite{li2020message} to prioritize the information with higher contributions and reduce communication costs}. 
A second research avenue is to learn resilient coordination that secures team performance against either the malicious team members~\cite{blumenkamp2020emergence} 
or adversarial outsiders~\cite{zhou2018resilient,zhou2020distributed}. A third research direction is to explore decentralized reinforcement learning methods~\cite{zhang2018fully,omidshafiei2017deep} for multi-robot target tracking. 
Further, our learning-based approach can be extended to data-based learning for more complicated tasks that do not have well-performing, decentralized solutions yet~\cite{sharma2022d2coplan}. The model could also be used to bootstrap the GNN-based planner for such tasks, which cannot be done using classical counterparts~\cite{sharma2022d2coplan}.

%%%%%%%%%%%%%%%%%%%%%%%%%%%%%%%%%%%%%%%%%%%%%%%%%%%%%%%%%%%%%%%%%%%%%%%
%%%%%%%%%%%%% D2CoPlan %%%%%%%%%%%%%%%%%%%%%%%%%%%%%%%%%%%%%%%%%%%%%%%%
%%%%%%%%%%%%%%%%%%%%%%%%%%%%%%%%%%%%%%%%%%%%%%%%%%%%%%%%%%%%%%%%%%%%%%%

%Chapter 7

\section{Learning to Track and Coordinate with Differentiable Planner}

%%%%%%%%%%%%%%%%%%%%%%%%%%%%%%%%%%%%%%%%%%%%%%%%%%%%%%%%%%%%%%%%%%%%%%%%%%%%%%%%
\subsection{Introduction}
\cleanfootnote{The work presented here is a result of equal contribution from Vishnu Dutt Sharma and Dr. Lifeng Zhou. Further details and results for this work are available at \url{http://raaslab.org/projects/d2coplan/}.}

As stated in the previous subsection, multi-robot coverage and tracking is a challenging problem for many reasons. Coordination amongst the robots is critical as you want to avoid overlap and maximize the coverage. This is easier in a centralized setting; however, our focus is on \emph{decentralized} strategies where the robots can communicate directly only with their immediate neighbors. Decentralization is also harder since each robot only knows of the targets in their own fields of view. Finally, since we need to cover mobile targets, we need to predict their motion over the planning horizon. However, the motion model of the targets itself may be unknown making the problem even more challenging.

Building upon the method in the previous subsection, we investigate the question: \emph{Can the robots \emph{learn} to plan and coordinate in a decentralized fashion for target coverage problems?} Recently, there has been significant work on learning-based approaches to multi-robot planning. However, most of this work is restricted to coordination for pathfinding (where each robot needs to find the shortest path to its own goal position in an unknown environment)~\cite{gama2020graphs,li2020graph,li2021message} and formation control (such as flocking)~\cite{Tolstaya19-Flocking,khan2019graph}. We build on this to study a more complex task that requires planning, coordination, and prediction.

Our contribution is a decentralized, differentiable coverage planner (\DTWOCOPLAN{}) for multi-robot teams. \DTWOCOPLAN{} consists of three differentiable modules, namely map encoder, decentralized \VS{feature} aggregator, and local action selector. The input to \DTWOCOPLAN{} is a coverage map that represents predictions of where the targets are going to be in the next time step. This map comes from another differentiable module we call Differentiable Map Predictor (\DMP{}). The map encoder takes the predicted maps and turns them into a compact representation which is shared with the other agents using a Graph Neural Network (GNN)~\cite{scarselli2008graph}. The GNN aggregates information from neighboring agents and uses that for selecting the ego robot's action. \DTWOCOPLAN{} is trained on an expert strategy (a centralized optimal algorithm that has global information) but is executed in a decentralized fashion (following the Centralized Training, Decentralized Execution paradigm~\cite{kraemer2016multi}). We show that \DTWOCOPLAN{} is a scalable, efficient approach for multi-robot target coverage. In particular, we show that \DTWOCOPLAN{} is able to achieve $~93\%$ of the centralized optimal algorithm in up to $~150$x less time but in a decentralized fashion. 

A typical approach for this problem is to frame it as a submodular maximization problem with a uniform matroid constraint~\cite{calinescu2011maximizing}. A decentralized greedy (\DG{}) algorithm gives theoretical performance guarantees and works well empirically~\cite{qu2019distributed,williams2017decentralized}. We show that \DTWOCOPLAN{} performs as well as \DG{} when the ground truth positions of the targets are known and better when the robots have to predict the motion of the targets. Further, the running time of \DTWOCOPLAN{} scales better compared to \DG{}. A key advantage of \DTWOCOPLAN{} is that it consists of two differentiable modules, where the observation processor module can be trained to be compatible with the planner module. We investigate several ways of combining the two modules as well as ablation studies for the design of \DTWOCOPLAN{}'s architecture.

The rest of the chapter is organized as follows:
We first discuss the related work on this topic in subsubsection~\ref{subsec:d2coplan:related_work}. Then we formulate the problem in subsection~\ref{subsec:d2coplan:problem_formulation}. subsection~\ref{subsec:d2coplan:methods} describes the design of \DTWOCOPLAN{}. subsection~\ref{subsec:d2coplan:experiment_results} first provides the implementation details and then describes various experiments and the results obtained. We conclude by summarizing our findings in subsection~\ref{ref:conclusion} and discuss the avenues of future work. We also share these findings on our project webpage.\footnotemark[3] \footnotetext[3]{Project Webpage: \href{http://raaslab.org/projects/d2coplan.html}{http://raaslab.org/projects/d2coplan.html}}

\subsection{Related Work}\label{subsec:d2coplan:related_work}
Multi-robot coordination problems have largely relied on using classical, non-leaning-based approaches. The centralized approaches assume the presence of a single entity that can access observations from all the robots and plan accordingly. Since finding optimal solutions may be practically intractable, centralized approaches often utilize greedy formulations to find approximate solutions. Many multi-robot tracking and coverage objectives are submodular i.e., they have diminishing return property, and greedy solutions provide constant factor approximation guarantee for them~\cite{zhou2021multi}. 

Finding solutions with centralized approaches is still computationally expensive and the runtime rapidly increases with the increase in the number of robots. Decentralized approaches provide an efficient solution at the cost of a lower, but acceptable drop in the task performance, by distributing the task of computation to cliques\cite{zhou2021multi,zhou2022distributed,shi2021communication}. The communication could be expanded to multiple hops to increase the information horizon, but it comes at the cost of increased runtime~\cite{qu2019distributed,williams2017decentralized}.

Neural networks provide an avenue to improve upon classical solutions through their ability to model complexities using data. Furthermore, a differentiable approach can be combined with other differentiable methods to enable efficient learning with end-to-end learning from data~\cite{chaplot2021differentiable}. Introduction of GNNs~\cite{scarselli2008graph} to solve the problem with graph representations opened doors to the application of neural networks to decentralized multi-robot tasks by facilitating feature sharing between robots~\cite{gama2020graphs}. Recent works have successfully employed GNNs to solve multi-robot problems such as path planning~\cite{li2020graph, li2021message}, persistent monitoring~\cite{chen2021multi}, and formation control~\cite{Tolstaya19-Flocking, khan2019graph} among others. Specifically for multi-robot coverage problems, Tolstaya et al.~\cite{tolstaya2021multi} and Gosrich et al.~\cite{gosrich2022coverage} used GNNs in different training paradigms to learn control policies. Many of these works show that apart from achieving near-expert solutions, GNNs can help scale well to larger robot teams.

Unlike these works, we specifically focus on target coverage. Recently, Zhou et al.~\cite{zhou_graph_2022} proposed a planner for the coverage problem using GNN and show such a planner performs on par with the classical counterpart and scales marginally better. However, their approach requires hand-crafted features and uses only 20 closest targets as input. This design makes the network non-differentiable at the input layer and thus can not be used in conjunction with other learning methods. We address both these issues in our work by using a richer map representation, while also improving the coverage performance and scalability. 

\subsection{Problem Formulation}\label{subsec:d2coplan:problem_formulation}
\begin{figure}
\centering
\includegraphics[width=0.65\columnwidth]{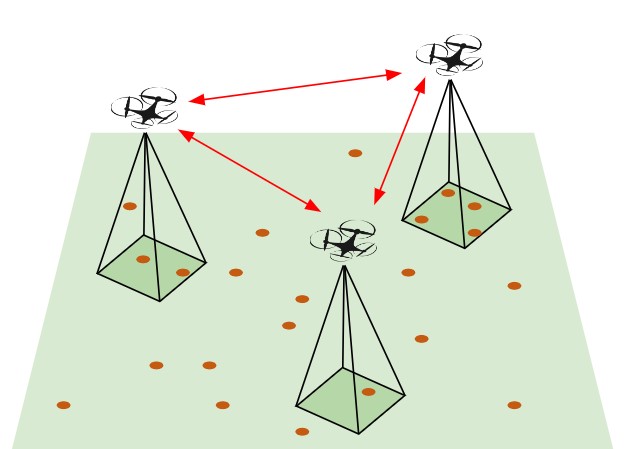}
\caption[An overview of the multi-robot target coverage setup. A team of aerial robots aims at covering multiple targets (depicted as red dots) on the ground.
The robots observe the targets in their respective field of view (green squares) using down-facing cameras and share information with the neighbors through communication links (red arrows).]{Multi-robot target coverage: a team of aerial robots aims at covering multiple targets (depicted as red dots) on the ground.
The robots observe the targets in their respective field of view  (green squares) using down-facing cameras and share information with the neighbors through communication links (red arrows).}
\label{fig:d2coplan:uav_tracking}
\end{figure}

In this work, we investigate the problem of decentralized, multi-robot action selection for joint coverage maximization. Consider the scenario in Figure~\ref{fig:d2coplan:uav_tracking}. 

A set of $N$ robots are tasked to cover targets moving in a grid of size $G\times G$. Every robot $R_i$ has a set of actions $A_i$ that it must select from at each time step. All the targets that fall within the sensing range $r_s$ (e.g., camera footprint) are said to be \textit{covered} by the robot. The objective is to maximize the total number of targets covered \VS{in the next step} by selecting the actions for each robot. 

We assume that the robots do not collide with each other (e.g., by flying at different altitudes). A robot $R_i$ can communicate with another robot $R_j$ if it is within the \textit{communication range} $r_c$.  
The robots need to select their actions based on only local information. 

\begin{figure}[t]
% \vspace{2 mm}
\centering{
{\includegraphics[width=0.6\columnwidth,trim= 0cm .0cm 0 0cm,clip]{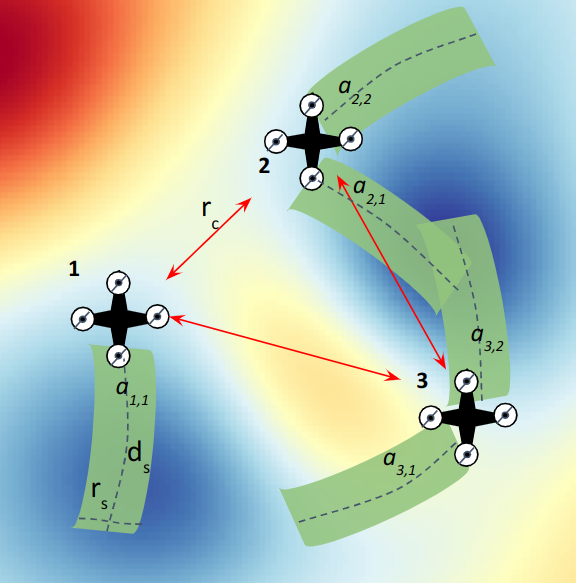}}
\caption{An illustrative example showing multi-robot action selection for joint coverage maximization.}{An illustrative example: at a given time step, each robot $R_i$ must choose a motion primitive $a_{i,k}$ (dashed curves). The background map shows areas with high target density in blue and low target density in red. Here, $R_1$ has one motion primitive $\{a_{1,1}\}$, $R_2$ has two primitives $\{a_{2,1}, a_{2,2}\}$, and $R_3$ also has two motion primitives $\{a_{3,1}, a_{3,2}\}$. The size of the \textit{coverage map} depends on the robot's sensing range $r_s$ and moving distance $d_s$. As $R_2$ and $R_3$ have overlapping coverage maps, they must communicate with each other using communication links (red arrows) of range $r_c$, to choose actions that can maximize the total coverage. \label{fig:d2coplan:uav_trajectory}}
}
\end{figure}

Each robot has access to a local \textit{coverage map}, which gives the \emph{predicted} occupancy of targets near the robot (specifically, targets that can be covered by its motion primitives). Any overlap in covering the same set of targets results in the targets being counted as \textit{covered} only once. We show an example in Figure~\ref{fig:d2coplan:uav_trajectory} where robot 2 and robot 3 may end up tracking the same target.  Thus, a robot must collaborate with others to minimize overlap in motion for efficient coverage. To do so, the robot must also share its local coverage map with others. It is also important to share a compact representation of the map to reduce the bandwidth requirement of the algorithm. 

Our main contribution is \DTWOCOPLAN{}, which solves both problems simultaneously. It consists of a map encoder that comes up with a compact representation of each robot's coverage map, an information aggregator, followed by an action selector. Furthermore, since \DTWOCOPLAN{} is differentiable, we can combine it with a Differentiable Map Predictor (\DMP{}), that takes as input the history of observations from a robot and predicts the coverage map of where the targets are going to be when the robots move.

\subsection{Proposed Approach: Differentiable, Decentralized Coverage Planner\texorpdfstring{\\} ((D2CoPlan)}\label{subsec:d2coplan:methods}
We present a differentiable, decentralized coverage planner called \DTWOCOPLAN{} to efficiently solve the multi-robot coverage problem by predicting the best action for a robot given its local coverage map. It can be integrated with any differentiable map predictor (\DMP{}), to solve tasks where direct observations are not available. We design \DTWOCOPLAN{} as a combination of three sub-modules:

\subsubsection{Map Encoder} This module takes the robot's coverage map as input and transforms it into a feature vector that can be shared with the robot's neighbors. We implement this module using a multi-layer Convolutional Neural Network (CNN), consisting of convolution, pooling, and ReLU activation layers. The input to the encoder is the coverage map as a single channel image of size $G \times G$. The output features from the CNN are flattened into a vector of size $H\times 1$ before sharing with the neighbors. This also allows for compressing the local maps making it efficient to communicate them to other robots. We choose CNN as the encoder here over a fully connected neural network as it allows for a richer representation than the pre-processed inputs required for the latter as used in prior work~\cite{zhou_graph_2022}. Furthermore, we do not need to limit the maximum number of targets as input in our representation.

\subsubsection{Distributed Feature Generator} This part of the network enables the sharing of the map encoding features with a GNN. GNN enables feature aggregation for each graph node through neural networks, allowing distributed execution. The information can be shared with $K$-hop communication to the neighbors identified using the adjacency matrix. The output of this module summarizes the information from the neighbors as a vector, enabling informed decision-making in the next step. 

\subsubsection{Local Action Selector} The last module of \DTWOCOPLAN{} is responsible for prescribing the best action to the robot based on the information gathered from the neighbors in the previous step. We implement this module as a Multi-Layer Perception (MLP) which outputs a $|A|$-dimensional output, denoting the fitness of each action, $a_i\in A$. During the training, the loss is calculated as cross-entropy over these outputs with the ground truth actions. Thus, this module enables the gradient flow for end-to-end training for \DTWOCOPLAN{}.

For training \DTWOCOPLAN{}, we use a centralized greedy algorithm as the expert algorithm to generate the target actions. The centralized greedy algorithm has access to global information (i.e., the global coverage map) and can therefore make much more informed decisions. In fact, it is known that the centralized greedy algorithm is within $\sim 66\%$ of the centralized optimal which eliminates the need to run an optimal, brute-force search algorithm that is infeasible for generating training data for a large number of robots. The expert algorithm evaluates the coverage of each robot-action pair and selects the pair with the highest value. The selected robot and the covered targets are removed from consideration and the process is repeated till each robot is assigned an action. The algorithm has a time complexity of $\mathcal{O}(n^2)$ for $n$ number of robots. We refer to this algorithm as \EXPERT{}.

\begin{figure*}[t]
\centering{
{\includegraphics[width=0.95\linewidth,trim= 0cm .0cm 0 0cm,clip]{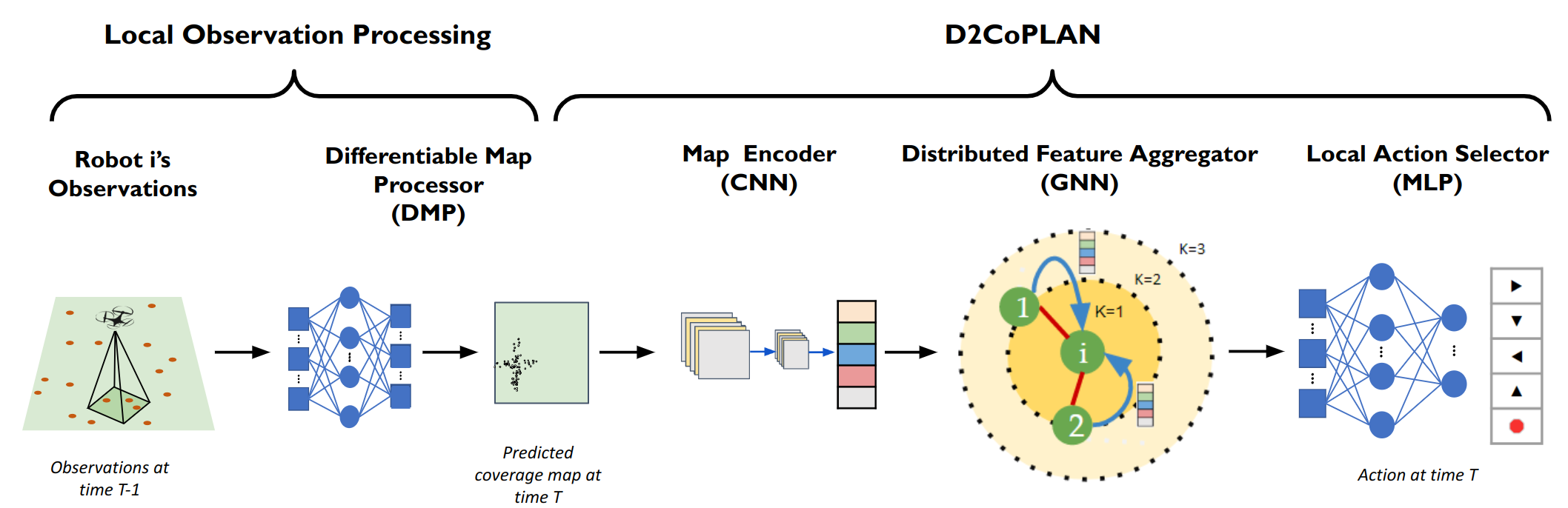}}
\caption{\small Overview of our approach from a robot's perspective: first the local observations are processed to generate the current coverage map. This can be done with the Differentiable Map Processor (\DMP{}). \DTWOCOPLAN{} takes the coverage map as the input and processes it to first generate compact feature representation, with \textit{Map Encoder}; shares the features with its neighbors, using the \textit{Distributed Feature Aggregator}; and then selects an action using the aggregated information, with the \textit{Local Action Selector}. The abbreviations in the parentheses for \DTWOCOPLAN{}'s sub-modules indicate the type of neural network used in their implementation. \label{fig:d2coplan:overview}}
}
\end{figure*}

\subsubsection{Differentiable Map Predictor}
To transform the robot's observation into coverage maps, we introduce a map predictor module. To allow integration with \DTWOCOPLAN{} in order to learn the transformation we use a differentiable map predictor (\DMP{}). The design of \DMP{} depends on the task at hand and can be realized with neural networks. For example, if the task is defined as maximizing coverage with moving targets, \DMP{} can be implemented as a recurrent neural network. We use CNN to solve this task by stacking the historical observations as a multidimensional image and training it with a pre-trained \DTWOCOPLAN{} over the expert actions. This module is optional and we can use the ground truth coverage map for action selection, if available.

\subsection{Experiments and Evaluation}\label{subsec:d2coplan:experiment_results}
\subsubsection{Experiment Setup}\label{subsec:d2coplan:setup}
In our experiments, We use \DTWOCOPLAN{} trained over $N=20$ robots. To generate the training data, we use a grid with $G=100$ i.e., a grid with size $100 \times 100$. The target coverage maps are generated using a mixture of Gaussian to simulate low and high-density areas. The intuition is to mimic real-life situations such as animals' density being higher closer to water holes and lower around ditches in a forest. For this, we choose a random number of Gaussian components in the range $[10,30]$ with the standard deviation for each uniformly sampled from the set $\{20, 30, 40, 50\}$. The locations of the means are selected uniformly at random on the grid. Some of the components are randomly inverted by multiplying by $-1$ to simulate lower-density regions. The probability density obtained by summing up the components is then normalized to obtain a categorical probability density function over the grid. As the last step, we sample locations using this density function to fill $15\%$ of the grid cells to represent the target locations. We simulate linear motion for the targets with a randomly chosen initial velocity.

The robot locations are randomly selected on the grid. The action set for each robot consists of 5 actions, one per cardinal direction and one to stay in place. In the sensing range $r_s=6$, each action moves a distance of $d_s=20$, and the communication range is $r_c=20$. With our choice of action primitives, the coverage map looks like a rectangular field on the grid of size $G \times G$, where only the target within the coverage map is visible. The communication is limited to 1-hop only. We generate a total of 40000 maps and run  \EXPERT{} on each to obtain the target actions. From this dataset, $60\%$ instances are used for training, $20\%$ are used for validation and the rest are used for testing. 

\textit{Map encoder} is implemented as a 3-layer CNN (Conv$\rightarrow{}$ReLU$\rightarrow{}$Maxpool) with intermediate output features of size $4$, $8$ and $16$. The final output is flattened to a vector of size $1600$. This vector acts as a compressed map representation. For \textit{Distributed feature aggregator}, we use implementation by Li et al.~\cite{li2020graph} with 2 graph layers of $512$ and $128$ nodes and ReLU activation. \textit{Local action selector} is implemented as a single layer fully connected network, directly predicting the actions. We also use dropout of $20\%$ in the CNN layers and after the GNN to regularize the network. We train the network on \VS{a 32-core, 2.10Ghz Xeon Silver-4208 CPU and a} Nvidia GeForce RTX 2080Ti GPU with 11GB of memory for 1500 epochs and use the network weights with the minimum validation loss for evaluation. 

\subsubsection{Evaluation}\label{subsec:d2coplan:evaluation}
An efficient distributed planner must have some desirable properties: it should run faster than the centralized algorithm while achieving coverage within a reasonable margin of the centralized algorithm; and it should scale well with a varying number of agents by generalizing beyond the settings it is trained on. In this subsection, we present empirical evidence that \DTWOCOPLAN{} has the aforementioned desirable properties. We go one step further and show that \DTWOCOPLAN{} scales better than even \DG{}. Finally, we demonstrate the advantages of a \textit{differentiable} design. Specifically, we show that \DTWOCOPLAN{} performs better when combined with \DMP{} than \DG{}. 

\subsubsection{Comparisons with \EXPERT{}} We begin by comparing the coverage performance (number of targets covered) and runtime of \DTWOCOPLAN{} with the \EXPERT{} which is the centralized greedy algorithm that \DTWOCOPLAN{} is trained using. In this set of experiments, we use the ground truth coverage map as input since our focus is on evaluating the planner. In subsequent experiments, we will evaluate the effect of the map predictor on the coverage task.

\DTWOCOPLAN{} was trained on a dataset of 20 robots in a grid of size $100 \times 100$. We compare the two algorithms with an increasing number of robots (from $4$ to $50$) in the same grid. We run 1000 Monte-Carlo simulations for each setting.

\begin{figure*}[t]
\centering
\begin{subfigure}[c]{.48\textwidth}
    \includegraphics[width=1.0\columnwidth]{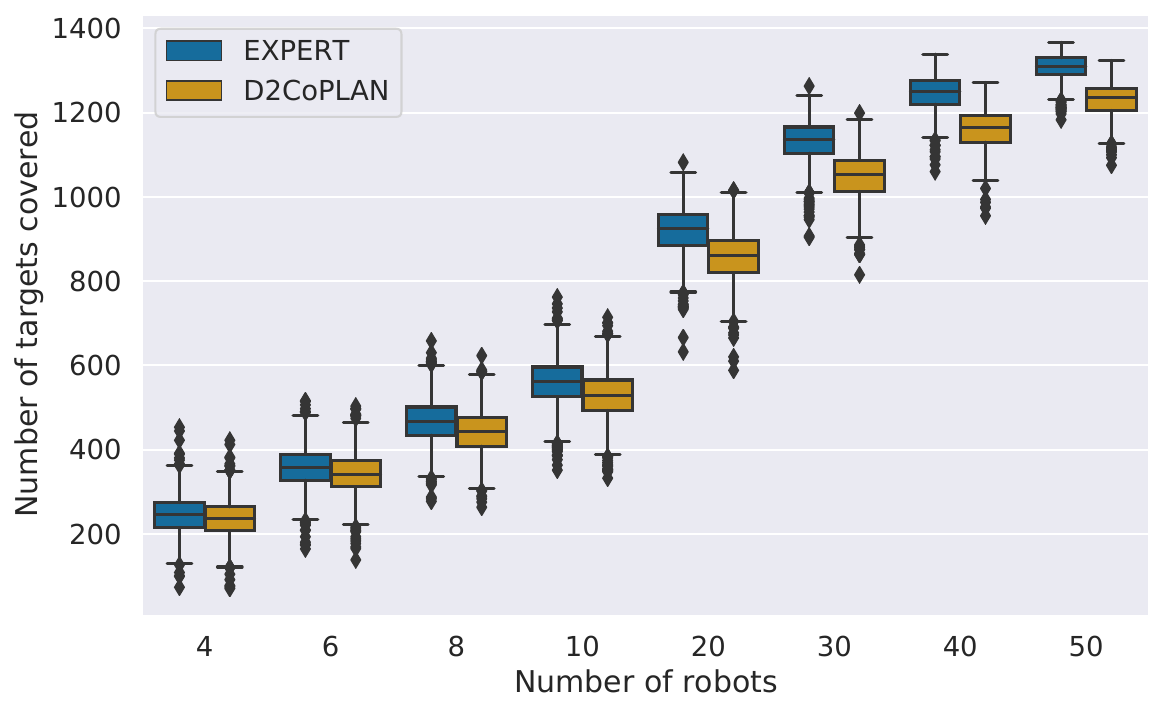}
    \caption{Scaling in terms of Coverage}
\end{subfigure}
\begin{subfigure}[c]{.48\textwidth}
    \includegraphics[width=1.0\columnwidth]{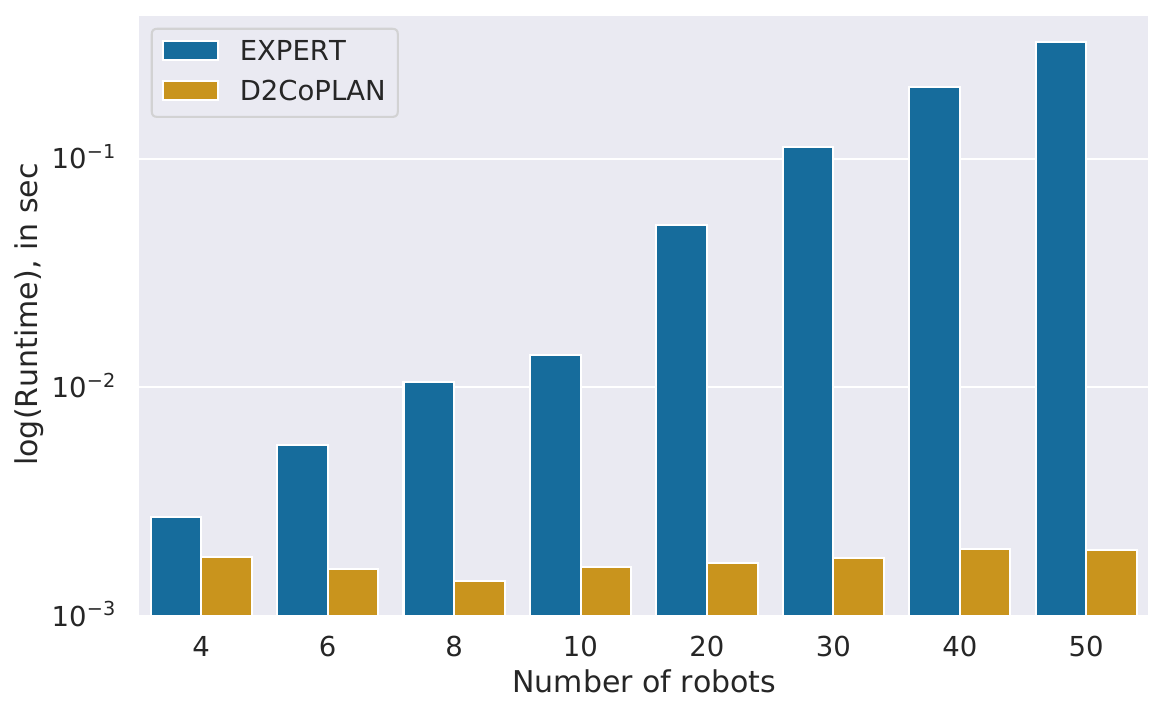}
\caption{Scaling in terms of time}
\end{subfigure}
\caption{\small Comparison of \EXPERT{} and \DTWOCOPLAN{} in terms of running time (plotted in $\log$ scale) and the number of targets covered, averaged across 1000 Monte Carlo trials. \DTWOCOPLAN{} was trained on 20 robots. \DTWOCOPLAN{} is able to cover ~92\%-96\% of the targets covered by \EXPERT{}, while running at a much faster rate.}
\label{fig:d2coplan:expert_vs_gnn}
\end{figure*}

The results for this evaluation are shown in Figure~\ref{fig:d2coplan:expert_vs_gnn}. \DTWOCOPLAN{} has a clear advantage in terms of runtime and the advantage increases as the number of robots increases. For example, with 50 robots, \DTWOCOPLAN{} is more than two orders of magnitude faster than \EXPERT{}. This is not surprising since \EXPERT{} is a centralized algorithm whose runtime scales quadratically with the number of robots whereas \DTWOCOPLAN{} is a decentralized algorithm. In addition to being significantly faster, we also observe that \DTWOCOPLAN{} covers $~92\%$ of the targets as the \EXPERT{}, despite each robot having only a limited amount of information.

\subsubsection{Comparisons with \DG{}}
\begin{figure*}[t]
\centering
\begin{subfigure}[c]{.48\textwidth}
    \includegraphics[width=1.0\columnwidth]{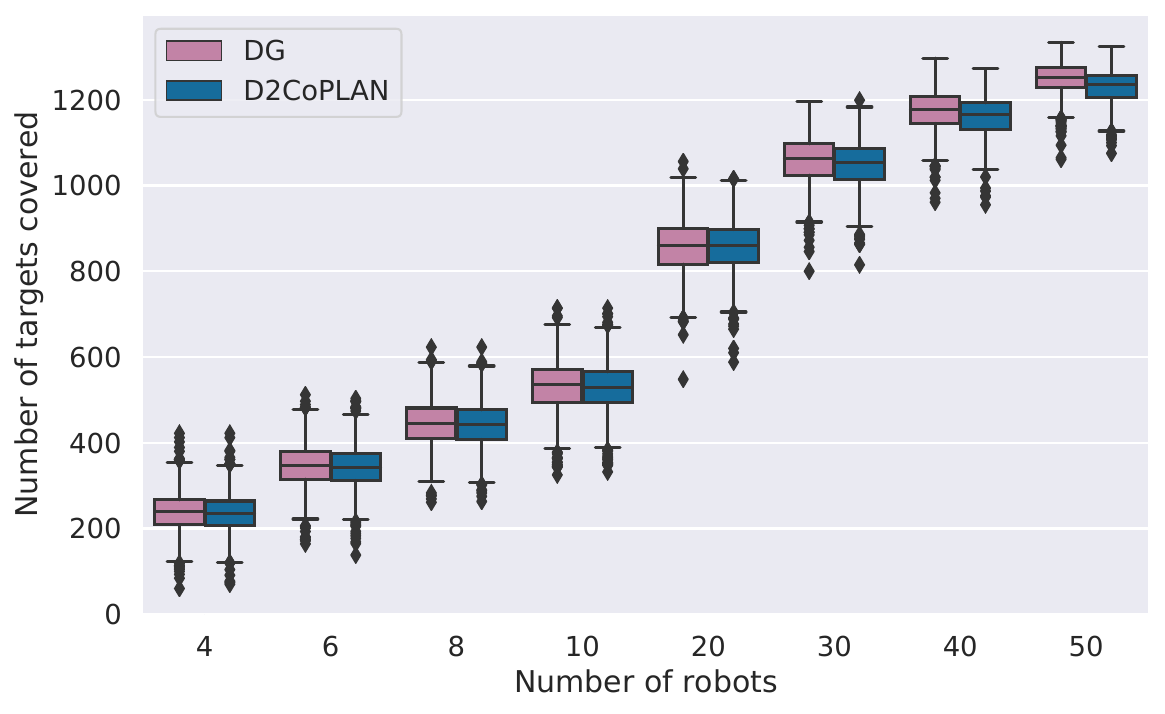}
    \caption{Scaling in terms of Coverage}
\end{subfigure}
\begin{subfigure}[c]{.48\textwidth}
    \includegraphics[width=1.0\columnwidth]{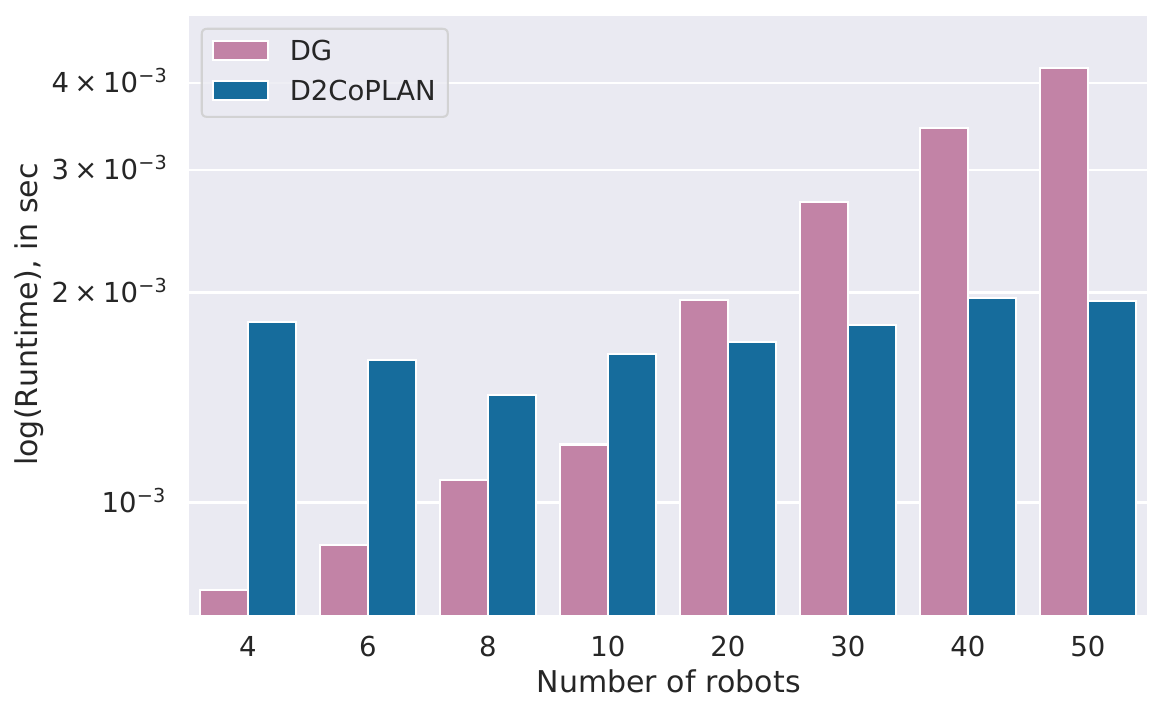}
    \caption{Scaling in terms of time}
\end{subfigure}
\caption{\small Comparison of \DTWOCOPLAN{}, and \DG{} in terms of running time (plotted in $\log$ scale) and the number of targets covered, averaged across 1000 Monte Carlo trials. \DTWOCOPLAN{} was trained on 20 robots. \DTWOCOPLAN{} is able to cover almost the same number of targets as \DG{}. \DG{} is faster for fewer number of robots, but as the number of robots increases, \DTWOCOPLAN{} scales better than it.}
\label{fig:d2coplan:dg_vs_gnn}
\end{figure*}

Next, we compare \DTWOCOPLAN{} with a classical decentralized algorithm, \DG{}. In \DG{}, each robot chooses its own action by running a greedy algorithm but only on the set that includes itself and its immediate neighbors (hence, decentralization). As shown in Figure~\ref{fig:d2coplan:dg_vs_gnn}, \DTWOCOPLAN{} and \DG{} perform almost the same in terms of the number of targets tracked. However, the real advantage of \DTWOCOPLAN{} comes in the runtime where we observe it becomes much faster than \DG{} as the number of robots increases (e.g., with 50 robots, \DTWOCOPLAN{} is almost twice as fast). While both algorithms are decentralized, \DG{} still requires running a greedy algorithm over the local neighborhood of each robot which increases the runtime as the density of the robots increases. Furthermore, in subsection~\ref{subsec:d2coplan:dmp_d2coplan} we show that \DTWOCOPLAN{} outperforms \DG{} even in terms of coverage performance when the true coverage map is not given. 

\subsubsection{Generalization}
Next, we evaluate the generalization capability of \DTWOCOPLAN{} beyond the scenario it has been trained on. We test two types of generalization: (1) across the number of robots; and (2) across the density of the targets (i.e., coverage map) in the environment. For both tests, we train on a specific number of robots (1) or target density (2) and test with a different number of robots (1) or target density (2). We summarize these results in Table~\ref{tab:d2coplan:general_robots} and Table~\ref{tab:d2coplan:general_density} obtained over 1000 Monte-Carlo runs.

We observe that \DTWOCOPLAN{} generalizes well in both cases. Table~\ref{tab:d2coplan:general_robots} shows the coverage performance when trained on the number of robots given in the row and tested on the number of robots given in the column. We see that in most cases, the performance remains unchanged. The network trained on $10$ robots sees a slight drop in performance on other test configurations but still covers around $~90\%$ of the targets covered by \EXPERT{}. 

\DTWOCOPLAN{} also generalizes well across varying target density as shown in Table~\ref{tab:d2coplan:general_density}. We observe that \DTWOCOPLAN{} trained with a target density of $15\%$ performs almost the same when tested on other target densities. The performance is $\sim 93\%$ of the \EXPERT{} in all cases but $5\%$ density (where it is $\sim 91\%$), which we believe is caused by fewer number of available targets, increasing the gap in the performance of the compared algorithms. These results validate the claim that \DTWOCOPLAN{} trained under one type of scenario generalizes to other deployment scenarios. 

\begin{table}
% \vspace{1mm}
\centering
{\renewcommand{\arraystretch}{2.0}
\begin{tabular}{|l|*{3}{c|}}\hline
\backslashbox{Train}{Test}
&\makebox[4em]{10 Robots}&\makebox[4em]{20 Robots}&\makebox[4em]{30 Robots}\\\hline
\makebox[6em]{10 Robots} &93.95\% &91.12\% &89.71\% \\\hline
\makebox[6em]{20 Robots} &94.38\% &93.17\% &92.60\% \\\hline
\makebox[6em]{30 Robots} &93.25\% &93.47\% &93.73\% \\\hline
\end{tabular}}
% \vspace{2mm}
\caption{Percentage of the targets covered (the average across 1000 trials) with respect to \EXPERT{} by \DTWOCOPLAN{} trained and tested with varying numbers of robots.}
\label{tab:d2coplan:general_robots}
% \vspace{-2mm}
\end{table}

\begin{table}
\vspace{1.5 mm}
\centering
\begin{tabular}{ | c | c|} 
\hline
{Target Density} & {Relative coverage}\\
\hline
$5\%$ & 91.40\%\\
\hline
$15\%$ & 93.20\%\\
\hline
$25\%$ & 93.46\%\\
\hline
$50\%$ & 93.46\%\\
\hline
\end{tabular}
\caption{Percentage of the targets covered (the average across 1000 trials) with respect to \EXPERT{} by \DTWOCOPLAN{} across varying target density maps.}
\label{tab:d2coplan:general_density}
\end{table}

\subsubsection{Prediction and Planning}\label{subsec:d2coplan:dmp_d2coplan}
A key advantage of \DTWOCOPLAN{} is its differentiability, allowing \DTWOCOPLAN{} to be combined with other gradient-based learning methods to solve challenging problems in an end-to-end manner. In this subsection, we evaluate how the differentiable map predictor can be trained along with the differentiable planner (\DTWOCOPLAN{}) and compare it with \DG{}.

So far, we have used the ground truth coverage map as input to the planners. Now, we consider a scenario where the input consists of the observations of the targets over the past timesteps. The true motion model of the robots is not known to the robots. Therefore, they need a predictor to estimate the positions of the targets over the planning horizon which can then be used by \DG{} or \DTWOCOPLAN{}.

Here, we use a \DMP{} to learn the motion model. \VS{All the targets move with the same linear velocity} selected randomly \VS{in the range $[-3, 3]$} at the start of the episode (unknown to the planner). To show the advantage of having a decentralized planner, we compare three methods: (1) an \ORACLE{} i.e., the ground truth map as the mapper along with \EXPERT{} as the planner; (2) \DMP{} as the learnable mapper with \DG{} as the planner; and (3) \DMP{} as the learnable mapper with \DTWOCOPLAN{} as the planner. In (2), \DMP{} is trained from scratch where\VS{as} in (3) \DMP{} is trained by backpropagating the loss from \DTWOCOPLAN{}. \DTWOCOPLAN{} itself is frozen and aids \DMP{} in learning better representations for action prediction. The three settings present different combinations \VS{of} classical and learning-based approaches. 

Coverage maps observed over the last 3 time steps are used as input to \DMP{} and it predicts the map at the next time step. We use a 4-layer CNN with 8, 16, 4, and 2 channels as \DMP{}. We keep the map size the same across each layer to avoid information loss and predict the occupancy probability of each cell as a two-channel map. The probability map thus obtained is used as input to the planner.  We trained \DMP{} over 2000 epochs with 5 examples of 20 robots (i.e., 100 training instances) in each. Given that most of the cells in the coverage map will be zero, we weigh the cross-entropy loss by a ratio of 1:10 for free and occupied cells. The action prediction loss for (3) is the unweighted cross-entropy loss. 

\begin{figure}[t]
\centering
{\includegraphics[width=0.90\columnwidth]{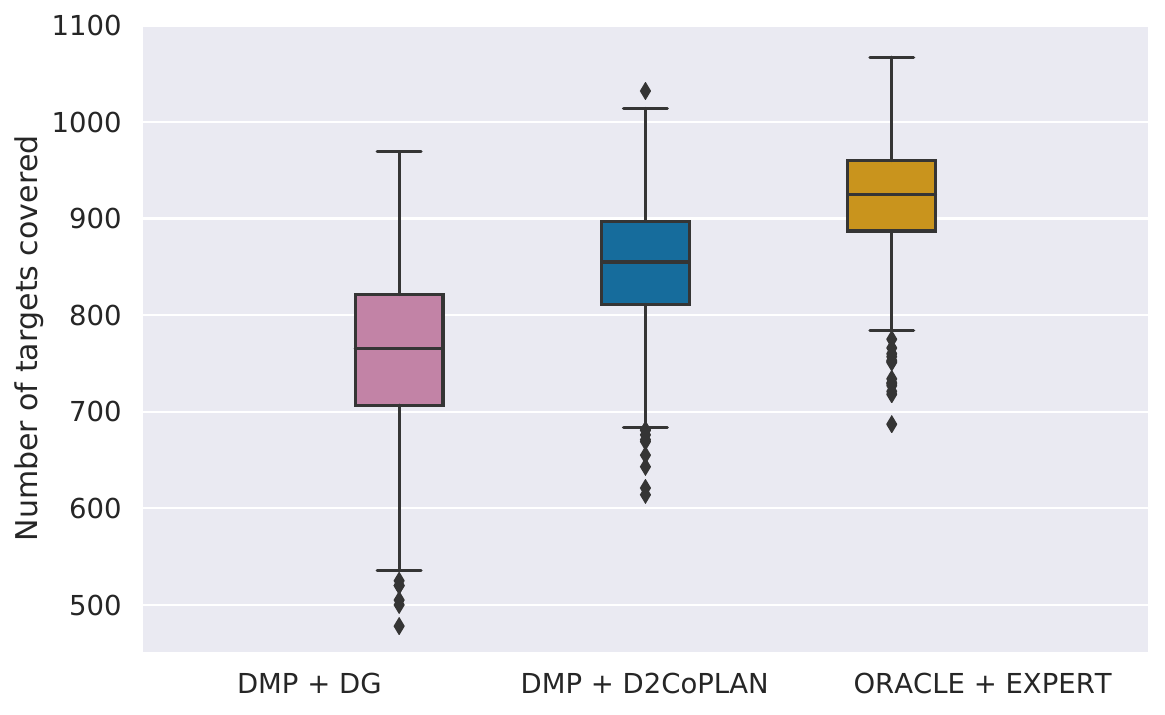}
}
\caption{\small Comparison of coverage highlighting the effect of using \DTWOCOPLAN{}, a differentiable planner to aid learning for a differentiable map predictor (\DMP{}), which works better than the \DMP{} trained standalone.}
\label{fig:d2coplan:downstream_comp}
\end{figure}

Figure~\ref{fig:d2coplan:downstream_comp} shows a comparison of the three approaches and provides evidence for the benefit of using a differentiable planner to realize end-to-end learning. The combination of \DTWOCOPLAN{} and \DMP{} is better compared to \DG{} and \DMP{}, despite \DMP{} in the latter being trained on ground truth. We attribute this to the fact that \DTWOCOPLAN{} and \DMP{} form a differentiable chain which allows \DMP{} to be trained directly on the downstream task (action selection) rather than on just map prediction. \DG{} and \DMP{}, on the other hand, are not a differentiable chain, and thus \DMP{} cannot be trained on the downstream task directly.

We further explore this by comparing 3 ways of training \DMP{} when used in conjunction with \DTWOCOPLAN{}: (1) \DMP{} and \DTWOCOPLAN{} are trained together from scratch; (2) \DMP{} and \DTWOCOPLAN{} are trained individually and then used together; and (3) \DTWOCOPLAN{} is first trained and then \DMP{} is trained on loss from \DTWOCOPLAN{} while \DTWOCOPLAN{} is frozen.

\begin{figure}
\centering
{\includegraphics[width=0.90\columnwidth]{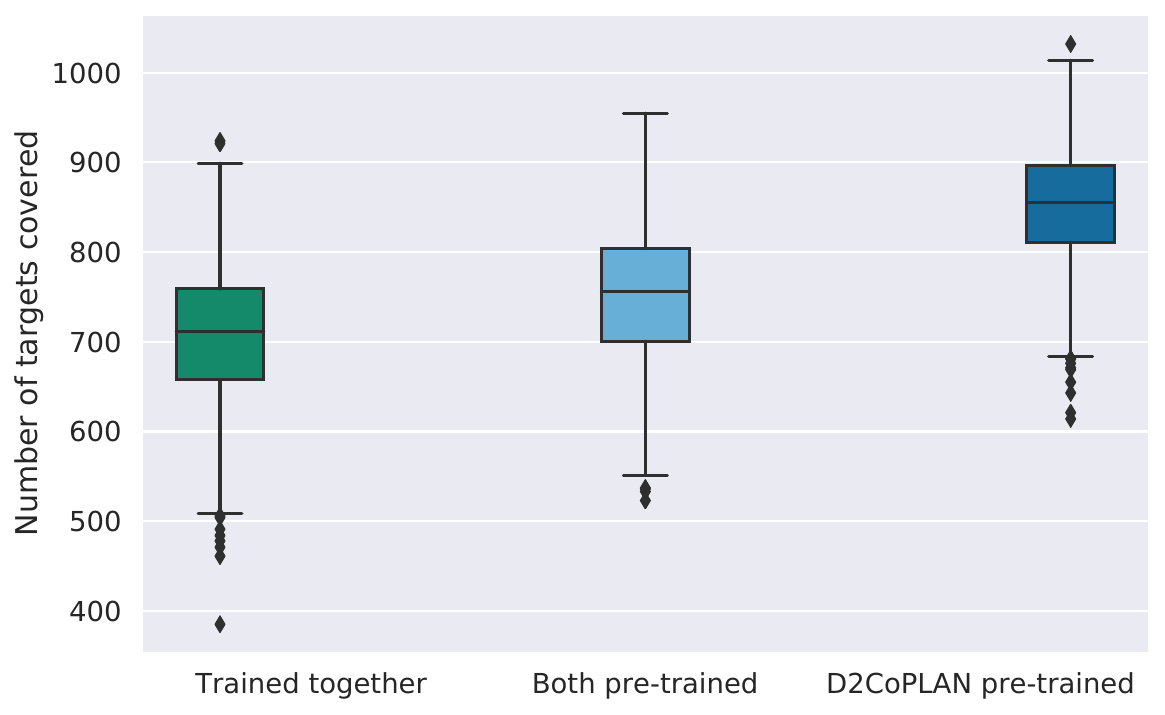}}
\caption{An ablation study for \DMP{} and \DTWOCOPLAN{}. The plot shows results for the scenarios where there parts and trained together or in isolation.}
\label{fig:d2coplan:downstream_ablation}
\end{figure}

Figure~\ref{fig:d2coplan:downstream_ablation} shows the comparison of all three methods. The third approach outperforms the other two. This demonstrates the advantage of having a differentiable planner. Using a pre-trained and frozen \DTWOCOPLAN{} and training directly on the downstream task loss, allows \DMP{} to learn patterns beneficial for action prediction and not just for map prediction. If both modules are trained from scratch in an end-to-end manner, they may need more time to learn the same behavior. The third approach also does not require ground truth motion models for the targets to be available for training \DMP{}. While in this work, we use the ground truth to generate the expert solutions used in training \DTWOCOPLAN{}, in general, one can use any other expert algorithm such as human inputs to train \DTWOCOPLAN{} which does not need ground truth target motion.

\subsection{Conclusion}\label{ref:conclusion}
We presented \DTWOCOPLAN{}, a differentiable, decentralized target coverage planner for multi-robot teams. Our experimental results show that \DTWOCOPLAN{} is more scalable than the classical decentralized algorithm that is used for such tasks while performing closer to the centralized algorithm. Furthermore, due to the fact that it is a differentiable planner, we can combine this with other differentiable modules (e.g., a coverage map predictor) to yield better performance than the classic counterparts. These results present an encouraging path forward for multi-robot coordination tasks. Our immediate work is evaluating \DTWOCOPLAN{} for more complex tasks. In this work, we train \DTWOCOPLAN{} in a supervised setting. We are also working on training \DTWOCOPLAN{} with reinforcement learning. Finally, an interesting avenue for extension is where we learn not just \emph{what} to communicate with other robots (as we do in this work) but also \emph{who} to communicate with.

%% file: Chapter7.tex
%Chapter 7

\renewcommand{\thechapter}{7}

\chapter{Semantic Pattern Prediction for Assistive Robot Perception and Planning }
\label{chapter:llm_assist}

In the previous chapters, we show how robots can use the structural, geometric, and spatiotemporal patterns in their surroundings to work efficiently with the help of predictions from learning-based approaches. These patterns emerge mainly from the abstract concepts of morphology and motions. A robot capable of making predictions about these concepts can thus predict how to go around a rectangular box on the ground or avoid collision with a person walking in front of the robot. But to help the humans, it may require \textit{semantic scene understanding}. Such a robot may decipher that a rectangular box is a grocery bag and can carry it to the kitchen. If it sees the person walking towards the laundry room with a basket full of clothes, it can offer help carrying the basket to the laundry room. Building such \textit{assistive agents} thus requires \textit{semantic predictions} about the environment. To maximize the observations about the environment overhead cameras can be used. To this end, we first present a pipeline for \textit{assisting humans} by observing the ongoing activities and making semantic predictions with vision-language models (VLMs) about what the human may need in the near future and using a robot to help with these tasks. Then we present an approach to using the overhead observations for \textit{assisting robots} with the help of semantic communication enabled by VLMs.

\section{Semantic Pattern Prediction for Assisting Humans}

\subsection{Introduction}
\label{subsec:assist:intro}
Embodied agents must understand the surrounding environment and make informed decisions to complete a task. Over the past few years, the reasoning capabilities and subsequent embodied task performance of these agents have improved with the help of generative models. This can be attributed to the greater semantic understanding of the scenes they provide. VLMs in particular allow for open-set scene understanding, witnessed by a proliferation of works on developing VLM-based embodied agents~\cite{shah2022lm,clipnav,shah2023navigation,pan2023langnav,gadre2023cows}. Most of these works, however, are aimed at single robot systems, where the limited field-of-view of the onboard camera also limits the planning horizon.

A traditional setup to alleviate the problem of limited observations is to add another agent that is limited in navigation and interaction capabilities but can provide a more comprehensive view of the scene. One such popular setup is to have an agent with an overhead view of the scene. In such a setting, the overhead view helps the ground robot with long-horizon planning while it uses the ego view for local planning and execution. Such settings are not unrealistic in today's world, where we can easily find monitoring cameras in public buildings and warehouses aimed at providing a large coverage of the environment.

\begin{figure}
    \centering
    \includegraphics[width=\textwidth]{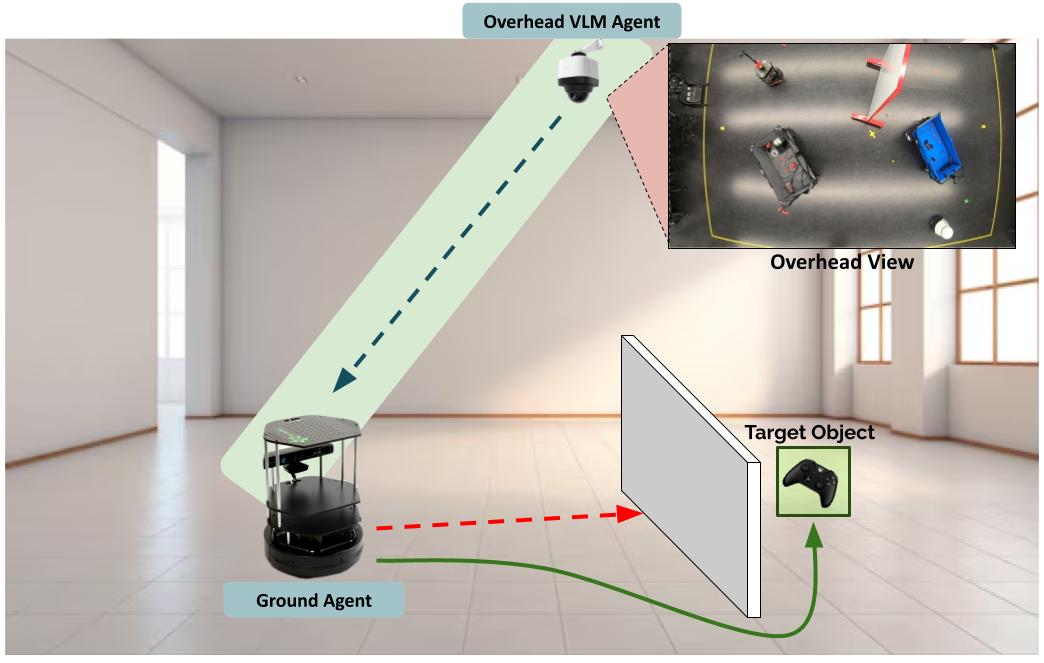}
    \caption{Overview of the assistive robot setup: an overhead agent equipped with the VLM directs a ground robot to help with the tasks based on semantic predictions}
    \label{fig:assist: pipeline}
\end{figure}

In this chapter, we present a pipeline utilizing the overhead view with a VLM to drive the robot around and assist the human with daily tasks. Here, we use GPT-4 VLM with the overhead view to observe the activities in the scene, predict what is likely to happen next, and direct a ground robot to perform tasks to help the human based on these predictions. We implement this pipeline on a Turtlebot2 robot and present qualitative results showing the utility of such an assistive robot such an \textit{assistive robot}. 

\subsection{Related Works}
\label{subsec:assist:related_works}
LLMs and VLMs have captured the attention of the robotic community since their introduction. These models are trained on internet-scale data and thus exhibit the ability to act as world models for robotic task planning and scene understanding without fine-tuning. This has led to traditional and deep learning-based approaches being replaced by zero-shot, LLMs/VLMs-based approaches for navigation~\cite{yokoyama2023vlfm,chang2023goat,shah2023navigation,shah2022lm,xie2023reasoning} and manipulation~\cite{zhang2023bootstrap,gao2023physically,ha2023scaling} among others~\cite{hu2023toward}. These models can also act as a natural language to code translators, making human-robot collaboration more feasible than prior approaches. Such approaches have been explored to obtain a plan as a PDDL~\cite{silver2022pddl}, as python code ~\cite{liang2023code}, or to simply build the plan as a sequence of predefined function~\cite{vemprala2023chatgpt} resulting in accessible and simple interfaces for human instruction to robot movements. This has also led to a proliferation in the solutions for vision-and-language navigation task~\cite{gu-etal-2022-vision}, where a robot is given instructions in natural language that it must follow using visual inputs. 

Given the success of VLMs and LLMs for the aforementioned tasks, the research community has been exploring ways to leverage them to build robots that can assist humans in daily tasks~\cite{wang2024mosaic,hu2024deploying}. This requires focusing on features beyond navigation and manipulation. TidyBot~\cite{wu2023tidybot} learns personalized choices for pre-defined repeated tasks such as household cleanup; GG-LLM~\cite{graule2023gg} tries to ensure minimum disturbance to the human in the environment; TaPA~\cite{wu2023embodied} ensures the plan abides by the availability of the objects in the environment. We focus on the VLMs and LLM's capability of predicting future~\cite{lin2023learning} to help the humans in the scene. Unlike VLaMP~\cite{patel2023pretrained}, which learns to predict a sequence of actions from videos to help a human, we aim for a generalizable, zero-shot approach. GG-LLM~\cite{graule2023gg} is similar to our method, but they use predictions to minimize interference with humans which also results in minimal interaction. Our approach, on the other hand, aims to work alongside humans and interact by bringing them objects that they may need soon.

\subsection{System Details}
\label{subsec:assist:system}
\begin{figure}
    \centering
    \includegraphics[width=\textwidth]{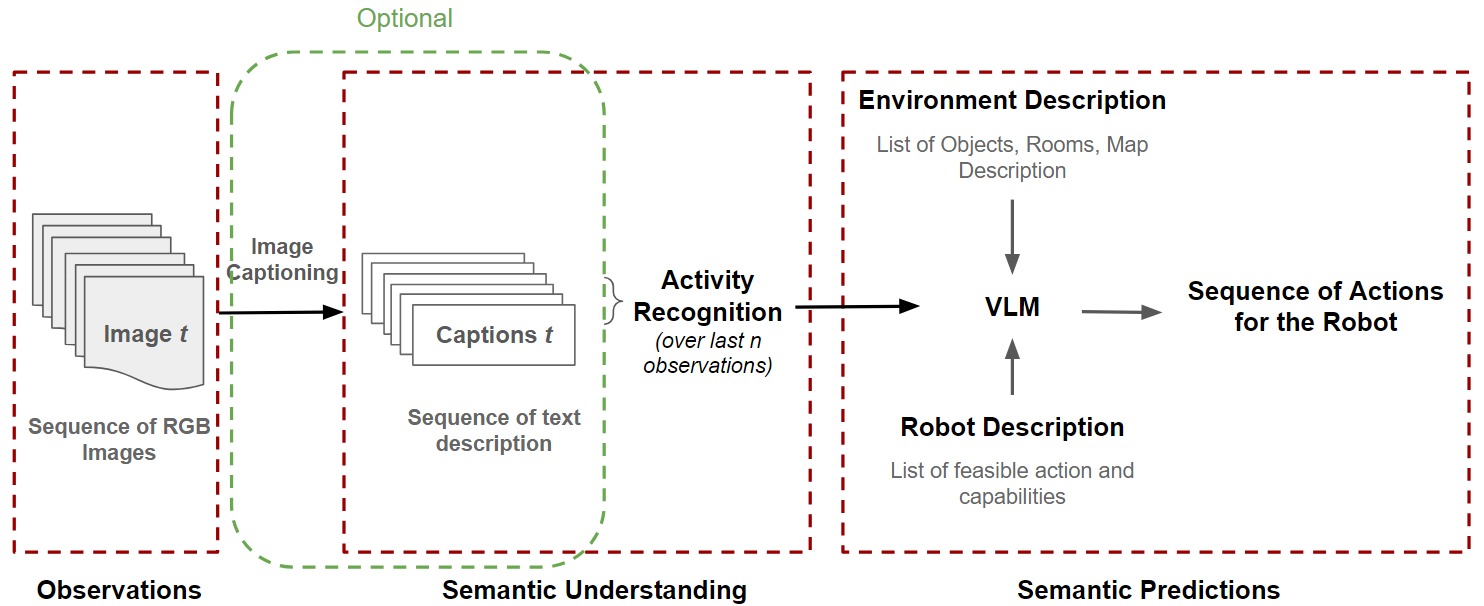}
    \caption{Overview of the proposed pipeline for the assistive robot}
    \label{fig:assist:pipeline}
\end{figure}

Figure~\ref{fig:assist:pipeline} shows an overview of the proposed pipeline composed of an overhead VLM agent and a ground robot. In the following subsection, we discuss the constituents of the pipeline in detail. In this work, we consider a scenario of an indoor environment, but with appropriate modifications, the pipeline can be extended to outdoors as well.

\subsubsection{Semantic Understanding}
In order to predict the future state of the environment, we must understand the past and current states. In our setup, the overhead camera keeps continuously monitoring the environment as a sequence of images. We can either use these images directly as input to the VLM or process them first. To obtain a longer history of the observations as inputs to the VLM while keeping the size of content smaller due to token limits and the cost associated, it might be useful to caption the images using VLM to identify people, objects, and their interactions in the scene. The sequence of such captions can then be fed to VLM in place of images. 

Using the sequence of the images or the captions, we can query the VLM to perform \textbf{activity recognition}, which can help in predicting the future requirements of the human and guide the ground robot accordingly.

\subsubsection{Semantic Predictions}
Once the activity has been recognized from the observations, the VLM can be used to predict future actions. However, without any other information about the environment, the VLM may predict actions that may not be feasible. For example, if the recognized activity is that the person is washing a fruit, the VLM may suggest bringing a knife from the dishwasher. But the house may not have a dishwasher making the suggestion futile. If the knife is around, the VLM may ask the robot to cut it, even though the robot may not have the actuators to perform this task.

To address these challenges rooted in \textit{grounding} and \textit{affordances}, we provide further context in the form of the \textbf{environment description} and \textbf{robot description}. The environment description tells the VLM what rooms are in the house and some objects in them. This can be accomplished by building a map of the house beforehand. The robot description is used to inform the robot about the tasks it can perform, helping the VLM  generate feasible actions only. The combination of these inputs to VLM is used to generate a sequence of feasible actions for the ground robot to assist the human with the tasks.

\subsection{Implementation Details}
We implement the proposed pipeline in a house-like environment in our lab. We create partitions using curtains as the walls and designate them in different rooms. The details for each subsystem are provided in the following subsections. 

\begin{figure}
    \centering
    \includegraphics[width=0.5\textwidth]{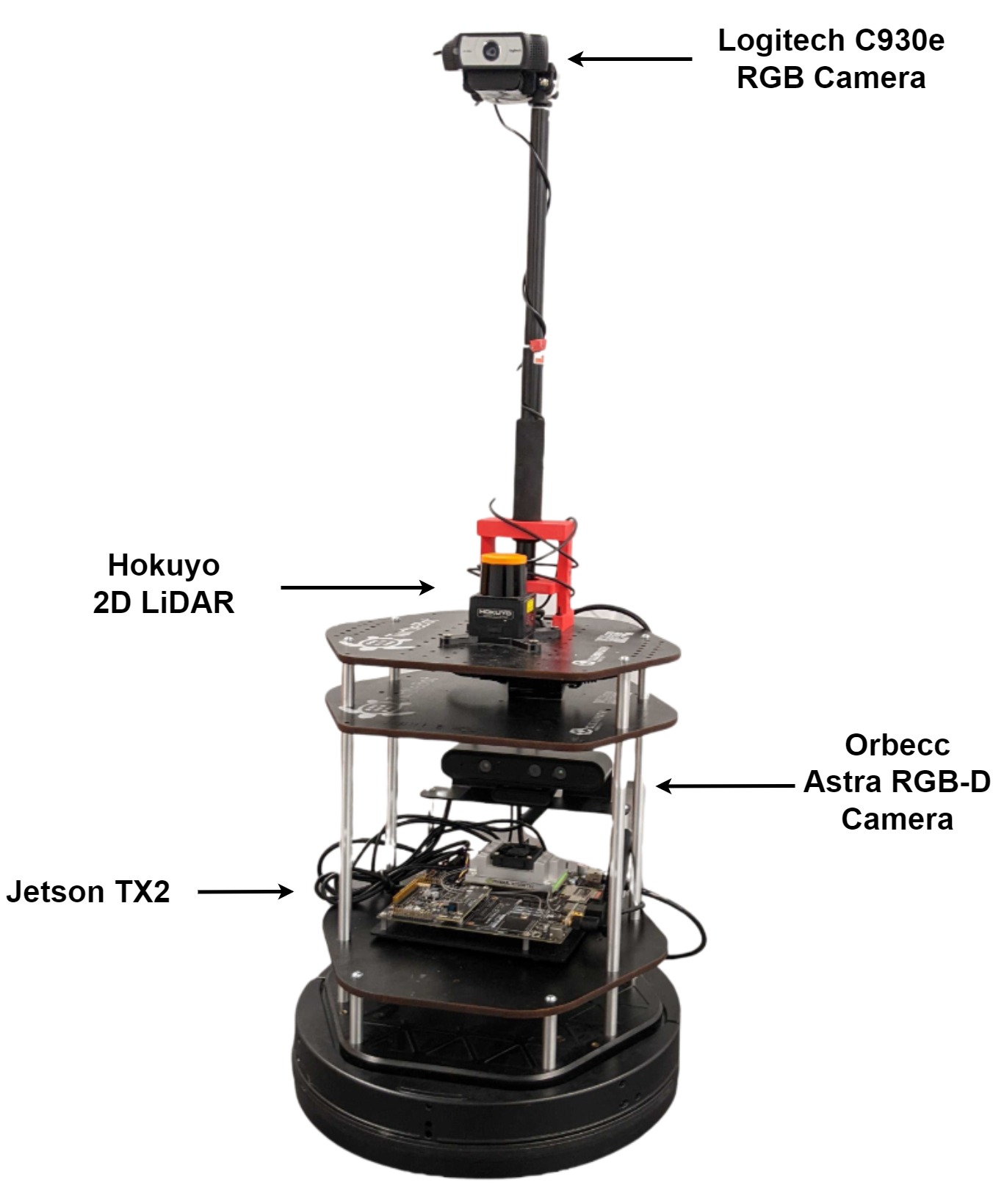}
    \caption{Turtlebot2 and the sensors used for the assistive robot setup}
    \label{fig:assist:turtlebot}
\end{figure}

\textbf{Hardware}: Figure~\ref{fig:assist:turtlebot} show the hardware setup used for this implementation. As the ground robot, we use a Turtlebot2 robot equipped with an Orbbec  Astra RGB-D camera to get depth maps, a Logitech C930e mounted on a stick to observe a wide area in front of the robot, and a Hokyo range sensor for mapping. The robot uses ROS Kinetic running on a Jetson TX2 board and communications with a Lenovo laptop acting as the remote host. We use a Logitech C930e camera mounted on the ceiling as our overhead camera. The camera observes a platform with a sink. 

\begin{figure}
    \centering
    \includegraphics[width=\textwidth]{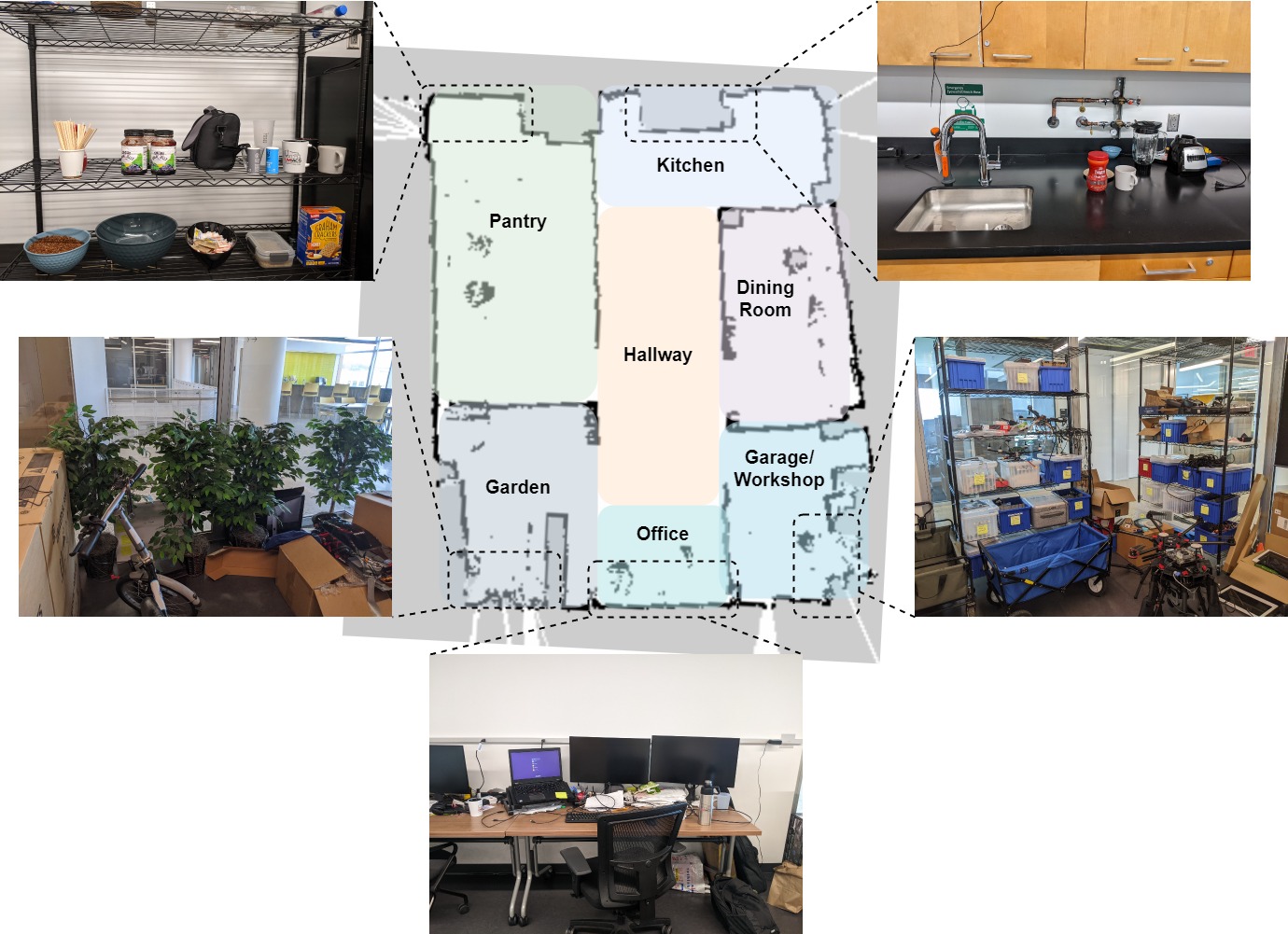}
    \caption{The occupancy map and the labels assigned to the rooms for the house-like environment. The images on the sides show example observations from the rooms.}
    \label{fig:assist:curtain_house_map}
\end{figure}

\textbf{Map Description}: We use the Gmapping~\cite{gmapping} package to build a 2D occupancy grid map of the house. On this map, we designate the partitioned areas as the following rooms: \textit{kitchen, pantry, dining room, office, garage/workshop, garden, and hallway}. The map thus obtained and example observations for the rooms are shown in Figure~\ref{fig:assist:curtain_house_map}. Inside each room, we select a point manually to act as the reference point for the room. We also manually list some objects inside each room to be used as context for the VLM.

\textbf{Robot Description}: Our ground robot, a Turtlrbot2 robot, can only navigate on the map. Thus we define two actions for this robot: (1) \textit{PointNav(room name)} and (b) \textit{ObjectNav(object name)}. The former allows the robot to perform PointGoal navigation from its current location to a room, using the corresponding reference point as the goal location. The latter enables the robot to perform ObjectGoal navigation to find an object. Since the overhead camera may not see the inside of the rooms due to the walls, the ground robot can use the onboard RGB camera for this task. Since the VLM may ask for an object, we use YOLO-World~\cite{Cheng2024YOLOWorld} for open-set object detection and perform visual serving to find and reach the object. YOLO-World is also a VLM and acts as a zero-shot object detector. While other alternatives such as OWL-ViT~\cite{minderer2022simple}, PaliGemma~\cite{model_development_contributors_lucas_beyer__andreas_steiner__andr__susano_pinto__alexander_kolesnikov__xiao_wang__xiaohua_zhai__daniel_salz_maxim_neumann_ibrahim_alabdulmohsin_et_al__2024}, and even GPT-4 could be used in its place, we choose YOLO-World as it can perform detection in real-time. 

In addition to the actions described above, we also allow a \textit{Pick(object)} action to simulate picking an object and keeping it on the robot. This is done to ensure that the VLM can generate meaningful plans as the sequence of actions for tasks such as \textit{fetch an object}. Our hardware does not have the manipulation capability for this task, so we manually put the object on the robot. We also add a \textit{StandBy()} action to allow the VLM to make the ground robot wait and not interfere with the human.

\subsection{An Examples Case Study}
We present an example case showing the pipeline in actions with our implementation. In this task, the person in the scene is making coffee in the kitchen. We provide the following prompts to the agent equipped with GPT-4o: 
\promptbox{System Prompt}{
You are a helpful assistant to a human. You will be given access to a sequence of observations. Based on these observations, you should guess what the person is doing, and what they may need next. The person is in a house with the following rooms: kitchen, pantry, garden, dining table, garage/workshop, office. You have access to a robot that can perform some predefined actions with some arguments (shown below in angular brackets <>). The following are the actions that the robot can take:\\
* PointNav(<room name>): the robot navigates to the room (<room name>) in the house\\
* ObjectNav(<object name>): the robot finds the objects (<object names>) in its surroundings and navigates to a position close to them\\
* Pick(<object name>): the robot picks up the object (<object name>) and place on itself\\
* StandBy(): the robot stays at the same place. This could be used when no robot movement is needed at the moment. 

Using the observations, first, identify a likely activity, then predict what object the person may need next and where they can find it, and then suggest a plan for the robot as a sequence of actions with appropriate inputs to help the human.
}

\begin{figure}[t]
\centering
\begin{subfigure}[c]{.48\textwidth}
    \includegraphics[width=1.0\columnwidth]{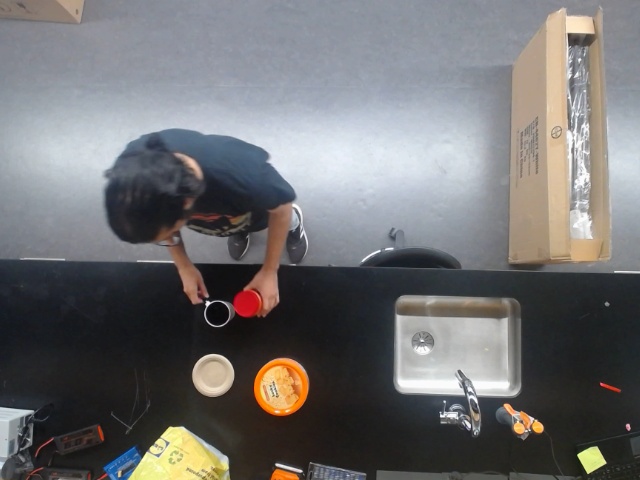}
    % \caption{Scaling in terms of Coverage}
\end{subfigure}
\begin{subfigure}[c]{.48\textwidth}
    \includegraphics[width=1.0\columnwidth]{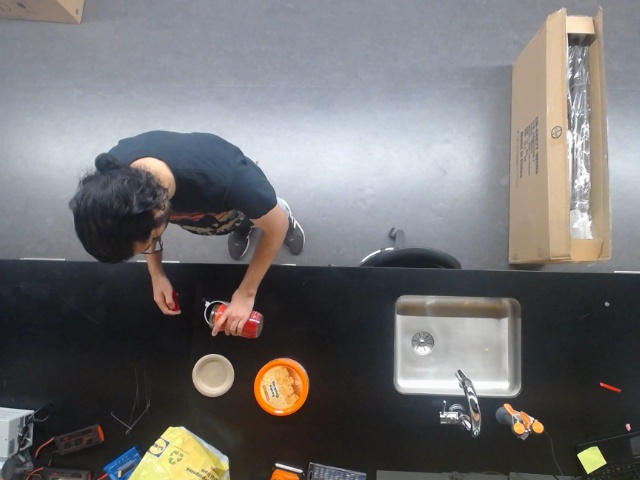}
% \caption{Scaling in terms of time}
\end{subfigure}
\caption{A sequence of observations from the overhead camera while making coffee. There is no spoon or stirrer nearby, which the person may need next.}
\label{fig:assist:coffee_making}
\end{figure}

\promptbox{User Prompt}{
The human is currently in the kitchen. Using these image sequences, predict the likely activity, the required objects, and their locations and generate an action plan for the robot.

<Sequence of observations>
}
As input, we use the sequence of two images instead of captions in this study. Figure~\ref{fig:assist:coffee_making} shows the two images captured for this task. Using these observations, the agent responds with the following:
\responsebox{Agent Response}{
\textbf{Required Objects}
\begin{itemize}
    \item Spoon or Stirrer - For stirring the beverage
    \item Kettle - For heating water if making tea or instant coffee
    \item Water - For the beverage, which might be in the kettle or a separate container
\end{itemize}

\textbf{Locations}
\begin{itemize}
    \item Spoon or Stirrer - Likely in the pantry
    \item Kettle - Likely in the kitchen or pantry
    \item Water - Likely in the kitchen
\end{itemize}

\textbf{Action Plan for the Robot}
\begin{itemize}
    \item ObjectNav('spoon, stirrer') - The robot finds the spoon or stirrer in the pantry.
    \item Pick('spoon, stirrer') - The robot picks up the spoon or stirrer.
    \item PointNav('kitchen') - The robot navigates to the kitchen.
\end{itemize}
}

\begin{figure}[h]
    \centering
    \includegraphics[width=\textwidth]{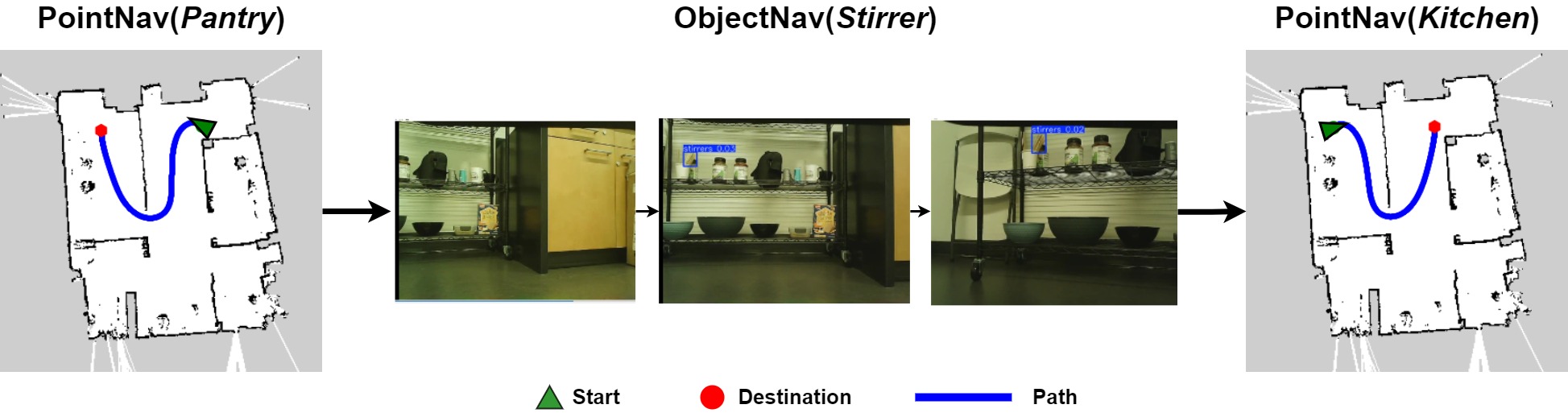}
    \caption{Executing the action suggested by the LLMs: The robot goes to the pantry, finds the stirrer, and brings it to the kitchen. }
    \label{fig:assist:stirrer_detection}
\end{figure}

As shown in the response, of all the rooms given in the map description, the VLM chooses \textit{pantry} as the most appropriate destination and generates the following actions to fetch a stirrer: (1) \textit{PointNav(pantry)}, (2) \textit{ObjectNav(spoon, stirrer)}, (3) \textit{Pick(spoon, stirrer)}, and (4) \textit{PointNav(kitchen)}. 

Given this plan, the Turtlebot2 navigates to the pantry using the stored reference point and then executes ObjectNav to find and reach the stirrer with the help of YOLO-World. We put the stirrer on the robot and signal that the object has been retrieved. Then the Turtlebot2 goes back to the kitchen to deliver the stirrer. This process is displayed in Figure~\ref{fig:assist:stirrer_detection}.

The proposed pipeline shows promising results in semantic predictions-based assistance to humans. These predictions are driven by the VLM-based semantic understanding of the objects and the human interactions with them. The presented implementation is effectively a centralized approach for semantic predictions. The overhead agent uses its view to understand the environment and directs the robot on the ground on a pre-defined map. In many situations, the map may not be known already. Consider a situation where a ground robot is tasked with finding an object in an unknown environment. In such cases, a ground agent with VLM may look around and identify where it is and explore areas where the object is likely to be present (e.g., a toaster is likely to be in the kitchen). However, the limited camera field of view and occlusions from walls and other objects mean the ground agent must rely on limited observation to find the object. 

The overhead agent with a larger view of the environment may \textit{assist} the ground robot to find the target object by directing it in the right direction. This requires effective coordination to identify where the ground robot is on the map. VLMs can be helpful here for cross-view localization, but it requires \textit{semantic conversation} between the two agents. We investigate this further next.

\section{Semantic Communication for Assisting Robots with ObjectNav}
\subsection{Introduction}
\cleanfootnote{The work presented here is a result of equal contributions from Vishnu Dutt Sharma and Vishnu Sashank Dorbala.}
\label{sec:llm_comm:intro}
Finding and navigating to a given object without a pre-build map, or \textit{ObjectNav}, is a well-studied problem in Embodied AI \cite{anderson2018evaluation, objnavsurvey1}. To accomplish this task an embodied agent must interpret its surroundings based on its observations and move towards areas with a higher likelihood of containing the target object. Recently, several works have explored the zero-shot variation for this task \cite{yokoyama2023vlfm, dorbala2023can, kuang2024openfmnav}, coming up with various solutions for navigation without any fine-tuning or pre-training. This has involved using foundation models including vision-language models (VLMs) and large language models (LLMs) for commonsense reasoning and planning~\cite{sarch2023helper,padmakumar2023multimodal}. An embodied agent equipped with such models can thus reason about the possible locations of the target object with open-set scene understanding; evidenced by a proliferation of works on developing single-agent VLM-based embodied agents~\cite{shah2022lm,clipnav,shah2023navigation,pan2023langnav,gadre2023cows}.

 \begin{figure*}[ht]
    \centering
    \includegraphics[width=1.0\linewidth]{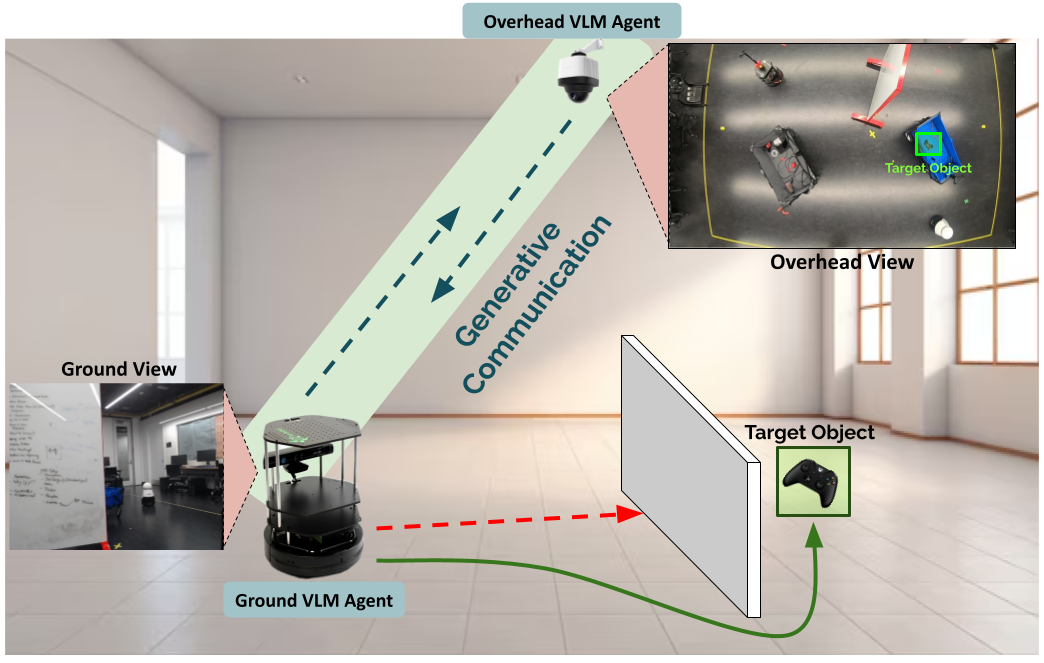}
    \caption[Overview of the assistive ObjectNav setup. The overhead and ground VLM agents communicate to coordinate and find the target object.]{\textbf{Overview}: Our work investigates \textit{Generative Communication (GC)} between VLM-based embodied agents performing ObjectNav. Specifically, we look at how factors such as \textit{preemptive hallucination}, \textit{communication length}, \textit{dialogue similarity}, and \textit{agent cooperation} influence overall ObjectNav performance. In the figure, both the overhead and the ground agent are equipped with VLMs that take their respective egocentric views as visual input. Given the target object label, the overhead and ground agents iteratively \textit{communicate} and provide actions for the ground agent to \textit{execute}. Note that the target object here is absent from the view of the ground agent, requiring concise navigational guidance from the overhead agent, and cooperation from the ground agent to reach the target.}
    \label{fig:llm_comm:overview}
\end{figure*}

While several approaches for ObjectNav have been introduced, zero-shot task performance remains poor, due to the difficulty in accurately estimating a target's location in a completely unknown environment~\cite{objnavsurvey1}. This challenge is further exacerbated in single-agent setups \cite{vlmobjnav2,vlmobjnav3}, with only one agent exploring the environment independently using a local view limited by the camera field-of-view and objects acting as occlusions. To improve zero-shot ObjectNav performance, we consider making better use of potentially available additional percepts in the environment in the form of an \textit{overhead agent} with a constrained global view, providing \textit{assistive} guidance to a \textit{ground agent} via generative natural language. Given the growing presence of indoor cameras often used for security, monitoring, and home assistants, it is both feasible and advantageous to integrate such inputs.

In collaborative setups such as ours, effective communication is necessary for meaningful collaboration ~\cite{commbase1}. The translation of complex perceptual cues into a shared and sufficiently descriptive language is a key skill in this regard. To this end, we consider a VLM-equipped ground and overhead agent in our setup. This allows agents to understand the scene from their respective views and share information through natural language to find a target object.  We study zero-shot ObjectNav performance through the lens of \textit{Generative Communication} (GC) between these agents.

A few prior works have studied a collaborative setup among embodied agents, particularly dealing with emergent communication \cite{emergentnatural, emergentsurvey1, emergentcontrol1}. These works however mainly deal with developing symbolic language with limited vocabularies for effective communication as opposed to using natural language, which offers a broader and interpretable vocabulary. \cite{guo2024embodied-mengdi} study emergent communication in embodied agents equipped with LLMs, but consider a predefined leadership hierarchy on \textit{homogeneous} agents of the same type. 

\noindent {\bf Main Results:} 
We present a \textit{coordinated} scheme for a zero-shot ObjectNav task, utilizing GC between a dynamic ground agent and a static overhead agent for generating actions. We investigate various \textit{traits} from the communication including \textit{preemptive hallucinations}, \textit{dialogue similarity}, \textit{communication length}, and \textit{agent cooperation}; evaluating the impact of these traits on zero-shot ObjectNav performance. Our work makes the following key contributions:
\begin{itemize}
    \item We tackle the zero-shot ObjectNav task in a novel \textit{assistive} setup, where the embodied ground agent seeks help from an overhead agent with a constrained global view.

    \item We establish two \textit{assisted} navigation schemes using Fully and Selective Cooperation on this setup, establishing a benchmark for ObjectNav in an \textit{assisted}, zero-shot setting. Our navigation schemes utilize \textbf{Generative Communication (GC)} between VLM-equipped ground and overhead agents.
    
    \item When compared with a non-assistive baseline, we observe a $\textbf{10\%}$ increase in performance while using Selective Cooperation. Conversely, we notice a $\textbf{13\%}$ drop in performance in a Fully Cooperative setup, suggesting that an embodied agent utilizing GC for improving ObjectNav must do so in a \textit{selective} manner. 

    \item We analyze dialogues in the GC, identifying and measuring a novel trait of ``\textit{\textbf{Preemptive Hallucinations}'' ($\mathcal{H_{PE}}$}) which directly impacts task performance. $\mathcal{H_{PE}}$ steadily increases from $\textbf{48.20\%}$ to $\textbf{81.60\%}$ with increasing lengths of GC and helps explain the behavior of our GC-driven assisted navigation schemes.  

    % \item 
    \item Finally, we perform a real-world inference on this task using a Turtlebot 2 as a Ground Agent and a GoPro Hero 7 as the Overhead Agent. We show that environment-specific fine-tuning of VLM prompts can help alleviate $\mathcal{H_{PE}}$ for real-world success. %reflecting this.
\end{itemize}

\subsection{Related Works}
\subsubsection{Ground-Overhead Agent Collaboration}
A team of embodied agents with different viewpoints can enhance team capabilities by combining their respective strengths. For example, in a team of a UAV and a UVG deployed for rescue in a disaster area, the UAV can act as a scout, observing a large area and guiding the UGV to reach affected people. The UGV can then perform long-range navigation and carry a heavy payload in the form of supplies, which a UGV may not be able to carry for a long time. Thus, effective collaboration would depend upon communication. Given the natural language-based communication capability of VLM-powered agents, it begs the question: \textit{How would a team of multiple VLM-equipped robots behave?} Historically, the information exchanged between agents in a multi-agent team has been handcrafted~\cite{commbase1}, or learned during task-specific training~\cite{zhou_graph_2022}. Natural language-based, generative communication for multi-agent teams of embodied agents is a fairly recent and relatively unexplored domain. Guo et al.~\cite{guo2024embodied-mengdi} study this situation and show that communication leadership improves team efficiency for such LLM-powered agents. However, they focus only on a team of homogeneous agents.

Chen et al.~\cite{chen2023scalable} studied natural language-based communication coming from LLMs for heterogeneous agents in various situations. Their focus, however, is on scalability and they perform experiments in simplistic grid-like environments. Our work instead focuses on embodied agents in indoor environments and specifically on how communication influences ObjectNav performance. Along with studying the communication length, similar to Chen et al.~\cite{chen2023scalable}, we also develop metrics to measure hallucination in generating communication originating from LLMs and its effect on collaboration. 

\subsubsection{Communication in Cooperating Agents}
When multiple agents communicate with each other to accomplish a task together, they effectively engage in a referential game~\cite{lazaridou2016multi}. Earlier approaches studying referential games used symbolic communication~\cite{havrylov2017emergence,lazaridou2018emergence,kottur2017natural}, but with the advancements in deep learning, attention shifted to communicating with embeddings, often learned over time with reinforcement learning~\cite{lazaridou2020emergent}. The development of language models then spurred an interest in natural language-driven methods ~\cite{lazaridou2020multi,kim2024knowledge}. Visual signals have also been explored to enhance emergent communication~\cite{evtimova2017emergent,mu2023ec2}. These works, however, involve learning to communicate, which forms the basis for emergent traits. Would we observe such traits when an LLM or a VLM is used in a zero-shot manner? \cite{guo2024embodied-mengdi} looked at some such traits but for homogeneous agents with a predefined leadership hierarchy. We aim to identify and analyze these traits for homogeneous agents in our approach. 

Patel et al.~\cite{commbase1} is the closest to our work, addressing the issues above by proposing a framework using limited pre-defined vocabulary acting as a black box for communication rendering limited specific to limited environments. Furthermore, they use specific types of objects (colored cylinders) as goals. In our work, we deal with the case of fully generated natural language not limited by vocabulary or environments, to find objects.

\begin{figure*}[t!]
    \centering
    \includegraphics[width=\linewidth]{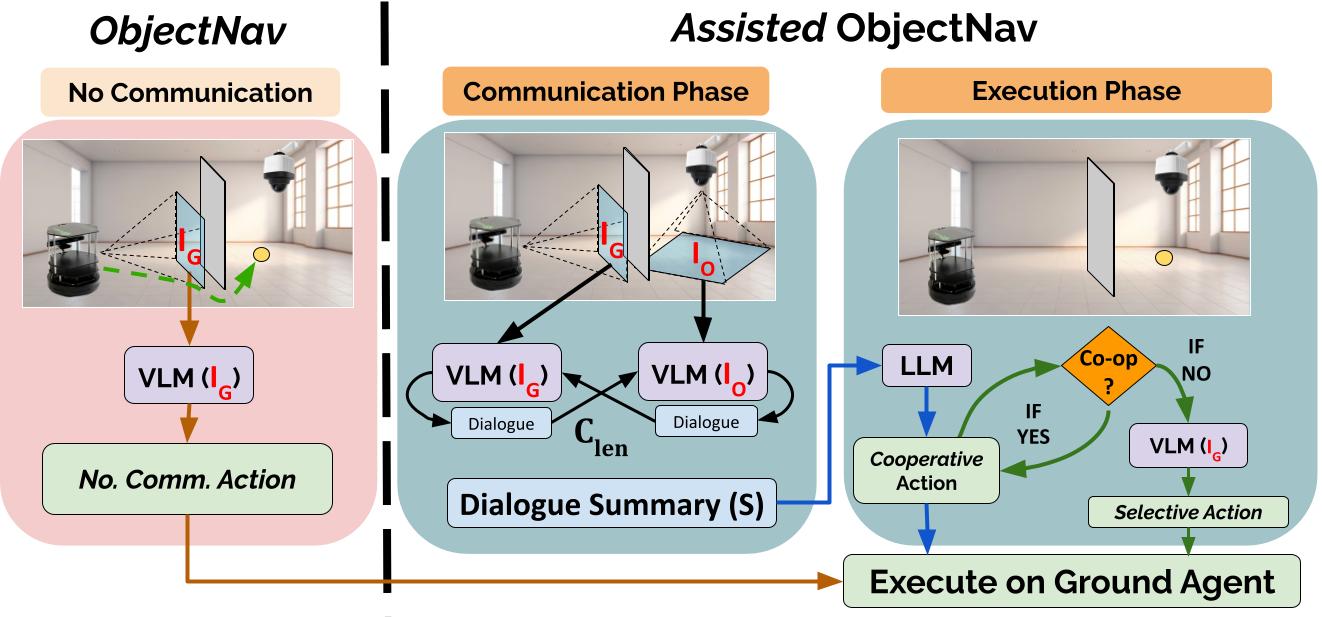}
    \caption[Overview of the proposed approach for assisted ObjectNav. The GA and OC first communicate with each other. A summary of the communication is then used to recommend an action for the GA. The GA then can either cooperate or decline cooperation and decide the next action on its own.]{\textbf{Approach}: We consider 3 different setups to measure \textit{generative communication} between a ground agent (GA) and an overhead agent (OA) performing ObjectNav. The \textbf{No Comm.} case (brown arrows) is a \textit{baseline} ObjectNav setup where the OA or the GA is directly prompted by a VLM for navigation actions for the ground agent. This is illustrated in the lefthand figure. For the remaining two cases, both agents first go through a Comm. phase $\mathcal{C}$) for a fixed number of interactions $C_{\text{len}}$. We then summarize the dialogues for decision-making and pass the dialogue summary $\mathcal{S}$ to the OA agent with an LLM for an action prediction. In the \textbf{Cooperative Action} case (blue arrows), the GA always follows the predicted/recommended action from the OA, thus fully \textit{cooperating} with it. In the \textbf{Selective Action} case (green arrows), the GA uses its view and the action recommended by the OA to decide whether it wishes to \textit{co-operate}. We later analyze the dialogue generated to measure generative communication traits.}
    \label{fig:llm_comm:flowchart}
\end{figure*}

\subsection{Proposed Approach: Assisted ObjectNav}
We propose VLM-based generative communication between two cooperating agents with access to different perspectives of the environment. We look at several cooperation setups and show how communication unfolds and affects ObjectNav. For this, we consider the following setup: \\

\noindent\textbf{Task}: We consider a variation of the ObjectNav task, called \textit{assisted} ObjectNav \cite{commbase1}, where the embodied agent on the ground is assisted by another agent with a different viewpoint of the target. In this dual-agent setup, both agents are expected to communicate and cooperate to identify and navigate toward a target object.

\noindent\textbf{Agents}: We consider two VLM-equipped agents: \\
1) A \textbf{Ground Agent (GA)} that has a \textit{local view} of the environment, and can perform physical maneuvers to reach the target. This is the embodied agent physically performing the ObjectNav task.
2) An \textbf{Overhead Agent (OA)} that has a constrained \textit{global view} of the environment containing the target object. This agent is incapable of physical maneuvers and aims to assist the GA reach the target.\\

As a baseline, we consider a no-communication setup, the GA independently aims to perform ObjectNav with VLM. In the assisted ObjectNav setup, our approach involves an interplay between two phases: a \textbf{Communication Phase, $\mathcal{C}$}, where both agents generate dialogue with information about their respective egocentric views, and an \textbf{Execution Phase, $\mathcal{E}$}, where the GA gets receives a maneuvering \textit{recommendation} to the other agent, which must then decide what action the ground agent must execute.

Figure \ref{fig:llm_comm:overview} presents an overview of our setup for the baseline ObjectNav and the assisted ObjectNav tasks.

\subsubsection{No Communication Baselines}

We consider a baseline case where the GA uses its VLMs to guide the ground agent independently toward the target object. The agent processes its visual input via a VLM along with a prompt to get a directional command to execute. The prompts used are present in Appendix \ref{sec:llm_comm:append:llm_comm:prompt}. Additionally, we present a \textbf{Random} decision-making approach where the GA randomly chooses and executes a directional command.

\subsubsection{Communication Phase (\texorpdfstring{{$\mathcal{C}$}}{C})}
Improving collaborative task success requires agents to both gather information about the environment with their percepts as well as effectively translate perception data to robot language. Both the GA and the OA are equipped with VLMs capable of generating natural language conversations. The GA seeks assistance from the OA and prompts the OA for a dialogue. We specify to the OA that it must communicate for a length of $C_{\text{len}}$, asking the GA queries that would help guide it to the target object. The GA then responds to the generated dialogue, and this cycle happens back and forth for $C_{\text{len}}$ times. At the end of the communication period, we concatenate all the dialogues to produce a summary, $\mathcal{S}$, containing potentially useful information about the environment for making navigation decisions. $\mathcal{S}$ represents the generative communication, GC, in our approach.

\subsubsection{Execution Phase (\texorpdfstring{$\mathcal{E}$}{E})}
While successful communication protocols have traditionally been learned using RL and improve performance on the heterogeneous ObjectNav task (\cite{commbase1}), to investigate the performance of \textit{generative communication} in our fully generative setup, we consider the two zero-shot execution variants: \textbf{Cooperative Action} and \textbf{Selective Action}.

In the \textbf{Cooperative Action} setup, the agents exchange dialogue of varying lengths ($C_{\text{len}}$) in $\mathcal{C}$. After this, in $\mathcal{E}$, a summary of the dialogues, $\mathcal{S}$, is passed to the OA looks only at $\mathcal{S}$, and suggests an action for the GA to take. The GA thus \textit{fully cooperates} with the OA and blindly executes the prescribed action.

In the \textbf{Selective Action} setup, the GA is presented with a choice on whether it wishes to cooperate with the OA's decision. After the comm. phase, $\mathcal{C}$, The GA is prompted asking for a ``Yes/No'' on whether it should cooperate with OA's recommendation. If it decides to cooperate (Yes), then the action recommended by the OA is executed. If the GA declines to cooperate (No), then we again prompt it to ask for its recommendation on the action to take given its image input. The former allows us to measure a \textit{cooperation rate} or $\text{Co}_{r}$ of the agents, while the latter is similar to the no comm. phase, where the action is taken directly from either VLM agent.

Both \textbf{Cooperative Action} and \textbf{Selective Action} are situations where the GA must communicate with the OA. We additionally investigate a scenario where the GA first decided whether to initiate a conversation with the OA. If it chooses to ask the OA for assistance then the rest of the process is the same as \textbf{Cooperative Action}. If the GA declines to seek help, then it makes decisions on its own, effectively acting in a \textit{No Communication} setup. We use the dialogue summary to measure generative communication traits between the VLM-based agents.

\subsubsection{Generative Communication Traits}
\label{subsec:llm_comm:GC_traits}
In this section, we look at quantifying various qualities that emerge during the VLM-based generative communication between the agents. We derive these from the dialogue summary $\mathcal{S}$ and use them to assess the influence of GC on ObjectNav performance.

\begin{figure*}[t!]
    \centering
    \includegraphics[width=\linewidth]{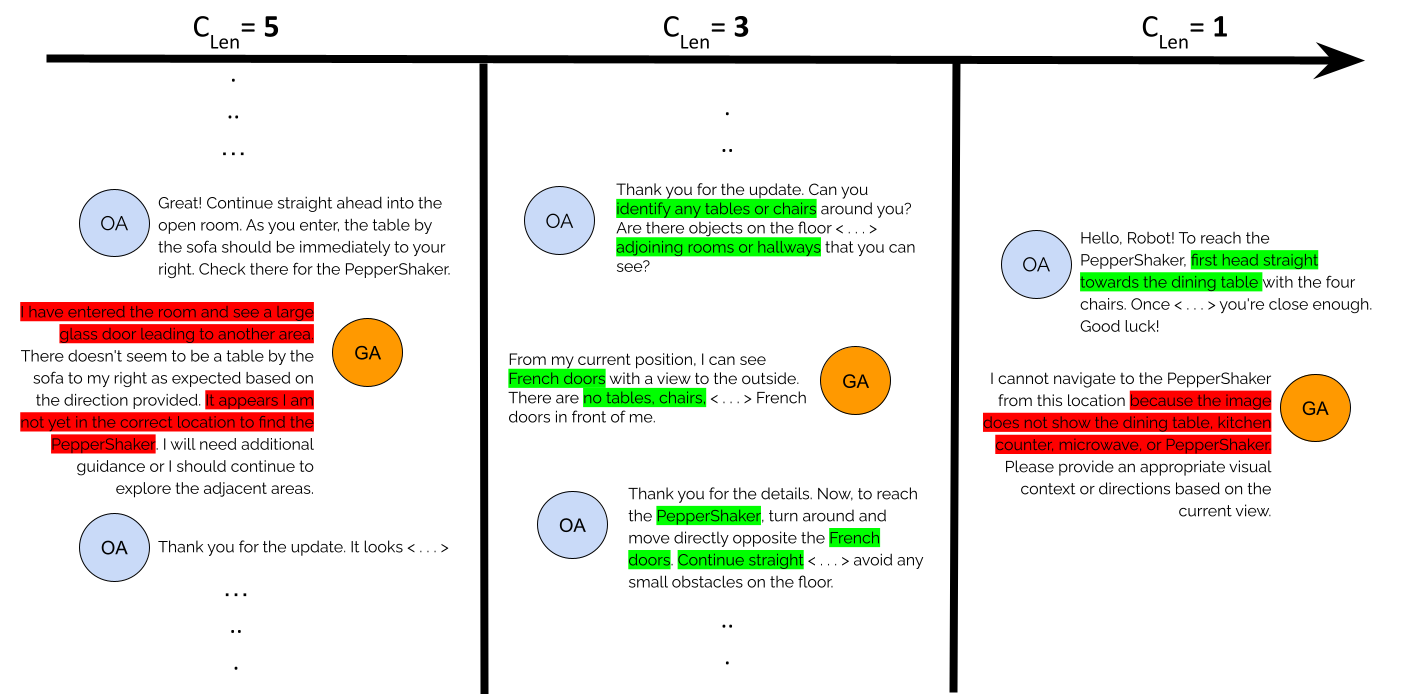}
    \caption[An example showing dialogues hallucinations across different lengths of communication between the ground and overhead agents.]{\textbf{Dialogue hallucinations $\mathcal{H}$}: We study the hallucination in the dialogues to explain the improved ObjectNav performance. In this figure, the target object is a \textit{PepperShaker}. On the $\mathcal{C}_{Len} = 5$ case on the left, note the pre-emptive hallucination on the GA, where it claims to have entered a room. In a $\mathcal{C}_{Len} = 3$ example, we see an ideal conversation about visual observations (\textit{French Doors, tables, chairs}), and a discussion about how to get to the target object. In the $\mathcal{C}_{Len} = 1$ case, being aware of the shorter $\mathcal{C}_{Len}$, the OA attempts to directly command the GA. This fails because of poor OA localization in estimating the GA's position.}
    \vspace{-0.5cm}
    \label{fig:llm_comm:hallucination}
\end{figure*}

\subsubsection{Dialogue Hallucinations - \texorpdfstring{$\mathcal{H}$}{H}}
Motivated by the high human-verified GPT-4 classification accuracy reported by \cite{guo2024embodied-mengdi} on dialogue analysis, we adopt a GPT-4 classifier for finding hallucinations in $\mathcal{S}$. 
We look at the following - \\
\noindent\textbf{Pre-emptive Actions}  ($\mathcal{H}_{PE}$): We notice in many cases that one of the agents assumes during $\mathcal{C}$ that the GA has moved, even though the GA can only move after the completion of $\mathcal{C}$. We refer to these as \textit{Pre-emptive Actions}. To quantify it, we prompt a GPT-4 agent to classify whether the dialogue in an episode indicates a preemptive action hallucination. The labels obtained are averaged over all the conversations. This can be defined as - 
\[
\mathcal{H}_{PE} = \frac{1}{N}\sum^{N}_{i=1}\frac{\mathcal{L}(S)_{i}}{k}
\]
where $\mathcal{L}(S)$ is the GPT-based classifier for pre-emptive actions. The prompts we use are discussed in Appendix \ref{sec:llm_comm:append:llm_comm:prompt}.

\noindent\textbf{Ghost Objects} ($\mathcal{H}_{GO}$): Here, we use BERT to identify the perceptual hallucination of the objects that are not present in the environment. We present BERT with a vocabulary consisting of objects in the environment along with the dialogue generated and look for overlap between the two. We then average this number across all conversations and report its complement, i.e. 100 - \textit{overlap}\%. This quantifies the conversation about objects that are not in the environment and thus can not be observed by either agent.\\

\subsubsection{Communication Length - \texorpdfstring{$\mathcal{C}_{Len}$}{Clen}}
The length of communication, $\mathcal{C}_{Len}$, sets the limit on the amount of information that can be exchanged between the two agents (especially with a fixed number of generated tokens). We ablate on $\mathcal{C}_{Len}$ of $1$, $3$, and $5$ for our approach and study its effect on other metrics.

\subsubsection{Cooperation Rate - \texorpdfstring{$\mathcal{CR}$}{CR}}
In the \textit{Selective Action} setting, the GA may choose to not comply with the OA's suggested action, instead choosing to act independently. We measure this cooperation rate as the number of times the agent chooses to cooperate, averaged over an episode. 
\[
\mathcal{CR} = \frac{1}{N}\sum^{N}_{i=1}\frac{c_{i}}{k}
\]

where $c_{i}$ is the number of times the GA decided to cooperate in the $i^{\text{th}}$ episode with $k$ steps, and $N$ is the total episode count.

A lower cooperation rate would indicate more independent exploration, with the GA choosing to ignore OA's suggested actions. Further, this could be an indirect indicator for hallucinated communication resulting in unfeasible actions.

\subsubsection{Dialogue Similarity - \texorpdfstring{$\mathcal{DS}$}{DS}}
A higher communication length would only prove useful if the agents exchange unique information over time to improve their understanding of the environmental state and make sound decisions.

We quantify this uniqueness by computing the cosine similarity between dialogues across episodes at each step. This metric gives us insight into the general trend of agent conversations, allowing us to gauge the \textit{structure} of the GC. We measure this as - 
\[
\mathcal{DS}_{k} = \frac{1}{N}{\sum_{i=0}^{N} CS(\text{E}_{0},\text{E}_{i})}
\]

where $\mathcal{DS}_{k}$ represents the dialogue similarity at the $k^{th}$ step across all episodes, and $\text{E}_{i}$ represents the BERT embedding of the $i^{th}$ sentence.

\subsection{Experiments and Evaluation}
We measure ObjectNav and GC performance via experiments in RoboTHOR as well as a real-world setup. RoboTHOR allows us to conduct large-scale experiments in a simulated environment, allowing us to objectively measure task performance via Oracle Success Rate \textbf{(OSR)} and Success Weighted by Path Length \textbf{(SPL)} as defined by \cite{anderson2018evaluation}. In the real world, we present some unique qualitative findings on differences in the GC and visual localization of the overhead agent in contrast with simulator experiments. In both cases, we compute and validate the previously defined GC Traits.

\begin{table*}[ht]
\centering
\begin{tabular}{cccrrr}
\toprule
\textbf{Comm. Length ($\mathcal{C}_{Len}$)} & \textbf{Execution} & \textbf{OSR \%} $\uparrow$ & \textbf{SPL \%} $\uparrow$ \\
\midrule
\multirow{2}{*}{0} & Random (No VLM) & 15.00 & 11.69\\
 & No Comm. & 22.00 & 21.50\\
% \midrule

\specialrule{.3em}{.05em}{.05em}

\multirow{2}{*}{1}
                    & Cooperative Action & 19.00 & 18.38 \\
                    & Selective Action & 26.00 & 24.13 \\
\midrule
\multirow{2}{*}{3}
                    & Cooperative Action & 19.00 & 17.94 \\
                    & Selective Action & 24.00 & 22.19 \\
\midrule
\multirow{2}{*}{5}
                    & Cooperative Action & 9.00 & 8.50 \\
                    & Selective Action & \textbf{32.00} &\textbf{29.15} \\
\bottomrule
\end{tabular}
\vspace{0.3cm}
\caption[Oracle Success Rate (OSR) and Success Weighted by Path Length (SPL) for the assisted ObjectNav in simulation.]{\textbf{Simulator ObjectNav Performance}: We measure Oracle Success Rate (OSR) and Success Weighted by Path Length (SPL) across different $\mathcal{C}_{Len}$. Note the improved performance in the Selective Actions case, where the GA is given a choice to reject or accept OA recommendations. This suggests the best-performing agent is one that shows a balance between fully \textit{independent} and fully \textit{cooperative} exploration.}
\label{tab:sim_objectnav}
\end{table*}

\begin{table*}[ht]
\centering
\resizebox{\textwidth}{!}{
\begin{tabular}{cccrrr}
\toprule
\textbf{Comm. Length ($\mathcal{C}_{Len}$)} & \textbf{Execution} & \textbf{$\mathcal{H}_{PE}$ \%} $\downarrow$ & \textbf{$\mathcal{H}_{GO}$ \%} $\downarrow$ &\textbf{ $\mathcal{CR}$ \%} &\textbf{$\mathcal{DS}$\%} \\
\midrule
\multirow{3}{*}{1}   
    & Cooperative Action &         \textbf{48.20} & 34.55 & 100 & 91.5 \\
    & Selective Action &         50.50 & 34.55 & 23.20 & 92.2 \\
\midrule
\multirow{3}{*}{3}
    & Cooperative Action &         77.90 & \textbf{32.73} & 100 & 91.6\\
    & Selective Action &         78.00 & 34.55 & 22.80 & 91.6 \\
\midrule
\multirow{3}{*}{5}
    & Cooperative Action &         81.60 & \textbf{32.73} & 100 & 92.4 \\
    & Selective Action &         80.70 & \textbf{32.73} & 18.90 & 92.1 \\
\bottomrule

\end{tabular}
}

\caption[Generative communication traits in simulations. We find that while the VLMs do not hallucinate much in describing images, they progressively perform worse in assuming pre-emptive agent motion during communication.]{\textbf{GC Traits in Simulations}: We measure 4 different GC traits. Note the increasing $\mathcal{H}_{PE}$ as $\mathcal{C}_{Len}$ increases while the $\mathcal{H}_{GO}$ remains almost constant. We can infer that while the VLMs do not hallucinate much in describing images, they progressively perform worse in assuming pre-emptive agent motion during communication. The decreasing cooperation rate $\mathcal{CR}$ in the Selective Action case further suggests that when given a choice to cooperate, the GA would not trust actions produced from hallucinated summaries. This also explains the superior performance of using Selective Actions over Cooperative Actions.}
\label{tab:sim_gc}
\end{table*}

\begin{figure*}[t!]
    \centering
    \includegraphics[width=\linewidth]{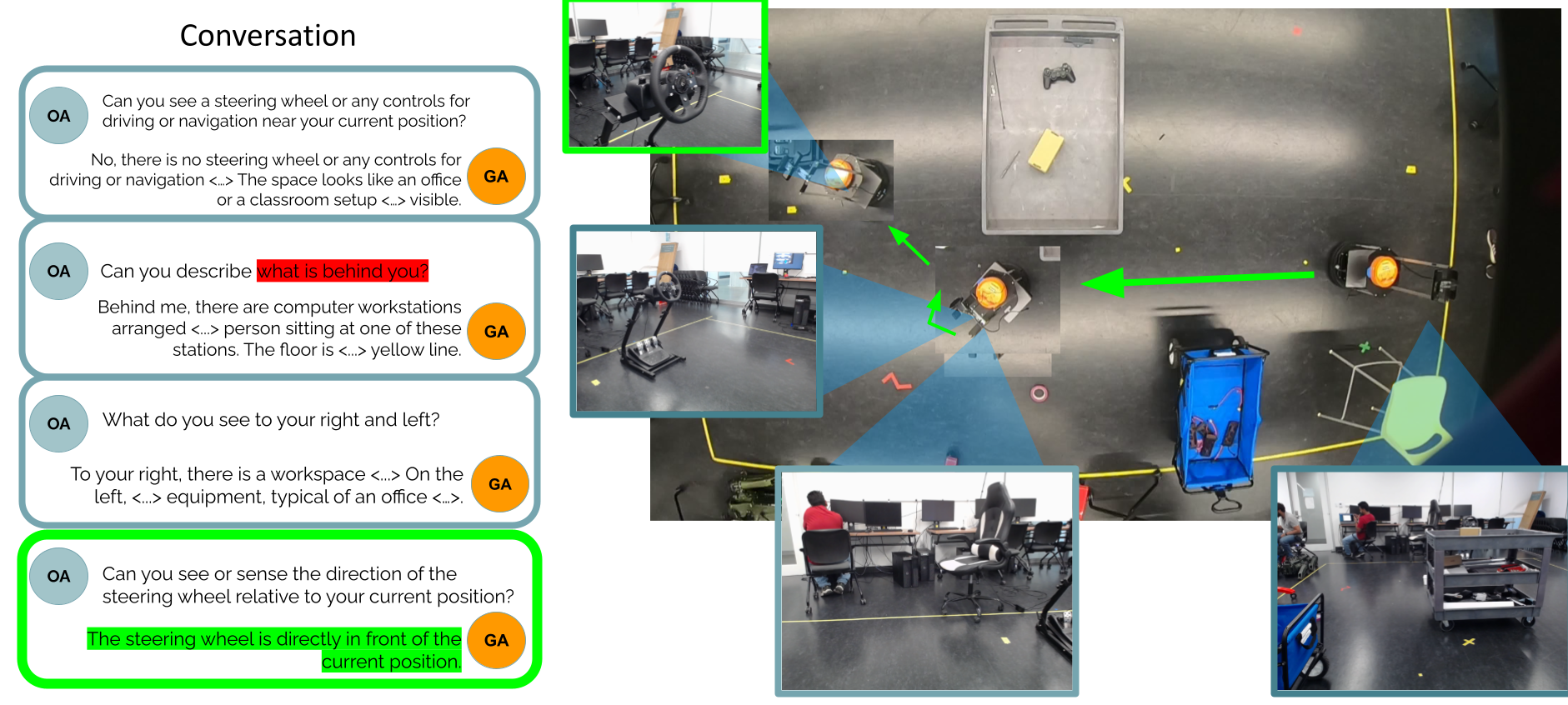}
    \caption[Overview of the real-world experimental setup consisting of a Turtlebot as a Ground Agent (GA) and a GoPro camera mounted to the roof as an Overhead Agent (OA).]{\textbf{Real World Setup}: We carry out a real-world experiment on a Turtlebot as a Ground Agent (GA) and a GoPro camera mounted to the roof as an Overhead Agent (OA). Notice the second dialogue on the left (in red), where the OA asks the ground agent an infeasible query about describing what is behind it, following which it describes what is actually in front of it. We notice these types of real-world hallucinations and fine-tune the VLM prompts for better results.. Appendix \ref{sec:llm_comm:append:llm_comm:localization_fail} covers these cases.}
    \label{fig:llm_comm:rw_setup}
\end{figure*}

\subsubsection{Simulation: RoboThor}
\label{subsec:llm_comm:simualtion_setup}
\noindent\textbf{Setup}:
We use 100 rooms from the training split of ProcTHOR~\cite{deitke2022️} for our experiments. In each room, we place a LoCoBot robot with an RGB camera as the GA and add an overhead camera. The overhead camera, our OA, is located at a height of 7.25m. We crop the images from the overhead camera to minimize the views outside the house before passing them to the VLM. We use \textit{GPT-4-turbo} as the VLM in all the experiments. 

At the start of each episode, the GA spawns at the default location for the room in the ProcTHOR setup. As the target object, we randomly select an object from the scene such that only one instance of it is available. We take 10 action steps during each episode and keep track of its outcome. The ground robot is allowed to take one of the following actions: \textit{MoveAhead}, \textit{MoveBack}, \textit{RotateLeft}, \textit{RotateRight}, or \textit{DoNothing}. All the movements result in the GA moving by 0.25m in the indicated direction. The rotation actions result in a $90^\circ$ rotation while staying at the same place. \textit{DoNothing} lets the GA stay where it is without any change.

During the communication phase, $\mathcal{C}$, the agents use their respective camera views and talk to each other to devise a plan to reach the target object. Then we share the dialogue, $\mathcal{S}$, between the agents, the past action, and the result of it (success, collision, etc.) with the decision-making agent during the execution phase $\mathcal{E}$. In a non-cooperative setup, the agent uses its camera view to decide the appropriate action to reach the target object. 

\noindent\textbf{Inference}:
Table \ref{tab:sim_objectnav} presents simulation results on Zero-Shot ObjectNav. We make two inferences: \textbf{1)} An agent using Selective Actions performs far better than one that blindly follows OA suggestions (Cooperative Action). Further, blind following performs even worse than if the agent were not communicating at all. This holds true across all $\mathcal{C}_{Len}$, showcasing that even when assisted, an agent must be provided with the choice to reject suggestions and independently explore to improve performance. \textbf{2)} Note the drastic drop in Cooperative Action performance at $\mathcal{C}_{Len} = 5$, especially with SPL. We can infer here that actions predicted from longer communications tend to be worse and negatively impact the performance of a fully cooperative agent.

Table \ref{tab:sim_gc} presents our inspection of the dialogue summaries $S$ obtained for the GC traits discussed in subsection \ref{subsec:llm_comm:GC_traits}. We can clearly see the increasing pre-emptive hallucination behaviour $\mathcal{H}_{PE}$ with increasing $\mathcal{C}_{Len}$, while $\mathcal{H}_{GO}$ remains almost constant. This indicates that while GPT-4V is consistent with regards to describing the content in images, it tends to hallucinate a lot more when it comes to generating \textit{pre-emptive actions}. Examples of such conversations are provided in Figure \ref{fig:llm_comm:hallucination}. Despite being explicitly prompted to converse only about visual elements during the conversation, and being told that the agent cannot execute actions, GPT-4 tends to hallucinate in assuming that the agent has moved.

Further, note the decreasing Cooperation Rate $\mathcal{CR}$, with increasing hallucinations. This relation explains the improved ObjectNav performance, showing that an agent with Selective Actions cooperates less with hallucinated instruction. Note the dialogue similarity $\mathcal{DS}$ is almost similar across all approaches, and as such is not a good metric to evaluate the dialogue sensitivity. Our Pre-emptive Hallucination metric $\mathcal{H}_{PE}$ is far more informative in allowing us to gauge the performances of these approaches. 

\subsubsection{Real World Experiments}

\noindent\textbf{Setup}: We recreate our dual agent setup in the real world using Turtlebot 2 as the ground agent and a GoPro Hero 7 as the overhead agent. Figure \ref{fig:llm_comm:rw_setup} showcases our setup, and more examples are in Appendix \ref{sec:llm_comm:append:llm_comm:localization_fail}. We arrange objects in our lab environment as obstacles asking the agent to find targets of various shapes and sizes such as \textit{white basket}, \textit{red ball}, \textit{steering wheel}, \textit{yellow chair} etc. We make the following inferences:
\begin{itemize}
    \item \textbf{VLM Localization}: We notice that GPT-4 is not good at localizing target objects from an overhead view. This often hinders guidance that is being given, as the overhead agent does not have a good intuition about where the robot or the target object lies, and motivates our heterogeneous two-way communication setup. For instance, the OA might learn from the GA while communicating about an object in its vicinity, and this information would help it localize. Examples of this are provided in Appendix \ref{sec:llm_comm:append:llm_comm:localization_fail}.
    \item \textbf{Finetuning Prompts}: Pre-emptive hallucinations lead to poor task performance in the simulator, as the conversation turns defensive with agents hallucinating over pre-emptive actions. In the real world, we finetune prompts to alleviate some of these hallucinations by explicitly specifying pose conditions to the agent. These are discussed in Appendix \ref{sec:llm_comm:append:llm_comm:localization_fail}.
\end{itemize}

\subsection{Conclusion}
We propose enhancing the zero-shot ObjectNav performance for an embodied ground agent through assistance from an overhead agent via VLM-based generative communication. We show that a naive communication-based solution does not work well and presents a selective action-based approach, improving the ground agent's ability to find the target object. Additionally, we develop a novel metric to detect `pre-emptive action' hallucinations within the communication between the embodied agents and demonstrate its significant impact on the ObjectNav performance compared to other language-based communication challenges. Finally, we implement our framework in a real-world scenario with a TurtleBot 2 robot, assisted by a GoPro Hero 7 camera as the overhead agent, and utilize our insights to enhance ObjectNav in the real world with prompt fine-tuning. We hope our findings will lead to the development of more robust methods for zero-shot, cooperative embodied agent tasks.

% \section{Limitations and Future Work}
Hallucination is an often-encountered challenge in LLMs and VLMs and affects our setup as well. We quantify a specific type of hallucination  $\mathcal{H}_{PE}$ for embodied agents, which captures the incorrect assumption about the motion of the GA even though it has not moved. We believe there may be other types of hallucinations specific to embodied AI that need identification and address. We aim to investigate these in our future work through extensive experimentation.

Additionally, we found that the lack of grounding in VLMs may cause localization failures, hampering the OA's ability to correctly guide the GA to the target object.  We will explore methods to address this issue in our future work with the help of unique and distinguishable markers, such as AR tags which are often used in robotic applications. 

%% file: Chapter8.tex
%Chapter 8

\renewcommand{\thechapter}{8}

\chapter{Meta-Reasoning for Risk Management with Implicit Measures}
\label{chapter:risk_aware}

\section{Introduction}\label{sec:intro}
\cleanfootnote{The work presented here is a result of equal contributions from Vishnu Dutt Sharma and Lifeng Zhou.}
In the previous chapters, we present methods grounded in learning-based predictions to enhance a robot's efficiency in completing the task at hand. A basic assumption behind the methods presented in these works is that the predictions can be trusted. However, when a prediction-based method encounters out-of-distribution inputs, it may result in incorrect predictions and cause harm to the people and itself. 

Consider scenarios where an autonomous ground vehicle must navigate in an unknown environment. Examples include search-and-rescue, space exploration, and disaster response. For instance, consider a disaster response scenario where ground vehicles must supply resources to specific demand locations as soon as possible. In such settings, prior GPS or satellite maps of the environment may no longer be valid. Instead, an aerial robot may be employed to take live aerial images which can then be used to plan the paths of the ground vehicles towards the demand locations. However, due to the inherent uncertainty of aerial images, the paths that are found may not actually represent the situation on the ground. Therefore, there is a risk of the vehicles taking longer to reach the demand positions than planned. In safety-critical situations, one way to mitigate the risk is to assign multiple vehicles to the same demand, with the earliest arriving one actually responding to the demand.

Motivated by this scenario, we study the problem of how to find risk-aware paths for multiple vehicles to serve multiple demand locations. There are two problems to be solved --- assigning the vehicles to the demand locations and finding risk-aware paths from start to demand locations. We present a risk-aware framework to solve both problems simultaneously.

The environment where the vehicles navigate is captured by an overhead image. We implement a deep learning technique for semantic segmentation of the overhead image. Due to the uncertainty from segmentation, the travel cost of the vehicle turns out to be a random variable. Built on our previous work~\cite{Toubeh2018}, our first contribution is to show how to utilize Bayesian deep learning techniques to handle the risk from the planning and perception level. After risk-aware planning and perception, we generate a set of candidate paths corresponding to different risk levels from each vehicle's start position to each demand location. Our second contribution is to assign each vehicle to a risk-aware path from its candidate path set to a demand. We utilize a risk measure, Conditional-Value-at-Risk (CVaR), to manage the risk from the uncertainty. Our assignment framework provides the flexibility to trade-off between risk and reward, which builds on our previous work~\cite{zhou2018approximation}, with risk here being assessed at multiple levels of the algorithm.
 Our empirical results show that this risk-aware framework results in safer path planning and assignment. 

\paragraph*{Related Work} 
Deep learning has shown significant improvements in perception capabilities for many robotics applications. However, the potential of the positive impact deep learning may have on real-world scenarios is inevitably proportionate to their interpretability and applicability to imperfect environments. In these cases, deep neural networks can even misrepresent data outside the training distribution, giving predictions that are incorrect without providing a clear measure of certainty associated with the result \cite{gal2017phd}. The extraction of uncertainty information, as opposed to the reliance on point estimates, is crucial in safety-critical applications, such as autonomous navigation in an urban, unstructured setting. Methods like Natural-Parameter Networks \cite{wang2016naturalparameter} propose modeling the network parameters and inputs as Gaussian distributions. However, these modifications impose huge computation costs on the model due to the increment in the number of trainable parameters. Lightweight Probabilistic Deep Networks \cite{Gast_2018} alleviate this concern to some extent by making the weights deterministic. Large size networks are also unsuitable for real-time robotics applications which may have constraints on inference time and memory. \cite{gal2016dropout} and \cite{Loquercio_2020} propose methods that allow uncertainty extraction from deep learning models, specifically those that do not interfere with the overall structure or training process. In this work, we leverage \cite{gal2016dropout}, which shows that dropouts, which are often used as a regularization enhancement in neural networks, can provide approximate Bayesian inference and thus help in uncertainty estimation for the deep learning models.   

In addition to considering uncertainty at the perception level, we also utilize some popular risk measures to handle uncertainty at the planning and assignment levels. A typical measure for optimization under uncertainty is the expectation of a stochastic function. However, the expectation is a risk-neutral measure and may not be desirable, especially in critical tasks, like search-and-rescue operations~\cite{majumdar2020should}. For example, we may find a path with a lower expected cost but with high variance. It is quite likely that the vehicle may encounter a much larger cost (as compared to the expected one) when traveling on this path. Thus, instead of using expected cost, we utilize some other risk-aware measures, such as mean-variance~\cite{marcus1997risk,chung2019risk} and conditional-value-at-risk (CVaR)~\cite{rockafellar2000optimization,majumdar2020should}. 

In particular, we use mean-variance as the risk-aware cost measure in the A* planner~\cite{lavalle1998rapidly} for planning paths for the vehicles. The mean-variance measure allows us to balance the mean cost and uncertainty (variance) when planning paths. Also, for the path assignment, we use CVaR to deal with the uncertainty on the path level. $\text{CVaR}_{\alpha}$ explicitly takes into account the risk associated with bad scenarios~\cite{rockafellar2000optimization, majumdar2020should}. Specifically, $\text{CVaR}_{\alpha}$ measures the expectation of a random variable in the $100\alpha$--percentile worst scenarios. Here, the user-defined risk level, $0<\alpha\leq 1$, provides a user with the flexibility to choose a risk that they would like to take. Setting $\alpha=1$ makes $\text{CVaR}_{\alpha}$ equal to the expectation whereas $\text{CVaR}_{\alpha}\approx 0$ is akin to worst-case optimization.

Risk in autonomous systems ultimately involves the risk of failure, whether due to damage to the agents or their environment or mission failure. Several risk-aware planning frameworks have been proposed for collision avoidance, such as the dynamic risk density function in ~\cite{pierson2019dynamic} which infers congestion cost from sensor inputs. Our navigation cost function observes the risk associated with noisy sensor perception and does not assume sensor perception is always correct. Uncertain semantic maps have also been proposed for perception-based planning. Both ~\cite{kantaros2019optimal} and ~\cite{fu2016optimal} model an unknown environment as a semantic map by assuming Gaussian distributions over landmark positions for simultaneous localization and planning (SLAM). These address the uncertainty in planning, but not in perception, i.e. uncertain landmark classification.

\textbf{Contributions.} In this chapter, we have three main contributions. 
\begin{itemize}
    \item We present a framework that plans and assigns risk-aware paths for robots that navigate in unknown environments. 
    \item We utilize the Bayesian deep learning technique to learn an unknown environment whose information is only available by an overhead, georeferenced image. 
    \item We deal with the uncertainty at both path planning and assignment levels by optimizing the corresponding risk measures. In the end, we assign each vehicle a risk-aware path to a demand location, and the path assignment is guaranteed to have a bounded approximation performance of the optimal assignment.
\end{itemize}
This work builds on our prior work where we studied these two problems (uncertainty extraction from deep learning and CVaR optimization) individually. Here, we investigate the joint problem. We find that utilizing the uncertainty extraction from deep learning and managing the risk from uncertainty by CVaR optimization provide the vehicles with safer and risk-aware paths in unknown environments.

\section{Preliminaries}\label{sec:background}
We start by defining the notations used in the chapter. We give background on a risk measure, conditional-value-at-risk (CVaR). We then provide a formal definition of the joint problem of planning and assignment.

\subsection{Conditional Value at Risk}~\label{subsec:CVaR}
Let $X$ be a random variable. $\text{CVaR}_{\alpha}(X)$ denotes the expectation on the $\alpha$-worst scenarios of the utility or cost function $f$ with $\alpha \in (0,1]$. More specifically, if $X$ indicates reward or benefit,  $\text{CVaR}_{\alpha}(X)$ denotes the expectation on the left $\alpha$-tailed scenarios. While, if $X$ represents loss or penalty, $\text{CVaR}_{\alpha}(X)$ is the expectation on the right $\alpha$-tailed cases. Here, $\alpha$ is the confidence level or the risk level. If $\alpha$ is close to $0$, $\text{CVaR}_{\alpha}$ is close to the worst-case. If $\alpha$ is equal to $1$, $\text{CVaR}_{\alpha}$ is same as the expectation. 

In this chapter, the utility function $f(\mathcal{S},y)$ defined on set  $\mathcal{S}$ is a random variable with randomness induced by parameter $y$. Since $f(\mathcal{S},y)$ is utility that indicates benefit, $\text{CVaR}_{\alpha}[f(\mathcal{S},y)]$  denotes the expectation on the left $\alpha$-tailed cases, as shown in  Figure~\ref{fig:risk_aware:cvar}. 

\begin{figure}[h]
  \centering
\includegraphics[width=0.75\textwidth]{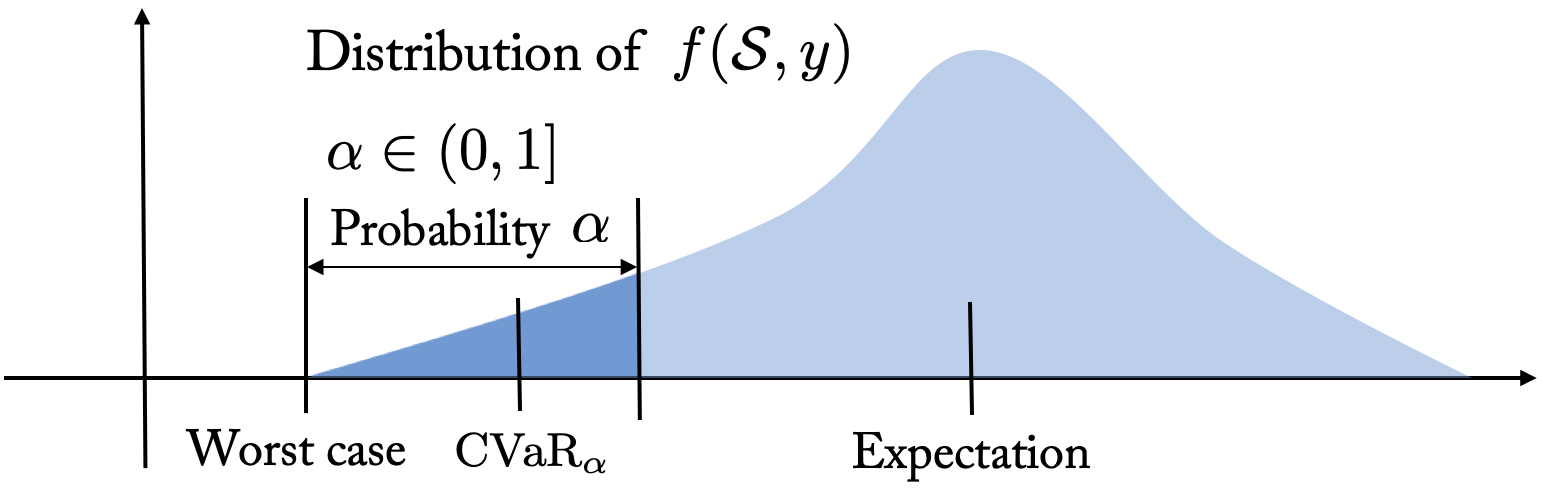}
  \caption[An illustration showing $\text{CVaR}_{\alpha}$ of function $f(\mathcal{S},y)$.\label{fig:risk_aware:cvar}. It denotes the expectation on the left $\alpha$-tailed cases.]{$\text{CVaR}_{\alpha}$ of function $f(\mathcal{S},y)$.\label{fig:risk_aware:cvar}.}
\end{figure}

We generally maximize $\text{CVaR}_{\alpha}[f(\mathcal{S},y)]$ by solving 
\begin{equation}
  \max_{\mathcal{S}, \tau} ~\tau - \frac{1}{\alpha}\mathbb{E}[(\tau-f(\mathcal{S},y))_{+}],
  \label{eqn:cvar_auxiliary}
\end{equation}
where $\mathcal{S}$ is a decision set (or solution set), $(t)_+ = \max(t,0)$, and $\tau\in \mathbb{R}_+$ is an auxiliary parameter. For the ease of expression, we define 
\begin{equation}
    H(\mathcal{S},\tau) = \tau - \frac{1}{\alpha}\mathbb{E}[(\tau-f(\mathcal{S},y))_{+}]
    \label{eqn:aux_fun}
\end{equation}

\subsection{Problem Formulation}~\label{sec:prob_form}
We consider the problem of finding paths for multiple vehicles to serve multiple demand locations. In particular, we are given $N$ vehicles' start positions, $\mathcal{V}=\{1,\cdots, N\}$ and $M$ demand locations, $\mathcal{D}=\{1, \cdots, M\}$ in the environment. The environment is represented by an overhead, georeferenced RGB image as shown in Figure~\ref{fig:risk_aware:overview}. The goal is to find offline paths for each vehicle such that they collectively serve all the demands using navigation costs derived from the overhead images. 

The cost of a path in the environment can be estimated by first performing a semantic segmentation of the overhead image. However, semantic segmentation is typically imperfect~\cite{gal2017phd}, and as such the estimated cost of a path may not be accurate. The problem we address in this chapter is that of finding paths for vehicles to collectively serve all demands under travel-cost uncertainty.  

We are motivated by tasks that are urgent and time-critical, such as fighting fires~\cite{harikumar2018multi} and delivering medical supplies in emergencies~\cite{ackerman2018medical}. When the number of vehicles is more than the demands, assigning multiple redundant vehicles to demands helps counter the effect of uncertainty~\cite{prorok2019redundant}. The waiting time at a demand location is the time taken by the earliest vehicle to arrive at that location. If the travel times are deterministic, then it is known in advance which vehicle will arrive first. When travel times are uncertain, as in this work, the arrival time of the earliest vehicle itself is a random variable. The goal is to assign vehicles to demand locations and find corresponding paths for the vehicles from the start to the assigned demand locations.

For convenience, we convert the minimization problem into a maximization one to make it submodular by taking the reciprocal of the travel cost. Specifically, we use the travel \textit{efficiency}, the reciprocal of travel cost, as the measure. Thus, the travel efficiency of a demand location is the maximum of the travel efficiencies of the vehicles that reach this demand location. 
The overall travel efficiency, denoted by $f$ is the sum of the travel efficiencies of all demand locations. Notably, $f$ is also a random variable. 

Our goal is to find risk-aware paths from vehicles' start positions to demand locations given a user-defined risk level $\alpha$. We formulate a risk-aware pathfinding problem by maximizing $\text{CVaR}_{\alpha}$ on the travel efficiency (Problem~\ref{prob:risk_plan_find}).  
\begin{problem}[Risk-Aware Path Finding]~\label{prob:risk_plan_find}
\begin{equation}
\underset{\mathcal{S}\subseteq \mathcal{X}}{\max}~ \emph{CVaR}_{\alpha}[f(\mathcal{S},y)]
\label{eqn:cvar_max}
\end{equation}
\label{pro:cvar_max}
\end{problem}
where $\mathcal{S}$ is a set of paths for vehicles (``per path per vehicle''), $\mathcal{X}$ is a ground set of paths from which $\mathcal{S}$ is chosen, and $f(\mathcal{S}, y)$ is the travel efficiency on the path set $\mathcal{S}$, with randomness induced by $y$.  

\section{Proposed Approach: Risk-Aware Path Planner}~\label{sec:alg_ana}
\begin{figure*}[h]
  \centering
  \includegraphics[width=0.9\columnwidth]{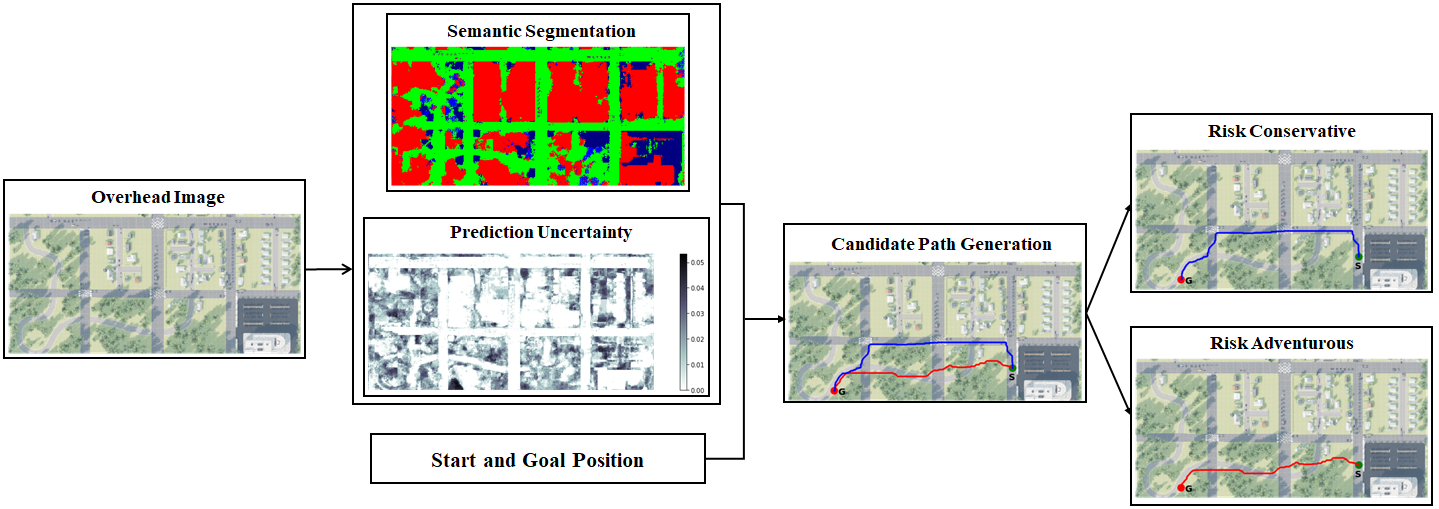}
  \caption{The breakdown of the risk-aware planning framework. Given an overhead image input, the algorithm provides a semantic segmentation and uncertainty map, then generates candidate paths, and finally performs the risk-aware path assignment of vehicles to demands.}
  \label{fig:risk_aware:overview}
\end{figure*}
The framework we propose consists of three main parts: semantic segmentation, candidate path generation, and risk-aware assignment. The overall pipeline of the framework and its parts are shown in Figure~\ref{fig:risk_aware:overview}. The inputs to the framework are a single overhead aerial two-dimensional image, the vehicles' start positions, and the demand locations. The output is a risk-aware assignment of paths from vehicles' start positions to demand locations. 

The input is first semantically segmented into per-pixel labels. These labels are assigned a cost proportionate to the risk involved in traversing them. The cost map and the uncertainty associated with the segmentation are then used as inputs to a path planner which generates candidate paths for assignment. Finally, the candidate paths from each vehicle's start position to each demand location are computed by maximizing CVaR, for risk-aware path assignment.

\subsection{Semantic Segmentation}
To plan a path for a vehicle in the environment, we need to first recognize features in the image, such as road, person, car, and so on. Given an input aerial image provided by an unmanned aerial vehicle (UAV), a deep learning model is used to provide a semantically segmented map. We then utilize an approximation technique, using dropout, to learn the uncertainty in the semantic labels provided by the model~\cite{gal2016dropout}. The advantage of using uncertainty has been studied~\cite{gal2017phd}. After semantic segmentation and uncertainty extraction, we assign different cost values for the robot traveling on different terrain features, similar to the cost function in~\cite{Toubeh2018}. For example, the cost of traveling on the road is less than that of traveling on vegetation. Traveling in a person or car is impossible, and thus, the corresponding cost can be set to infinity. Notably, the cost can be the energy or time spent by a vehicle traveling on the terrain. Since there exists uncertainty in the terrain features provided by the deep learning model, the cost of traveling on them is also a random variable with some uncertainty.

\subsection{Candidate Path Generation}~\label{subsec:path_gen}
Once a risk-aware cost map is generated based on the semantic segmentation of the terrain, augmented by the confidence in prediction, candidate paths are generated from each vehicle location to each demand location by an A* planner. For the different combinations of vehicles and demands, a candidate path represents a potentially feasible route given on the map. Relying on deep learning segmentation alone can be risky. Given confidence information, we expect paths to avoid regions of high uncertainty which could involve out-of-distribution data or misclassifications, as shown in Figure~\ref{fig:risk_aware:segnet_out1}.

For the A* planner, a risk-aware cost function $C(x)$ is defined on each pixel $x$. It combines the cost associated with the classified terrain type and the variance.

\begin{equation}
 \hat{C}(x)  :=  C(l_x) + \lambda \texttt{Var}(l_x),
 \label{eqn:cost_assign}
\end{equation}
where $C$ refers to the risk-neutral mean cost associated with the most likely predicted label $l_{x}$ for a pixel $x$ given multiple softmax outputs. The risk-aware function $\hat{C}$ assigns a cost to each pixel $x$, characterized by a user-defined cost mapping and a weight parameter $\lambda$ to quantify emphasis on the uncertainty. We use the variance in the prediction as a measure of uncertainty.
% }
\subsection{Risk-Aware Path Assignment}~\label{subsec:path_assign}
Consider a set of $K$ candidate paths $\mathcal{P}=\{1,\cdots, K\}$  generated from each vehicle's start position to each demand location in Section~\ref{subsec:path_gen}. We then assign each vehicle a path to a demand location. As mentioned in Section~\ref{sec:prob_form}, we follow a redundant assignment setting~\cite{prorok2019redundant}, where each vehicle can be assigned to at most one demand location and multiple vehicles can be assigned to the same demand location. Only the vehicle that arrives through a path with the highest efficiency is chosen at each demand location.

The travel efficiency is denoted for a vehicle starting at  $i\in \mathcal{V}$ arriving at demand location $j\in\mathcal{D}$ through path $k \in \mathcal{P}$ as $e_{ijk}$. Correspondingly, we take the tuple $(i,j,k)$ as an assignment where the vehicle-path pair $(i,k)$ is assigned to the demand location $j$. The total efficiency is denoted at all $M$ demand locations as 
\begin{equation}
 f(\mathcal{S}, y)  = \sum_{j\in M} \max_{(i,j,k)\in \mathcal{S}_j} ~e_{ijk} 
 \label{eqn:fsy_assign}
\end{equation}
where $\mathcal{S}_j$ denotes the assignment set $\{(i,j,k)\}$ to the demand location $j$. The total assignment set $\mathcal{S}:= \bigcup_{j=1}^{M} \mathcal{S}_j$ is a collection of the assignment sets at all demand locations. Notably, since each vehicle-path pair $(i,k)$ can be assigned to at most one demand location, all $\mathcal{S}_j$(s) are disjoint, i.e., $\mathcal{S}_j \cap \mathcal{S}_{j'} = \emptyset, ~\forall j\ne j',  j, j'\in\mathcal{D}$. This is called a partition matroid constraint in the literature~\cite{fisher1978analysis}, denoted by $\mathcal{I}$. 

We use the ``max" operator to capture the selection rule that only the vehicle-path pair $(i,k)$ with the maximum efficiency is chosen at each demand location. Due to the ``max" operator, the total efficiency $f(\mathcal{S},y)$ turns out to be monotone submodular in $\mathcal{S}$. If there is no assignment, we set $f(\emptyset, y)=0$ to normalize $f$. Here, $y$ indicates the randomness of $f(\mathcal{S},y)$ due to the uncertainty in travel efficiency. 

Then, by utilizing Equation~\ref{eqn:cvar_auxiliary} and the candidate paths generated in Section~\ref{subsec:path_gen}, we transform Problem~\ref{prob:risk_plan_find} to a risk-aware path assignment problem below.  
\begin{problem}[Risk-Aware Path Assignment]~\label{prob:risk_plan_assign}
\begin{align}
\underset{\mathcal{S}\subseteq \mathcal{X}}{\max} ~~\tau - \frac{1}{\alpha}\mathbb{E}[(\tau-\sum_{j\in M} \max_{(i,j,k)\in \mathcal{S}_j} ~e_{ijk})_{+}]\nonumber\\
s.t.~\mathcal{S}= \bigcup_{j=1}^{M} \mathcal{S}_j, ~\mathcal{S} \in \mathcal{I} ~~~and %\,nonumber\\
~~~~\tau\in[0, \Gamma],
\label{eqn:cvar_max_desc}
\end{align}
\label{pro:cvar_max_desc}
\end{problem}
with $\mathcal{S}$ the path assignment set (``per path per vehicle''), $\mathcal{I}$ the partition matroid constraint~\cite{fisher1978analysis}, $\mathcal{X}$ the ground set of all possible assignments, $\{(i,j,k)\},~k\in\mathcal{P},~i\in\mathcal{V},~j\in\mathcal{D}$,  and $\Gamma\in \mathbb{R}^{+}$ the upper bound of the parameter $\tau$. $\Gamma$ can be set as an upper bound on $f(\mathcal{S}, y)$ (Eq.~\ref{eqn:fsy_assign}).

Building on the sequential greedy algorithm (\texttt{SGA}) from our previous work \cite{zhou2018approximation}, we present Algorithm~\ref{alg:sga} for solving Problem~\ref{prob:risk_plan_assign}.

\begin{algorithm}[ht]
\caption{Risk-Aware Path Assignment}  
\begin{algorithmic}[1]
\Require
\begin{itemize}
    \item Vehicles' initial positions $\mathcal{V}$, demand\\ locations $\mathcal{D}$, and path set $\mathcal{P}$
    \item User-defined risk level $\alpha \in [0, 1]$
    \item Range of the parameter $\tau \in [0, \Gamma]$ and discretization stage $\Delta \in (0, \Gamma]$
    \item An oracle $\mathcal{O}$ that approximates $H(\mathcal{S}, \tau)$ \\as $\hat{H}(\mathcal{S}, \tau)$
\end{itemize}
\Ensure
\begin{itemize}
    \item Path assignment $\mathcal{S}$
\end{itemize}

\State $\mathcal{M}\leftarrow\emptyset$ \label{line:initiliaze}
\For{$ ~i \in \{0,1,\cdots, \ceil{\frac{\Gamma}{\Delta}}\}$}
\label{line:search_tau_forstart}
\State $\tau^i = i\Delta$\label{line:search_tau_separation}
\State $\mathcal{S}^{i}\leftarrow\emptyset$ \label{line:gre_empty}
\For{$l=1:|\mathcal{D}|$}
\label{line:D_round}
\State $(i^\star, j^\star, k^\star) = \underset{i\in \mathcal{V}, j \in \mathcal{D}, k \in \mathcal{P}}{\text{argmax}}~\hat{H}(\mathcal{S}^{i}\cup (i,j,k), \tau^i) - \hat{H}(\mathcal{S}^{i}, \tau^i)$ \label{line:max_margin}
\State $\mathcal{S}^{i}\leftarrow \mathcal{S}^{i} \cup (i^\star,j^\star,k^\star)$
\label{line:add_in_set}
\State $\mathcal{V} = \mathcal{V}\setminus i^\star$
\label{line:delete_vstar}
\EndFor
\label{line:greedy_end}
\State $\mathcal{M} = \mathcal{M} \cup \{(\mathcal{S}^{i}, \tau^i)\}$\label{line:pair_collection}
\EndFor\label{line:search_tau_forend}
\State $(\mathcal{S}, \tau^{\star}) = \underset{(\mathcal{S}^{i}, \tau^i) \in \mathcal{M}}{\text{argmax}}~{\hat{H}(\mathcal{S}^{i}, \tau^i)}$ \label{line:find_best_pair}
\end{algorithmic}
\label{alg:sga}
\end{algorithm}

There are four stages in Algorithm~\ref{alg:sga}: 
\paragraph{Initialization (line~\ref{line:initiliaze})} We initialize a storage set $\mathcal{M}$ to be empty. We use $\mathcal{M}$ to store the assignment $\mathcal{S}$ with the corresponding $\tau$ when searching all possible values of $\tau$. 

\paragraph{Searching for $\tau$ (\textbf{for} loop in lines~\ref{line:search_tau_forstart}--\ref{line:search_tau_forend})} We sequentially search for all possible values of $\tau$ within $[0, \Gamma]$ by a user-defined separation $\Delta$ (line~\ref{line:search_tau_separation}). 
$\Gamma$ is an upper bound on $\tau$ and can be set based on the specific problem at hand. We show how to compute $\Gamma$ in specific scenarios in Section~\ref{sec:sim}. 

\paragraph{Greedy algorithm (lines~\ref{line:gre_empty}--\ref{line:greedy_end})} For a specific $\tau$, e.g., $\tau_i$, we use the greedy approach~\cite{fisher1978analysis} to choose the corresponding assignment set $\mathcal{S}^{i}$. We first initialize set $\mathcal{S}^{i}$ to be the empty set (line~\ref{line:gre_empty}). Then we execute the greedy algorithm in $|\mathcal{D}|$ rounds (line~\ref{line:D_round}) since the total number of demand locations to be served is $\mathcal{D}$. At each round, the assignment $(i^\star,j^\star,k^\star)$ which gives the maximum marginal gain of $\hat{H}(\mathcal{S}^{i}, \tau^i)$ is selected (line~\ref{line:max_margin}) and added into set $\mathcal{S}^{i}$ (line~\ref{line:add_in_set}). Then, we remove the vehicle position $i^\star$ from  $\mathcal{V}$ (line~\ref{line:delete_vstar}) to make sure vehicle position $i^\star$ and the corresponding paths will never be selected in the following rounds. 

\paragraph{Finding the best assignment (line~\ref{line:find_best_pair})} From the collection of $(\mathcal{S},\tau)$ pairs, $\mathcal{M}$ (line~\ref{line:pair_collection}), we pick the one that maximizes $\hat{H}(\mathcal{S}^{i}, \tau^i)$ as the output $\mathcal{S}$ with the corresponding $\tau$ denoted by $\tau^\star$ (line~\ref{line:find_best_pair}). 

\textit{Oracle Design:} We use an oracle $\mathcal{O}$ to calculate the value of $H(\mathcal{S}, \tau)$. The oracle uses a sampling-based method to approximate $H(\mathcal{S}, \tau)$~\cite{zhou2018approximation}. 

It has been shown in \cite{zhou2018approximation} that Algorithm~\ref{alg:sga} generates an assignment that has the bounded approximation performance of the optimal assignment. 

\section{Experiment and Evaluation}~\label{sec:sim}
In this section, we report the results from empirical studies evaluating the proposed risk-aware perception, planning, and assignment framework. We start by describing the experimental setup and then describe the results.

\textbf{Setup.} We use the AirSim~\cite{shah2018airsim} simulator as it offers photo-realistic images along with the semantically segmented ground truth in various pre-defined environments. We use the \textit{CityEnviron} environment which contains city-like and suburban landscapes. We collect the downward-looking aerial images at an altitude value of 200m in the simulator. The dataset thus generated has 480 images, which are divided in the ratio of 10:3:3 as training, validation, and testing data for the Bayesian SegNet~\cite{kendall2015bayesian}. We use the PyTorch implementation of a basic version of this network~\cite{mshahsemseg} and add dropout layers in accordance with the original architecture. 
In our ground truth, we reduce the ground truth to 12 classes consistent with the original model~\cite{kendall2015bayesian}. 

\begin{figure}[ht]
    \centering
    \begin{subfigure}[c]{0.8\textwidth}
      \includegraphics[clip,width=0.9\columnwidth]{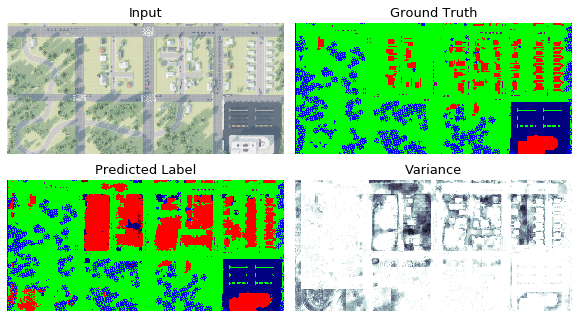}
      \caption{Output from training on shuffled data}
    \end{subfigure}
    \hfill
    \begin{subfigure}[c]{0.8\textwidth}
    \includegraphics[clip,width=0.9\columnwidth]{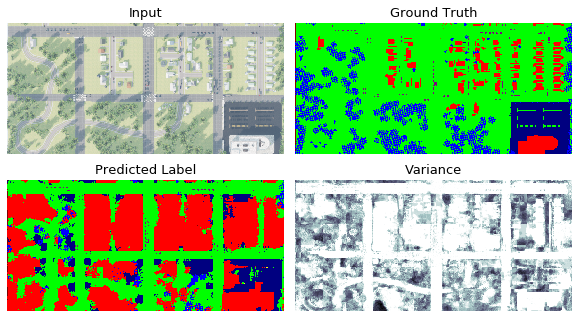}
        \caption{Output from training on unshuffled data}
    \end{subfigure}
    \caption{Difference in variance in semantic segmentation outputs due to data distribution. \label{fig:risk_aware:segnet_out1}}
\end{figure}

Due to limitations in the simulator, roads and grass patches are indistinguishable in the ground truth as depicted in Figure~\ref{fig:risk_aware:segnet_out1}. We keep such images limited to the test dataset. Hence, the grass patches act as unknown objects (similar to out-of-distribution) to the model and provide interesting observations on the uncertainty in prediction for such objects.

 We produce 20 outputs (or \textit{stochastic samples}) for each image using the trained Bayesian SegNet. For each pixel, the predicted label is found as the most frequent label among all of the most likely labels for each pixel, i.e., $l_{x} = \texttt{Mode}_\text{samples}(\text{arg} \max\limits_c P(c|x))$ where $x$ is the pixel input to the model, $c$ is the class label, and $\texttt{Mode}$ stands for the statistical mode. Uncertainty for each pixel is defined as the average variance in the probabilities of labels i.e., $\texttt{uncertainty}(x) = \frac{1}{c} \sum_c \texttt{Var}( P(c|x) )$. The cost for each pixel is calculated using these two quantities, which is further passed to the path planning algorithm (Sec.~\ref{subsec:path_gen}). To provide an orthographic view of the results, we make predictions over partially overlapping images with perspective projections. Then we combine subsets around the center of the images.

\textbf{Results.} 
\textit{a) Out-of-Distribution Data.}
The Bayesian SegNet model used in our path planning algorithm has dissimilar data distribution for training and testing. The effect of this is shown in Figure~\ref{fig:risk_aware:segnet_out1}. When the data distributions are the same across training and test datasets (shuffled data), the model is able to make predictions with very low variance (implied by brighter shades). However, in the case of dissimilar distribution, some part of the image acts as unseen data, and thus the model prediction has high variance. This highlights the importance of uncertainty. For example, the prediction is precise for already observed data and thus the decision involving such input would be less risky, whereas the opposite is true for unobserved data.

\begin{figure}
  \centering
  \includegraphics[width=0.9\columnwidth]{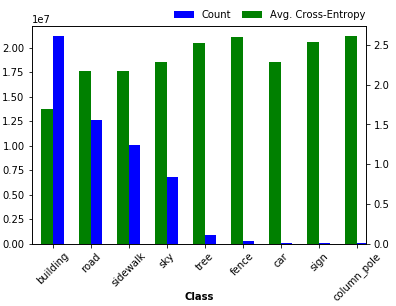}
  \caption{Average Cross-Entropy for each class/label. It is inversely correlated with the frequency of the class/label in the data.}
  \label{figs:avg_cross_ent}
\end{figure}

\textit{b) Average Cross-Entropy on the Test Dataset.}
In order to understand the correlation between the quality of prediction and the training data, we look at the cross-entropy of the model prediction (over unshuffled data) averaged over the 20 samples and the number of pixels in the training data for each class in Figure~\ref{figs:avg_cross_ent}. 

The model performs comparatively well in identifying obstacles like buildings. Objects which are rarely observed in the ground truth have very high cross-entropy. The lack of difference between the vegetation and the road in the ground truth also affects the performance of relevant classes. As expected, high cross-entropy is observed for classes where the pixel count is low. 

\begin{figure}
  \centering
  \includegraphics[width=0.7\columnwidth]{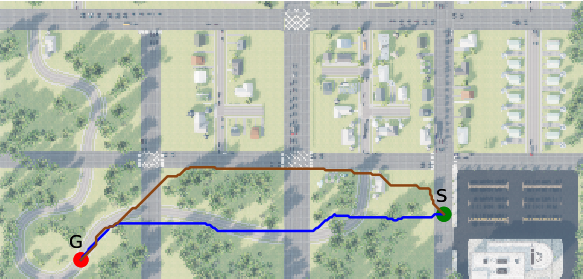}
  \caption{Effect of $\lambda$ on path planning. {\color{blue} Smaller $\lambda$ (=10)}  gives a shorter path with high uncertainty; {\color{brown} larger $\lambda$ (=50)} gives a longer path with low uncertainty.}
  \label{fig:risk_aware:path_lambda}
\end{figure}

\textit{\c) Risk-Aware Path Planning and Assignment.}
The value of $\lambda$ decides the risk-awareness of path planning. This effect is shown in Figure~\ref{fig:risk_aware:path_lambda} where the cost is 1 for the navigable classes and 3 for non-navigable classes (except for the tree class, where it is 2). For a small value of $\lambda$, the algorithm plans a short path passing through the vegetation. For a high value of $\lambda$, this area is avoided due to the high uncertainty in the prediction for this part of the scene.

By varying the values of $\lambda$, we generate $K=2$ candidate paths for each pair of start and demand locations. We consider assigning $N=3$ supply vehicles to $M=2$ demand locations in this unknown environment.

\begin{figure}[ht]
\centering{
\begin{subfigure}[c]{0.49\textwidth}
    \includegraphics[width=1.0\columnwidth]{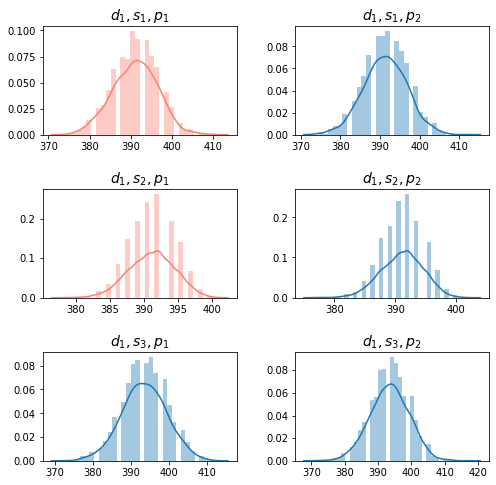}
    \caption{Demand location $d_1$}
\end{subfigure}
\hfill
\begin{subfigure}[c]{0.49\textwidth}
    \includegraphics[width=1.0\columnwidth]{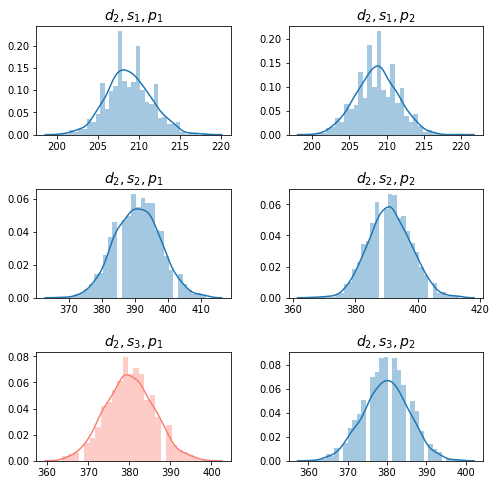}
    \caption{Demand location $d_2$}
\end{subfigure}
\caption{Efficiency distributions of paths and the path assignment when $\alpha = 0.01$. The assigned path for each robot is marked in red. \label{fig:risk_aware:risk_001}}
}
\end{figure}

\begin{figure}[ht]
\centering{
\begin{subfigure}[c]{0.49\textwidth}
    \includegraphics[width=1.0\columnwidth]{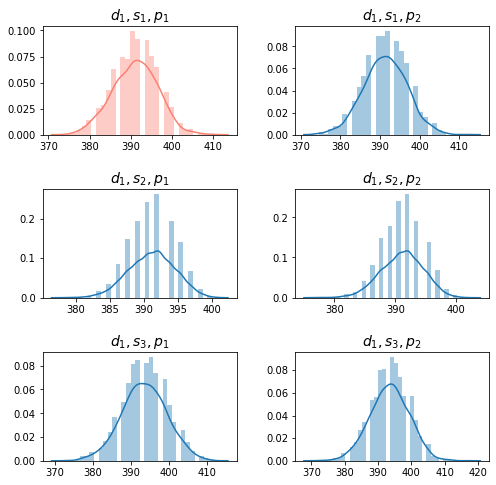}
    \caption{Demand location $d_1$}
\end{subfigure}
\hfill
\begin{subfigure}[c]{0.49\textwidth}
    \includegraphics[width=1.0\columnwidth]{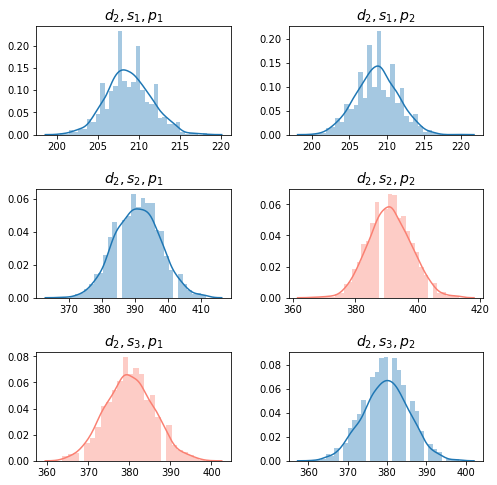}
    \caption{Demand location $d_2$}
\end{subfigure}
\caption{Efficiency distributions of paths and the path assignment when $\alpha = 1$. The assigned path for each robot is marked in red. \label{fig:risk_aware:risk_1}}
}
\end{figure}

Due to the imperfect semantic segmentation, the cost or efficiency of the path is a random variable. We show the efficiency distributions of the paths from the vehicles' start positions to the demand locations in Figure~\ref{fig:risk_aware:risk_001} and Figure~\ref{fig:risk_aware:risk_1}. 

We use Algorithm~\ref{alg:sga} to assign each vehicle a risk-aware path to a demand location.  
For example, in Figure~\ref{fig:risk_aware:risk_001}-(a), vehicle $v_2$ is assigned path $p_1$ for demand location $d_1$ when the risk level is small, e.g., $\alpha = 0.01$. In contrast, as seen in Figure~\ref{fig:risk_aware:risk_1}-(a), when the risk level is high, e.g., $\alpha=1$, the assignment is more adventurous, and thus the path with a larger mean efficiency and a larger variance is selected. As shown in Figure~\ref{fig:risk_aware:risk_1}-(b), vehicle $v_2$ switches to path $p_2$ for demand location $d_2$, because the efficiency of this path has a larger mean. The path assignment changes follow the comparison of CVaR values, i.e., $\text{CVaR}_{0.01}[e(p_1)] > \text{CVaR}_{0.01}[e(p_2)]$ and $\text{CVaR}_{1}[e(p_1)] < \text{CVaR}_{1}[e(p_2)]$ with $e(p_1)$ and $e(p_2)$ denoting the efficiency of path $p_1$ and $p_2$, respectively. Thus, the risk level, $\alpha$, provides the flexibility to trade-off between risk and total efficiency (reward). 

\textit{d) Quantitative results.}

\begin{figure}[t]
  \centering
  \includegraphics[width=0.8\columnwidth]{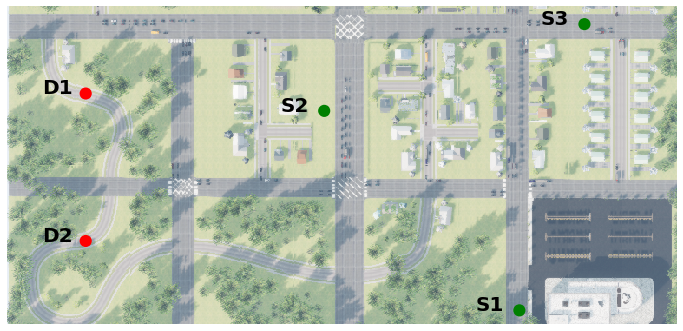}
  \caption{Start and demand positions for \textit{surprise} calculation setup.}
  \label{fig:risk_aware:city_start_goal}
\end{figure}

\begin{figure}[t]
  \centering
  \includegraphics[width=0.65\columnwidth]{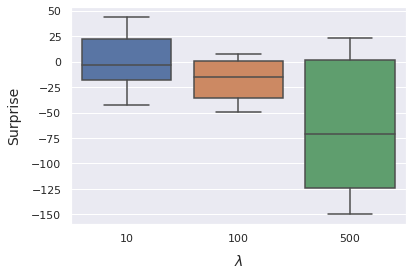}
  \caption{\textit{Surprise} vs $\lambda$. A higher $\lambda$ may result in a longer path, increasing the cost of traversal. For large values of $\lambda$, even a few pixels with high variance may greatly increase the cost.}
  \label{fig:risk_aware:all_surprise}
\end{figure}

In order to quantify the effect of $\lambda$, we define a metric called \textit{surprise}  for a path as the difference between the cost of the ground truth labels and predicted labels. We take 6 combinations of start and demand positions given in Figure~\ref{fig:risk_aware:city_start_goal}  and find the average value of \textit{surprise} as shown in Figure~\ref{fig:risk_aware:all_surprise}. 
Ideally, we expect the \textit{surprise} to reduce with emphasis on variance. However, for large values of $\lambda$, even a few pixels with high variance may greatly increase the cost. In general, a higher value of $\lambda$ may cause the robot to choose a longer path, which may increase the cost of traversal. However, sometimes the path may pass through only navigable regions, resulting in a small value of \textit{surprise}. This causes the \textit{surprise} to have a larger variance. Thus, the value of $\lambda$ should be chosen after careful consideration of the variance in predictions and the range of the cost mapping.

\begin{figure}[t]
  \centering
  \includegraphics[width=0.9\columnwidth]{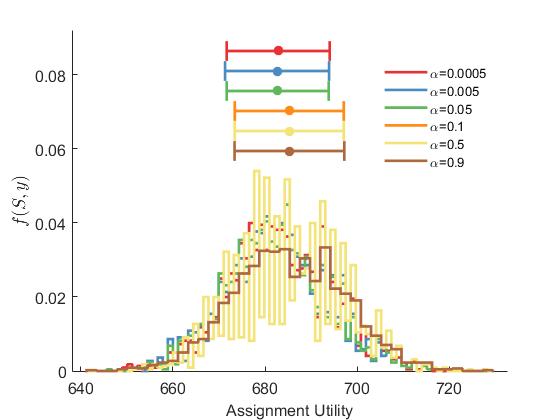}
  \caption{Distribution of the total travel efficiency $f(\mathcal{S}, y)$ by Algorithm~\ref{alg:sga}.}
  \label{fig:risk_aware:dis_total_effi}
\end{figure}

We also plot the distribution of the total travel efficiency (Eq.~\ref{eqn:fsy_assign}) in Figure~\ref{fig:risk_aware:dis_total_effi}.
$\mathcal{S}$ is the path assignment selected by Algorithm~\ref{alg:sga}. With small risk levels $\alpha$, paths with low efficiencies (equivalently, low uncertainty) are mostly selected. This is because a small risk level indicates the assignment is conservative and only willing to take a small amount of risk. Thus, the path with lower efficiency and lower uncertainty is assigned to avoid the risk induced by the uncertainty. In contrast, when $\alpha$ is large, the assignment is riskier. In such a case, the paths with high efficiencies (equivalently, high uncertainty) are selected.

\section{Conclusion} \label{sec:conslude}
In this chapter, we propose a risk-aware path planning and assignment framework for vehicles navigating in an unknown environment. We consider a scenario in which the information of the unknown environment is only available by an overhead, georeferenced image that is taken by an aerial robot. To deal with this challenge, we utilize Bayesian deep learning to learn the environment through semantic segmentation. Since the output of this segmentation is noisy, the cost of the vehicle traversing in the environment is uncertain. To deal with the cost uncertainty, we optimize some popular risk measures to generate and assign risk-aware paths for the vehicles. We use extensive simulation results to show the effectiveness of our risk-aware strategy.

Future work will focus on the online coordination of aerial and ground vehicles to achieve long-term, real-time risk-aware navigation in unknown environments.

%% file: Chapter9.tex
%Chapter 9

\renewcommand{\thechapter}{9}

\chapter{Meta-Reasoning for Risk Management with Explicit Measures}
\label{chapter:hybrid}

\section{Introduction}
\cleanfootnote{The work presented here was done as an intern at Nokia Bell Labs, Murray Hill, NJ, under the guidance of  Matthew Andrews, Jeongran Lee, and Ilija Had\v{z}i\'{c}.}

In chapter~\ref{chapter:risk_aware} we show how the uncertainty inherent in the predictions can be utilized to facilitate risk-aware planning. Extracting uncertainty in this manner is time-consuming as we need multiple forward passes through a neural network. This can be disastrous for real-time applications where fast decisions are crucial. In this chapter, we address this issue in the context of mobile robot navigation equipped with a classical and a prediction-based local planner. We show that the prior knowledge about the strengths and weaknesses of each can be used to build an explicit method to design an efficient and safe risk-aware planner.

A mobile robot navigation stack is broadly responsible for safely (and desirably optimally) getting the robot from its present position to the goal while respecting externally or internally imposed constraints.
Components of a path- and motion-planning and control subsystem can be broadly categorized into global planners, local planners/controllers, and motion controllers, which are typically deployed in concert. Global planner finds the path toward the goal location, often expressed as a set of waypoints that the robot must visit. The local planners are responsible for generating the velocity vectors to lead the robot towards the next waypoint.

In a known map, global planners are optimal as they utilize the global costmap, but are brittle in the presence of unknown (and discovered after the fact) dynamic obstacles, such as humans, clutter, unmapped fixtures, and other vehicles. Local planners, on the other hand, can react well in such situations. Additionally, local planners take less time to compute and thus process the data at a higher frequency.

Local planning in velocity space can be characterized as an optimization process (which may in practice produce suboptimal, but acceptable solutions), whose optimization criteria include distance to the next waypoint (or the goal), clearance around the obstacle, smoothness of motion, energy efficiency, and the like. For this discussion, we broadly classify the implementations into classical and learning-based approaches. Classical planners explore the velocity space and evaluate each proposed velocity against the constraints and the optimization criteria in real time. To find an optimal solution a classical planner must often search the entire space of admissible velocities, which, depending on the size of the planning window, the number of degrees of freedom, and the complexity of constraints, can be a challenging process. 

Learning-based planners are exposed to various situations offline and trained to map the robot state to the deemed best velocity, typically using a neural network. The complexity of searching and evaluating the solution is moved to an offline training process. The real-time computation becomes the model inference and does not involve explicit search. The performance of these planners strongly depends on how the training environment was set up, the variety of situations the robot has been exposed to, and how well the dynamics of the robot were captured during the training. Typically, reinforcement learning (RL) techniques are used here. As with all learning-based algorithms, false results are possible and it is impossible to guarantee that the RL-planner will always produce optimal or even correct solutions. Nevertheless, RL-planners have been shown to produce useful results that generalize well\cite{guldenring2020learning, van2017extended, liu2020robot, KastnerZBLSLM21, patel2021dwa, nakhleh2023sacplanner}. In our previous work\cite{nakhleh2023sacplanner}, we designed an RL-planner with superior obstacle-avoidance performance compared to a widely used Dynamic Window Approach (DWA) planner\cite{fox1997dynamic}, but the price of this improvement was an uneven and jerky motion, even when no dynamic obstacles were present in the robot path.

This lack of smoothness limits the attractiveness of RL-based local planners as a general solution. If the robot is moving through a large open space, or if it is moving in a maze-like structure with known, fixed walls, it can stay close to the global plan. Classical local planners typically excel at generating smooth motion toward the goal. In this case, the instantaneous decisions of an RL-based planner are overkill and can lead to rapid changes in velocity that do not provide any benefit. Although reworking the training process to penalize uneven motion may lead to improved behavior, it is unclear how the two opposing criteria would reflect on general performance. Further, conceiving a new training process and designing an improved reward function is an arduous effort that is often subject to trial and error.

Alternatively, one can simply recognize that an RL-based planner performs better when confronted with an unexpected or dynamic obstacle, whereas a classical planner performs better when the robot simply needs to track the global plan. In this context, a pragmatic solution is to conceive a decision tree that recognizes the current situation and switches to using the planner known to produce a better solution. The existing works~\cite{dey2023learning, raj2023targeted} have proposed learning the switching criteria with a neural network, which requires further training and may suffer from the typical shortcomings of the learning-based approaches, such as generalizability.

In this chapter, we propose a simple hybrid planner that detects if the global plan is obstructed by an unexpected obstacle and picks the solution provided by a (more responsive) RL-planner. Otherwise, it takes the solution provided by a classical planner. We demonstrate via experiments that this hybrid approach responds well in the obstructed case while maintaining smooth performance in the non-obstructed case.

\section{Related Work}
Local planners play an important role in obstacle avoidance and have been a topic of interest for a long time~\cite{sanchez2021path}.
Classical planning approaches, which do not employ learning, are widely used across robotics applications. Reactive replanning~\cite{fox1997dynamic,rosmann2017kinodynamic}, artificial potential field~\cite{bin2011research}, and fuzzy logic-based approaches~\cite{yan2016mobile} are examples. One such widely used method, proposed by Fox et al.~\cite{fox1997dynamic} and called Dynamic Window Approach (DWA) planner, uses reactive replanning and has been frequently used as the baseline planner by the Robot Operating System (ROS) navigation stack\cite{rosnavigation}. Because of its widespread use and availability in the open-source community, ROS implementation of DWA has often been used as the baseline, despite the algorithm being relatively old. For this reason, we baseline our results to DWA.

An alternate way to design a local planner is to learn the system model using data and fine-tune the learning model in a new environment. Such learning-based approaches have been introduced in the past few years and have been growing rapidly in number. A deep reinforcement learning (DRL) framework is often used for training in such approaches as it allows the robot to interact with the environment without needing data collection and annotation~\cite{van2017extended, Gldenring2019ApplyingDR,guldenring2020learning,liu2020robot,KastnerZBLSLM21,patel2021dwa,nakhleh2023sacplanner}. The framework proposed by G{\"u}ldenring et al.~\cite{Gldenring2019ApplyingDR}, which uses 2D local map and waypoints from the global plan for state representation, was used as the base for the development of many subsequent works. In our previous work\cite{nakhleh2023sacplanner}, we studied and compared classical planners and different RL network architectures and proposed a method that used a polar representation of the costmap in state representations. This network, named SACPlanner, was trained in a simulation environment and tested on live robots. SACPlanner outperformed other approaches, including DWA, in safely avoiding collisions with static and dynamic obstacles. The practical result was a more responsive planner, but slower and jerky motion caused by the robot trying to move cautiously even when the path ahead was clear. Arguably, this behavior can be improved with training in a higher-fidelity simulation environment, but at the risk of breaking other
desirable properties achieved during the original training.

One way to get the benefits of different types of planners is to use an ensemble of methods with user-defined control. The use of such \textit{hybrid} planning strategies to harness both classical and learning-based approaches is a fairly recent development~\cite{von2020combining}. Existing work in the literature has explored both hybrid robotic planners consisting of classical approaches~\cite{orozco2019hybrid} and planners using learning-based approaches~\cite{lu2020hybrid}. Existing hybrid planners combining classical and learning-based approaches lie in the middle of this spectrum and aim to combine the model-based classical approaches and data-based learning approaches by switching between them.

Almadhoun et al.~\cite{almadhoun2021multi} use heuristics-based criteria to switch between a  classical and a learning-based approach to generate viewpoints for 3D reconstruction. Linh et al.~\cite{linh2022all} and Dey et al.~\cite{dey2023learning} study ground robot navigation but they rely on neural networks for learning and focus on high-level planning. Raj et al.~\cite{raj2023targeted} also proposed a neural network-based switch, but they focused on social navigation only. In contrast, our work contributes towards the development of a local planner that uses a hybrid approach that combines classical and learning-based methods. We design a heuristics-based logic for switching between a DWA planner and SACPlanner, enjoying the benefits of both. This hybrid planner exhibits a superior performance with a simple design which forgoes the need to train another neural network for switching.

\section{Preliminaries}
\label{sec:hybrid:background}
The local planner/controller is responsible for generating the velocity vector that makes progress toward the goal or the next waypoint. Some implementations explore the velocity space and score candidate velocities based on forward simulation in the configuration space (which, strictly speaking, makes them planners), whereas others solve a constrained optimization problem that maps the state to an action (which, strictly speaking, makes them controllers). These planners/controllers can either generate the motion in the velocity space and leave it to a lower-level motion controller to generate the actuation or directly solve for actuation. A motion controller (if present separately from the local planner/controller) generates the actuation that delivers the desired velocity vector. In this chapter, we focus on local planning/control in velocity space, and for simplicity, we use the term ``local planner" to mean any subsystem that generates the desired velocity vector based on the present robot configuration (specifically, the robot pose) and the state of the surrounding environment (specifically, the next waypoint pose, goal pose, and perception of obstacles). In the following subsections, we describe the classical and the learning-based local planners used in our work, DWA and SACPlanner respectively. 

\subsection{Dynamic Window Approach (DWA)}
\label{subsec:hybrid:dwa}
DWA planner generates a set of admissible velocities, which are the velocities that can be reached given the present velocity and the robot dynamic constraints (i.e., acceleration limits). For each admissible velocity, DWA performs a forward simulation to calculate the resulting trajectory should the robot use this velocity. Finally, each simulated trajectory is scored and the one with the lowest cost is selected. The objective function reflects
progress towards the goal, clearance from the obstacle, adhering to the global plan (distance to the waypoint) and twirling.

DWA considers the robot's dynamics and the overall motion is a series of arcs determined by the angular and linear velocity, where each planning step produces one such arc. If there are no obstacles in the path, the planner will pick the arc that best advances the robot toward the next waypoint, as the distance from the global plan is part of the cost function. In an obstacle-free environment, the selection of the best velocity will be the balance between sticking to the global plan (advancing to the nearest waypoint) and advancing towards the global (cutting corners in the global plan to reach the goal sooner). Parameters allow the user to tune the planner to balance between one behavior and the other. While this single-arc planning works well in general, situations, as described below, may need a complex velocity profile, making DWA ineffective in the scenarios.

If there is an obstacle in the path, the obstacle-distance component of the cost function will start to dominate and the arcs that point away from the global path will have a lower cost, consequently making the robot deviate from the global plan or the goal.
As the robot steers away, the plan-distance and goal-distance components of the cost function will equalize and the robot will gravitate back to the plan. Three cases are possible next: 1) the robot may have made sufficient forward progress that the next waypoint is behind the obstacle, in which case the local planner will return the robot to the path determined by the global plan; 2) the robot may turn back into the obstacle make a motion towards it, and steer away from the obstacle again, but this time being in a more difficult situation due to obstacle proximity; 3) the global planner may trigger and generate a new set of waypoints that will guide the robot around the obstacle.

Ideal local planners should always result in the first case, which would make it able to deal with obstacles on its own.  The second case can often lead to a live-lock that manifests itself by a robot approaching the obstacle and indecisively oscillating without making progress. In some cases, the collision may occur due to sensor limitations. Namely, in our experiments, we saw collision because the LiDAR sensor that we used has the minimum-range distance. Once the robot gets too close to the obstacle, the reflections are not registered and the robot charges into the obstacle. Augmenting the robot with the second, short-distance sensor to prevent these collisions resulted in described live-locks.

\begin{figure}[ht!]
% \vspace{3mm}
  \centering
  \includegraphics[height=0.25\textwidth]{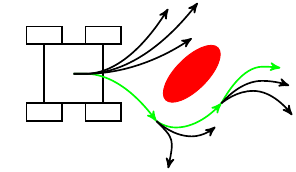}
  \caption[An illustrative example showing various ways in which a robot can react to an obstacle with a series of arc-motions.]{Confronting an obstacle in a series of arc-motions.}
  \label{fig:hybrid:arcmotion} 
\end{figure}

We argue that these shortcomings are direct consequences of single-arc
motion planning that DWA uses. Successful obstacle avoidance requires three consecutive arcs as shown in green in Fig.~\ref{fig:hybrid:arcmotion}. The first arc pushes the robot away from the obstacle, the second sends it back on track once the obstacle has been successfully navigated around, and the third realigns the direction to the plan. DWA planner simply does not explore the space beyond one velocity vector and longer simulation time simply extends the arc into the space that is not relevant for evaluating the motion. We confirmed this with a series of experiments, tuning one parameter at a time while tracing the DWA code to find the root cause. All tests pointed to the lack of visibility into the subsequent arcs that may follow the one being scored.
% \vspace{-1mm}

Extending the planner to explore a series of velocity vectors scales exponentially with the number of composite arcs to explore. The sequence of arcs shown in green in Fig.~\ref{fig:hybrid:arcmotion} successfully navigates around the obstacle, but to select it, all three arcs in the sequence must be scored. At each step, there are multiple candidates (arcs shown in black) that must also be scored to find the optimal path around the obstacle.

The third case is commonly used in practice to counter the above-described problem. Replanning on the global level is achieved either by running the global planner periodically at a low rate (e.g., once every few seconds) or by having ``patience" timers built into the navigation stack that trigger the global planner when deemed necessary.
Careful tuning of cost-function weights, timer values, and other constraints, can result in satisfactory and safe performance of the navigation stack, but this process is arduous and practitioners often resort to trial and error.

More cases the local planner can deal with when left on its own, more robust navigation stack will be when the assistance from the global planner is turned on. In our evaluation, we disallow global replanning, because we are interested in the performance of the local planner alone, rather than the whole navigation stack. This results in collision-avoidance performance that some practitioners may find surprisingly poor, but this is due to confusing the performance of the whole navigation stack as opposed to the performance of the local planner alone.

\subsection{SACPlanner}
\label{subsec:hybrid:sacplanner}
SACPlanner, which we previously developed\cite{nakhleh2023sacplanner}, is a RL-based planner that outperforms DWA in challenging situations. We have experimentally shown that it successfully resolves the problem described in Section~\ref{subsec:hybrid:dwa}. An intuitive explanation is that the arc motion it selects is statistically the most likely to be the correct first step in the chain of velocity vectors that will avoid the obstacle and put the robot back on the planned path. There is no velocity-space exploration and although a single compute step is more complex, it eliminates the problem of exponential scaling.

SACPlanner uses a polar representation of the local costmap as the input to the neural network (see Fig.~\ref{fig:hybrid:wp_polar}) and outputs an angular and linear velocity pair as the action for the robot. It uses the Soft Actor-Critic~\cite{haarnoja2018soft} method for training with a mixture of dense and sparse rewards that quantify the robot's progress towards the goal and collision-avoidance, similar to DWA's objective function. Even though it is trained in a simulation environment, we have shown that it generalizes well\cite{nakhleh2023sacplanner} and uses polar representation of the local costmap as the state helps in sim-to-real transfer without fine-tuning. We demonstrated that a real robot can successfully execute PointGoal navigation in complex mazes and with unexpected obstacles, whereas DWA typically ends up in a state from which it does not make meaningful progress toward the goal or in some cases collides. We have experimentally determined that when the collision occurs, it is typically due to the sensor limitation. Namely, the LiDAR we use has the minimum range below which it becomes ``blind''. Whenever the collision occurred, it would be because the DWA planner pushed the robot too close to the obstacle to provoke the sensing problem. We believe that if the sensing were augmented to resolve this problem the problem would simply morph into stalling the robot in front of the obstacle. SACPlanner, on the other hand, never brought the robot into such a situation and successfully avoided the obstacles despite the sensing limitation.

A learning-based planner effectively retains the mapping between the input and the output as network weights. This results in SACPlanner potentially learning how to behave in a conservative fashion to safely avoid obstacles. Whereas DWA is limited to executing motion on circular arcs, SACPlanner can traverse complex trajectories. However, as SACPlanner looks at the costmap in an instant only, the robot's motion is jerky and it moves at a slower speed, making it inefficient even when there is no obstacle ahead. 

These two planners represent two seemingly contrasting planning approaches. Choosing one planner from these is essentially a tradeoff between \textit{smoothness} and \textit{responsiveness}. While DWA is more suitable for moving on a static map, SACPlanner is better equipped for successful navigation in complex and dynamic environments. We use this idea to propose a hybrid approach that uses both planners for safer and more efficient planning.

\section{Proposed Approach: Hybrid Local Planner}
\label{sec:hybrid:approach}
\begin{figure}
    \centering
    % \vspace{2mm}
    \includegraphics[width=0.98\linewidth]{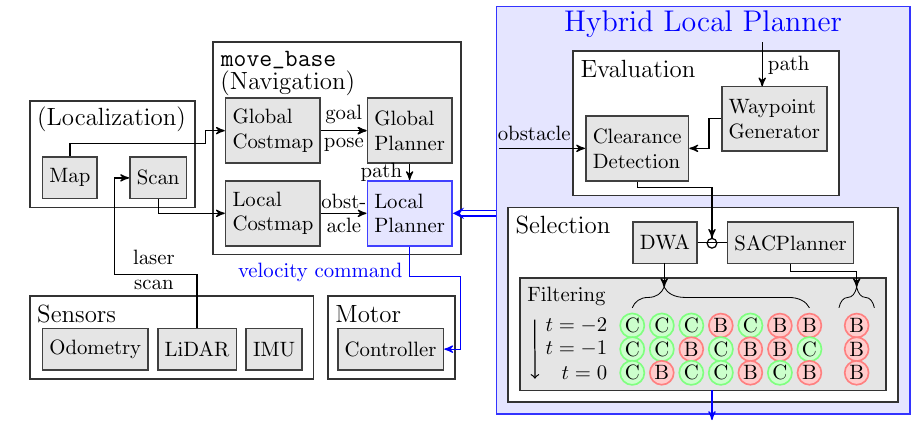}
    \caption{ROS framework and the architecture of our hybrid local planner.}
    \label{fig:hybrid:ros}
    % \vspace{-3mm}
\end{figure}

We propose a hybrid local planning approach that combines the benefits of a classical planner and a learning-based planner. Specifically, we run DWA and SACPlanner in parallel and switch between them based on the clearance ahead of the robots. Fig.~\ref{fig:hybrid:ros} shows the architecture of our implementation. The box labeled {\tt move\_base} comes from the standard ROS navigation stack and we modified the local planner plugin to include the DWA code verbatim from the ROS navigation stack, our SACPlanner implementation, along with the code that implements the switching policy. This is illustrated by the box on the right labeled Hybrid Local Planner.

\subsection{Waypoint Generation}
\label{subsec:hybrid:waypoint}
\begin{figure}[!h]
    \centering %\vspace{-4mm}
    \includegraphics[width=0.7\linewidth]{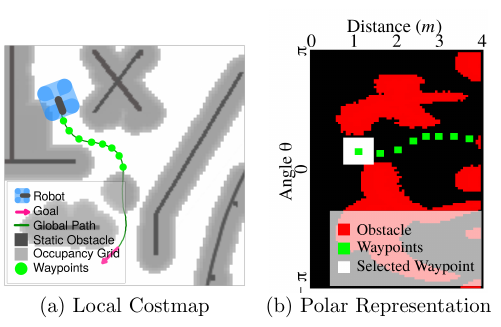} %\vspace{-2mm}
    \caption[An overview of the waypoint generation scheme for the hybrid local planner.]{Waypoint generation.} %\vspace{-4mm}
    \label{fig:hybrid:wp_polar}
\end{figure}

First, we use the method proposed by G\"{u}ldenring et al.~\cite{Gldenring2019ApplyingDR} to find waypoints on the local map. We use the waypoints both to decide which local planner to use and also to create the goal in case the SACPlanner is selected. To generate the waypoints, the global plan leading to the goal, generated with Dijkstra's algorithm, is downsampled and a fixed number of waypoints, 8 in our case, are selected on the local costmap, as shown in Fig.~\ref{fig:hybrid:wp_polar}(a).  This set of waypoints helps the robot align with the global plan and thus also avoids local minima. The first waypoint not on the obstacles is fed to SACPlanner as the goal in the polar image as Fig.~\ref{fig:hybrid:wp_polar}(b).%\vspace{-3mm}

\subsection{Clearance Detection}
\label{subsec:hybrid:linearization}
\begin{figure}[!h]
% \vspace{1mm}
    \centering
    \includegraphics[width=\linewidth]{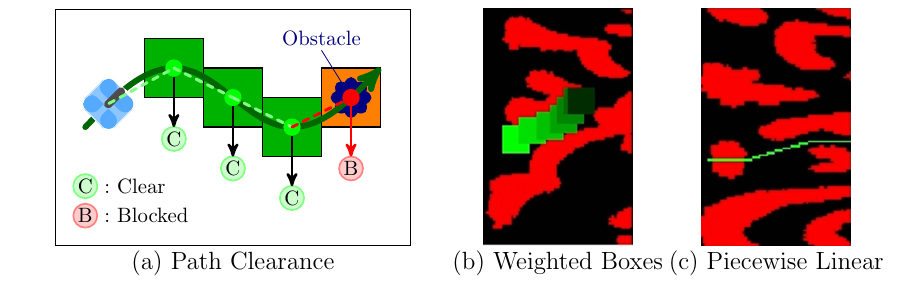} %\vspace{-2mm}
    \caption[An overview of the clearance detection module for the hybrid planner.]{Clearance detection.}
    \label{fig:hybrid:wpc}
\end{figure}

To switch between the planners, the robot needs to determine the clearance ahead. In order to enable an early response, we find if the path without any dynamic obstacle can be traversed without collision. We use the waypoints generated on the local costmap for clearance detection. We check if this path is obstructed anywhere on the local map. If the whole path is unobstructed, we consider the path to be \textit{clear}. Otherwise, this path is considered as \textit{blocked}. Fig.~\ref{fig:hybrid:wpc}(a) demonstrates this approach. The clearance detector can be defined as weighted boxes around the waypoints as in Fig.~\ref{fig:hybrid:wpc}(b). However, the size of the box should be tuned since a smaller box can miss obstacles residing in between gaps, whereas a bigger box cannot get through a narrow pathway smoothly. To avoid this, we created a piecewise linear trajectory to approximate the path the robot would have followed if there were no dynamic obstacles in the environment. The path shown in the example Fig.~\ref{fig:hybrid:wpc}(c) is detected as not clear since the initial part of the trajectory is blocked by an obstacle (shown as a red blob).
% \vspace{-2mm}

% \vspace{-4mm}

\subsection{Filtering} %\vspace{-1mm}
\label{subsec:hybrid:filtering}
Noise in the sensor data could result in the clearance detector rapidly flip-flopping between the two planners if only the latest clearance is used for planner selection. For stabilization, the switch should take place only when we are confident about the presence of an obstacle on the path. The typical way to tackle noise in such a situation is to check the likelihood $\mathcal{L}(b | O_{t-n:t})$ of the path being blocked $b$ based on the past $n$ observations till the current time $t$. If the likelihood of obstacles is higher than a user-defined threshold $\tau$, we consider the path to be blocked.

We implement this strategy as a filter that keeps track of the last $n=3$ path clearance statuses from the detector. If all the statues indicate a blocked path, we use the SACPlanner, effectively using $\tau = 1$. Otherwise, DWA is used. This scheme is visualized in the \textit{Filtering} step (right bottom box) in Fig.~\ref{fig:hybrid:ros}. This design helps in using the SACPlanner when the sensors strongly indicate the presence of an obstacle on the path and results in efficient navigation as the comparatively smoother and faster approach, DWA is used most of the time and the switching occurs only if necessary. 

We use Robot Operating System (ROS) to implement this pipeline in C++ and Python. Our approach runs DWA and SACPlanner in parallel and switches between them by using the velocity prescribed by the selected planner. %\vspace{-1mm}

\subsection{Implementation Details} %\vspace{-1mm}
\subsubsection{SACPlanner}
We now provide some more details about the SACPlanner. See \cite{nakhleh2023sacplanner} for the full description. SACPlanner is a Reinforcement Learning (RL) based planner with a state space $\cS$, an action space $\cA$, and a reward function $R(\cdot,\cdot)$. 
The actions are simply the linear/angular velocity pairs $(v,\omega)$. 
For the state space, we use an image representation that allows the RL machinery that has been developed for video games. Specifically, the RL state is an image that combines a goal point and all the obstacles that are either derived from the static map or sensed by LiDAR. 

The goal point is one of the waypoints already discussed in Section~\ref{sec:hybrid:approach}. In particular, we select the first waypoint that does not coincide with an obstacle. We combine this waypoint with the Occupancy Grid representation of the ROS costmap (that represents the nearby obstacles). We then create a polar representation of the waypoint and obstacles, where the horizontal axis represents the distance from the robot and the vertical axis represents the angle. (See Fig.~\ref{fig:hybrid:wp_polar}(b) for an example.)
% \vspace{-3mm}

\begin{figure}[h]
	\centering
	\begin{subfigure}[c]{0.49\textwidth}
            \centering
	    \includegraphics[height=0.50\textwidth]{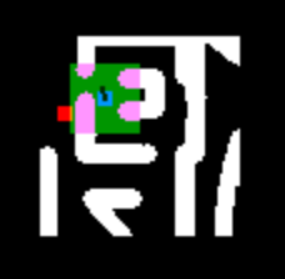}%\hspace{-1mm}
            \label{fig:hybrid:dummy_cart}
        \end{subfigure}
	\begin{subfigure}[c]{0.49\textwidth}
            \centering
	    \includegraphics[height=0.50\textwidth]{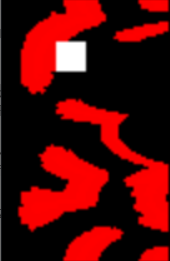}
            \label{fig:hybrid:dummy_polar}
        \end{subfigure}
	\caption{Dummy training environment for SACPlanner (left) and the associated polar costmap (right).}\label{fig:hybrid:dummy}
\end{figure}

To train SACPlanner it is convenient and faster to train it offline than in real time, so we utilize a simulated ``dummy environment''. For each training episode, we pick a synthetic obstacle map and place a robot starting point and a waypoint as in Fig.~\ref{fig:hybrid:dummy} (left). The episode is successful if the robot reaches the waypoint. The RL state during the training is the associated polar costmap, as described above and shown in Fig.~\ref{fig:hybrid:dummy} (right). 

We train SACPlanner using a Soft Actor-Critic (SAC) approach~\cite{haarnoja2018softa, haarnoja2018softb}, where the actor is a policy function and the critic evaluates the actor-value function. SAC augments the standard RL objective with an additional entropy maximization term. We also use the RAD~\cite{laskin2020reinforcement} and DrQ~\cite{kostrikov2020image} methods that apply a variety of image augmentations when training the actor/critic functions. 

The reward function $R(s, a)$ for taking action $a$ in state $s$ is defined as follows. Let $(d_{\mathrm{old}}, \theta_{\mathrm{old}})$ be the distance and bearing
to the next waypoint in state $s$, let $s'$ be the new state after taking action $a$, 
and let $(d_{\mathrm{new}}$, $\theta_{\mathrm{new}})$
be the distance and bearing in state $s'$. 
$$
\begin{array}{rl}
    & R(s,a)=\left(d_{\mathrm{old}}- d_{\mathrm{new}}\right) \cdot\left(1 \mbox{~if } d_{\mathrm{old}}- d_{\mathrm{new}} \ge 0, \mbox{~else } 2 \right) \\
    &+\left(|\theta_{\mathrm{old}}|-|\theta_{\mathrm{new}}|\right)\cdot\left(1 \mbox{~if } |\theta_{\mathrm{old}}|-|\theta_{\mathrm{new}}| \ge 0, \mbox{~else } 2 \right) \\
    &-R_\mathrm{max} \cdot\left( 1 \mbox{~if collision, else } 0 \right)  \\
    &+R_\mathrm{max} \cdot\left( 1 \mbox{~if } d_{\mathrm{new}}=0, \mbox{~else } 0\right)\\&-G(s'),
  \label{eq:rew}
\end{array}
$$
where $R_\mathrm{max}$ is the reward/penalty for reaching the waypoint or hitting an obstacle, and $G(s')$ is the product of a truncated Gaussian kernel centered at the robot location and the occupancy grid in state $s'$. (The kernel is represented by the green square in Fig.~\ref{fig:hybrid:dummy}.) We incentivize direct navigation by doubling the penalty for moving away from the waypoint vs.\ moving towards it.
After 10000 training episodes, we achieve a 98\% episode success rate. For more details on the training performance see \cite{nakhleh2023sacplanner}.

\subsubsection{Running the planners in parallel}
To implement the hybrid planner we used the \texttt{move\_base} ROS package \cite{rosmovebasewiki}. We instantiate three planners using its base planner class: (1) DWA, (2) SACPlanner, and (3) Hybrid Planner. While the first two compute the appropriate velocity profile, only the latter can send the velocity commands to the motion controller. The hybrid planner calls both DWA and SACPlanner functions for its planning functions, effectively running them in parallel. In the output function, responsible for generating the velocity vector, the planner runs the decision logic described in Section~\ref{subsec:hybrid:sacplanner} and publishes the velocity computed by the selected planner only.

\section{Experiments and Evaluation}
\label{sec:hybrid:experiemnts}

\begin{figure}[!ht]
    \centering
    \begin{subfigure}{1.00\textwidth}
        \centering
        \includegraphics[height=0.43\textwidth]{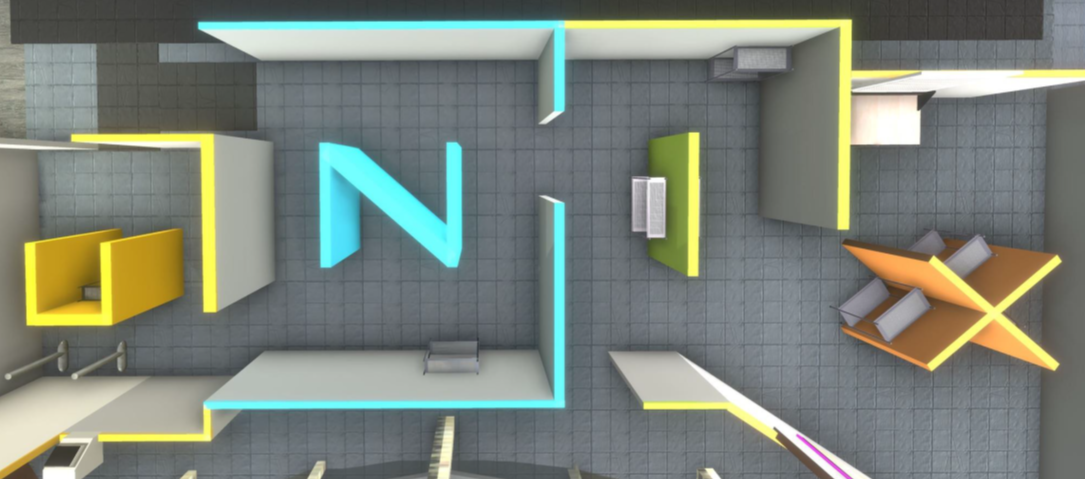}
        \caption{UNIX maze testbed}
        \label{fig:hybrid:maze}
    \end{subfigure}
    \hfill
    \begin{subfigure}{0.24\textwidth}
        \centering
        \includegraphics[height=1.0\textwidth]{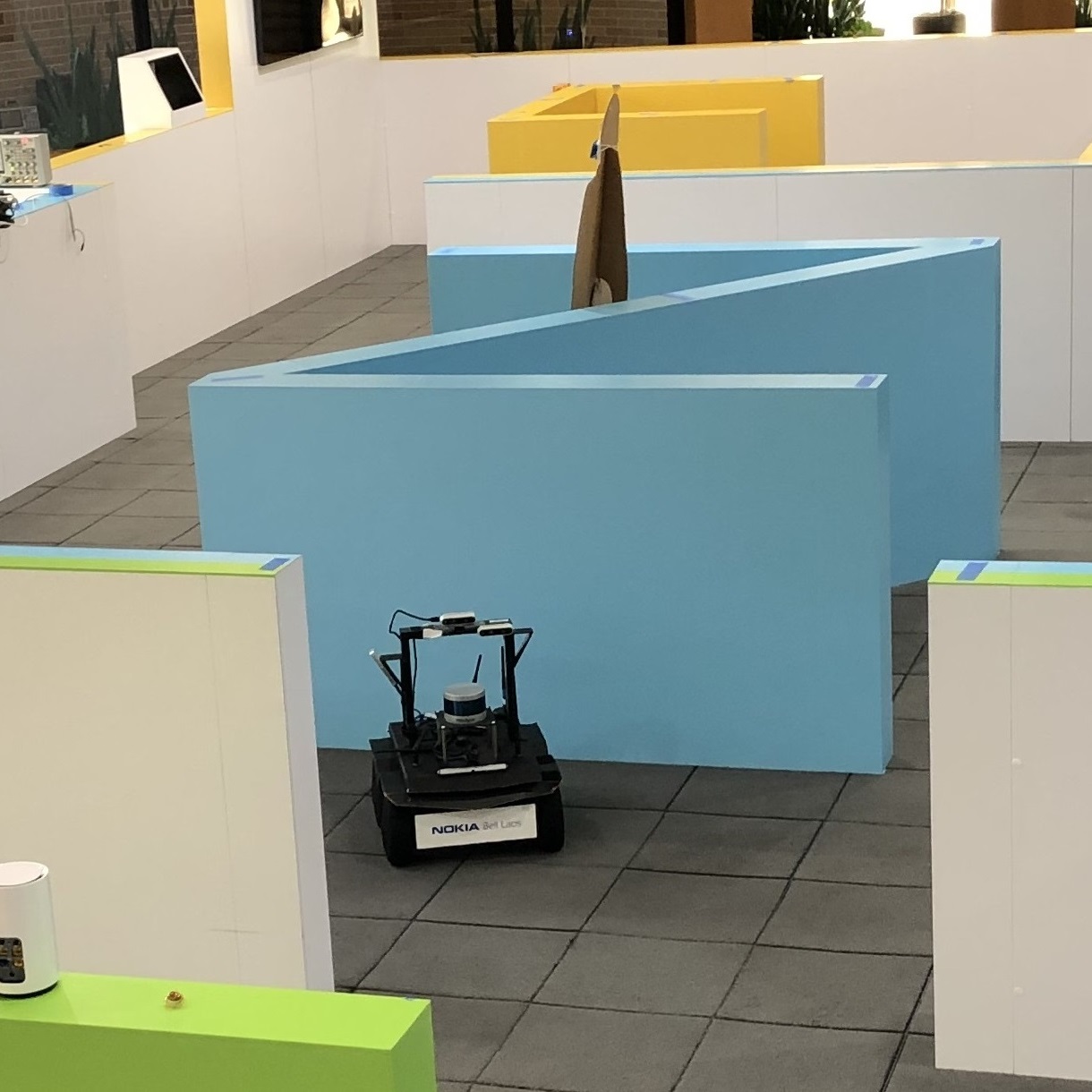}
        \caption{N-I room doorway}
        \label{fig:hybrid:c1}
    \end{subfigure}
    \begin{subfigure}{0.24\textwidth}
        \centering
        \includegraphics[height=1.0\textwidth]{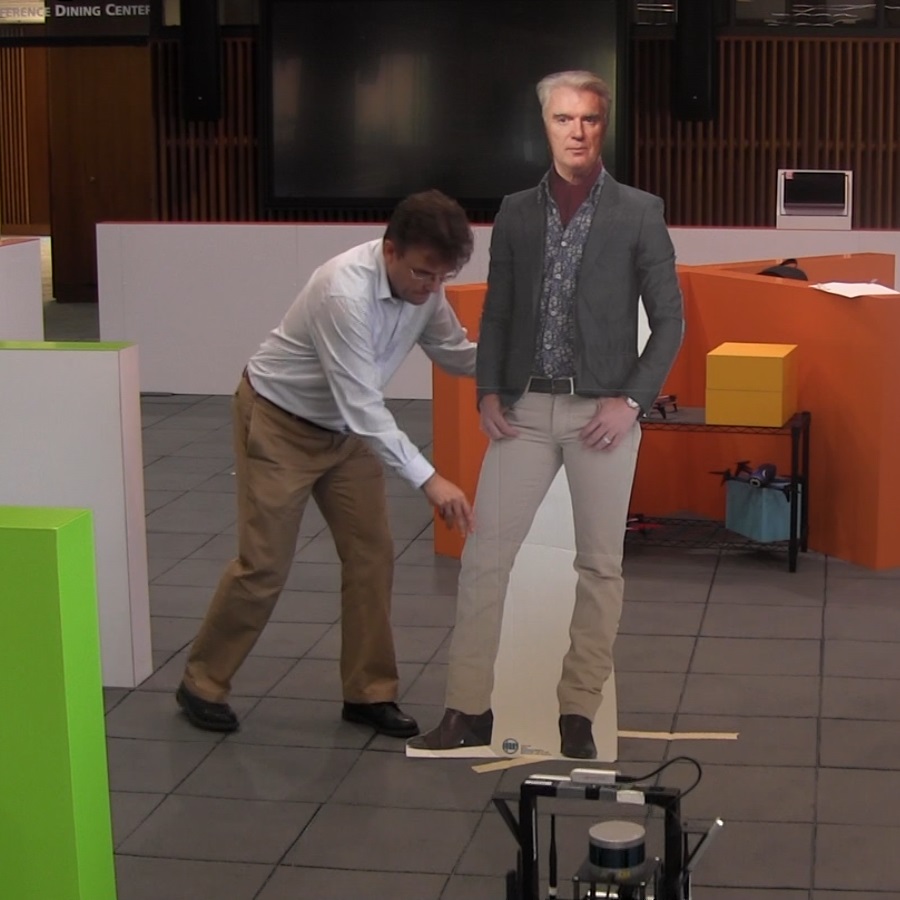}
        \caption{I-X room cardboard}
        \label{fig:hybrid:c2}
    \end{subfigure}
    \begin{subfigure}{0.24\textwidth}
        \centering
        \includegraphics[height=1.0\textwidth]{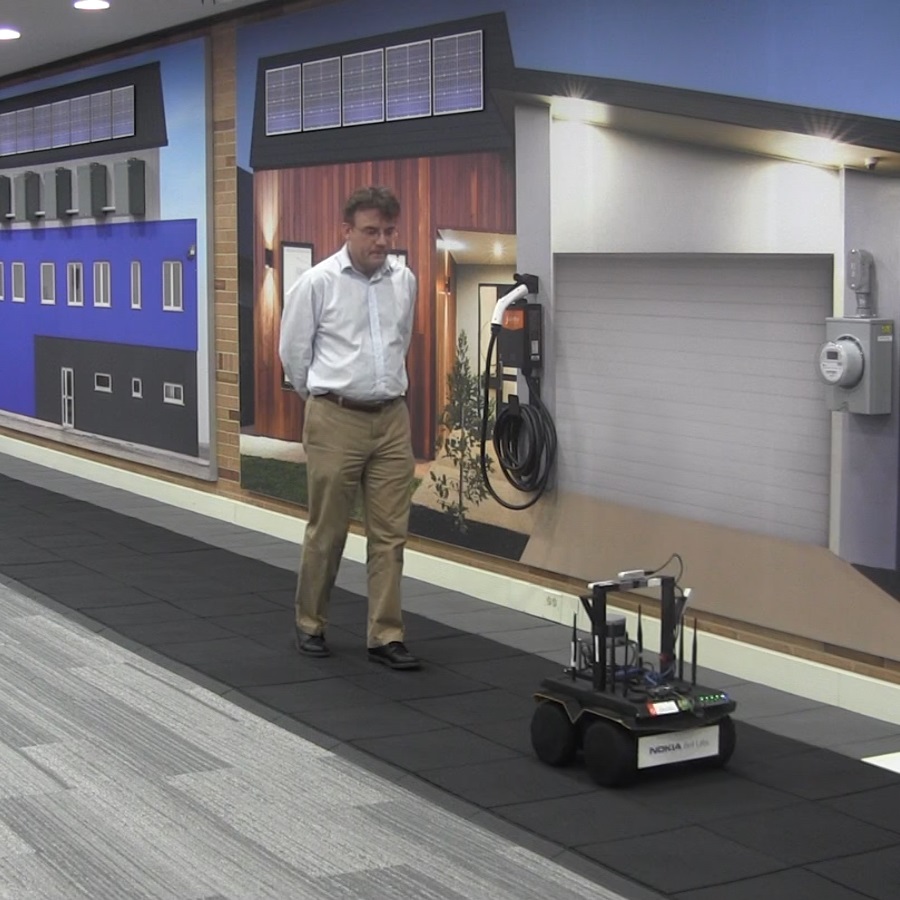}
        \caption{Approaching front}
        \label{fig:hybrid:c3}
    \end{subfigure}
    \begin{subfigure}{0.24\textwidth}
        \centering
        \includegraphics[height=1.00\textwidth]{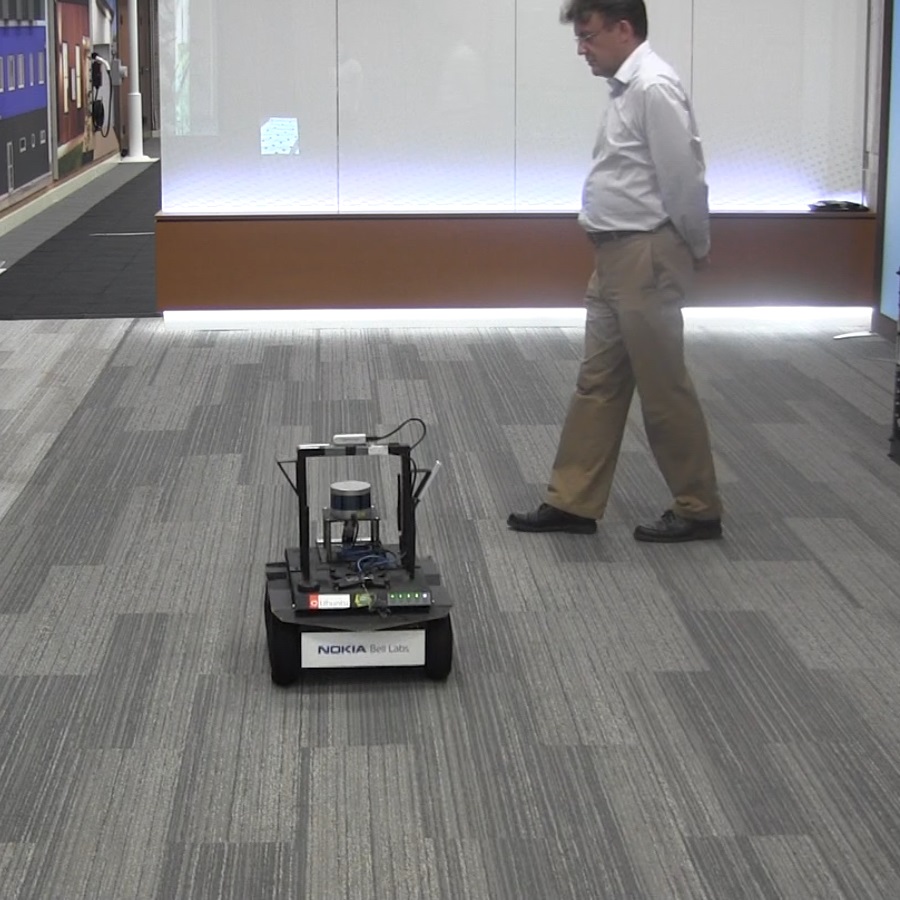}
        \caption{Crossing path}
        \label{fig:hybrid:c4}
    \end{subfigure}
    \caption{Real experimental environment and 4 test case scenarios (C1-4) from left to right.}
    \label{fig:hybrid:scenarios}
\end{figure}

\subsection{Experiment Setup}
Our experimental setup is similar to \cite{nakhleh2023sacplanner} for fair comparison. We run experiments using a ClearPath Robotics Jackal robot~\cite{clearpath} in an indoor facility with an open room and a maze with narrow pathways and tight corners, as shown in Fig.~\ref{fig:hybrid:maze}. We refer to the maze as the \textit{UNIX maze room} after the letters that constitute the walls inside the maze. In the discussion ahead, we also refer to these letters to indicate the location of the experiment. The robot uses a Velodyne LiDAR running at 10Hz for perception and the planner runs at 5Hz (half the sampling rate of the LiDAR). We study different challenging scenarios in a known map as follows:\\ 
% \vspace{-2mm}

\textbf{(C1)}
    \textbf{Obstacle-Free Intricate Trajectory:} 
    \label{subsection:c1}
    This task evaluates if the robot is able to traverse on a serpentine trajectory passing through a narrow doorway. Moving on this trajectory requires that the robot make a $180^\circ$ turn. For this setup, we move the robot from Room I to Room N through a narrow doorway as shown in Fig.~\ref{fig:hybrid:c1}. Successful traversal requires that the robot closely follows the global plan on the known map. The challenge for the local planners lies in adjusting their speed timely while accounting for the inertia to avoid collision with the walls.\\%\vspace{-1mm}

    \textbf{(C2)}
    \textbf{Unexpected Static Obstacle on Path:} 
    \label{subsection:c2}
    In this case, we test if the robot is able to react well to an unexpected object on the path that appears after the global planning is done and stays at a fixed location for the rest of the experiment. This experiment is realized by moving the robot between Room I and Room X, as shown in Fig.~\ref{fig:hybrid:c2}. Here we use a life-sized cutout of a person as the static obstacle and place it on the robot's global path after the robot starts moving. This setup is similar to \textit{Doorway} setting in Raj et al.~\cite{raj2023targeted}. Successful execution requires that the robot moves past the obstacle from the side.\\

    \textbf{(C3)} 
    \textbf{Dynamic Obstacle on Path:}
    \label{subsection:c3}
    Here we test the robot's ability to dynamic obstacles on the robot's global path. For this, we move the robot in a straight line in an open area and a pedestrian walks quickly toward the robot after the robot starts moving on the global path, in a straight line. An obstacle moving at a high speed makes it difficult for the local planner to react in time as the obstacle only shows up after it has entered the robot's local map and keeps changing the location. This situation is shown in Fig.~\ref{fig:hybrid:c3} and is similar to the \textit{Frontal} setting in Raj et al.~\cite{raj2023targeted}. To achieve success in this case, the robot must react early and back up or move around the pedestrian, or else it will collide with the pedestrian.\\

    \textbf{(C4)}
    \textbf{Dynamic Obstacle Crossing the Path:}
    \label{subsection:c4}
    While C3 checks the situation when the dynamical obstacle moves directly towards the robot, here we test if the robot can react well when a pedestrian crosses the robot's straight line path perpendicularly. Fig.~\ref{fig:hybrid:c4} shows this test case. This is similar to the \textit{Intersection} case in Raj et al.~\cite{raj2023targeted}. In this situation, even if the robot observes the pedestrian on its local map, it may not react in time as the obstacle is not yet on the global path. A successful execution requires the robot to back up to turn away from the pedestrian before moving ahead.
    
We compare the hybrid planner with DWA and SACPlanner across all these situations for 10 runs for C1, C2, and C3, and for 3 runs for C4. In C1 and C2, we also switch the start and goal location for half of the runs. As we focus on task efficiency, we compare the average distance traversed, velocity, time taken to navigate, and the number of collisions (in percentage) for each planner. 

\subsection{Results}
\label{sub sec:hybrid:results}

\begin{figure*}[!ht]
    % \vspace{3mm}
    \centering
    \begin{subfigure}{0.49\textwidth}
        \centering
        \includegraphics[width=\linewidth]{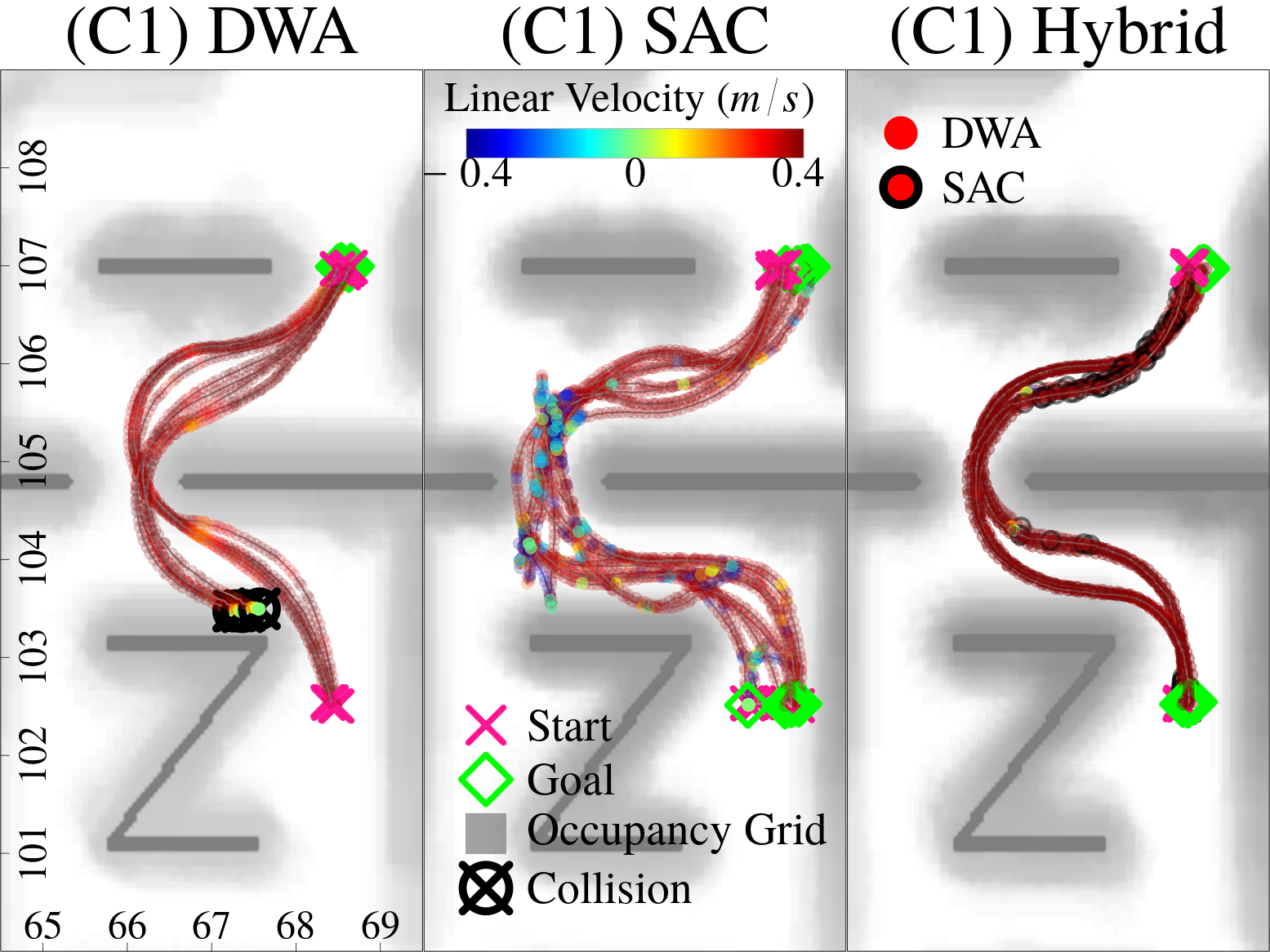}
        \caption{} % Optionally add a caption here
        \label{fig:hybrid:case1_hybrid}
    \end{subfigure}\hspace{-1mm}
    \begin{subfigure}{0.49\textwidth}
        \centering
        \includegraphics[width=\linewidth]{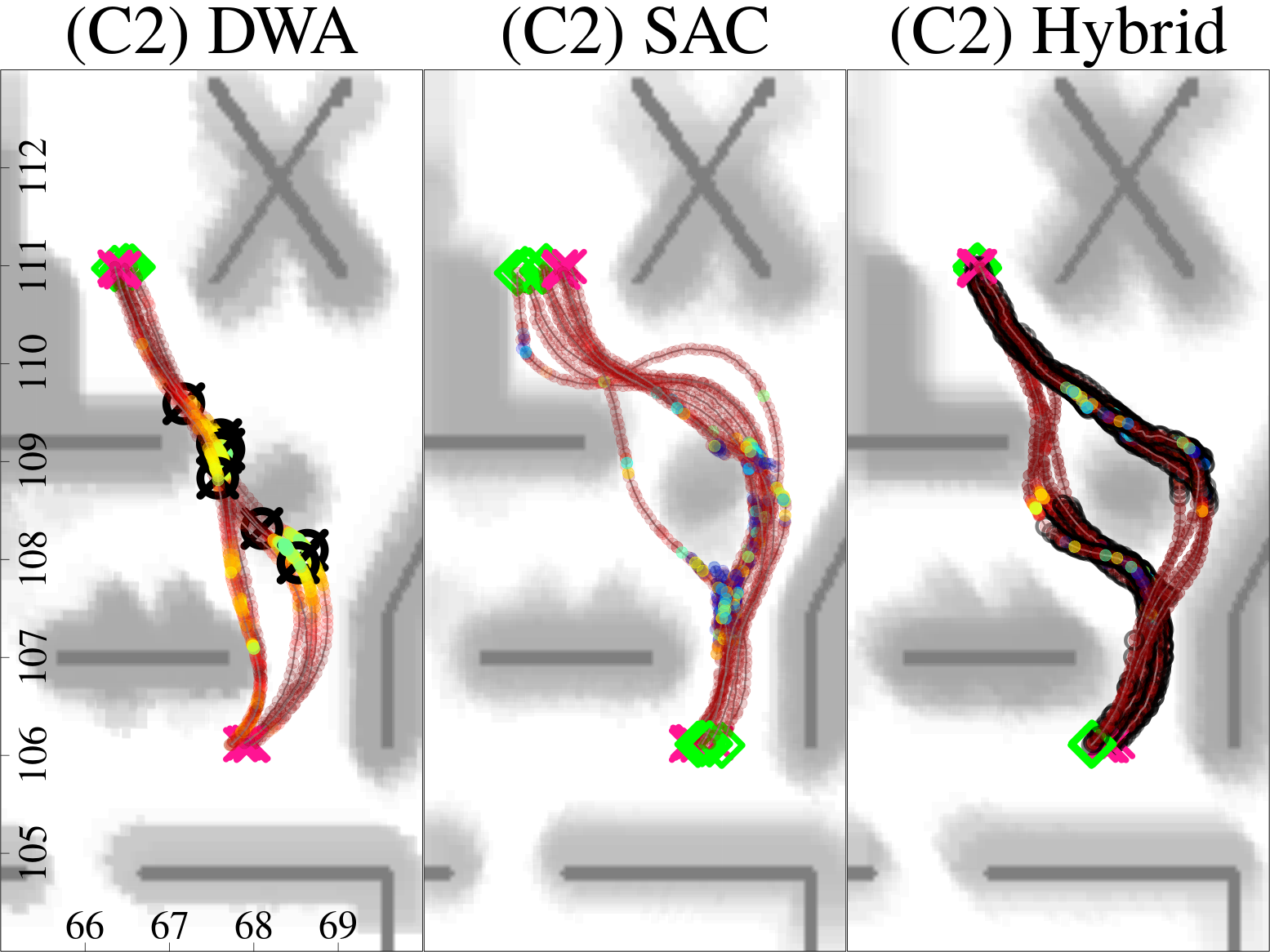}
        \caption{} % Optionally add a caption here
        \label{fig:hybrid:case2_hybrid}
    \end{subfigure}\hspace{-1mm}    
    \vfill
    \begin{subfigure}{0.49\textwidth}
        \centering
        \includegraphics[width=\linewidth]{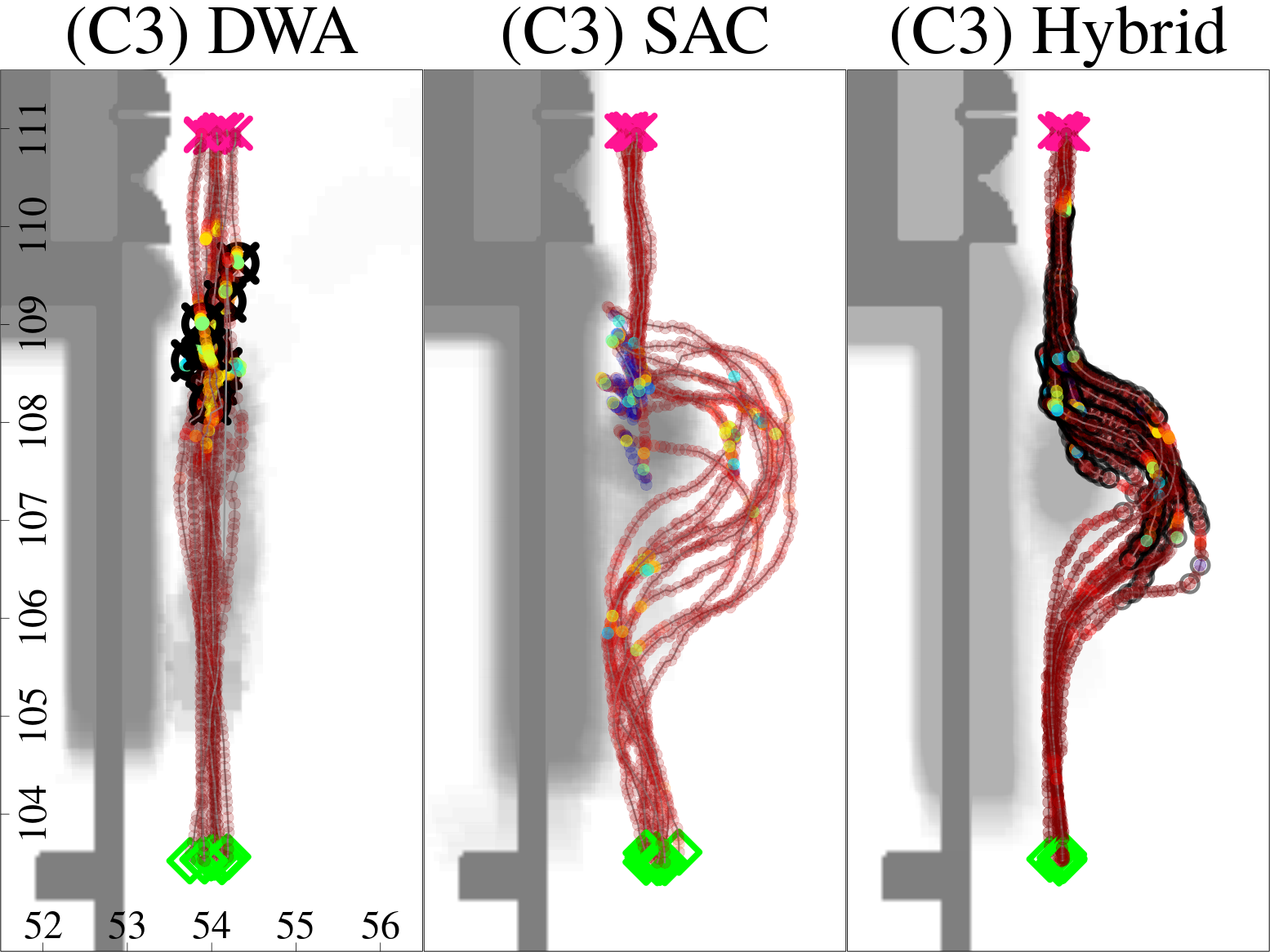}
        \caption{} % Optionally add a caption here
        \label{fig:hybrid:case3_hybrid}
    \end{subfigure}\hspace{-1mm}    
    \begin{subfigure}{0.49\textwidth}
        \centering
        \includegraphics[width=\linewidth]{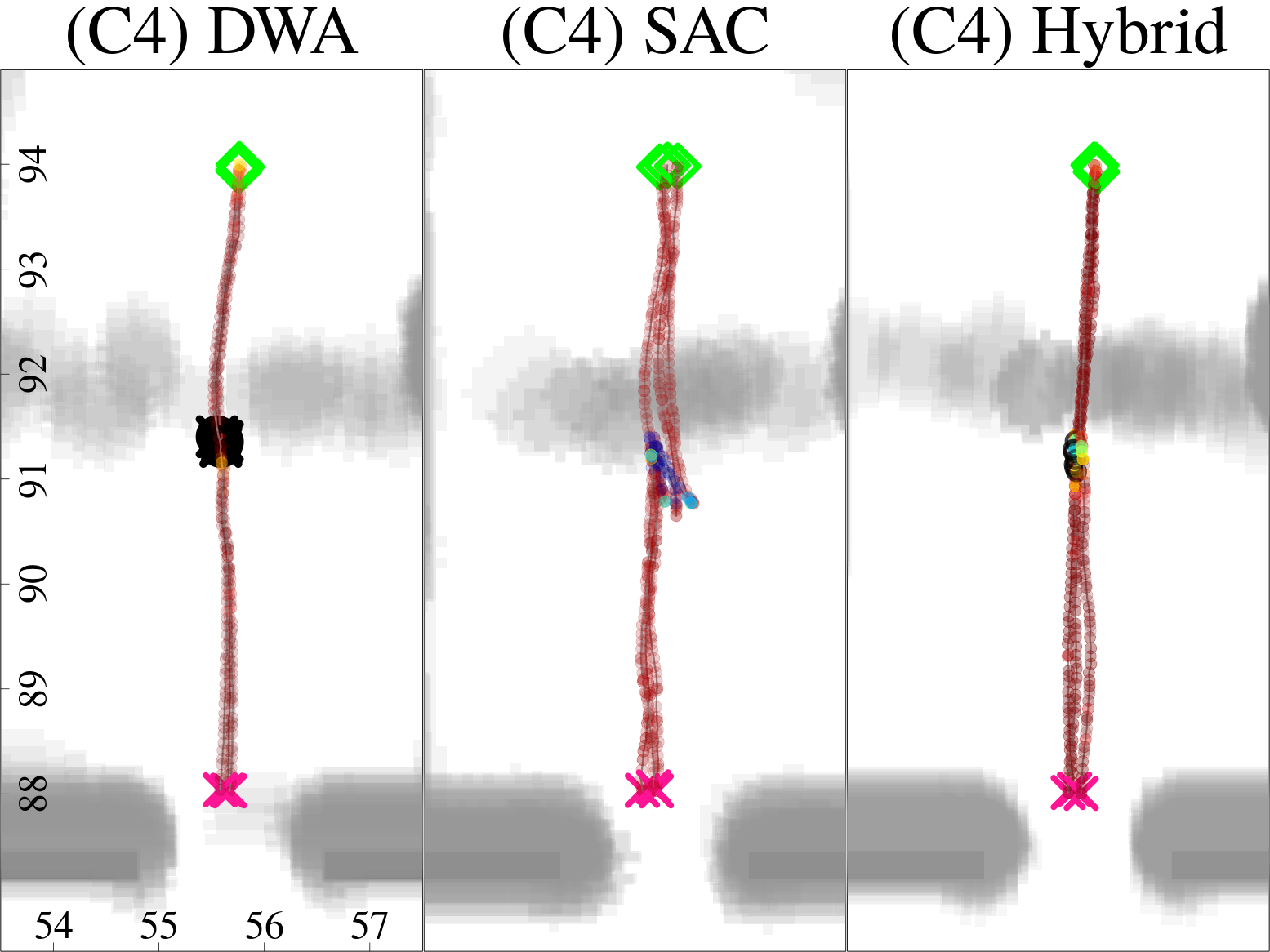}
        \caption{} % Optionally add a caption here
        \label{fig:hybrid:case4_hybrid}
    \end{subfigure}
    % \vspace{-2mm}
    \caption{Trajectory comparison between DWA, SACPlanner vs. Hybrid planner agent for each test case.}
    \label{fig:hybrid:case_hybrid}    
\end{figure*}

\begin{figure*}[!th]
    \centering
    \begin{subfigure}{0.49\textwidth}
        \centering
        \includegraphics[width=\linewidth]{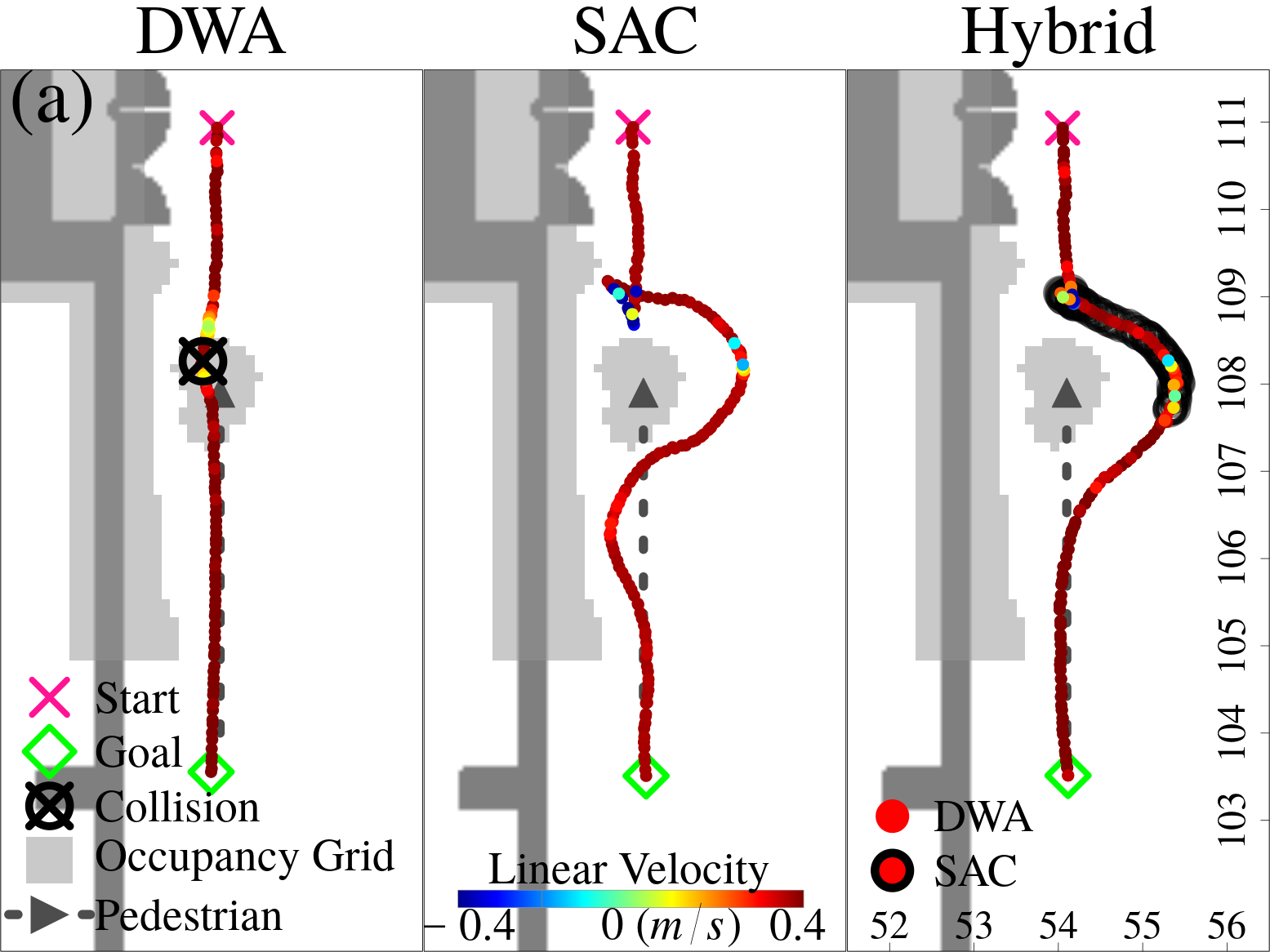}
        \caption{} % Optionally add a caption here
        \label{fig:hybrid:ped_traj}
    \end{subfigure}\hspace{-2mm}
    \begin{subfigure}{0.49\textwidth}
        \centering
        \includegraphics[width=\linewidth]{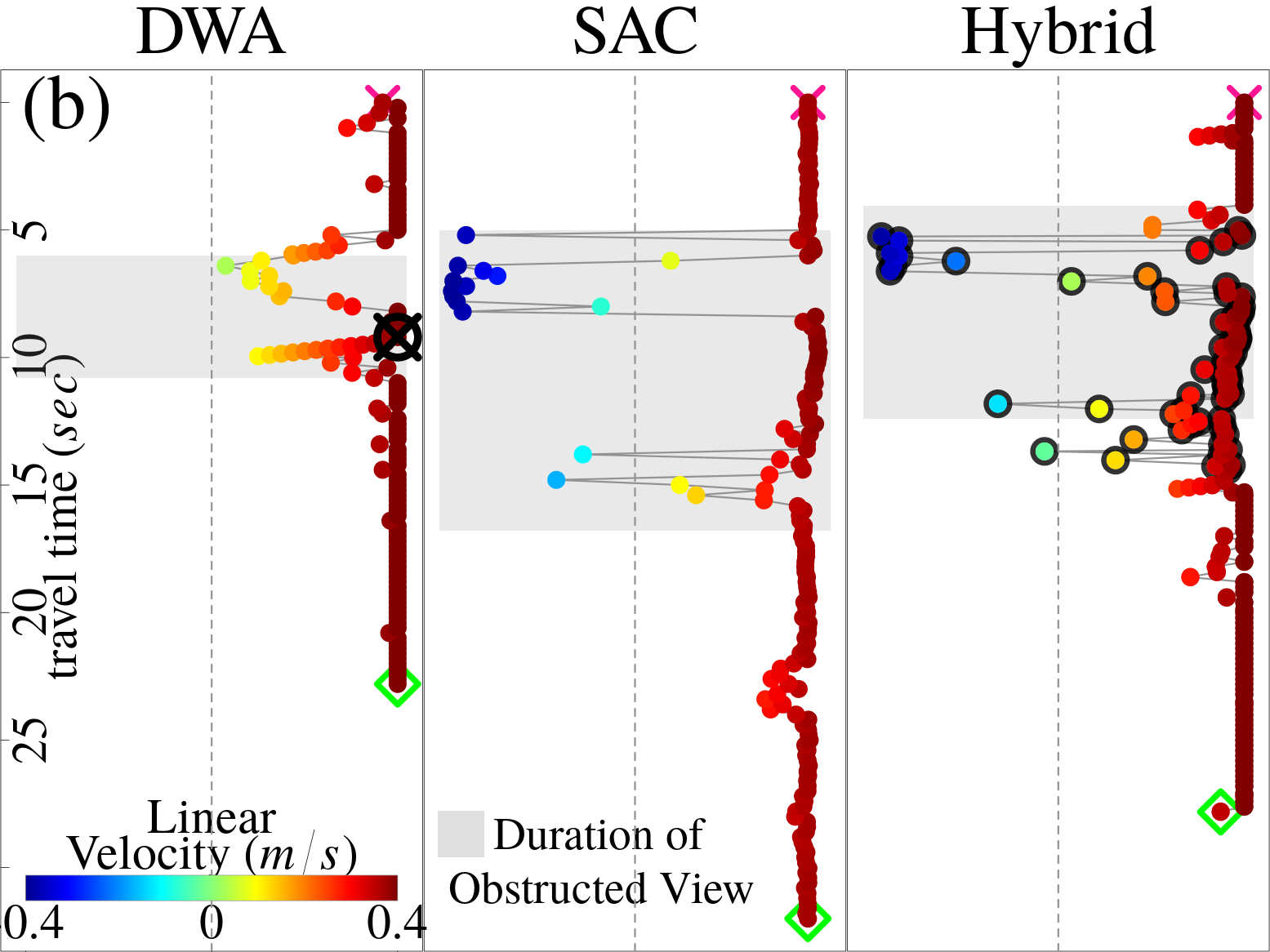}
        \caption{} % Optionally add a caption here
        \label{fig:hybrid:ped_linv}
    \end{subfigure}\hspace{-2mm}
    \vfill
    \begin{subfigure}{0.49\textwidth}
        \centering
        \includegraphics[width=\linewidth]{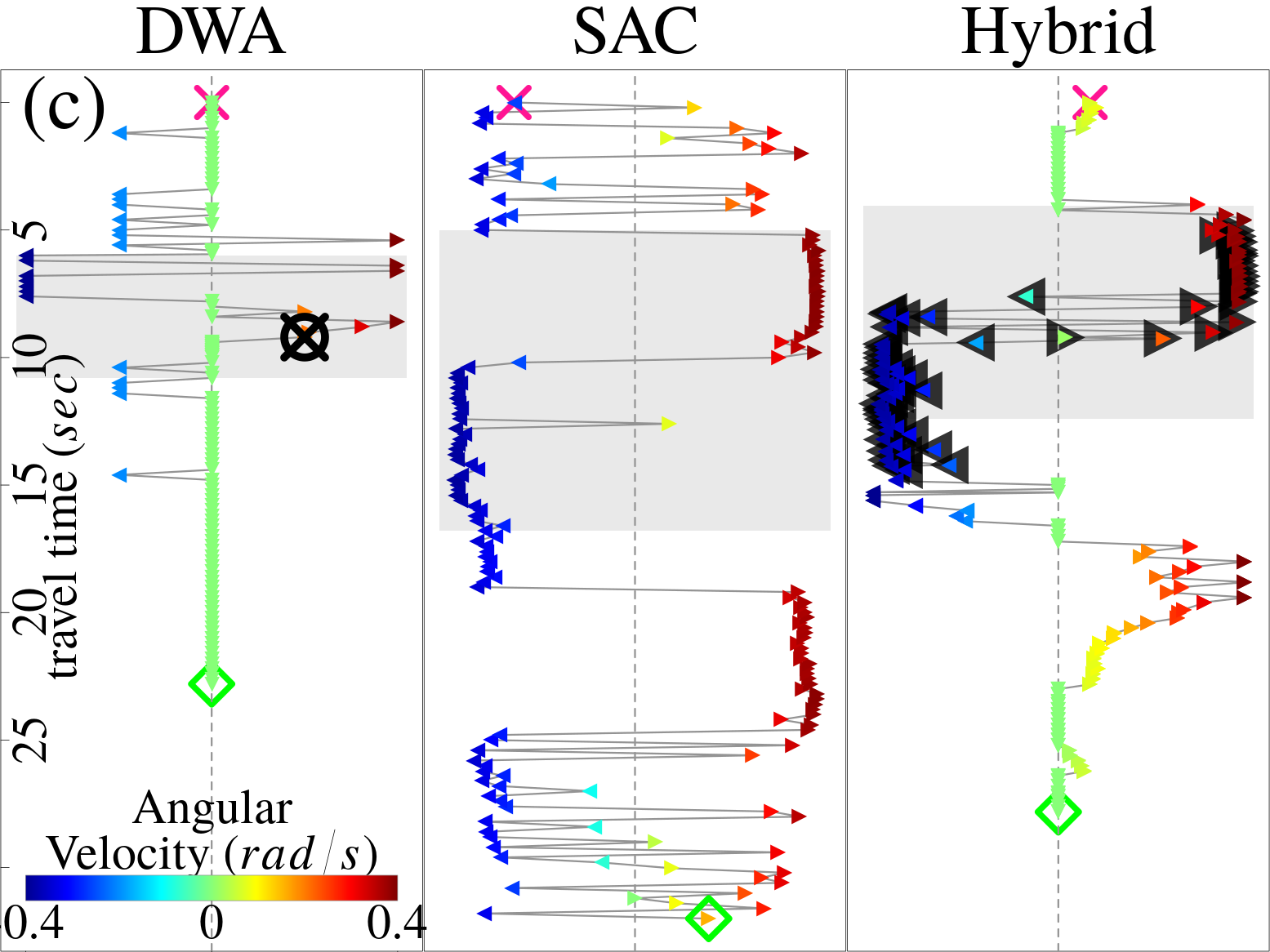}
        \caption{} % Optionally add a caption here
        \label{fig:hybrid:ped_angv}
    \end{subfigure}\hspace{-2mm}
    \begin{subfigure}{0.49\textwidth}
        \centering
        \includegraphics[width=\linewidth]{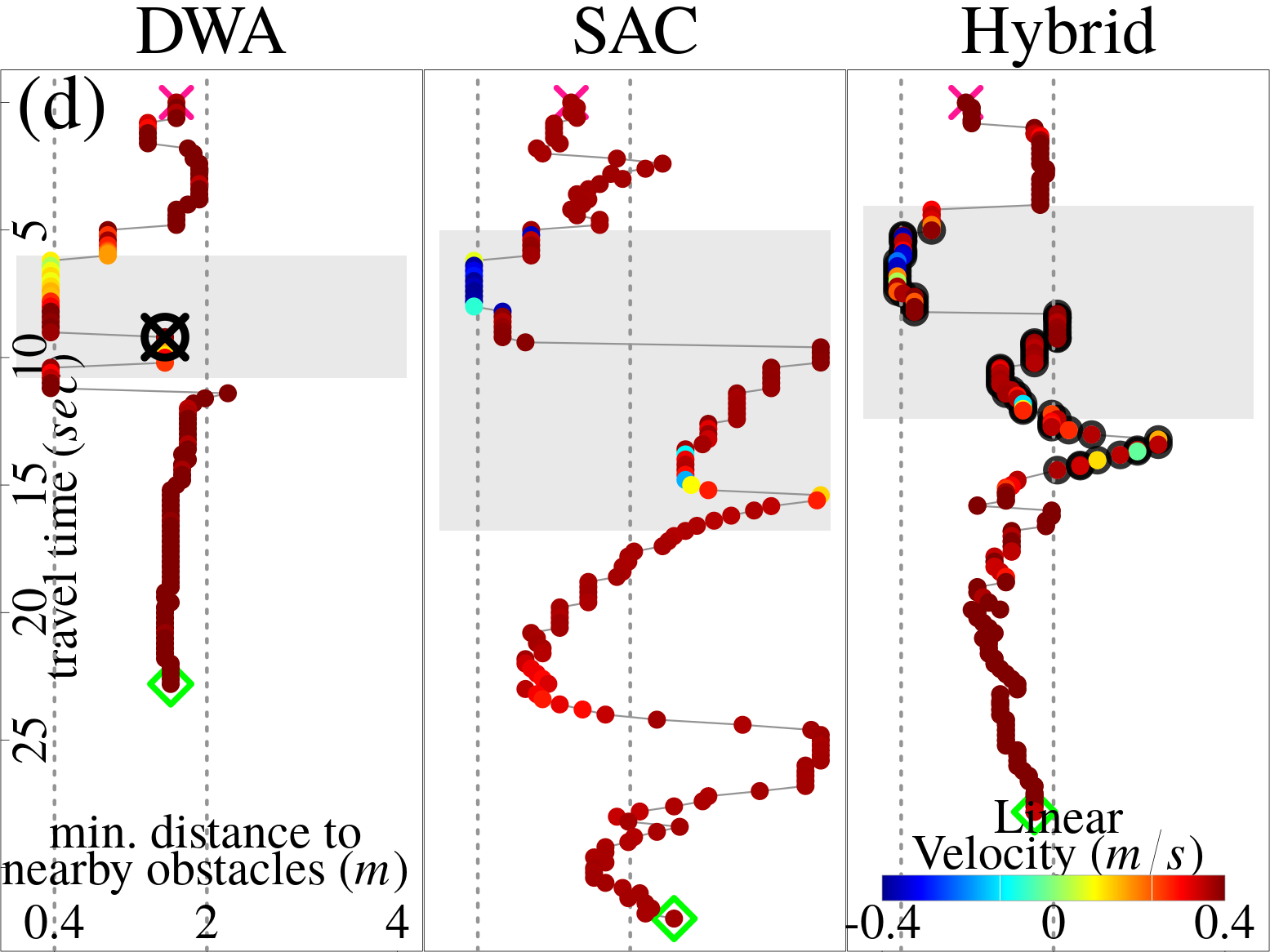}
        \caption{} % Optionally add a caption here
        \label{fig:hybrid:ped_mdist}
    \end{subfigure}
    % \vspace{-2mm}
    \caption{Trajectory comparison between DWA, SAC, and Hybrid planners based on logs from the scenario (C3).}
    \label{fig:hybrid:ped_ex}
\end{figure*}

The robot trajectories for each of C1-C4 are shown in Fig.~\ref{fig:hybrid:case_hybrid}. We denote the start and goal along with the collision points. 
The color of the trajectory represents linear velocity and the circles with a thick black border represent where SACPlanner has been used for the hybrid planner. We also show the Occupancy Grid values in gray (taken from the map and the LiDAR). For C3 \& C4, the gray shading captures all the positions of the unexpected obstacle over time. The three local planners have qualitatively different behaviors. DWA collides with the walls or obstacles in all cases except a few in C1. SACPlanner allows the robot to circumnavigate the obstacles but it results in the robot moving slowly, even with negative velocity in some cases, and usually results in a long detour.  
In each case, the hybrid planner helps the robot avoid obstacles successfully, while moving on a smooth trajectory with high speed, making it more suitable than either individual planner.

Table~\ref{tab:hybrid:exp_results} summarizes the quantitative comparison averaged over 10 runs for C1-C3. (For brevity we refer to SACPlanner as SACs in the table.) The hybrid planner is faster than both DWA and SACPlanner, as shown by the higher average speed. Collisions when DWA is used, result in the robot covering a shorter distance without success. SACPlanner has the same success rate as the hybrid planner, but the hybrid planner results in a relative improvement of \textbf{26\%} in the navigation time with \textbf{18\%} shorter path length. Notably, our planner exhibits safe and efficient navigation in situations similar to prior works~\cite{raj2023targeted}, without the need to learn when to switch with a neural network.

To understand more deeply why the hybrid planner performs better, we show in Fig.~\ref{fig:hybrid:ped_ex} the behavior of each planner in a single run from the test case (C3). The beginning and ending behavior of the hybrid planner is closer to a straight line, since DWA is selected using full-speed (dark red) linear velocities as in Fig.~\ref{fig:hybrid:ped_traj}, \ref{fig:hybrid:ped_linv}. The shaded area in Fig.~\ref{fig:hybrid:ped_angv}-\ref{fig:hybrid:ped_mdist} represents the duration of time when LiDAR first captures the pedestrian in its view in the polar costmap until he stops walking at the location $x=54$m, $y=108$m. From the overall travel time, the hybrid planner gets the robot to the goal faster than the SACPlanner without any collisions. The reaction time (in seconds) to begin turning starts from when the robot first enters the shaded area, 4.05s for hybrid planner, 5s for SACPlanner, and 5.99s for DWA planner. In addition, the hybrid planner gets around the pedestrian about 3.5 seconds faster than SACPlanner (8.36s $<$ 11.8s). The transition in rotational velocities is much smoother in the hybrid case since it reverts to DWA after passing around the pedestrian as in Fig.~\ref{fig:hybrid:ped_angv}. Moreover, when the robot is far from the pedestrian the angular velocity is zero (green). This explains how the hybrid planner almost eliminates the jerky motion caused by SACPlanner. Fig.~\ref{fig:hybrid:ped_mdist} shows the distance to the nearest `front obstacle' (within $\pm\frac{\pi}{4}$rad range from the current yaw). The hybrid planner manages both safe and efficient distance during the whole travel time.  

The results highlight that the hybrid planner makes appropriate use of both planners for navigation in various scenarios. It moves smoothly and quickly through clear areas and is responsive in the face of obstacles discovered along the path. This behavior is also safer, both for the robot and for the humans acting as the dynamic obstacles.%\vspace{-3mm}

\begin{table}[!h]
	\caption{Summary statistics of trajectories from the real-world experiments using DWA, SAC, and Hybrid planner.}
	% \vspace{-2mm}
	\label{tab:hybrid:exp_results}
	\centering{%\scriptsize
	\begin{tabular}%{@{\hspace{1pt}}c@{\hspace{1pt}}|r|r|r|r|r|r|r|r|r}
	{l|r|r|r|r|r|r|r|r|r}
	% \hline
	&	\multicolumn{3}{c|}{(C1)}					&	\multicolumn{3}{c|}{(C2)}					&	\multicolumn{3}{c}{(C3)}					\\\hline
	&	\multicolumn{1}{c|}{DWA}	&	\multicolumn{1}{c|}{SAC}	&	\multicolumn{1}{c|}{Hybrid}	&	\multicolumn{1}{c|}{DWA}	&	\multicolumn{1}{c|}{SAC}	&	\multicolumn{1}{c|}{Hybrid}	&	\multicolumn{1}{c|}{DWA}	&	\multicolumn{1}{c|}{SAC}	&	\multicolumn{1}{c}{Hybrid}	\\\hline
Time	&	21.80	&	37.20	&	{\bf 21.10}	&	30.70	&	28.50	&	{\bf 23.60}	&	27.50	&	33.10	&	{\bf 27.10}	\\\hline
Distance	&	{\bf 7.13}	&	10.70	&	7.67	&	{\bf 5.47}	&	8.57	&	7.41	&	{\bf 8.77}	&	10.80	&	9.40	\\\hline
Speed	&	0.33	&	0.29	&	{\bf 0.39}	&	0.18	&	0.30	&	{\bf 0.31}	&	0.32	&	0.33	&	{\bf 0.36}	\\\hline
Collision	&	50\%	&	0\%	&	{\bf 0\%}	&	100\%	&	0\%	&	{\bf 0\%}	&	100\%	&	0\%	&	{\bf 0\%}	\\\hline%\vspace{-3mm}

	\end{tabular}}
\end{table}

\section{Conclusion}
\label{sec:hybrid:conclusion}
We present a hybrid local planner that combines DWA, a classical planning method, and SACPlanner, a learning-based planning approach. Experiments on a ClearPath Jackal robot in various situations show that the proposed approach is safer and more efficient than the two constituent planners, showing a significant improvement in navigation time without any collision. The design of our switch forgoes the need to collect data and train another neural network, making it more suitable than learning-based switching from real-world development.

We focus on a heuristics-based approach to define the criteria for switching between the planners. Future work will explore more sophisticated approaches. A drawback of the hybrid approach is that the shortcomings of the constituent planners appear when the hybrid planner uses them. An example of this would be some jerky motion of the robot, owing to the SACPlanner, while the robot tried to avoid the obstacle. In the future, we intend to work on improving the constituent planners to further improve the overall performance of the hybrid planner.

%% file: Chapter10.tex
%Chapter 10

\renewcommand{\thechapter}{10}

\chapter{Conclusion and Future Work}
\label{chapter:conclusion}

\section{Summary of Contributions}
\label{sec:conclusions:contribution}
This dissertation addresses the challenge of limited perception and planning horizon of the robots, originating from the limited observations. Utilizing the regularities in the environment, this dissertation presents methods to learn and predict the structural and geometric patterns, spatiotemporal patterns, and semantic patterns in the robot's surroundings to enhance the perception and planning for robots. Considering the uncertainty in the predictions, the dissertation also presents risk-management approaches to pave the way for the safe deployment of these methods. In summary, we make the following contributions through this dissertations:

\begin{itemize}
    \item Using the structural and geometric patterns in the objects around us, we proposed methods to infer the yet-unobserved environment from partial observations, both in 2D, as well as 3D representations, and presented approaches, named, ProxMaP and PredNBV, to use these predictions and improve planning for single and multi-robot systems.
    
    \item Using the spatiotemporal patterns from moving entities, we proposed methods to learn to predict and coordinate in decentralized multi-robot tracking systems, in a scalable manner, and show that end-to-end learning can provide a performance advantage over traditional plug-and-play solutions.

    \item Using the semantic information inherent in natural language, along with world knowledge from large vision-language models, we show how they can be used to aid the robots in helping humans, and other robots accomplish tasks by acting as intelligent assistants. 

    \item To allow the safe deployment of prediction-based approaches, we present methods using implicit and explicit uncertainty measures, using uncertainty extraction and hybrid switching,  to provide risk-aware solutions.
\end{itemize}

\section{Future Directions}
\label{sec:conclusions:future_work}
This dissertation focuses on identifying, learning, and predicting three types of patterns in the environment, and presents approaches for risk-aware deployment of such predictive methods. Takeaways from the presented work, along with the latest developments in learning-based approaches, open the door to more future directions. Specifically, we identify the following directions to augment this dissertation:

\subsection{Large Models for Zero-Shot Robotic Applications}
\label{subsec:conclusions:large_models}
Chapter~\ref{chapter:fliptd} and Chapter~\ref{chapter:llm_assist} show that large models trained in large-scale data hold the potential for the robots to break free from the longstanding limitation in the form of the dearth of data. Not only do such models act as excellent replacements for the traditional learning models of robots, but they also seem to show an understanding of the fundamental concept of the world around us, which can help the robots reach the level of intelligence necessary to converse and interact with humans. However, the nature of robot observations does add some constraints that must be respected. For vision-based large models, the future works should address how to understand and use the \textit{uninteresting} observations, such as plain walls, which these methods may not have come across in the training data, which tend to focus on \textit{interesting} observations only. For language-based models, similar issues occur in the form of a lack of grounding and need further study for their meaningful extension to robots.

\subsection{Risk-Aware Methods for Large Models}
\label{subsec:conclusions:risk_large}
While the previous subsection looks toward the large models with an enthusiastic gaze, this subsection warns that one must proceed with caution. The large models only \textit{seem} to understand the fundamental concepts of the environment, and are hard to verify due to their black-box nature. We need to build risk-aware approaches for the deployment of these models as well. The first step in this direction would be to ensure that these methods are robust and do not fail due to rare, but natural observations, such as a robot walking into an empty room. Unless such situations are addressed, the deployment of robots in daily life may pose a danger to people around them, and themselves.

\subsection{Building End-to-End Methods}
\label{subsec:conclusions:end_to_end}
Chapter~\ref{chapter:gnn_d2coplan} shows that end-to-end methods can perform better than the simple combination of classical and learning-based approaches. This requires the classical approaches to be differentiable, which they often are not. This requires building learning-based solutions that can approximate their counterparts very well and then help the learning-based approaches train better by allowing the flow of gradients from the loss for task objective. Many planning algorithms used in practice are still classical, and hence non-differentiable. Developing differentiable versions of them would be a great step in this direction for efficient and improved solutions. 

%% file: Appendix.tex
%Appendix -- January 2015
\appendix
\renewcommand{\thechapter}{A}
\renewcommand{\chaptername}{Appendix}

\chapter{Prompts and Additional Experiments for Assisted ObjectNav}

% \chapter{Prompts for Assistive ObjectNav}
\section{Prompts}
\label{sec:llm_comm:append:llm_comm:prompt}
Following are the prompts used for our experiments. The variables passed to the prompts are indicated by chevrons ($<\cdot>$).

\subsection{No Comm. - Ground}
\label{subsec:llm_comm:no_comm_ground}
% \textbf{System Prompt:}\\
% \coloredbox{blue!10}{
\promptbox{System Prompt}{
I am a mobile ground robot equipped with a camera. You are a helper agent with access to the camera feed.\\
I am trying to reach an object in an indoor environment, and you should help me by giving me an action command to accomplish this task successfully.\\
You are only allowed to give me one of the following commands:\\
- MoveAhead: moves the robot in the forward direction by 0.25 meters.\\
- MoveBack: moves the robot in the back direction by 0.25 meters while looking forward.\\
- RotateRight: rotates the robot towards the right by 90 degrees while staying at the same location.\\
- RotateLeft: rotates the robot towards the left by 90 degrees while staying at the same location.\\
- DoNothing: the robot does not do anything and stays in place.\\
Give only an action command as the answer.
}

% \textbf{User Prompt:}\\
% \coloredbox{blue!10}{
\promptbox{User Prompt}{
The target object is $<$target object$>$.\\
The last action was $<$last action$>$. The agent executed it $<$successfully/unsuccessfully$>$ and the resulting distance to the target object is $<$shortest path distance to the target.$>$ 
}

If the agent is unsuccessful in executing the action, the following is also added to the user prompt:
% \coloredbox{blue!10}
\promptbox{}{The agent failed to execute it for the following reason: $<$blocking object$>$ is blocking the agent from moving.}

Additionally, the image from the ground agent's camera is provided.

\subsection{Communication}
\label{subsec:llm_comm:comm_dialogue}

% \textbf{Overhead Agent - System Prompt}:\\
% \coloredbox{blue!10}{
\promptbox{Overhead Agent - System Prompt}{
You are an overhead agent with a top view in an environment. You must guide a ground robot agent to reach the $<$target object$>$.
}

% \textbf{Overhead Agent - User Prompt}:\\
% \coloredbox{blue!10}{
\promptbox{Overhead Agent - User Prompt}{Give me a dialogue to pass to the ground agent to reach the target. Respond with just the dialogue.\\
You must query about the environment $C_{len}$ times before giving any form of instruction.\\
$<$Dialogue between the agents$>$
}

Additionally, the image from the overhead agent’s camera is provided.

% \textbf{Ground Agent - System Prompt}:\\
% \coloredbox{blue!10}{
\promptbox{Ground Agent - System Prompt}{You are a ground agent with an egocentric view. You need to reach the $<$target object$>$.\\
You will be questioned by an overhead agent. Be very concise with your answers.
}

% \textbf{Ground Agent - User Prompt}:\\
% \coloredbox{blue!10}{
\promptbox{Ground Agent - User Prompt}{
$<$Dialogue between the agents$>$
}

Additionally, the image from the ground agent’s camera is provided.

\subsection{Execution}
\subsubsection{Cooperative Action}
\label{subsubsec:llm_comm:cooperative_action}
% \textbf{Overhead Agent - System Prompt}\\
% \coloredbox{blue!10}{
\promptbox{Overhead Agent - System Prompt}{
You are an oracle agent with a top view in an environment. You must guide a ground robot agent to reach a target object.\\
You will be given your conversation with the ground agent which you can use to make the decision.\\
You may also be given the last action the ground agent took and its outcome.\\
You are only allowed to give only one of the following commands:\\
- MoveAhead: moves the robot in the forward direction by 0.25 meters.\\
- MoveBack: moves the robot in the back direction by 0.25 meters while looking forward.\\
- RotateRight: rotates the robot towards the right by 90 degrees while staying at the same location.\\
- RotateLeft: rotates the robot towards the left by 90 degrees while staying at the same location.\\
- DoNothing: the robot does not do anything and stays in place.\\
Give an action command as the answer.
}

% \textbf{Overhead Agent - User Prompt}\\
% \coloredbox{blue!10}{
\promptbox{Overhead Agent - User Prompt}{
The target object is $<$target object$>$.\\ 
Conversation: $<$Conversation between the agents$>$.\\
The last action was $<$last action$>$. The agent executed it $<$successfully/unsuccessfully$>$ and the resulting distance to the target object is $<$shortest path distance to the target.$>$ 
}

If the agent is unsuccessful in executing the action, the following is also added to the user prompt:
% \coloredbox{blue!10}
\promptbox{}{The agent failed to execute it for the following reason: $<$blocking object$>$ is blocking the agent from moving.}

The ground agent follows the action provided by the overhead agent.

\subsubsection{Selective Action}
\label{subsubsec:llm_comm:selective_action}

First, we obtain a command from the overhead agent using the prompt in subsection~\ref{subsubsec:llm_comm:cooperative_action}. Then we ask the ground agent whether it wishes to follow the overhead agent's suggestion using the following prompt. 

% \textbf{Ground Agent - System Prompt}\\
% \coloredbox{blue!10}{
\promptbox{Ground Agent - System Prompt}{
I am a mobile ground robot equipped with a camera. You are a helper agent with access to the camera feed.\\
I am trying to reach an object in an indoor environment. There is another agent with an overhead camera feed who is providing me with an instruction for this task.\\
Following are the valid action commands and their meaning for me:\\
- MoveAhead: moves the robot in the forward direction by 0.25 meters.\\
- MoveBack: moves the robot in the back direction by 0.25 meters while looking forward.\\
- RotateRight: rotates the robot towards the right by 90 degrees while staying at the same location.\\
- RotateLeft: rotates the robot towards the left by 90 degrees while staying at the same location.\\
- DoNothing: the robot does not do anything and stays in place.\\
I will tell you the target object and the action command that the overhead agent has given to me. You should tell me whether I should trust and follow the overhead agent's command (Yes) or whether I should disregard it and decide on my own (No).\\
You must answer in Yes or No.
}

% \textbf{Overhead Agent - System Prompt}\\
% \coloredbox{blue!10}{
\promptbox{Overhead Agent - System Prompt}{
The target object is  $<$target object$>$. The overhead agent prescribed the following action: $<$action command$>$.
}
Additionally, the image from the ground agent’s camera is provided.

If the ground agent decides not to follow the overhead agent's suggestion, we execute \textbf{No Comm.} scenario using the prompt given in subsection~\ref{subsec:llm_comm:no_comm_ground}.

\subsection{Preemptive Actions Classifier}
\label{subsec:llm_comm:append:llm_comm:preempt}
% \textbf{System Prompt}\\
% \coloredbox{blue!10}{
\promptbox{System Prompt}{
You will be given a conversation between an overhead agent and a ground agent. They are expected to talk about the environment around them and nothing else.\\
No actions should be assumed to have been taken during the period of the conversation.
Can you tell me if the conversation has any `preemptive motions'?\\
For example, dialogues such as <`I have moved towards the room'> are `preemptive motions', as they assume the agent has already executed a motion in the past.\\
If the motion is yet to be executed, and it is in the future, it is NOT a `preemptive motion'.
Answer only with Yes or No.
}

% \textbf{User Prompt}\\
% \coloredbox{blue!10}{
\promptbox{User Prompt}{
<Conversation between the agents>
 }

% \chapter{Adverserial Cases for Image Inputs to GPT-4}
\section{Adverserial Cases}
\label{sec:llm_comm:append:llm_comm:adverserial}
In our simulation experiments, we found that some observations may be perceived as adversarial or CAPTCHA-like inputs, resulting in error from \textit{GPT-4-turbo}.

\begin{figure}[h!]
    \centering
    \begin{subfigure}{0.48\textwidth}
        \includegraphics[width=\linewidth]{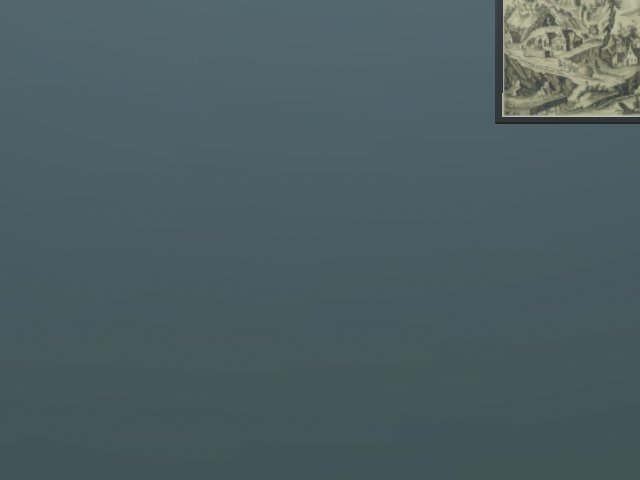}
        \caption{}
    \end{subfigure}\hfill
    \begin{subfigure}{0.48\textwidth}
        \includegraphics[width=\linewidth]{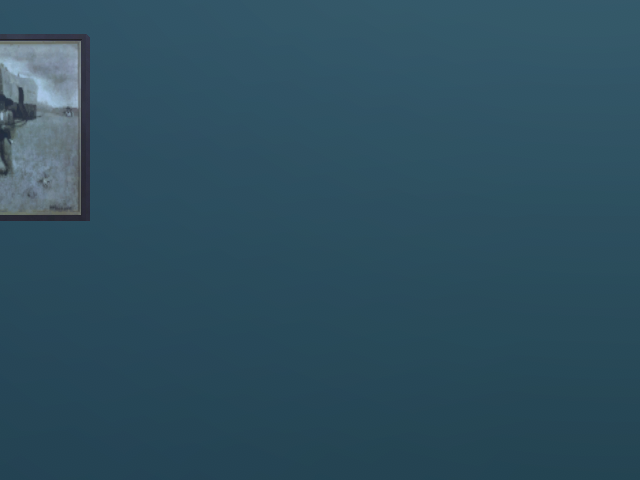}
        \caption{}
    \end{subfigure}\hfill
    \begin{subfigure}{0.48\textwidth}
        \includegraphics[width=\linewidth]{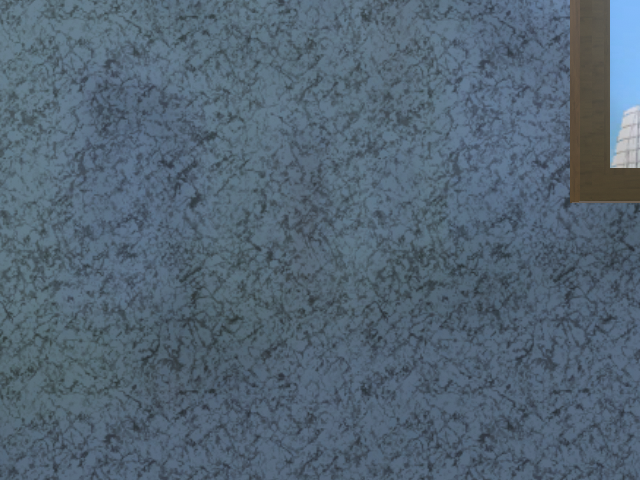}
        \caption{}
    \end{subfigure}\hfill
    \begin{subfigure}{0.48\textwidth}
        \includegraphics[width=\linewidth]{"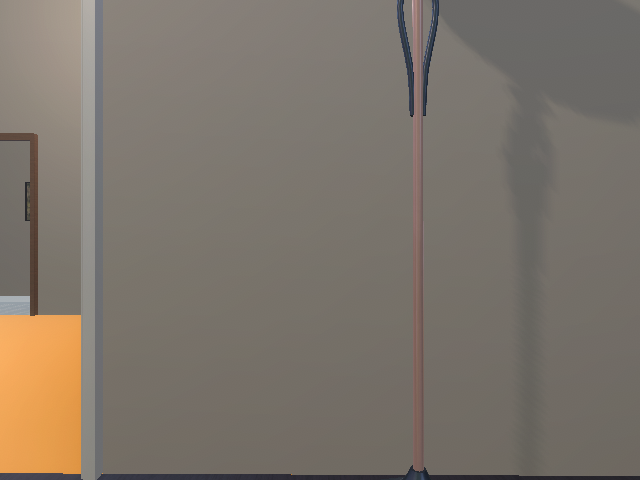"}
        \caption{}
    \end{subfigure}\hfill
    \begin{subfigure}{0.48\textwidth}
        \includegraphics[width=\linewidth]{"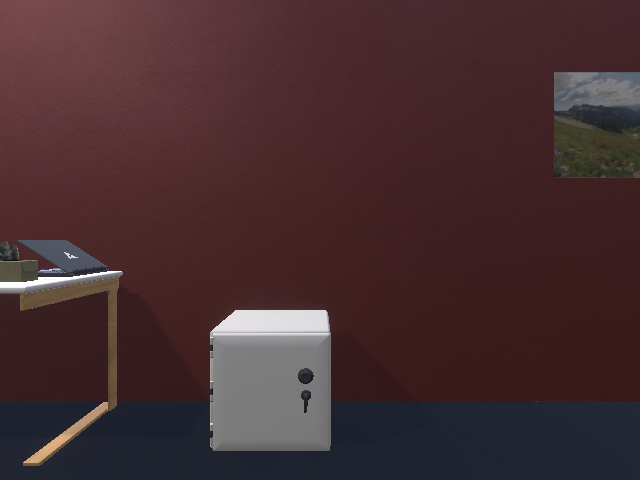"}
        \caption{}
        \label{fig:llm_comm:captcha}
    \end{subfigure}
    \caption{Ground agent observations which result in error with \textit{GPT-4-turbo}.}
    \label{fig:llm_comm:adverserial}
\end{figure}

Fig.~\ref{fig:llm_comm:adverserial} shows some examples of such image inputs. As evident from these, images mainly containing uninteresting features such as walls cause the error. This could be due to safety measures put in place for \textit{GPT-4-turbo} to avoid adversarial attacks. Regardless, these cases showcase a vulnerability for robots equipped with similar LLMs. It is not uncommon for the robots to face walls during their navigation. However, observations from such trivial cases may require adding measures for embodied agents to deal with potential failures. We did not use the episodes with such examples in our evaluation.

% \chapter{Real World Experiments for Assisted ObjectNav}
\section{Real World Experiments}
\label{sec:llm_comm:append:llm_comm:localization_fail}
\subsection{Localization Failures}
In a communication setup, one may expect the cross-view localization to be trivial as VLM can match the objects in the two images (from GA and OA) to guess the robot's location. However, we found that VLM struggles at doing so. 

\begin{figure}[h]
    \centering
    \begin{subfigure}{0.49\textwidth}
        \includegraphics[width=\linewidth]{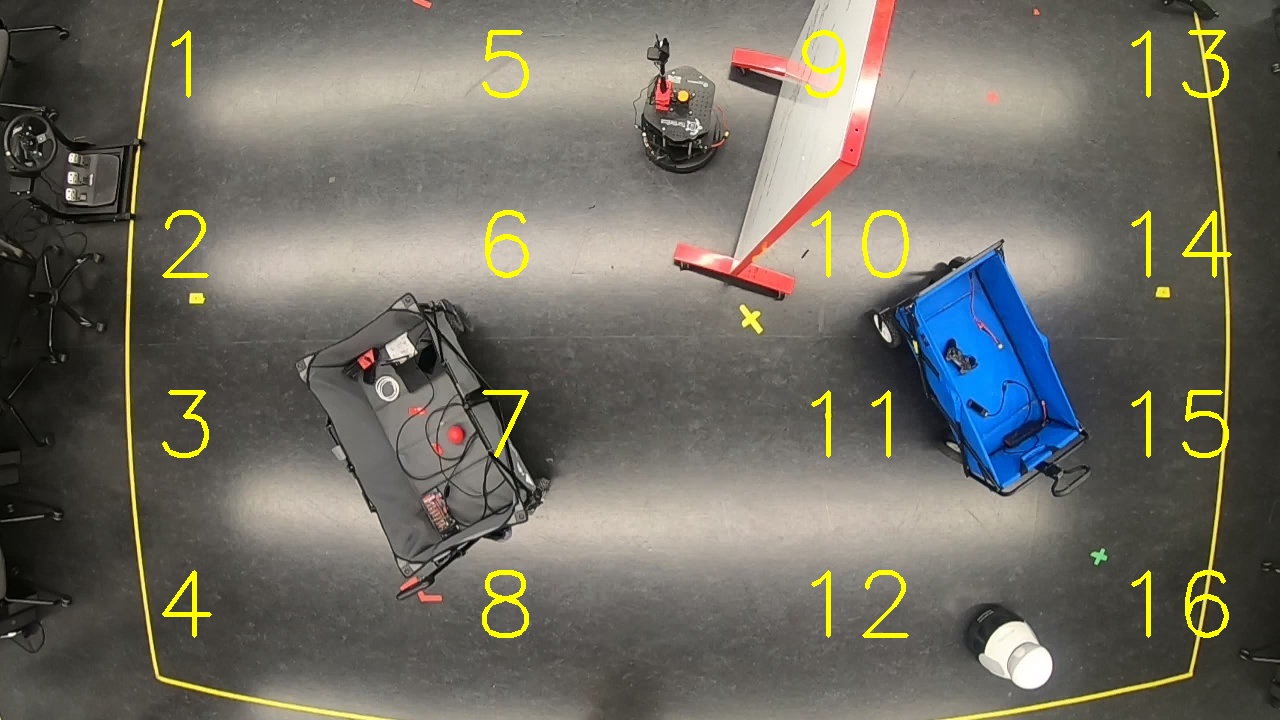}
        \caption{Predicted position: 6}
    \end{subfigure}\hfill
    \begin{subfigure}{0.49\textwidth}
        \includegraphics[width=\linewidth]{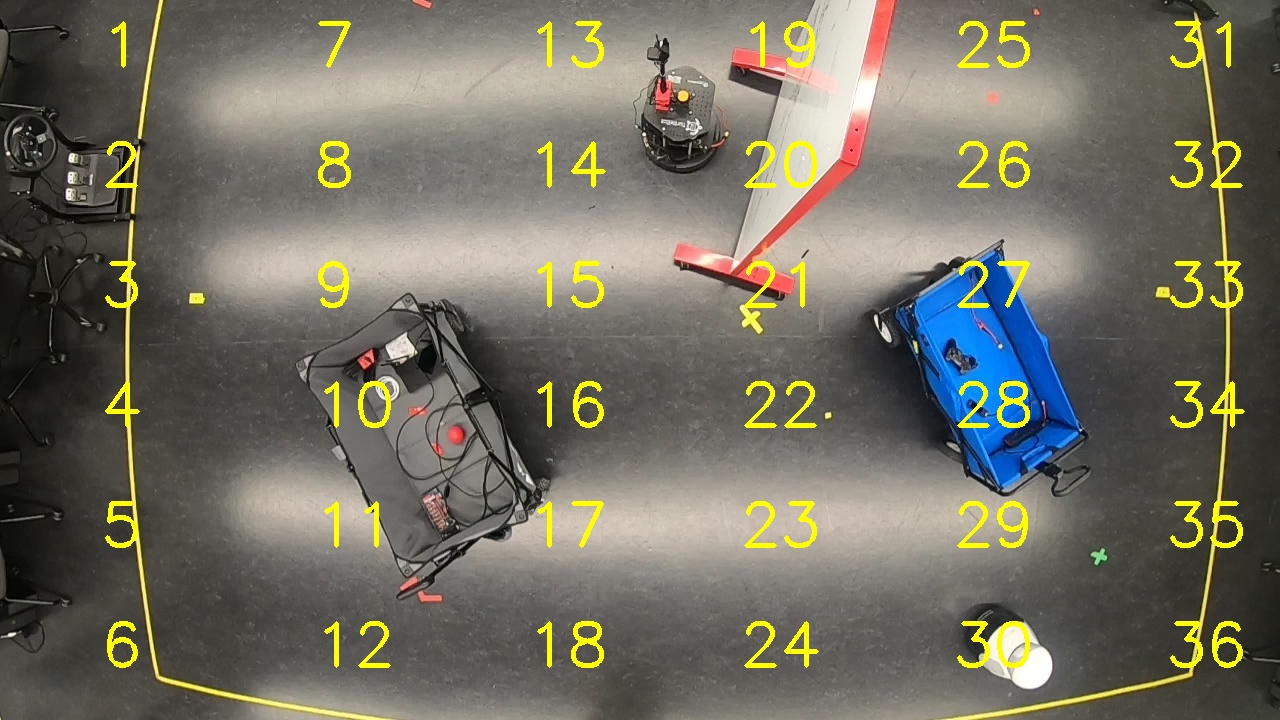}
        \caption{Predicted position: 16}
    \end{subfigure}\vfill
    \begin{subfigure}{0.49\textwidth}
        \includegraphics[width=\linewidth]{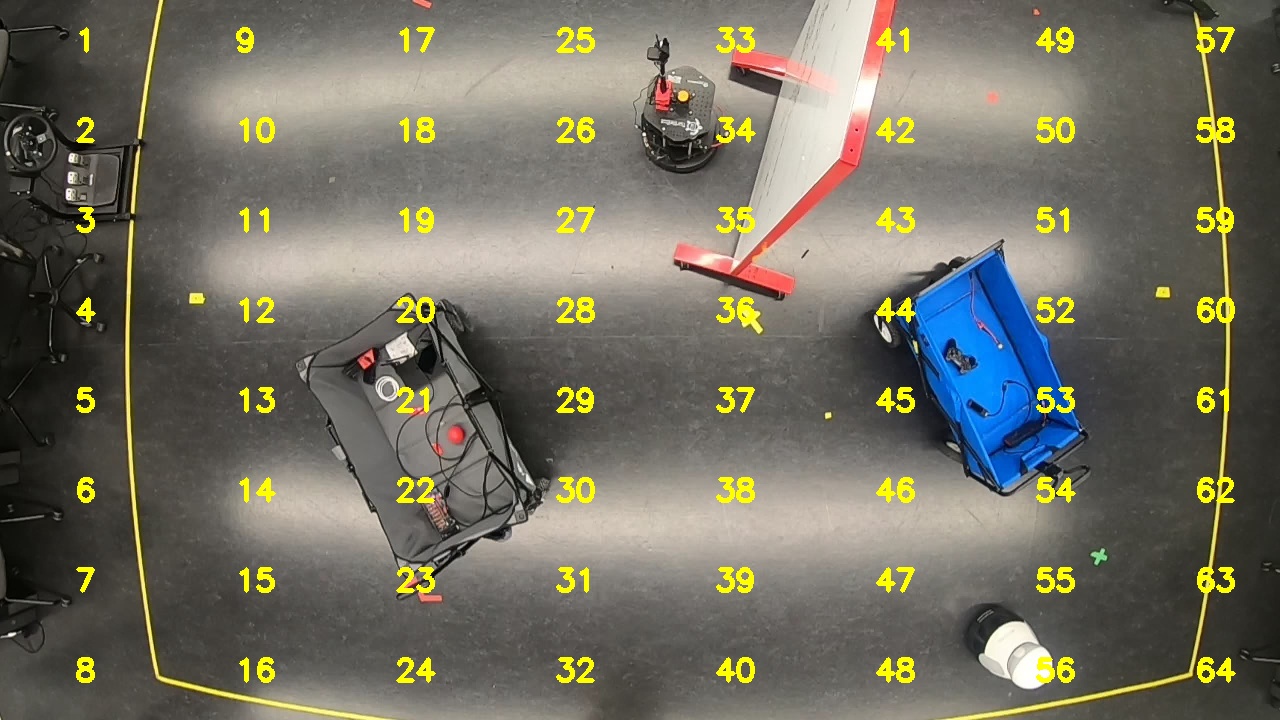}
        \caption{Predicted position: 21}
    \end{subfigure}\hfill
    \begin{subfigure}{0.49\textwidth}
        \includegraphics[width=\linewidth]{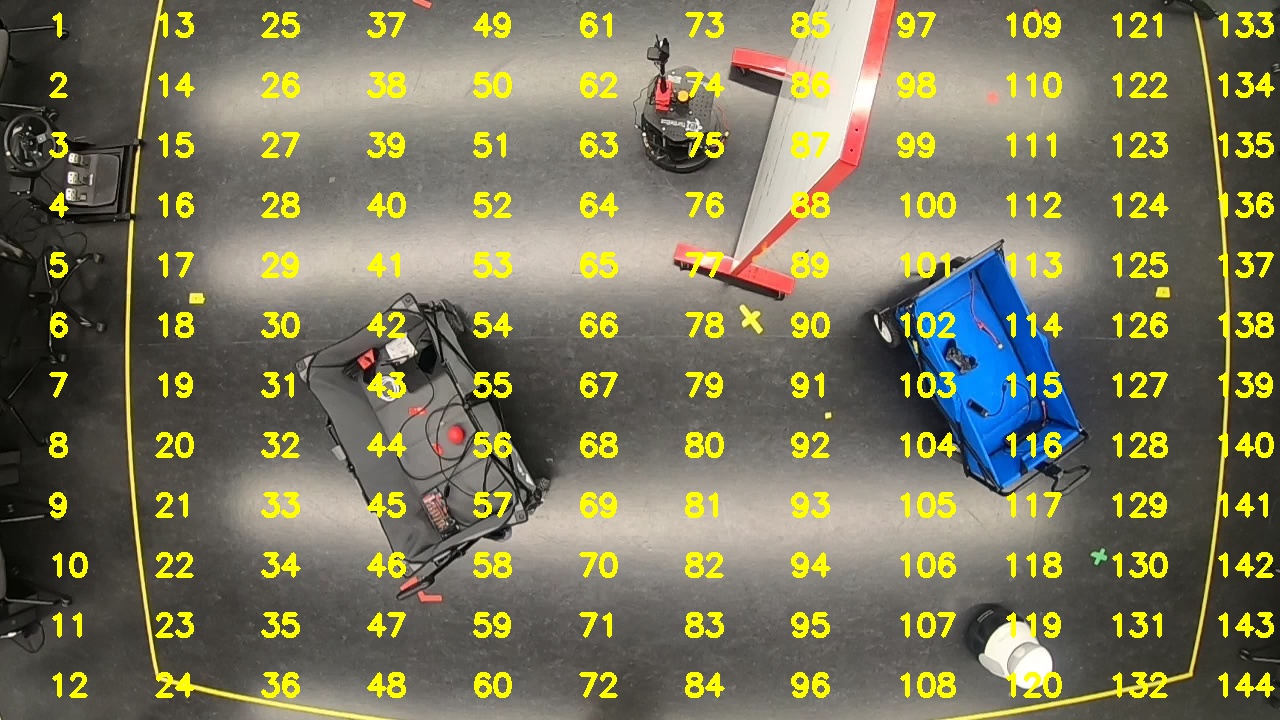}
        \caption{Predicted position: 42}
    \end{subfigure}\hfill
    \caption{Localization responses from \textit{GPT-4-turbo} for varying extent of labelling.}
    \label{fig:llm_comm:localization_size}
\end{figure}

\begin{figure}[h]
    \centering
    \begin{subfigure}{0.49\textwidth}
        \includegraphics[width=\linewidth]{figs/image_23_grid_6.jpg}
        \caption{Predicted position: 16}
    \end{subfigure}\hfill
    \begin{subfigure}{0.49\textwidth}
        \includegraphics[width=\linewidth]{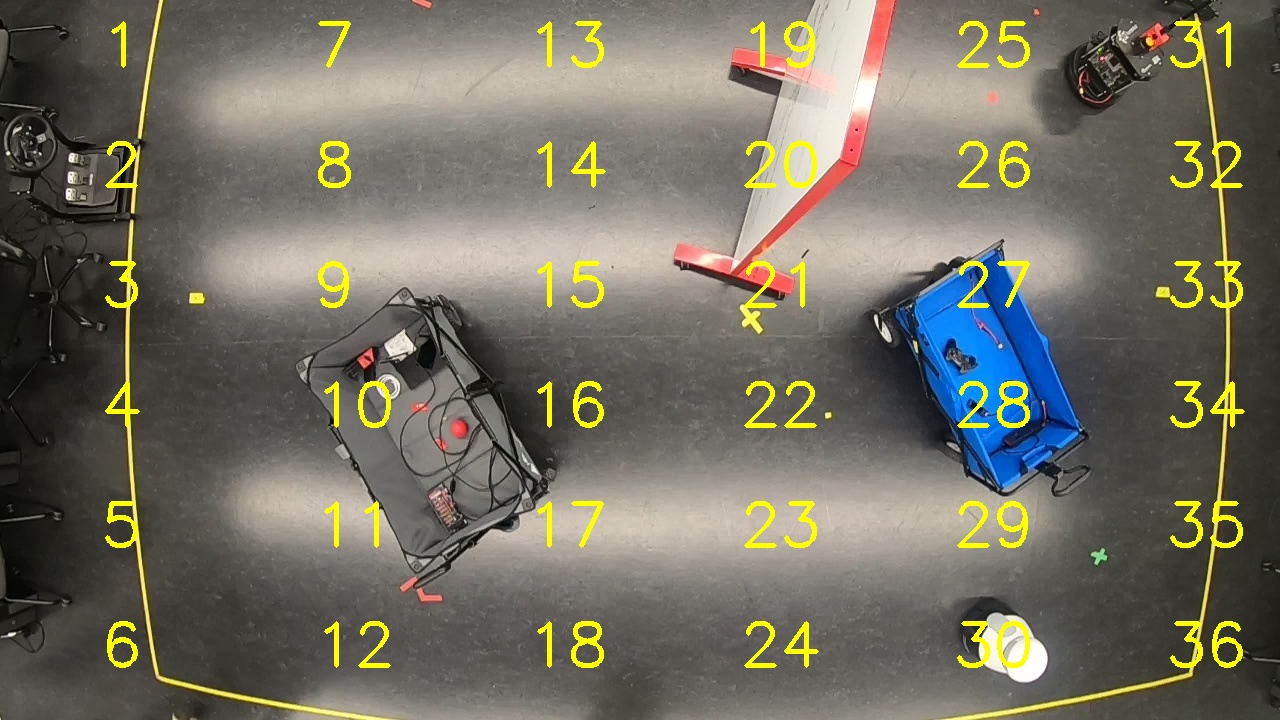}
        \caption{Predicted position: 30}
    \end{subfigure}\vfill
    \begin{subfigure}{0.49\textwidth}
        \includegraphics[width=\linewidth]{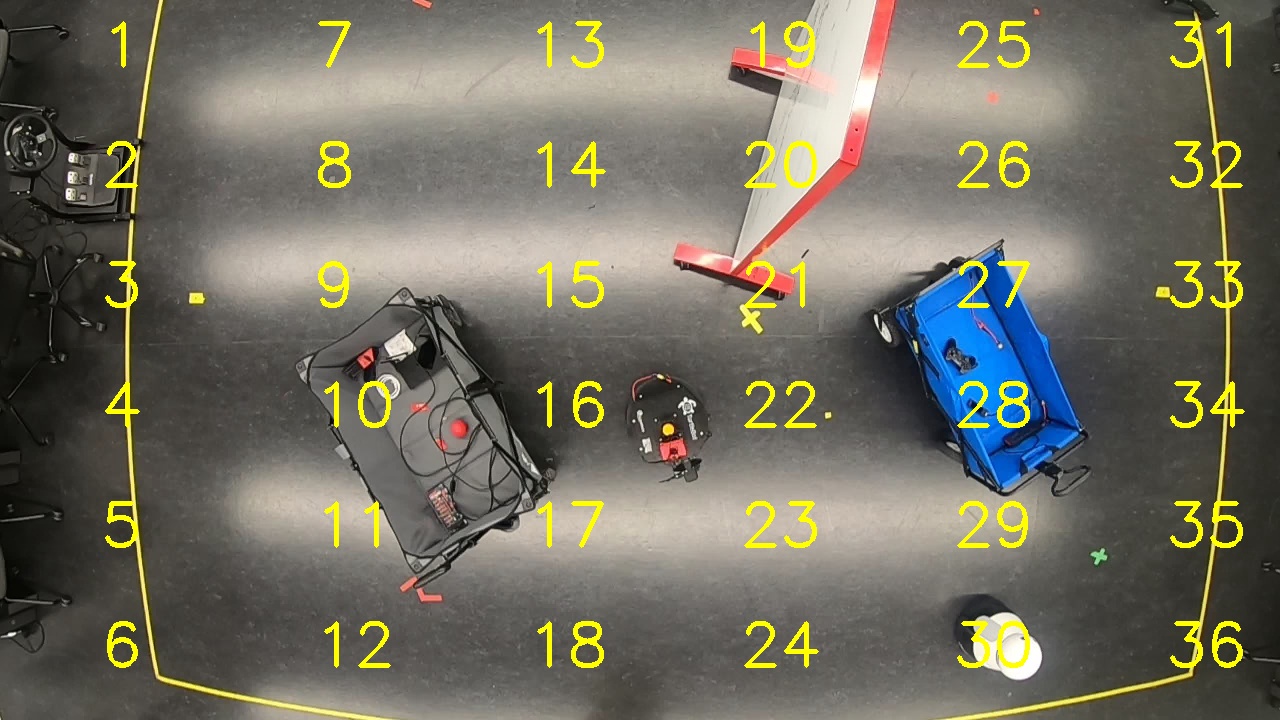}
        \caption{Predicted position: 16}
    \end{subfigure}\hfill
    \begin{subfigure}{0.49\textwidth}
        \includegraphics[width=\linewidth]{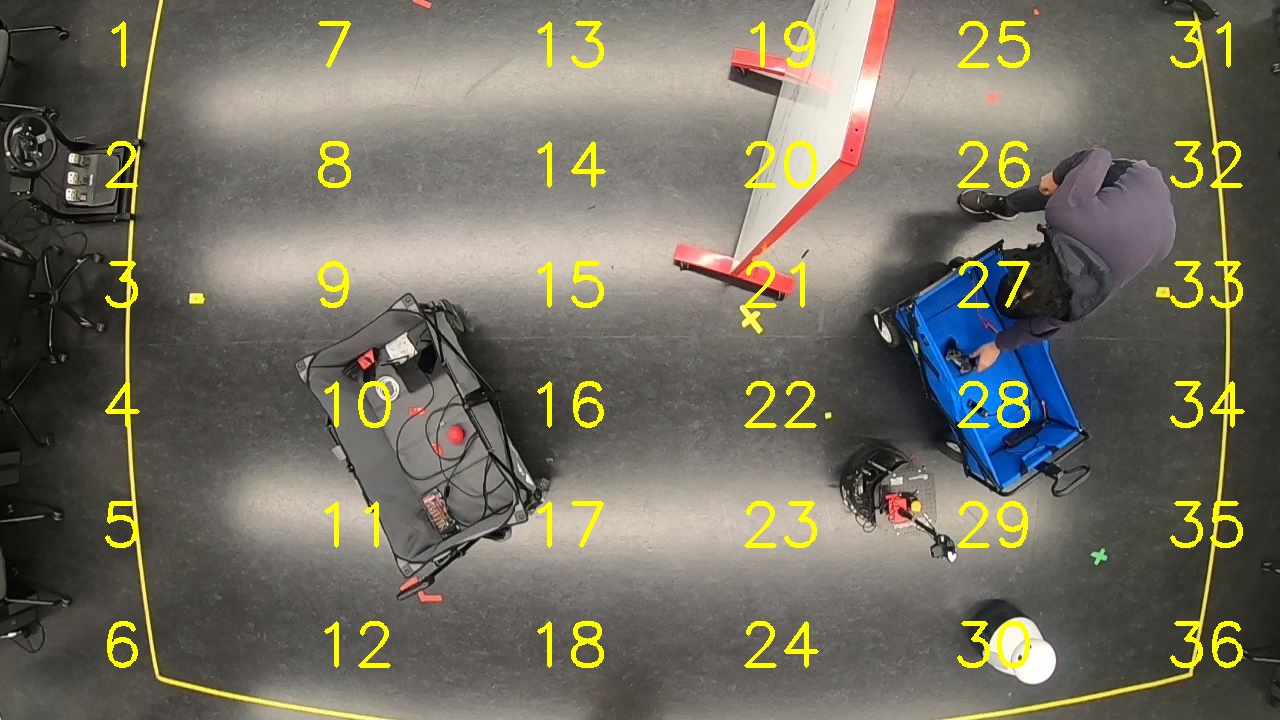}
        \caption{Predicted position: 9}
    \end{subfigure}\hfill
    \caption{Localization responses from \textit{GPT-4-turbo} for different locations of the Turtlebot2 robot.}
    \label{fig:llm_comm:localization_pose}
\end{figure}

For this experiment, we use a Turtlebot2 robot in an indoor environment with a few other objects (carts with objects inside, whiteboard, driving simulator, etc). The Turtlebot2 is black in color and has a range sensor with an orange cap on top.  For the overhead view, we put a GoPro 7 camera on top. Inspired by PIVOT~\cite{nasiriany2024pivot}, we added labels to the overhead image and asked the VLM agent to find `a black Turtlebo2 robot with an orange sensor on top'. To add the labels, we split the image into grids with 16, 36, 64, and 144 cells and varied the label sizes to ensure that the labels were equally spaced without any overlap. As shown in Figure~\ref{fig:llm_comm:localization_size}, the results are comparatively better with smaller grids but not none of the results are precise.
A closer look reveals that the VLM may be confused due to similar-looking objects in the scene (a grey cart with red and range objects inside). This is also shown in Figure~\ref{fig:llm_comm:localization_pose}, which shows the Turtlebot at different locations and the position predicted by the VLM. 

\begin{figure}[h]
    \centering
    \begin{subfigure}{0.4\textwidth}
        \includegraphics[width=\linewidth]{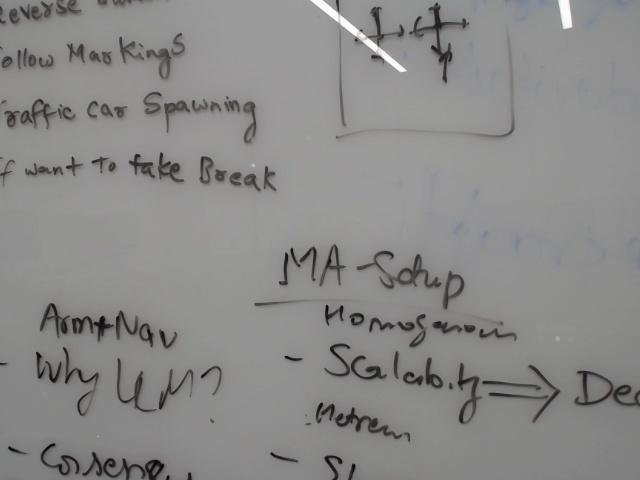}
        \caption{Predicted position: 16}
    \end{subfigure}\hfill
    \begin{subfigure}{0.4\textwidth}
        \includegraphics[width=\linewidth]{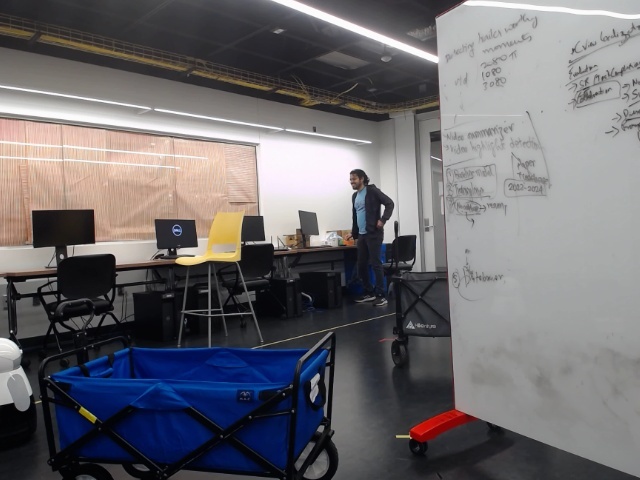}
        \caption{Predicted position: 10}
    \end{subfigure}\vfill
    \begin{subfigure}{0.4\textwidth}
        \includegraphics[width=\linewidth]{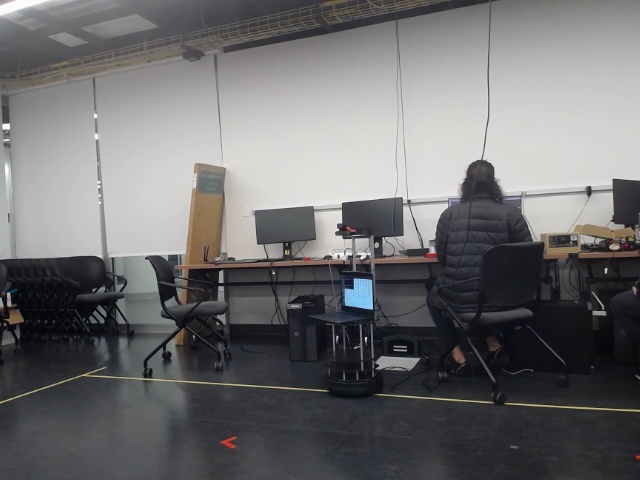}
        \caption{Predicted position: 16}
    \end{subfigure}\hfill
    \begin{subfigure}{0.4\textwidth}
        \includegraphics[width=\linewidth]{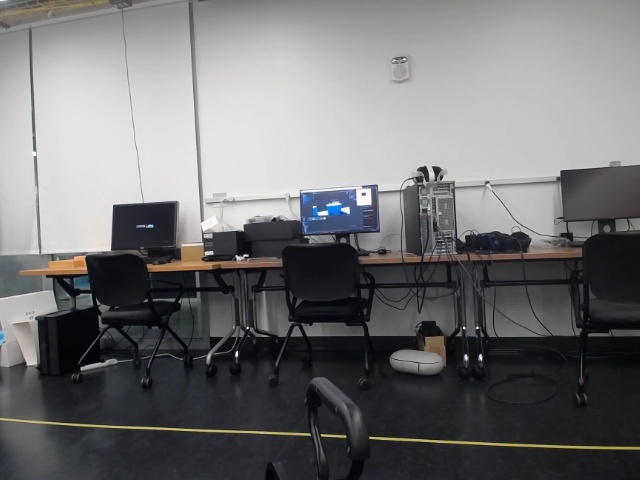}
        \caption{Predicted position: 10}
    \end{subfigure}\hfill
    \caption{Egocentric images from the camera on Turtlebot2. These images correspond to images of robot locations shown in  Figure~\ref{fig:llm_comm:localization_pose}}
    \label{fig:llm_comm:localization_ego}
\end{figure}

We also tried adding the egocentric images from the camera on Turtlebot, as shown in Figure~\ref{fig:llm_comm:localization_ego}, as additional inputs to VLM to help with cross-view localization, but they do not improve the results. 

\subsection{Finetuning Prompts}

Our prompt in the simulator is described in Appendix \ref{sec:llm_comm:append:llm_comm:prompt}. In the real world, we add additional constraints to allow the agent to localize the agent with the targets detected.

% \textbf{Overhead Agent - System Prompt}:\\
% \coloredbox{blue!10}{
\promptbox{Overhead Agent - System Prompt}{
You are an Overhead Agent (OA) with a top view in an environment. You must identify and guide a Ground Agent (GA) to reach the <target object>. \\
The GA is a turtlebot with a circular orange box on it. Be sure to use the ground agent's perspective. Also, note there is a reflection from multiple rectangular white lights on the black floor."
\\
$<$Dialogue between the agents$>$
}

% \textbf{Overhead Agent - User Prompt}:\\
% \coloredbox{blue!10}{
\promptbox{Overhead Agent - User Prompt}{
Give me a dialogue questioning the GA about its surroundings. The GA can not move while speaking with you. \\
You must query the GA about the environment and try to estimate its position and orientation with respect to the <target object>. You have {$\mathcal{C}_{Len}$ - 1} questions left to ask. \\
\\
$<$Dialogue between the agents$>$
}

% \textbf{Ground Agent - System Prompt}:\\
% \coloredbox{blue!10}{
\promptbox{Ground Agent - System Prompt}{
You are a Ground Agent with an egocentric view. You need to reach the <target object>.\\
You will be questioned by an Overhead Agent about things you see. Be very concise with your answers.
}

% \textbf{Ground Agent - User Prompt}:\\
% \coloredbox{blue!10}{
\promptbox{Ground Agent - User Prompt}{
$<$Dialogue between the agents$>$
}

Note here that we ask the OA to use environmental cues to figure out the position and orientation of the agent in the communication phase. This explicit mention helps curb pre-emptive hallucinations. Additionally, we also provide unique identifying details about the ground agent (with an orange box here, for instance).

\begin{figure}[H]
    \centering
    \includegraphics[width=\linewidth]{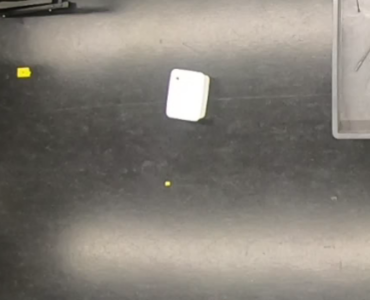}
    \caption[An example showing the need for environment-specific fine-tuning. Upon passing the simulator prompts for conversation to this image, the overhead agent would misidentify the lights as a table. Explicitly mentioning the presence of white lights on the floor alleviates this issue.]{\textbf{Environment-Specific Finetuning}: Upon passing the simulator prompts for conversation to this image, the overhead agent would misidentify the lights as a table. Explicitly mentioning the presence of white lights on the floor alleviates this issue.}
    \label{fig:llm_comm:reflection}
\end{figure}

Additional information such as \textit{``note there is a reflection from multiple rectangular white lights on the black floor''} is environment-specific, and allows the VLM to adapt better. Prior to providing such an instruction, the agent kept hallucinating while reflections on the floor to be a table. This is illustrated in Figure \ref{fig:llm_comm:reflection}.

\subsection{Identifying Correct Action from Dialogue}
\begin{figure}[H]
    \centering
    \includegraphics[width=\linewidth]{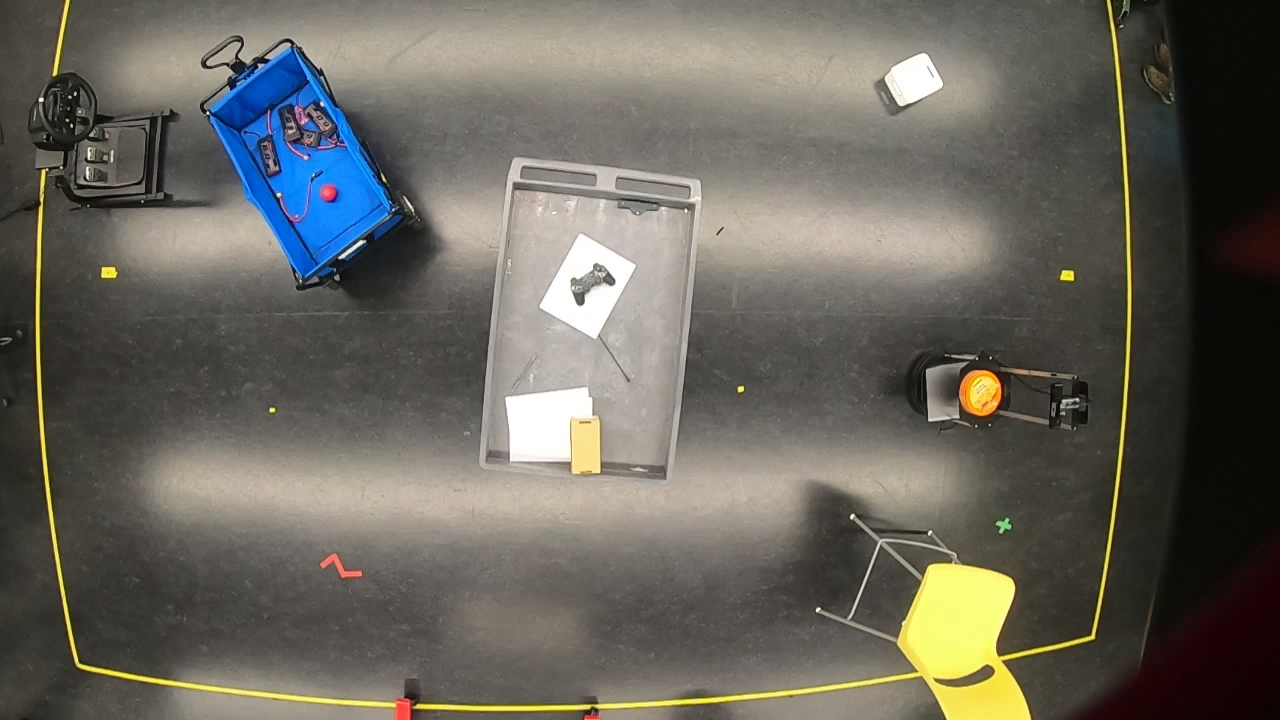}
    \caption[Corrective action with GA in the real-world experiments. In this case, the agent decides to rotate by 180 degrees first and then move towards the left. The OA in the execution phase correctly suggests that the same can be accomplished by rotating towards the right.]{\textbf{Corrective Action with GA}: In this case, the agent decides to rotate by 180 degrees first and then move towards the left. The OA in the execution phase correctly suggests that the same can be accomplished by rotating towards the right.}
    \label{fig:llm_comm:180deg_case}
\end{figure}

In many cases, the planning failure occurs due to the preemptive action hallucination in the conversation. During the execution phase, $\mathcal{E}$, the OA looks at this conversation and tries to execute an action to continue toward the target object. However, it may provide the right action if prompt fine-tuning reduces the preemptive action hallucination. Figure~\ref{fig:llm_comm:180deg_case} shows this case where the target is the white basket on the ground and the GA is currently looking towards the center of the image.

\textbf{Conversation Phase:}\\
\textbf{OA:} Can you describe any nearby objects or structures that are directly in front of you?\\
\textbf{GA:} Directly in front of me, there is a work area with several desks and office chairs. There's a person sitting at one of the desks, working on a computer. To the left, there's a door, and overhead are strip lighting fixtures and exposed ceiling infrastructure. The foreground shows a metal cart with various items, possibly tools or mechanical components.
\textbf{OA:} What do you see to your right and behind you?\\
\textbf{GA:} To my right, there is an office area with multiple desks equipped with computer monitors and an office chair. Behind me, there is more of the workspace visible, including additional desks and chairs. A whiteboard is visible on the right wall, and there's a large technical rigging above the workspace, likely part of the lab or office's infrastructure.\\
\textbf{OA:} Based on the information you provided and the overhead view, you are currently facing away from the small white basket on the floor. To reach the basket, first rotate 180 degrees, then move forward until you reach the basket positioned on your left side after the turn.\\
\textbf{GA:} Understood. I will rotate 180 degrees and then advance to locate the small white basket positioned to the left after the turn.\\

\textbf{Execution Phase:}\\
\textbf{Prescribed Action:} RotateRight

When asked to provide a reason, GPT-4 responded with the following:\\
\textbf{OA:} Based on the overhead image and the Ground Agent's plan to rotate 180 degrees, a single command of "Rotate Right" will initiate this maneuver. 

We also note that the GA described the objects behind it even though it can only get observations from the front-looking camera. OA was able to correctly estimate the GA's location relative to the white basket as it gathered information about its observations to correctly estimate its pose. 

\section{Selective Communication}
 While communication with the OA helps the GA with ObjectNav, bandwidth, and cost restrictions may require a selective approach toward initiating communication~\cite{liu2020when2com, arul2022dmca}. To investigate such situations, we developed a variation of \textit{Cooperative Action} where the ground agent is first asked if it wishes to seek help from the overhead agent. The view from the ground agent camera is used as an additional input here. We use the simulation setup as described in section~\ref{subsec:llm_comm:simualtion_setup} for 100 episodes and report our findings in Table~\ref{tab:selective_comm}. We observe that such a \textit{selective communication} setup results in an improvement over the Cooperative Action setup, suggesting a judicious assistance-seeking behavior. However, we note that it comes at the cost of additional prompting at every episode.
 
\begin{table}[ht]
\centering
\begin{tabular}{ccrrr}
\toprule
\textbf{Comm. Length ($\mathcal{C}_{Len}$)} & \textbf{OSR \%} $\uparrow$ & \textbf{SPL \%} $\uparrow$ \\
\midrule
1 & 20.00 & 19.20 \\
3 & 23.00 & 21.65 \\
5 & 26.00 & 24.49 \\
\bottomrule
\end{tabular}
\vspace{0.3cm}
\caption[Performance of the \textit{Selective Communication} setup. We find that letting the GA choose whether to communicate with the agent or not results in a better ObjectNav performance than the fully cooperative setup.]{\textbf{Selective Communication Performance}: We find that letting the GA choose whether to communicate with the agent or not results in a better ObjectNav performance than \textbf{Cooperative Action}, but requires additional prompting at each step.}
\label{tab:selective_comm}
\end{table}

\section{Communication Wordclouds}
\begin{figure}[ht]
    \centering
    \begin{subfigure}{0.49\textwidth}
        \includegraphics[width=\linewidth]{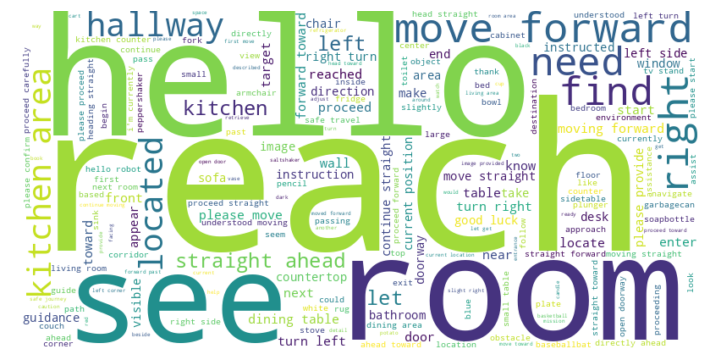}
        \caption{Word cloud for $\mathcal{C}_{Len} = 1$}
    \end{subfigure}\hfill
    \begin{subfigure}{0.49\textwidth}
        \includegraphics[width=\linewidth]{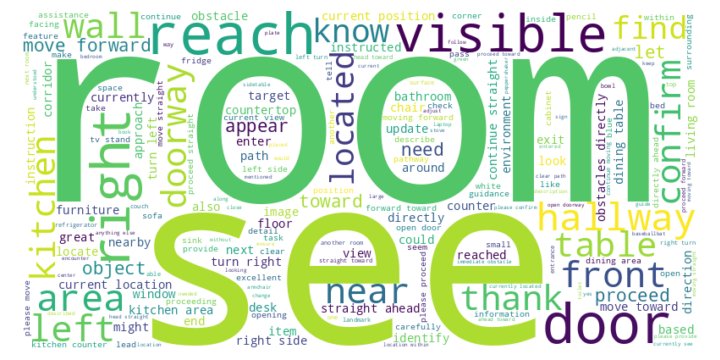}
        \caption{Word cloud for $\mathcal{C}_{Len} = 3$}
    \end{subfigure}\vfill
    \begin{subfigure}{0.49\textwidth}
        \includegraphics[width=\linewidth]{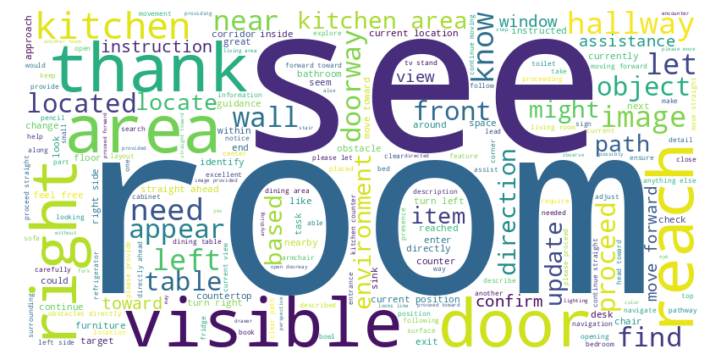}
        \caption{Word cloud for $\mathcal{C}_{Len} = 5$}
    \end{subfigure}\hfill
    \caption{World clouds over the communication between the GA and OA in \textit{Selective Action} setup. Communication in \textit{Cooperative Action} setup also exhibits similar patterns.}
    \label{fig:llm_comm:wordclouds}
\end{figure}

To identify the patterns in GC, we look at the word cloud over the communication between GA and OA. Figure~\ref{fig:llm_comm:wordclouds} shows the word clouds for the \textit{Selective Action} setup. We observe that across all  $\mathcal{C}_{Len}$, the `rooms', `see', and `visible' are prominent as the OA always asks GA questions about surroundings and whether it sees the target object. 

For $\mathcal{C}_{Len} = 1$, we also see more prominence of action commands such as `forward' and `move' as the limited communication length may make the OA may suggest actions to reach the target right away. Some of these suggestions may instruct the GA to `reach' certain objects before reaching the target object. We also see `hello' having more weight for this case compared to other communication lengths as the OA greets the GA at the beginning of the conversation in most episodes. 

$\mathcal{C}_{Len} = 3$ have similar word clouds for words related to queries about the surroundings. Due to the large communication length, the OA may have the chance to suggest going to other rooms through `doors'. We observe similar trends for the \textit{Cooperative Action} setup as well.

